Doctoral Thesis

# Monolingual Alignment of Word Senses and Definitions in Lexicographical Resources

Sina Ahmadi

M.Sc., Paris Descartes University, 2017

M.A., Sorbonne Nouvelle University, 2016

B.Eng., University of Kurdistan, 2014

**Supervisor**

Dr. John P. McCrae

**External Examiner**

Prof. Marie-Claude L'Homme

**Internal Examiner**

Dr. James McDermott

*Thesis Submitted in Partial Fulfillment of the Requirements for the Degree of*

***Doctor of Philosophy***

in the

School of Computer Science

College of Science and Engineering

National University of Ireland Galway

Spring 2022

# ABSTRACT

Dictionaries are fundamental resources for people to learn and document languages as well as for computers to process natural languages. A dictionary provides a fine-grained structure and description of the vocabulary of a language. With decades of advances in electronic lexicography, a significant amount of lexicographical resources are currently available. Such resources are the fruits of elaborate and strenuous efforts of lexicographers and oftentimes, are costly projects to initiate and maintain. Moreover, given the increasing number of lexical semantic resources thanks to community-driven initiatives such as Wiktionary, the alignment of such resources is of importance to promote interoperability and increase their exploitation more effectively. On the other hand, the significant progress in the field of computer science, artificial intelligence and the semantic web has been tremendously beneficial to various scientific fields, particularly language technology. Therefore, there is an opportunity to leverage the current techniques and resources to facilitate the automatic alignment, integration and enrichment of lexicographical data.

The focus of this thesis is broadly on the alignment of lexicographical data, particularly dictionaries. In order to tackle some of the challenges in this field, two main tasks of word sense alignment and translation inference are addressed. The first task aims to find an optimal alignment given the sense definitions of a headword in two different monolingual dictionaries. This is a challenging task, especially due to differences in sense granularity, coverage and description in two resources. After describing the characteristics of various lexical semantic resources, we introduce a benchmark containing 17 datasets of 15 languages where monolingual word senses and definitions are manually annotated across different resources by experts. In the creation of the benchmark, lexicographers' knowledge is incorporated through the annotations where a semantic relation, namely exact, narrower, broader, related or none, is selected for each sense pair. This benchmark can be used for evaluation purposes of word-sense alignment systems. The performance of a few alignment techniques based on textual and non-textual semantic similarity detection and semantic relation induction is evaluated using the benchmark. Finally, we extend this work to translation inference where translation pairs are induced to generate bilingual lexicons in an unsupervised way using various approaches based on graph analysis. This task is of particular interest for the creation of lexicographical resources for less-resourced and under-represented languages and also, assists in increasing coverage of the existing resources. From a practical point of view, the techniques and methods that are developed in this thesis are implemented within a tool that can facilitate the alignment task.

# CONTENTS

# DECLARATION

I declare that this thesis, titled "*Monolingual Alignment of Word Senses and Definitions in Lexicographical Resources*", is composed by myself, that the work contained herein is my own except where explicitly stated otherwise in the text, and that this work has not been submitted for any other degree or professional qualification.

Galway, March 7, 2022

Sina Ahmadi

# ACKNOWLEDGEMENTS

Writing this section has inspired me since the very beginning. Perhaps because I was wondering how my Ph.D. journey would shape this text at the end. And here is how it goes.

This thesis is the result of my past four years of work as a Ph.D. researcher. Although I am sure this work is not flawless, I must say that finishing up my Ph.D. during Covid-19 gave me immense satisfaction. Starting my Ph.D. on an exceptionally sunny day in April 2018 in Galway, I had a sweet episode of my life full of new ideas in research, rewarding experiences in engineering, papers, and conferences that were typical of Ph.D. life at the time. To my surprise, I even rarely struggled with the type of challenges that some of my fellows complained about, such as making progress, carrying out experiments, catching deadlines to submit papers, or having a life while doing a Ph.D. However, the impact of the physical restrictions and psychological burden that Covid-19 imposed on everyone, including myself, were undeniable.

2020 was a particularly harsh year for everyone but even worse for international postgraduate students living in Ireland. Doing a Ph.D. during Covid-19 did not only require taking care of regular activities but also learning to cope in the new mainly virtual world while living in a shared house and dealing with the emerging mental health issues far away from the loved ones. Now that I look back, I feel that I should be gleeful that those daunting days are gone, despite the long-lasting effects.

Regardless of the problems, I enjoyed every single day of this journey and will always remember it with sweet memories. Living in Ireland was not without challenge, but at least, it taught me to be more grateful for having things that we usually take for granted: the generous sun, more predictable weather, good food, fresh artisan bread, decent accommodation, museums, and a more affordable place to live. Nevertheless, I will miss this place for the friendly Irish people, the grasslands that make running more joyful, the biodiversity and the wildlife, particularly those robins singing at night!

This work could not have been accomplished without the help of many people. First and foremost, I would like to thank my esteemed supervisor, Dr. John McCrae, for his invaluable supervision, wise pragmatism, support, tutelage, and kindness during the course of my Ph.D. degree. My gratitude extends to the ELEXIS project for the funding opportunity to undertake my studies at the School of Computer Science at NUI Galway. My sincere thanks also go to Dr. Mathieu d'Aquin and Dr. Paul Buitelaar who played an important role as my graduate research committee members, and to the examination committee members, Dr. Marie-Claude L'Homme and Dr. James McDermott, for their constructive comments and invaluable insights.

This journey could not be this memorable without my friends at Insight: Tobias Daudert, Joana Barros, Omnia Zayed, Adrian Ó Dubhghaill, Oksana Dereza, Sarah Carter, Niki Pavlopoulou, and Brendan Smith. Many thanks to the administrative staff at Insight and ELEXIS for their support: Hilda Fitzpatrick, Christiane Leahy-Coen, and Anna Woldrich. My Ph.D. experience would have certainly been different without the compassionate friends who helped me in many ways, especially by hosting me with open arms when I had difficulty finding accommodation: Housam Ziad and his lovely family, Aftab Alam and Dr. Theodorus Fransen. Likewise, thanks to Jalal Sajadi and Daban Q. Jaff for their true friendship over the years. Special thanks to Dr. Sanni Nimb who hosted me during my visit to the Society for Danish Language and Literature and the Centre for Language Technology in Copenhagen in 2019 which was an enriching experience in many ways. I would also like to warmly thank Dr. Mathieu Constant for hosting me in Nancy in 2021. Throughout my visit, I had a great time sharing my office with Pauline Gillet and Charlène Weyh.

The accomplishment of my Ph.D. studies is important to me in two other ways too. First, it marks the end of my formal education which has been continuously going on since I remember. I partially owe this to France which provided me with free education and unique, eye-opening, rich, and unforgettable experiences. Also, I am indebted to Dr. Kyumars S. Esmaili who encouraged me during my bachelor's to follow my passion for languages, linguistics, and computer science by studying natural language processing. Second, it ends as the fourth decade of my life begins, so do my plans to contribute to society and have a more significant role "to make the world a better place", something that I whimsically told my parents when I left them and I am sometimes contemplative if ever it will be eventual! I will stay optimistic.

This brings me to the most important people in my life. There is no word to properly express my gratitude to my family back in Kurdistan, particularly to my parents who always believed in me, endorsed me in my choices, and even sacrificed themselves to make sure that their children have a fair chance to achieve their goals, more than their own generation. I understand that pursuing my studies imposed thousands of kilometers of distance between us and years of untogetherness that could not have passed this way without their support and unconditional love. *Spas û xoşewîstî*! I am also deeply grateful to my family in Greece who have always supported me, have given new meaning to my life with their kind hearts, positive vibes and love. *Άπειρα ευχαριστώ*!

Finally, I would like to thank the love of my life, Ioanna, from the bottom of my heart for being my best friend and companion and for her patience and encouragement over the years. It is impossible to imagine this journey getting to an end without her unending support and love. To her, I would say: *Je t'aime*.

# ACRONYMS

**BERT** Bidirectional Encoder Representations from Transformers
**BLI** Bilingual Lexicon Induction
**CL** Computational Linguistics
**CNN** Convolutional Neural Network
**ECD** Explanatory Combinatorial Dictionary
**FLN** French Lexical Network
**IAA** Inter-Annotator Agreement
**IC** Inverse Consultation
**LLOD** Linguistic Linked Open Data
**LSR** Lexical Semantic Resource
**LSTM** Long Short-Term Memory
**MRD** Machine-Readable Dictionary
**MTT** Meaning-Text Theory
**MWSA** Monolingual Word-Sense Alignment
**MWE** Multiword Expression
**NLP** Natural Language Processing
**NSM** Natural Semantic Metalanguage
**OED** Oxford English Dictionary
**OWL** Web Ontology Language
**RBM** Restricted Boltzmann Machine
**RDF** Resource Description Framework
**RNN** Recurrent Neural Network
**SKOS** Simple Knowledge Organization System
**SLR** Sentence Length Ratio
**STS** Semantic Textual Similarity
**SVM** Support Vector Machine
**TEI** Text Encoding Initiative
**TIAD** Translation Inference Across Dictionaries
**TLFi** *Trésor de la Langue Française informatisé*
**WBM** Weighted Bipartite Matching
**WSA** Word Sense Alignment
**WSD** Word Sense Disambiguation

# LIST OF FIGURES

# LIST OF TABLES

# 1 INTRODUCTION

> Dictionaries are treasure houses of data on the uses of words. They are also our best starting point for all questions regarding word sense distinctions, in natural language processing, the humanities or lexicography. But to reveal the dictionary's treasures in a systematic way is no simple task.
>
> Adam Kilgarriff (1992a, p. 381)

'Apple' used to refer to any type of fruit long before denoting what it means now. 'Welsh' used to mean 'foreigner' or 'slave' and 'Dutch' does basically mean 'German'. Anything stupid was 'nice' and sacred 'silly'. Yet more shockingly, 'Ethiopia' means 'burnt-face' and 'cretin' comes from 'Christian'! With the hope that these words could provoke your curiosity, I would like to highlight the importance of the broader topic on which this thesis focuses: *dictionaries*.

As one of the major components of the structure of a language, words play an essential role in understanding the nature of our languages to convey *meaning*. Depending on the level of analysis, words can be analyzed based on their pronunciation as in phonology, on their formation as in morphology, on their structure within a phrase as in syntax, on their meaning as in semantics and more broadly, on the context as in pragmatics. As the fundamental material of these linguistic fields, words are further studied, collected and maintained in a reliable and accurate way in dictionaries. Thanks to lexicography, we can argue about the etymology of words, like those mentioned above, and track the changes that some words have gone through over time. We also understand how civilizations and societies have evolved and influenced each other by examining the semantic changes that occur in some words as the distinctions between 'craft' and 'skill' under the influence of Old Norse and 'pig' and 'pork' under the influence of Old French on English.

With decades of advances in electronic lexicography, a significant amount of lexicographical resources are currently available, particularly in Europe. Such resources are the fruits of elaborate and strenuous efforts of lexicographers and oftentimes are costly projects to initiate and maintain. Therefore, there is an opportunity to leverage the current progress in computer science and information technology to facilitate electronic lexicography. On the other hand, the significant progress in the field of artificial intelligence has been tremendously beneficial to various scientific fields, particularly language technology and natural language processing (NLP). Despite the burgeoning advances in creating efficient methods with less reliance on manually-annotated data, various NLP applications are still heavily dependent on human-curated resources, such as dictionaries. Thus, current tools and applications can benefit from such reliable and systematic inventories of data to address data-driven approaches.

The focus of this thesis is broadly on the *alignment* of data of lexicographical nature, particularly expert-made ones such as dictionaries and lexical or conceptual resources such as WordNet (Miller, 1995), as well as collaboratively-curated ones such as Wikipedia[1] or Wiktionary[2]. The alignment task aims to identify identical words in resources and link various pieces of information within the lexical or semantic context. A dictionary provides a fine-grained structure of information about words, including entries that are in lemma form as the lemma BRING for 'brought' and 'brings', definitions, senses, usage examples and more elaborate properties such as semantic relationships between words. In a bilingual or multilingual context, such information can be found in a cross-lingual setup as well. Moreover, given the increasing number of lexico-semantic resources, thanks to community-driven initiatives such as Wiktionary and Open Multilingual WordNet[3], the alignment of such resources is of importance to promote interoperability and to facilitate the integration of various resources in a viable manner. This will ultimately pave the way for lexicographers to analyze lexical entries in various resources, verify the evolution of particular information, such as senses over time, and significantly reduce the lengthy editorial period and data collection burden.

Aligned resources have been shown to improve word, knowledge and domain coverage and increase multilingualism by creating new lexical resources such as YAGO (Suchanek et al., 2007), BabelNet (Navigli and Ponzetto, 2012a) and ConceptNet (Speer et al., 2017). In addition, they can improve the performance of NLP tasks such as word sense disambiguation (Navigli and Ponzetto, 2012b), semantic role tagging (Xue and Palmer, 2004) and semantic relations extraction (Swier and Stevenson, 2005). Despite the previous works addressing the alignment task, various challenges in this field require further studies. Some of these challenges are due to the various conceptual structures followed in different resources, e.g. network-based resources such as WordNet versus dictionary-based resources. Moreover, words have a non-static nature and their meanings may undergo changes over time as in old dictionaries of centuries ago in comparison to modern ones. In addition, the microstructure of a dictionary may contain phrasal texts – "a linguistic composition or utterance" according

1 https://www.wikipedia.org
2 https://www.wiktionary.org
3 http://compling.hss.ntu.edu.sg/omw

to (Pettersson, 2017) – particularly in sense definition and usage examples causing further complications and variations in meaning, as in metaphors and idioms.

Dictionaries are valuable resources that document the life of words in a language from various points of view. Senses, or definitions, are important components of dictionaries where dictionary entries, i.e. lemmata, are described in plain language. Therefore, unlike other properties such as references, cross-references, synonyms and antonyms, senses are unique in the sense that they are more descriptive but also highly contextualized. Moreover, unlike lemmata which remain identical through resources in the same language, except in spelling variations, sense definitions can undergo tremendous changes based on the choice of the editor, lexicographer and publication period, to mention but a few factors. Therefore, there is a need to create systems to detect semantic similarity between given definitions and propose alignment of word senses. Accordingly, this will facilitate the integration of various resources and the creation of inter-linked language resources.

Dictionaries are not only vast systematic inventories of information on words, they are also important as cultural and historical artifacts. Elaborate efforts are put into the development of lexicographic resources describing the languages of communities. Although confronted with similar problems relating to technologies for producing and making these resources available, cooperation on a larger European scale has long been limited. Consequently, the lexicographic landscape in Europe is currently rather heterogeneous. Many lexicographic resources have different levels of structuring and are not equally suitable for application in other fields. On the one hand, it is characterized by stand-alone lexicographic resources, which are typically encoded in incompatible data structures due to the isolation of efforts, prohibiting the reuse of this valuable data in other fields, such as NLP, linked open data and the semantic web, as well as in the context of digital humanities. On the other hand, there is a significant variation in the level of expertise and resources available to lexicographers across Europe. This forms a major obstacle to more ambitious, innovative and transnational approaches to dictionaries, both as tools and objects of research.

## 1.1 THE ELEXIS PROJECT

ELEXIS – the European Lexicographic Infrastructure[4] – proposes to integrate, extend and harmonize national and regional efforts in the field of lexicography, both modern and historical, with the goal of creating a sustainable infrastructure that will enable efficient access to high-quality lexical data in the digital age, and bridge the gap between more advanced and lesser-resourced scholarly communities working on lexicographic resources. The need for such infrastructure has clearly emerged out of the lexicographic community within the European network of electronic lexicography.

4 https://elex.is

The project develops strategies, tools and standards for extracting, structuring and linking lexicographic resources to unlock their full potential for language technology and linked open data, as well as in the context of digital humanities. The project also helps researchers create, access, share, link, analyze, and interpret heterogeneous lexicographic data across national borders, paving the way for ambitious, transnational, data-driven advancements in the field, while significantly reducing duplication of effort across disciplinary boundaries.

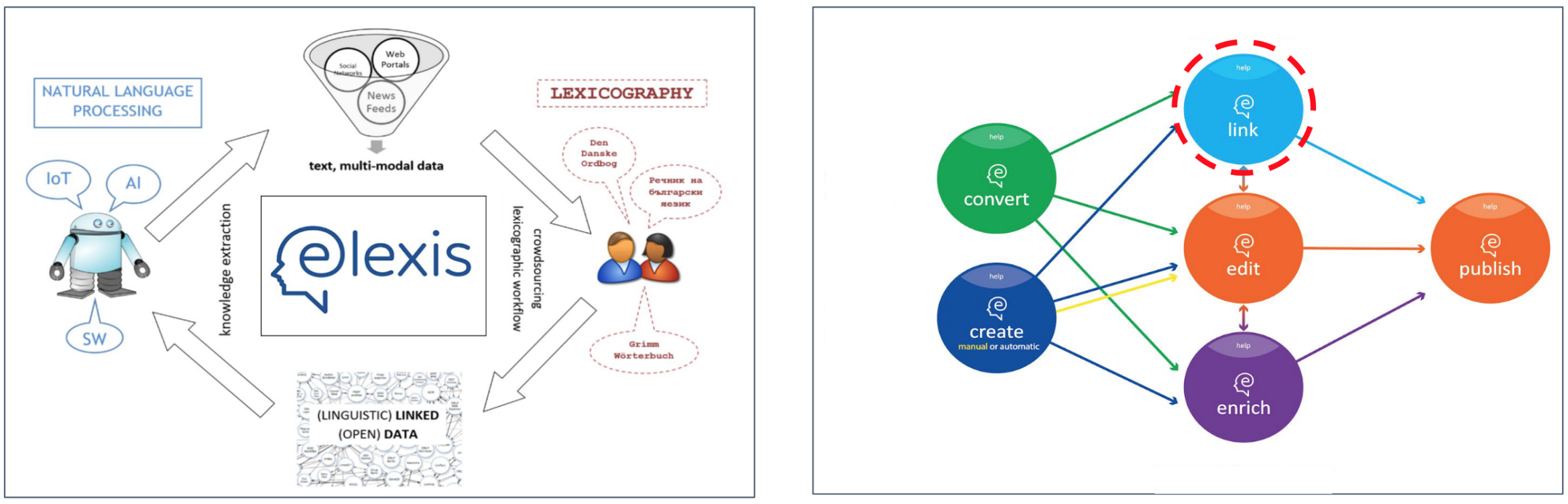

Figure 1.1: An overview of the objectives of the ELEXIS project. Our focus of linking within this project is encircled in dashed red.

ELEXIS is carried out by a balanced consortium with distributed geographical origins. It is composed of content-holding institutions and researchers with complementary backgrounds such as lexicography, digital humanities, language technology and standardization, a crucial feature required to address the multi-disciplinary objectives of the project. Figure 1.1 presents an overview of the objectives of the ELEXIS project (to the left) along with the outcomes of the project (to the right). In cooperation with CLARIN[5] and DARIAH[6], it also focuses on defining and providing common interoperability standards, workflows, conceptual models and data services as well as training and education activities focusing on user needs and cross-disciplinary advances.

Having the goal of making lexicographical data more accessible and processable, various research groups have come together within the ELEXIS project to address challenges in creating a lexicographic infrastructure. This PhD project is defined within the tasks of the ELEXIS project in general, and in particular, on one of its 10 work packages focused on interoperability and linked open data. This work package consists of the following tasks:

- Common models and protocols for lexicon access
- Semi-automatic linking of lexical resources
- Cross-lingual mapping through shared conceptualization
- Validation and quality assurance for lexical resources

5 https://www.clarin.eu
6 https://www.dariah.eu

Among these tasks, semi-automatic linking has been my topic of interest in this thesis. Therefore, my progress kept pace according to the objectives of ELEXIS.

## 1.2 MOTIVATION

In order to tackle some of the challenges in the semi-automatic linking of lexical resources, we address two main tasks of word sense alignment and translation inference, also referred to as bilingual lexicon induction, in a monolingual and cross-lingual setup, defined as follows:

- **Word sense alignment** (WSA): This is the task of finding the alignable senses among two entries with the same lemma and the same part-of-speech from two different dictionaries. While there is an increasing number of lexical resources, particularly expert-made ones such as WordNet (Miller, 1995) or FrameNet (Baker et al., 1998), as well as collaboratively-curated ones such as Wikipedia or Wiktionary, manual construction and maintenance of such resources is a cumbersome task. This can be efficiently addressed by NLP techniques. We demonstrate that what is perceived as a word sense in a dictionary is not essentially the same in another dictionary, particularly when it comes to defining such word senses. Therefore, we extend the alignment task to **semantic relationship detection** as well. In this sub-task, we not only determine that two word senses should be aligned but also specify the type of semantic relation that they would have. To this end, we use the following semantic relations between senses: exact, broader, narrower and related. In the WSA task, the focus is on the alignment of monolingual data, a task that we refer to as monolingual word sense alignment (MWSA).

- **Translation inference**: Given a set of words in two different dictionaries, the translation inference task aims to align dictionary headwords, regardless of spelling variations, that refer to the same senses or concepts in an *unsupervised* manner, i.e. without using any previously-aligned resource in the source and target languages or any bilingual dictionary. This task is deemed non-trivial and challenging due to the inconstant level of polysemy of words in a language and different coverage of senses in various resources (Søgaard et al., 2018). To this end, we focus on the data provided by the translation inference across dictionaries (TIAD) shared task (Gracia et al., 2019) and propose a few graph-based approaches. Addressing this task is beneficial not only to align existing lexicographical data but also to create new dictionaries for less-resourced and under-represented languages that lack such resources. In addition, inducing new translation pairs enables lexicographers to document words more efficiently and therefore, facilitates the dictionary compilation process. Ultimately, by creating an integrated, linked and interlinked resource, a huge amount of high-quality lexical data will not only become available to the linguistic, NLP and the semantic web community, it will also facilitate cutting-edge research in digital humanities.

## 1.3 RESEARCH QUESTIONS

This thesis attempts to answer the following research questions:

**RQ1.** Do graph-based methods improve the performance of lexicographical alignment systems?

**RQ1.1** What are the features in lexical and semantic data that can be effectively represented in the form of a graph?

**RQ1.2** Do path-based and cycle-based techniques in graph analysis capture polysemous items when inducing translations across bilingual dictionaries?

Graphs are powerful structures that have been widely used in various methods to represent data and the dependency between data instances. Modeling problems as bipartite graphs and aligning such graphs has been previously of interest in many problems where optimization of assignments within a task is required, as in the efficient assignment of reviewers to manuscripts (Liu et al., 2014) and donor organs to patients (Bertsimas et al., 2020). In the same vein, we would like to know how lexical and semantic data can be modeled as a graph. To this end, a network structure, named *lexicographic network*, is defined where nodes and edges respectively are lexicalized items and the relationship between them. This is then used for both of the tasks of word sense alignment and translation inference. This research question is addressed in Chapter 4.

**RQ2.** How can the output of a monolingual word sense alignment system be evaluated in a multilingual context?

**RQ2.1** What are the characteristics of senses and sense definitions in modern and historical dictionaries?

**RQ2.2** How can lexicographers' expertise be incorporated in the evaluation of alignment systems?

One of the conclusions of Chapter 4 where the first research questions are answered is the limitation of datasets that can be used for evaluating the performance of various systems in word sense alignment. To tackle this, the creation of a benchmark is described in Chapter 5. This benchmark contains 17 datasets covering 15 languages where sense definitions of expert-made resources, such as Webster's Dictionary, and collaboratively-curated resources, such as Wiktionary, are manually aligned and annotated by lexicographers and language experts. The focus of the benchmark is on the alignment of monolingual data of lexical semantic resources in such a way that sense definitions of identical lemmas with part-of-speech tags are aligned. We formalize linking as inferring links between pairs of senses as exact equivalents, partial equivalents, i.e. broader and narrower, or a more loose relation as related, or no relation between the two senses. This formulates the problem as a five-class classification for each pair of senses between the two dictionary entries. Furthermore, Chapter 5 sheds light on the major characteristics of sense definitions and the necessity to include semantic relations, namely exact, broader, narrower and related, to effectively capture the type of relationship that exists between sense definitions with nuances in meaning.

**RQ3.** What features in sense definitions can be used to create techniques for word sense alignment?

- **RQ3.1** What similarity metrics perform more effectively to estimate the similarity of sense definitions? Do the textual features extracted from definitions perform sufficiently efficiently in word sense alignment?
- **RQ3.2** What is the impact of additional information in sense definitions, such as cross-references or citations, on the alignment task? Does truncating sense definitions improve the estimation of similarity between a definitions pair?

This research question along with the sub-questions is addressed in Chapter 6 and concern more practical and experimental aspects of the alignment task. To this end, a set of similarity detection techniques based on textual and non-textual features are provided as well as linking constraints. These techniques are implemented and integrated in our data alignment tool—Naisc. Moreover, a few experiments are carried out to analyze the performance of various alignment techniques, particularly concerning semantic relation detection.

## 1.4 THESIS STRUCTURE

Following this introduction, a background is presented in Chapter 2 where some of the main lexical semantic resources are introduced. This chapter focuses on the challenging issues in aligning various resources, particularly polysemy, and describes different types of lexical semantic resources, such as dictionaries, network-based and ontological ones, along with their characteristics. Following this, a systematic literature review is provided in Chapter 3 where previous studies in linguistic data alignment in general, and monolingual word sense alignment, in particular, are reviewed. This chapter also presents the applications of lexical semantic resources in NLP, and sheds light on the limitations and hurdles in the field.

Having set the scene in the first three chapters, we proceed to the main contributions of the thesis that aim to answer and clarify the research questions. To do so, each chapter consists of a few sections. Firstly, we sketch the topics on which the chapter focuses where specific problems are defined and accordingly, the related work is provided. We then present our methodologies to address those problems along with the results of experiments and insights based on analyzing the experiments. At the end of each chapter, the main contributions and major limitations are discussed. This way, readers can benefit from an in-depth and rapid understanding of the progress that has been made and should be made in the future.

Finally, the thesis is concluded in Chapter 7 where the limitations and findings of the current work along with suggestions for future directions are discussed. Figure 1.2 schematizes the structure of the thesis.

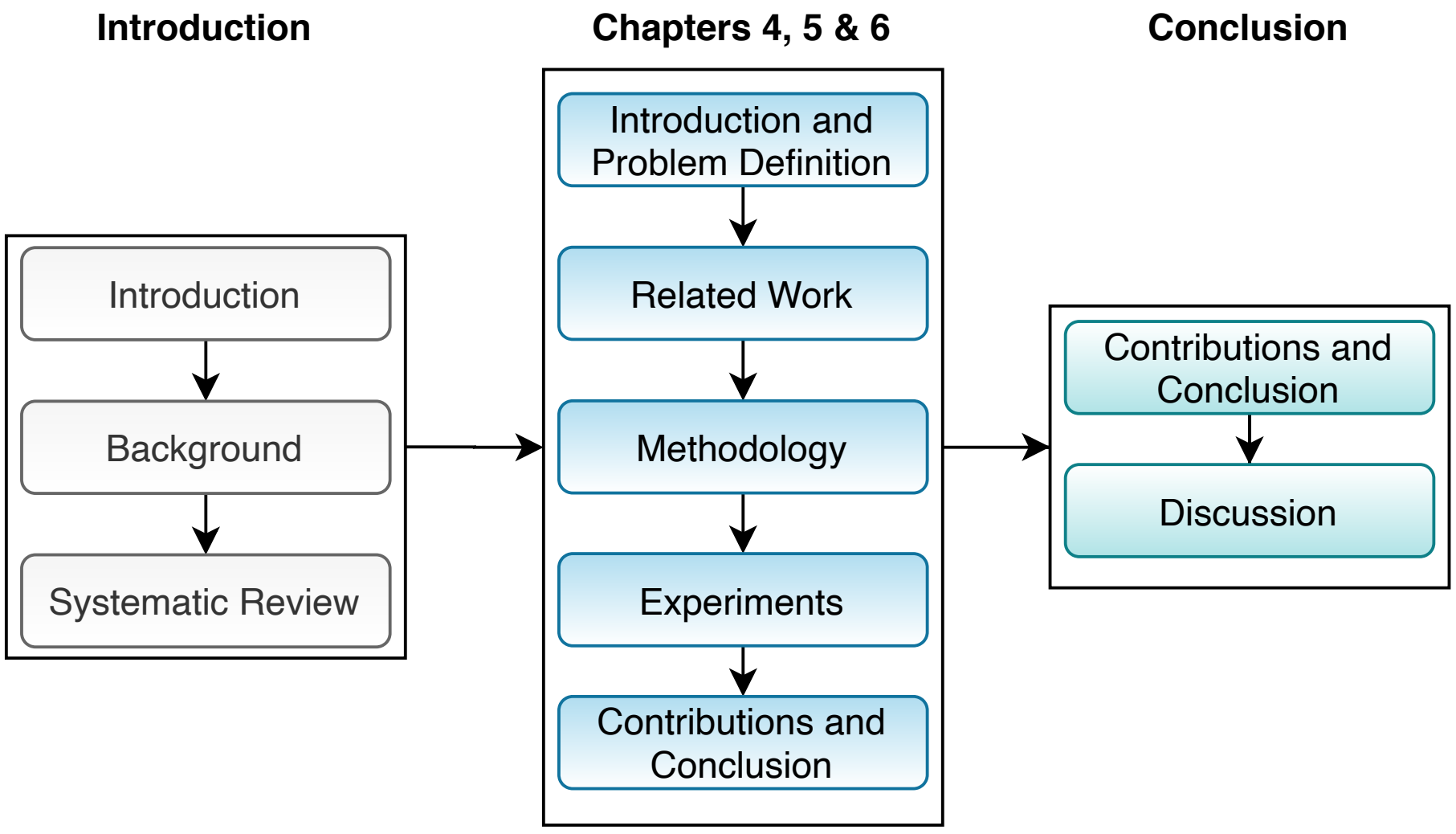

Figure 1.2: The structure of the thesis

The following summarizes the objectives of the remainder of this thesis:

**Chapter 2** broadly presents the background knowledge required for the topic of the thesis.

**Chapter 3** covers the previous studies in semantic similarity detection and induction with the current applications of lexical semantic resources in NLP.

**Chapter 4** entitled "leveraging the graph structure of lexicographical resources" provides solutions based on graph analysis to align word senses and dictionary entries.

**Chapter 5** presents a benchmark for monolingual word sense alignment.

**Chapter 6** describes the experiments results for monolingual word sense alignment.

## 1.5 PUBLICATIONS

This thesis is a manuscript-based document where every chapter is based on one or more peer-reviewed papers which were published during my Ph.D. Thanks to ELEXIS, I had the chance of collaborating closely with other partners, within ELEXIS or elsewhere, and address some of the research questions in collaboration with them. This being said, this thesis only provides the contributions and experiments that were carried out by myself. Where contributions of a collaborative nature are relevant to the current work, references will be explicitly provided.

The following list provides the publications, ordered based on importance, and the chapter where they appear:

**Chapter 4**

– **Sina Ahmadi**, Mihael Arcan, and John McCrae. Lexical sense alignment using weighted bipartite b-matching. In *Proceedings of the LDK 2019 Workshops*. 2nd Conference on Language, Data and Knowledge (LDK 2019), 2019a
– Mihael Arcan, Daniel Torregrosa, **Sina Ahmadi**, and John P McCrae. TIAD 2019 Shared Task: Leveraging knowledge graphs with neural machine translation for automatic multilingual dictionary generation. *Shared Task on Translation Inference Across Dictionaries*, 2019b
– **Sina Ahmadi**, Atul Kr. Ojha, Shubhanker Banerjee, and John P. McCrae. NUIG at TIAD 2021: Cross-lingual Word Embeddings for Translation Inference. *Shared Task on Translation Inference Across Dictionaries*, 2021b

**Chapter 5**

– **Sina Ahmadi**, John P. McCrae, Sanni Nimb, Fahad Khan, Monica Monachini, Bolette S. Pedersen, Thierry Declerck, Tanja Wissik, Andrea Bellandi, Irene Pisani, Thomas Troelsgård, Sussi Olsen, Simon Krek, Veronika Lipp, Tamás Váradi, László Simon, András Győrffy, Carole Tiberius, Tanneke Schoonheim, Yifat Ben Moshe, Maya Rudich, Raya Abu Ahmad, Dorielle Lonke, Kira Kovalenko, Margit Langemets, Jelena Kallas, Oksana Dereza, Theodorus Fransen, David Cillessen, David Lindemann, Mikel Alonso, Ana Salgado, José Luis Sancho, Rafael-J. Ureña-Ruiz, Kiril Simov, Petya Osenova, Zara Kancheva, Ivaylo Radev, Ranka Stanković, Andrej Perdih, and Dejan Gabrovšek. A Multilingual Evaluation Dataset for Monolingual Word Sense Alignment. In *Proceedings of the 12th Language Resource and Evaluation Conference (LREC 2020)*, Marseille, France, 2020a
– Ana Salgado, **Sina Ahmadi**, Alberto Simões, John P. McCrae, and Rute Costa. Challenges of word sense alignment: Portuguese language resources. In *the 7th Workshop on Linked Data in Linguistics: Building tools and infrastructure at the 12th International Conference on Language Resources and Evaluation (LREC)*, Marseille, France, 2020
– Patricia Martín-Chozas, **Sina Ahmadi**, and Elena Montiel-Ponsoda. Defying Wikidata: validation of terminological relations in the web of data. In *Proceedings of the 12th Language Resources and Evaluation Conference*, pages 5654–5659, 2020
– **Sina Ahmadi**, Mathieu Constant, Karën Fort, Bruno Guillaume, and John P. McCrae. *Convertir le Trésor de la Langue Française en Ontolex-Lemon : un zeste*

*de données liées. LIFT 2021 : Journées scientifiques de linguistique informatique, formelle & de terrain*, December 2021a
– **Sina Ahmadi**, Mihael Arcan, and John McCrae. On lexicographical networks. In *Workshop on eLexicography: Between Digital Humanities and Artificial Intelligence*, 2018

**Chapter 6**
– **Sina Ahmadi** and John P. McCrae. Monolingual word sense alignment as a classification problem. In *Proceedings of the 11th Global Wordnet Conference, GWC 2021, University of South Africa (UNISA), Potchefstroom, South Africa, January 18-21, 2021*, pages 73–80. Global Wordnet Association, 2021. URL https://aclanthology.org/2021.gwc-1.9/
– **Sina Ahmadi**, Sanni Nimb, John P. McCrae, and Nicolai H. Sørensen. Towards Automatic Linking of Lexicographic Data: the case of a historical and a modern Danish dictionary. In *The XIX EURALEX International Congress of the European Association for Lexicography*, Alexandroupolis, Greece, 2020b
– John P. McCrae, **Sina Ahmadi**, Seung-bin Yim, and Lenka Bajčetić. The ELEXIS system for monolingual sense linking in dictionaries. *Proceedings of Seventh Biennial Conference on Electronic Lexicography (eLex 2021)*, 2021
– Ilan Kernerman, Simon Krek, John P. McCrae, Jorge Gracia, **Sina Ahmadi**, and Besim Kabashi, editors. *Globalex Workshop on Linked Lexicography*, Language Resources and Evaluation LREC 2020, France, 2020b

Furthermore, the following articles were published during my Ph.D. on a topic related to the thesis, even though not included in the thesis as they don't address the research questions:

- John P. McCrae, Theodorus Fransen, **Sina Ahmadi**, Paul Buitelaar, and Koustava Goswami. Toward an integrative approach for making sense distinctions. *Frontiers in Artificial Intelligence*, 5, 2022. ISSN 2624-8212. doi: 10.3389/frai.2022.745626. URL https://www.frontiersin.org/article/10.3389/frai.2022.745626
- John P. McCrae, Carole Tiberius, Anas Fahad Khan, Ilan Kernerman, Thierry Declerck, Simon Krek, Monica Monachini, and **Sina Ahmadi**. The ELEXIS interface for interoperable lexical resources. In *Proceedings of the sixth biennial conference on electronic lexicography (eLex)*, pages 642–659, Sintra, Portugal, 10 2019b. URL https://elex.link/elex2019/wp-content/uploads/2019/09/eLex_2019_37.pdf
- Mihael Arcan, Daniel Torregrosa, **Sina Ahmadi**, and John P McCrae. Inferring translation candidates for multilingual dictionary generation with multi-way neural machine translation. In *Proceedings of the Translation Inference Across Dictionaries Workshop (TIAD 2019)*, 2019a
- **Sina Ahmadi**, Hossein Hassani, and John P. McCrae. Towards electronic lexicography for the Kurdish language. In *Proceedings of the sixth biennial conference on electronic lexicography (eLex)*, pages 881–906, Sintra, Portugal, 10 2019b

- Giulia Speranza, Carola Carlino, and **Sina Ahmadi**. Creating a multilingual terminological resource using linked data: the case of archaeological domain in the Italian language. In *Proceedings of the Sixth Italian Conference on Computational Linguistics (CLiC-it)*, Bari, Italy, 11 2019

# 2 BACKGROUND

## 2.1 INTRODUCTION

Language and linguistic resources are important components of many applications in NLP and computational linguistics. Over the past few decades, there have been considerable efforts in the development of language resources with a particular focus on increasing the coverage of the resources and the precision of the lexical and semantic information. The followings are some of the main language resources:

- Corpora, which are a digital collection of text, audio or multimedia
- Language descriptions such as grammar and formal modeling
- Lexical semantic resources which are curated following a lexical or conceptual paradigm such as dictionaries and network-based resources
- Ontologies, terminological resources and semantic lexica
- Language models which are trained on large corpora
- Knowledge graphs and knowledge bases

Considering language resources available in electronic form, there are a few distinctive properties that differentiate them. Such properties are mainly related to the content and the theoretical framework based on which the resource is conceptualized and created. In addition to conceptualization, resources vary in terms of the systematic coverage of lexical and semantic information. For instance, a dictionary does not systematically provide synonyms of all entry words, while WordNet (Miller, 1995) does. Another example is dictionary and encyclopedia; while a dictionary focuses on the linguistic properties of linguistic units represented by lemmas of any word class describing their use within a language, an encyclopedia provides the properties of objects of world knowledge designated by lemmas. Consequently, it is necessary to assess various properties of language resources, especially lexical semantic resources which are created according to an underlying theoretical framework.

Creating and maintaining language resources for a constantly changing phenomenon like a natural language requires much time and effort. Therefore, many of the expert-made resources have restrictions over open access and cannot be used within applications that voraciously require data to be trained and evaluated. As such, beyond the expert-made resources such as traditional dictionaries, thesauri and terminologies, more collaboratively-curated resources are being created and enriched nowadays thanks to the advances in information technology. Perhaps the most prominent examples of such collaboratively-curated resources are Wikipedia[1] as a free encyclopedia and Wiktionary[2] as a free dictionary. The extensive approach of describing entries in an encyclopedia makes it possible to translate them into other languages, while this may not be the case for all dictionary entries, and also act as a corpus for the language. As such, encyclopedias such as Wikipedia can be beneficial as a lexical semantic resource (LSR) (Zesch et al., 2007) but also as a corpus (Ghaddar and Langlais, 2018) and have widely contributed to the advances in language technology. This being said, encyclopedias encompass a vast range of encyclopaedic information beyond lexical resources (Zock and Biemann, 2020).

With the expansion of collaboratively-curated resources such as Wiktionary, processing lexicographical resources automatically and efficiently is of high importance recently in computational lexicography, computational linguistics and NLP. Nonetheless, the utility of language resources, in general, and the extent to which they contribute to specific tasks in languages is yet to be fully explored (Mirkin et al., 2009). This is partially due to the coverage of lexical and semantic information in a specific resource, but also the conceptualization that provides a more convenient structure for the computer to process linguistic data. For instance, the network-based structure of WordNet as well as its semantic richness has made it a popular resource in many language technology applications.

Our primary goal in this chapter is to provide a necessarily brief but essential background of language and linguistic resources. Our focus is particularly on those resources which have been used in NLP applications, such as WordNet, or are virtually addressing a specific challenge of interest in computational semantics, such as generative lexicons. This being said, there are many other language resources such as VerbNet (Schuler, 2005), PropBank (Kingsbury and Palmer, 2002) or Treebank (Marcus et al., 1993) that are not discussed in the chapter due to either the common features that are covered in other resources or their irrelevance to the context of this thesis.

To do so, we first introduce LSRs along with the crucial property of polysemy. Then, a few types of lexical semantic resources are described, namely dictionaries, network-based resources, Explanatory Combinatorial Dictionary (Mel’čuk, 2006), Generative Lexicon (Pustejovsky, 1995) and Natural Semantic Metalanguage (Wierzbicka, 1996). Moreover, a description of ontologies, terminologies, knowledge graphs and language models is provided. We believe that these language resources are of importance, given their current widespread usage and promising future in many NLP applications. It is important to note that polysemy and sense definitions in dictionaries are of particular interest to the context in which this thesis is defined. Therefore, particular attention is paid to those topics.

1 https://www.wikipedia.org
2 https://www.wiktionary.org

### 2.1.1 Lexical Semantic Resources

Lexical semantic resources (LSRs) are knowledge repositories that provide the vocabulary of a language in a descriptive, structured and conceptualized way. According to the literature, such resources are also often referred to as lexicosemantic resources. Lexical semantics focuses on the representation of meanings of lexical units and their variability based on the context. A lexical unit denotes a linguistic entity that can express a specific meaning either by lexemes, e.g. 'window', or non-compositional phrases as in idioms, e.g. "not someone's first rodeo". LSRs are differentiated based on the type of theory of lexicon, i.e. the theory that determines how to conceptualize and structure meanings of words. Therefore, their applications within language technology depend on the conceptual structure and the lexical encoding, i.e. lexicalization.

The conceptualization of many of the previously proposed LSRs has been motivated by various lexical semantic theories influenced by linguistic, psychological and cognitive concepts (Geeraerts, 2010). Such cognitively and psychologically motivated conceptualizations have created a wide range of resources beyond traditional dictionaries that have been deemed beneficial in many applications in language technology such as information retrieval (Shah and Croft, 2004), question answering (Pasca and Harabagiu, 2001), semantic role labeling (Zhou and Xu, 2015) and word-sense disambiguation (Pal and Saha, 2015).

The semantic analysis of LSRs can be carried out from various points of view, including cognitive, computational and philosophical. According to the structuralist tradition, two notions of paradigmatic and syntagmatic relations are taken into account. In this approach, the paradigmatic analysis focuses on the paradigmatic relation between words, such as synonymy, antonymy, meronymy, etc. On the other hand, syntagmatic relations consider syntactic collocations that are related to individual words as in "put someone on a pedestal".

Similarly, computational methods have been used for semantic analysis (Bird et al., 1995). The main goal in applying NLP methods to semantic analysis is to improve machine understanding and a wide range of related tasks, such as information retrieval, machine translation and human-machine interaction. To mention a few methods among the previously proposed ones, latent semantic analysis which uses statistical co-occurrence information between words and phrases (Kou and Peng, 2015), explicit semantic analysis which incorporates human knowledge, for instance using Wikipedia, to represent semantic associations (Milne, 2007; Yeh et al., 2009) and contextual and word embeddings by language modeling (Vulić et al., 2020b).

To further clarify the notion of polysemy, which plays a pivotal role in lexical semantics, we describe it in the following section.

### 2.1.2 Polysemy

Polysemy is the characteristic of different and multiple meanings of a word. Unlike monosemous words for which only a univocal meaning is available, polysemous lexical items emerge in a nuance of semantically related meanings which are identifiable to the speaker within context. According to Rey-Debove (2012, p. 256), polysemy is the "the main subject of semantics". Along with homonyms, polysemous items are considered as the major sources of lexical ambiguity. According to Lyons (1977), the lexical ambiguity continuum is delimited to homonymy and polysemy. Unlike homonyms which are often defined as separated lexical entries in a dictionary, polysemous entries are represented as a single entry. For example, in the following sentences based on the Collins Dictionary[3], different meanings of 'heart' are defined within one entry:

HEART (noun)
A The bullet had passed less than an inch from Andrea's *heart*.
B The only sound inside was the beating of his *heart*.
C Alik's words filled her *heart* with pride.
D The *heart* of the problem is supply and demand.

Even though not available to the same extent for all words, polysemy is a common phenomenon that has been traditionally defined depending on the relatedness of various meanings, register, domain and historical considerations regarding the etymology of word senses. The notion of relatedness in meaning has been the key to group word senses as polysemous. As such, homonyms are said to have a contrastive polysemy as they do not refer to the same concepts. Figure 2.1 shows the degree of polysemy in the French, German and Italian lexicographical data on Wiktionary[4]. Based on this analysis, multiword expressions (MWE) appear to be less polysemous on average in comparison to the set of other types of lexical entries including words and affixes. Similarly, among the grammatical categories of noun, adjective and verb, the latter has the highest number of senses available. It is important to note that even though Wiktionary is not a comprehensive dictionary and is being modified constantly by communities, it contains a considerable number of lexical entries making this analysis suggestive.

Defining various types of polysemy has remained a matter of theoretical discussions with not essentially all of them providing distinctive characteristics to determine polysemy (Geeraerts, 1993; Lyons and John, 1995). One of the prevailing theories is the classification of polysemy into *regular polysemy*, also known as *systematic polysemy*[5] (Dölling, 2020) and *irregular polysemy*. In regular polysemy (Apresjan, 1974),

3 https://www.collinsdictionary.com/dictionary/english/heart
4 https://www.wiktionary.org
5 Systematic polysemy is also defined on its own by Buitelaar (1998).

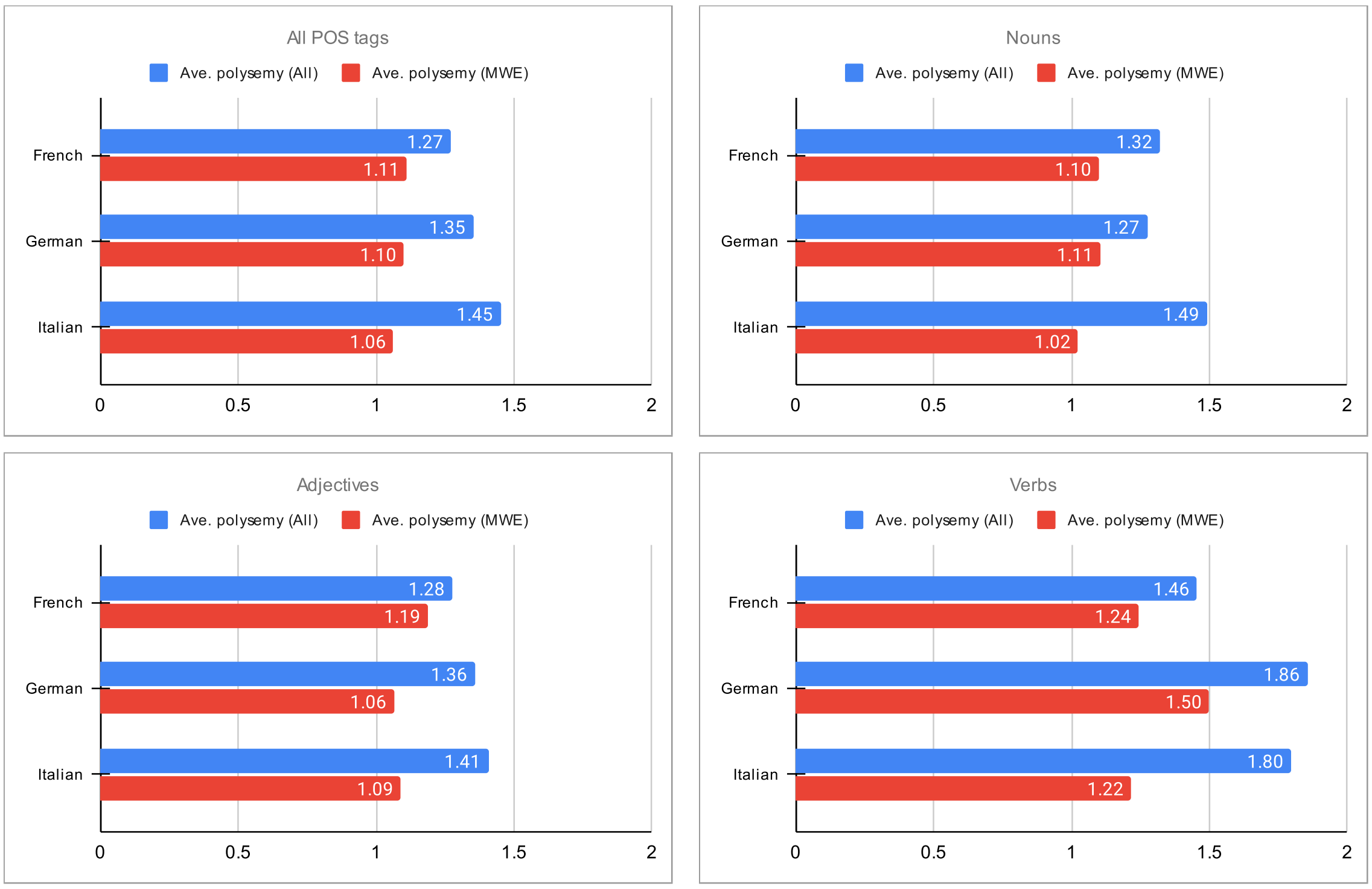

**Figure 2.1:** Degree of polysemy of French, German and Italian lexicographical data on Wiktionary (dump of early 2022 using Dbnary (http://kaiko.getalp.org))

polysemous items share a common range of lexical items that are semantically and systematically related. Regular polysemy is formally defined as follows:

> "Polysemy of a word A with the meaning $a_i$ and $a_j$ is called regular if, in the given language, there exists at least one other word B with the meaning $b_i$ and $b_j$ , which are semantically distinguished from each other in exactly the same way as $a_i$ and $a_j$ and if $a_i$ and $b_i$, $a_j$ and $b_j$ are non-synonymous." (Apresjan, 1974, p. 16)

In other words, regular polysemy may be modeled based on lexical rules and extended based on patterns of sense extension. In this sense, metonyms can be highly systematic, as in 'animal' for 'meat of animal', e.g. 'rabbit' and 'armadillo', 'fruit' for 'tree species', e.g. 'pear' and 'cherry', and 'flower' for 'plant', e.g. 'rose' and 'daffodil' (Cruse, 2010). In the example above, 'heart' in all the sentences represents regular polysemy, despite sentence C being a metaphor and sentence D being a metonym, as there is a word as 'B' in the definition like 'core' with two meanings of 'the tough central part of fruits' ($b_i$) and 'the part of something that is central to its existence' ($b_j$). Unlike simple metaphors as 'fill one's heart with something' to refer to 'making someone feel something', metaphors have generally the least systematic polysemy due to the unpredictable meaning of words, as in 'wear your *heart* on your sleeve' to say 'to openly display one's emotions or sentiments'. Although such irregular polysemy can affect all words, metaphors are widely known to be idiosyncratic and the most important type of irregular polysemy.

In the same vein, Cruse (2010, p. 110) describes two main relations between polysemes: *linear* and *non-linear*. In the first, there is a specialization-generalization relation where one of the polysemes refers to a more specialized meaning, i.e. hyponyms or meronyms, in comparison to the other one. For instance, the definitions of two senses of 'demand' (noun) in the Longman Dictionary of American English (Mayor, 2009) as '*the need or desire that people have for particular goods and services*' and '*a very firm request for something that you believe you have the right to get*' refers to a linear relation. This relation can be further divided into sub-categories according to semantic overlap, namely autohyponymy, automeronymy, autosuperordination and autoholonymy. On the other hand, non-linear relations denote polysemes that have a resemblance or relevance connection as in metaphors, e.g. 'to run *circles* around someone', and metonyms as in 'the *Pentagon*' to refer to the US military leadership. Therefore, metonymic polysemy seems to fit the best into the definition of regular polysemy while metaphorically-motivated polysemy appears closer to homonymy in the lexical ambiguity continuum (Klepousniotou and Baum, 2007, p. 11).

Polysemy is one of the most important notions based on which various LSRs are defined. In fact, lexical semantics has been driven by the compelling questions of how to define meaning to represent word senses in a systematic way, based on purely linguistic, cognitive, communicative notions or based on an interdisciplinary setup. This has motivated researchers to propose various theoretical frameworks where meanings may be considered as atomic concepts or of compositional nature, i.e. made of smaller parts. For instance, the enumerating approach of word senses in traditional dictionaries is called into question in the Generative Lexicon theory (Pustejovsky, 1995) by putting forward a compositional approach where senses can be generated. Furthermore, dealing with polysemy has been one of the main challenges in NLP, as demonstrated in word sense disambiguation (Stevenson and Wilks, 2003) and delineating lexical ambiguity (Aina et al., 2019). We will see in more detail in the following sections how different resources model polysemy and sense representation. Falkum and Vicente (2015); Vicente and Falkum (2017); Béjoint (1990) provide further discussions regarding the theoretical debates on polysemy and approaches to distinguish between monosemy and different types of polysemy.

## 2.2 DICTIONARIES

One of the best known LSRs are dictionaries which have been used as inventories for documenting human language since ancient times. There is a wide range of dictionaries that are published based on language, as in monolingual versus bilingual dictionaries, level of proficiency in the language, as in learner's dictionary, the field of expertise, such as the Oxford Dictionary of World Religions (Bowker, 1997), the etymology of words, such as the Etymological Dictionary of Greek (Beekes, 2009), and a specific period when the language was spoken, such as the *Ordbog over det danske Sprog*[6] ("Dictionary of the Danish Language") that documents the Danish language from 1700 until 1955. Although information technology has facilitated lexicography in a vast number of ways, the drudgery of collecting words and compiling words seems to have remained the same nonetheless.

According to Laporte (2013), there are two requirements for dictionaries to be used in NLP: readability of information and general architecture that facilities processing lexicographical content. We believe that these requirements can be generalized, to a lesser or greater extent, to all types of LSRs and are used as the leading factors in creating relevant standards to access resources and publish them.

### 2.2.1 Lexicon vs. Dictionary

Although the two terms of lexicon and dictionary are sometimes informally used interchangeably, the notion of *lexicon* refers to a different concept than that of a dictionary. According to (Ježek, 2016, p. 16),

> A dictionary is a concrete object, typically a book, in either printed or electronic format, whereas the lexicon is an abstract object, that is, a set of words with associated information, stored in our mind and described in the dictionary.

The difference between lexicon and dictionary is analogous to grammar and grammar books. While the book provides information based on a topic or a consultation method, such as providing entries in alphabetical order, the concept of the lexicon is understood as any type of word or morpheme based on lexical structures. Such lexical structures can be meaning associations such as *plow*, *sow* and *plant*, morphological properties such as *relate*, *relationship* and *unrelated*, syntactical behavior such as *nouns*, *verbs* and *prepositions* and semantically related words such as *ask*, *demand* and *inquire*. Therefore, a dictionary is a reference book that tries to document and collect the body of language, i.e. vocabulary, which is used in a language and may not always be complete and comprehensive with respect to a language lexicon.

Figure 2.2 provides an illustration created by Lehmann (2020) to highlight the differences between the aforementioned analogy. According to the same source, "the lexicon is that component of the significative system of a language which represents the relation between signifier and signified of signs insofar as it is not subject to rules". In addition, there are two polar ways of accessing language signs. Holistic access is made to idiosyncratic links between expression and content. Such idiosyncratic signs are inventoried in the lexicon. On the other hand, analytical access is made to regular links between expression and content and is formed in grammar. Furthermore, the horizontal dimension of the diagram indicates the more or less holistic and analytical

6 https://ordnet.dk/ods

access to linguistic signs in lexicon and grammar. Both lexicon and grammar are traversed by levels of complexity, which form the vertical axis of the diagram, in particular, the phrastic level and the word level.

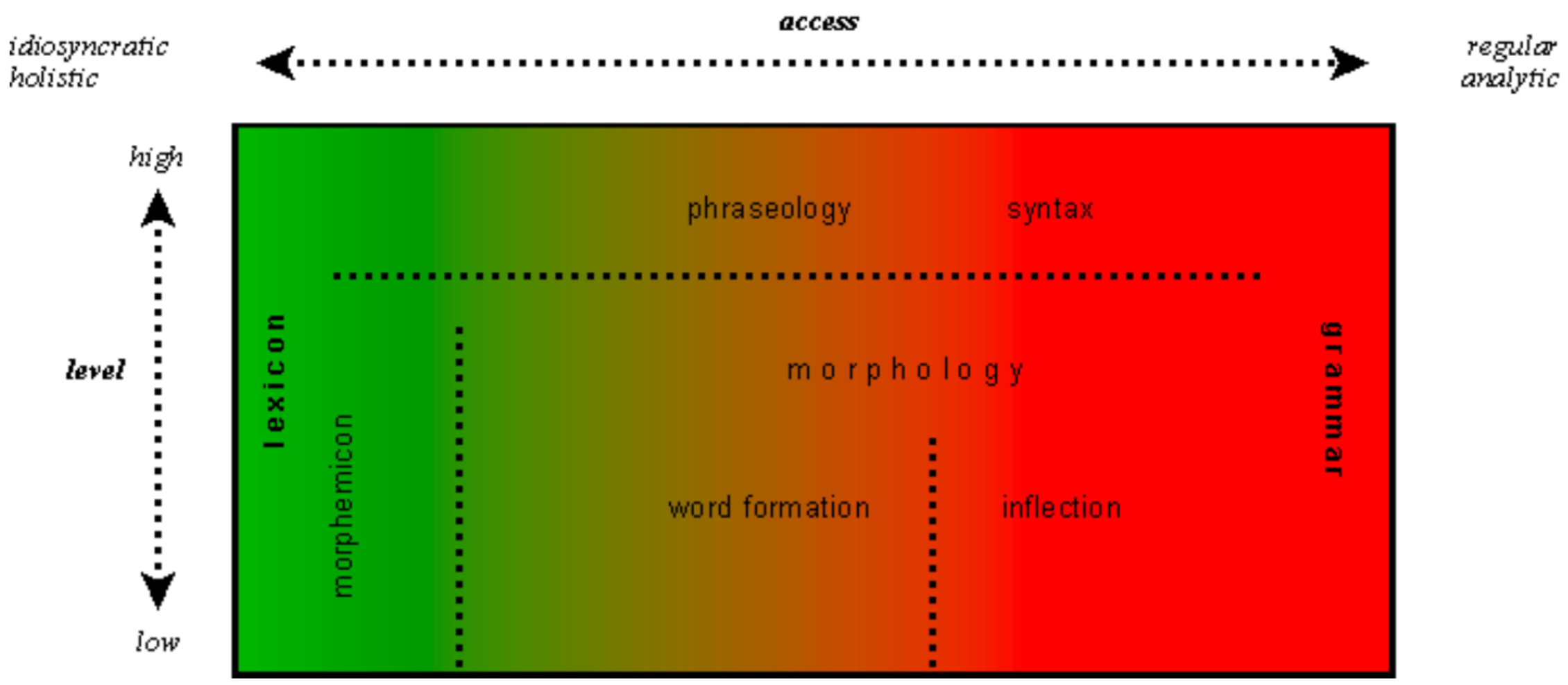

**Figure 2.2:** The relationship between lexicon and grammar according to Lehmann (2020)

This conceptual distinction between lexicon and grammar versus dictionary and grammar book is of importance in the discussions in the upcoming sections.

### 2.2.2 Content

The content of a dictionary is usually thought of from two structural points of view: *macro-structure* and *micro-structure* (Atkins and Rundell, 2008, p. 160). The first notion focuses on the organization of lexical items and the type of lemma headwords and vocabulary to be included in the body of the dictionary. Headwords can be simple words e.g. 'sacrilegious', abbreviations e.g. 'PDF', bound affixes e.g. *-ment* and multi-word expressions e.g. 'pang of conscience'. Moreover, the vocabulary type may vary based on domain, style, register, dialect, time and, etc. On the other hand, micro-structure refers to the internal structure and content of each lexical entry. Oftentimes, lexicographers take morphological forms based on rules of word formation, semantics and syntax into account in designing both these structures. Figure 2.3 illustrates the micro-structure of the entry 'beauty' in the Longman Dictionary of American English (Mayor, 2009) containing pronunciation, grammatical information, etymology, phraseology, definitions, and example labels such as 'old-fashioned', 'colloquial' and 'informal'. It should be noted that the richness of such information and the components of the micro-structure depend on the entry, the type of dictionary (monolingual vs. bilingual) and the lexicographer's approach.

Although not equally provided for all entries, semantic relationships, including lexical relations such as synonymy and antonymy and conceptual relations such as hyponymy and meronymy, are occasionally provided in a dictionary as well. Such information is not systematic and may not come within a specific data structure. For instance, synonyms or near-synonyms such as 'mistake' and 'slip' may be provided as cross-references, i.e. *cf*, or explicitly as synonyms. That said, there are other types of relationships that are usually not included in lexicographical resources; e.g. temporal relationships such as happens-before as in `enroll-graduate`, `marry-divorce` and `detain-prosecute` defined in VerbOcean (Chklovski and Pantel, 2004), associative relationships as in `cat-purr`, typical instrumentality as in `nail-hammer`, locative relationships as in `priest-church` and causal relationships as in `sand-snooze` (Hirst, 2009, p. 7).

Akin to a dictionary in structure but chiefly focused on semantic relations, thesauri are a specific type of dictionaries that provide words based on their semantic relationship. The initial attempt to create such a semantic resource for English was Roget's Thesaurus of English Words and Phrases (Roget, 2014). While Roget's thesaurus provides general-usage words, many other thesauri with controlled vocabularies adopted for classification of technical documents extend the relationships between terms to equivalent, broader, narrower and related, as well.

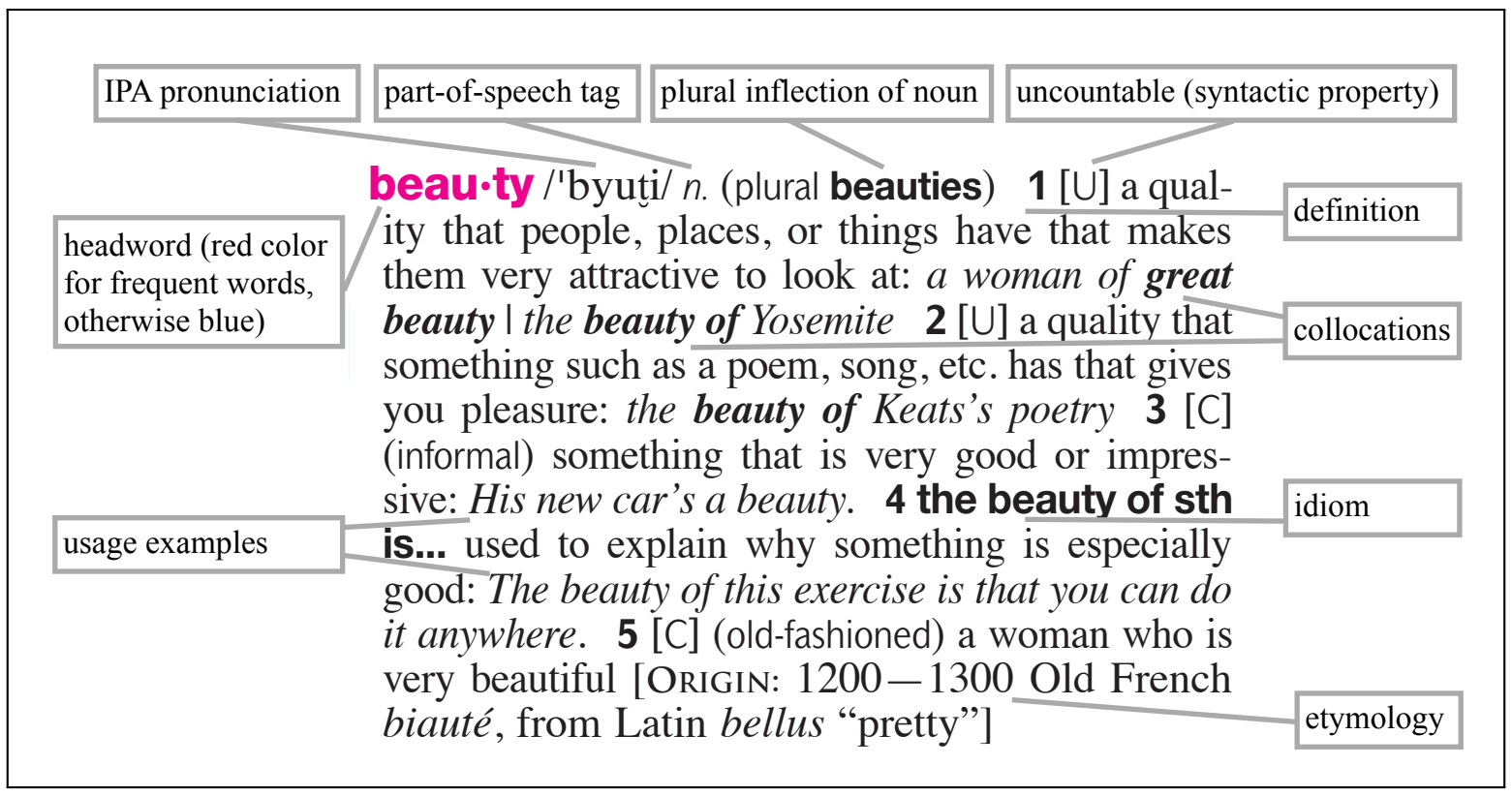

**Figure 2.3:** The micro-structure of the entry 'beauty' in the Longman Dictionary of American English

### *Headwords*

As a general rule, the lemma form of word forms and lexemes are used as headwords. Depending on the morphological complexity of a language, the identification of headwords can be a challenging task. For instance, retrieving the lemma of inflected word forms which are textually close to their lemma form is easier as in 'drinking' vs. 'drink' than in some word forms that are inflected based on allomorphs, as in the Kurdish word *bêje* (say.IMP.2SG) vs. *gotin* (say.INF). Furthermore, the dominance of the graphemic representation to access a word, described as "orthographic supremacy" by (Lew, 2012, p. 346), further poses challenges in providing headwords in a dictionary. Akin to this, spelling variation is another challenge, particularly for dictionaries compiled in different time periods, as in the modern and old Danish dictionaries, *Den Danske Ordbog*[7] and *Ordbog over det danske Sprog*. Moreover, there are different views and problems related to multi-word expressions and homographs as dictionary entries (Gantar et al., 2019). The latter is specified in both printed and electronic dictionaries, for example by using a superscript number as in LEAD$^1$.

### *Senses*

Word senses, or simply senses, form the basic unit of the organization of the microstructure of a dictionary. Sense indicators are provided to distinguish between various senses of a headword. In some dictionaries, such as the Longman Dictionary of American English (Mayor, 2009) or the *Trésor de la Langue Française*[8], senses are ordered according to their frequency and time period (old to modern senses) in the language.

Despite efforts to extract senses automatically from large corpora (McCarthy et al., 2004), there is no decisive way to determine sense distinctions from a theoretical point of view to such an extent that many do not believe in the nature of senses even, as Kilgarriff (1997) states in his article entitled 'I don't believe in word senses'. A very well-known example is the word 'bank' (noun) which has different meanings as in 'river bank' and 'financial institution'. This is referred to as the Bank Model by Kilgarriff (1992b) to highlight the complexities in polysemy and determining various senses of a word. Nonetheless, senses are a part of dictionaries and are listed using an enumerative approach (Lew, 2013). Traditionally, senses derived from the same etymological source are considered polysemous in dictionaries; otherwise, separate entries are defined.

Furthermore, senses in some comprehensive dictionaries are typically organized in a hierarchy where semantically related concepts are provided as subsenses to a main sense, as illustrated in Figure 2.4 in the monolingual modern Danish dictionary, *Den Danske Ordbog*. However, the sense granularity and the exact distinctions drawn between both main senses and subsenses of a lemma might differ quite a lot across monolingual dictionaries. Closely related concepts, e.g. the many cases of regular polysemy in the language as discussed by Buitelaar (2000); Pustejovsky (2017) and McCrae et al. (2022), might be expressed as separate subsenses, but might also be indirectly included in the main senses. This varies not only across dictionaries but also within the same dictionary. Furthermore, the sense granularity of a dictionary is influenced by specific editorial guidelines, such as the cases where formatting takes precedence over comprehensiveness considerations. That being said, determining which senses are to be included follows the subjective and individual judgments made by each lexicographer (Kilgarriff, 1997).

In a recent study, McCrae et al. (2022) shed light on sense distinction from a computational point of view where an approach for making such distinctions is proposed.

7 https://ordnet.dk/ddo
8 https://www.atilf.fr/ressources/tlfi

Based on this approach, sense distinction can be facilitated by integrating various approaches such as formal, cognitive, intercultural and distributional, among which the last one is widely used in practice nowadays. Thanks to such an approach, lexicographers can carry out quantitative and qualitative analyses of sense distinctions. Moreover, there are many techniques to distinguish and identify different senses of a word using topic modeling (Lau et al., 2014), word sense induction (Cook et al., 2013) and contextualized word embeddings (Zhou and Bollegala, 2021; Gessler and Schneider, 2021), to mention but a few.

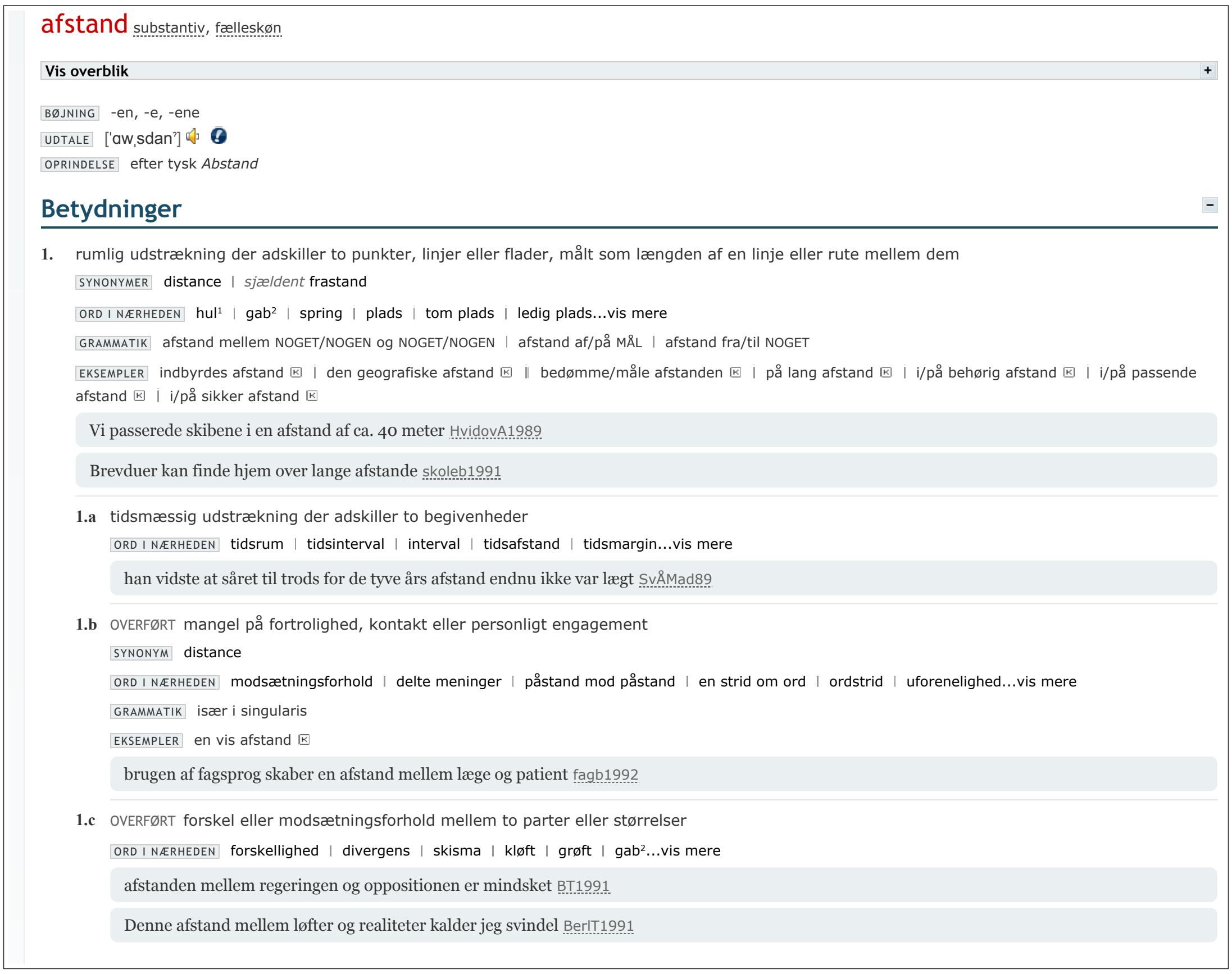
afstand substantiv, fælleskøn

Vis overblik +

BØJNING -en, -e, -ene
UDTALE [ˈɑwˌsdanˀ]
OPRINDELSE efter tysk *Abstand*

Betydninger

1. rumlig udstrækning der adskiller to punkter, linjer eller flader, målt som længden af en linje eller rute mellem dem
SYNONYMER distance | *sjældent* frastand
ORD I NÆRHEDEN hul¹ | gab² | spring | plads | tom plads | ledig plads...vis mere
GRAMMATIK afstand mellem NOGET/NOGEN og NOGET/NOGEN | afstand af/på MÅL | afstand fra/til NOGET
EKSEMPLER indbyrdes afstand | den geografiske afstand | bedømme/måle afstanden | på lang afstand | i/på behørig afstand | i/på passende afstand | i/på sikker afstand
Vi passerede skibene i en afstand af ca. 40 meter HvidovA1989
Brevduer kan finde hjem over lange afstande skoleb1991

1.a tidsmæssig udstrækning der adskiller to begivenheder
ORD I NÆRHEDEN tidsrum | tidsinterval | interval | tidsafstand | tidsmargin...vis mere
han vidste at såret til trods for de tyve års afstand endnu ikke var lægt SvÅMad89

1.b OVERFØRT mangel på fortrolighed, kontakt eller personligt engagement
SYNONYM distance
ORD I NÆRHEDEN modsætningsforhold | delte meninger | påstand mod påstand | en strid om ord | ordstrid | uforenelighed...vis mere
GRAMMATIK især i singularis
EKSEMPLER en vis afstand
brugen af fagsprog skaber en afstand mellem læge og patient fagb1992

1.c OVERFØRT forskel eller modsætningsforhold mellem to parter eller størrelser
ORD I NÆRHEDEN forskellighed | divergens | skisma | kløft | grøft | gab²...vis mere
afstanden mellem regeringen og oppositionen er mindsket BT1991
Denne afstand mellem løfter og realiteter kalder jeg svindel BerlT1991

**Figure 2.4:** The noun '*afstand*' (distance) in the Danish monolingual dictionaries, *Den Danske Ordbog* (https://ordnet.dk/ddo/ordbog?query=afstand)

### *Definitions*

In addition to senses, sense definitions are principal components of monolingual dictionaries describing various meanings of words in plain text. Since antiquity, there have been many theories and discussions on how to define a concept, i.e. *definiendum*, and the words and phrases which are used for this purpose, i.e. *definiens*. Hanks (2016) provides a description of such theories from historical, logical, and lexicographical points of view starting from Aristotle up to the latest theories, namely those revolving around the generative lexicon. Nevertheless, definitions are one of the major components of monolingual dictionaries.

| Type | Paradigm | Definition | Example |
|---|---|---|---|
| Purpose-based | Lexical | "that sort of word-thing definition in which we are explaining the actual way in which some actual word has been used by some actual persons" (Robinson, 2003, p. 35) | **atmosphere** (noun): the gaseous envelope of a celestial body (Merriam-Webster) |
| | Stipulative | attributes a meaning to a word depending on the context and what one intends a word to mean, sometimes without taking common usages into account (Bogacki et al., 2015) | **atmosphere (noun):** *J'ai besoin de changer d'atmosphère, et mon atmosphère, c'est toi.*[9] (I need to change atmosphere, and my atmosphere is you.) |
| Method-based | Ostensive | exploiting learner's knowledge to define words through examples, drawings and paintings (Kotarbinska, 1960) | pointing out objects by color, as in '**white**' (adjective) for yogurt |
| | Synonymous | define words by synonyms | **mean** (adjective): penurious, stingy (Merriam-Webster) |
| | Analytical | a formal descriptive sentence consisting of four main components, namely species, verb, genus, and *differentia* | **rodent** (noun): any of an order of relatively small gnawing mammals that have in both jaws a single pair of incisors with a chisel-shaped edge (Merriam-Webster) |
| | Relational | explain the meaning of a word in comparison to other entities | **extraneous** (adjective): not belonging to a thing (Webster 1913) |
| | Exemplifying | use examples to illustrate words | **living** (adjective): people, animals, and plants (Mayor, 2009) |
| | Contextual | defining a word by describing its properties within a context | "therefore, only those organisms that can grow without oxygen — anaerobic organisms — were able to live." (Spala et al., 2019) |
| | Reference | referring to an external resource by quoting or relying on a personal or historical belief | **virtue** (noun): According to Socrates, virtue is knowledge |
| | Rule-giving | providing rules to define words, e.g. grammatical function words | **whom** (pronoun): used as an interrogative or relative, as object of a verb or a preceding preposition |
| Corpus-based | Is-definitions | verb 'to be' is used as a connector | **variable** (noun): "a **variable** is any part of the experiment that can vary or change during the experiment." (Spala et al., 2019) |
| | Verb-definitions | a verbal connector is used to define a word | **pitch** (noun): "the perception of frequency is called pitch." (Spala et al., 2019) |
| Definition modeling | Word embeddings | generating a definition using static word embeddings | **reprint** (noun): "a written or printed version of a book or other publication" (Gadetsky et al., 2018) |
| | Contextual embeddings | generating a definition using contextual embeddings | **scoot** (verb): "cause to move along by pushing" (Bevilacqua et al., 2020) |

Table 2.1: A classification of some of the best known definition paradigms

9 A famous replica in the film *Hôtel du Nord* (1938)

Table 2.1 summarizes some of the major paradigms in defining the definition of a concept inspired by the classification of Westerhout (2010). Depending on the semantic complexity and the editorial choice, one or more of these definition paradigms are used by lexicographers in the micro-structure of a dictionary. For instance, while the exemplifying paradigm has limited usage for words that may be described using more representative examples, the analytical paradigm allows a more descriptive and therefore detailed description of the entry without sacrificing clarity to brevity. It is worth mentioning that definitions can not always cover all the senses of a word. In order to fill semantic gaps, oftentimes dictionaries provide usage examples and glosses. The latter are clearer descriptions of the definition and more elaborate explanations in bilingual dictionaries.

Furthermore, extracting and generating definitions of a given word using computational techniques has been previously studied (Westerhout, 2010; Silva et al., 2016). Recently, many methods have been proposed to generate definitions, a task called *definition modeling*. The main idea in this approach is to explore distributed statistical representations of words to generate definitions for words in context or as lemmas. Among the explored methods, we can mention recurrent neural networks with word embeddings (Noraset et al., 2017), latent variable modeling and soft attention mechanism with word embeddings (Gadetsky et al., 2018), and ultimately Generationary (Bevilacqua et al., 2020) which uses a span-based encoding scheme to fine-tune an English pre-trained Encoder-Decoder system to generate glosses using contextual embeddings.

### 2.2.3 Electronic Dictionaries

With the advent of information technology, electronic dictionaries have been extensively used where lexicographical data are available in an electronic form. In addition to electronic dictionaries, machine-readable dictionaries (MRDs) are created to specifically represent data under a machine-readable format, such as XML. This being said, the terms machine-readable dictionary, electronic dictionary and lexical database are often conflated and used interchangeably.

In addition to the electronic form and data format, researchers have been motivated to design LSRs in such a way that processing linguistic data would be facilitated by a computer. In this context, an electronic dictionary has been more specifically called a *virtual dictionary* (Selva et al., 2003; Polguère, 2012; Polguère, 2014) to be differentiated from an electronic dictionary and an MRD, not only based on the format but also the representation of data structures, such as graphs, that are more compatible with computers. In the same vein, one of the main objectives of the ELEXIS project is to provide the infrastructure to convert printed dictionaries into electronic form. For this, a service named ELEXIFIER[10] has been created.

10 https://elexifier.elex.is

## 2.3 NETWORK-BASED RESOURCES

Moving away from the theoretical discussions regarding modeling and defining various aspects of lexical and semantic data in LSRs, a crucial issue is the representation of such data. A popular solution is to organize lexical and semantic data as networks. This idea of representing dictionaries as lexical networks has received much attention previously, mainly to facilitate the representation of lexicographical data and also, leverage the structure of networks to create further associations and connections between words (Reuer, 2004; Gilquin, 2008).

The main characteristic of such resources is to provide lexical units in such a way that a relationship can be assigned to the relation between lexical units. A lexical network or lexical graph, as the name suggests, would refer to the network of word senses connected by semantic links such as synonymy, antonymy, hyponymy and also metonymic, metaphorical, and polysemy relations (Norvig, 1989). Therefore, many LSRs fall into the category of lexical networks to some extent, for instance, WordNet as a lexical network (Beckwith and Miller, 1990). Such networks are sometimes referred to as *lexical systems*, that is to say, resources modeled in the form of a graph (Polguere, 2009).

### 2.3.1 WordNet

WordNet is a widely known and used lexical database and reference that groups words based on linguistic and psychological accounts of the theory of lexical memory (Miller et al., 1988). The fundamental idea of WordNet is to represent a concept or word sense by its synonyms, also known as *synset*s. Beyond the vocabulary of a given language, WordNet provides different types of lexical relations to enrich synsets, such as hyponymy and meronymy. The idea of WordNet was initially implemented for the English language which is referred to as the Princeton WordNet (Miller, 1998).

Since its conception, the Princeton WordNet has gone through a few improvements and enrichment in information. For instance, some of the succeeding WordNets are enriched with information such as pronunciations (Declerck et al., 2020), strength of relationships (Boyd-Graber et al., 2006), marking semantic roles and causality of verbs (Dziob et al., 2017), sense definitions (Mihalcea and Moldovan, 2001), logical form axioms (Moldovan and Rus, 2001) and increasing multilinguality. In order to facilitate the latter, many studies have focused on creating a centralized or shared database to connect synsets, such as inter-lingual-index (Ellman, 2003) and the Collaborative Interlingual Index (Bond et al., 2016).

In addition, new WordNets are created for other languages by following the same theory of the original Princeton WordNet, as in Rudnicka et al. (2015); Sagot and Fišer (2012); Pedersen et al. (2009); Isahara et al. (2008), or by translating the Princeton WordNet, also known as the Expand Method (Vossen, 2002, p. 52), as for Sanskrit (Kulkarni et al., 2010), Latin (Franzini et al., 2019), Kurdish (Aliabadi et al., 2014) and Mongolian (Batsuren et al., 2019), or by merging existing ones based on domains, also known as the Merge Method, as in Hanoka and Sagot (2012); Zafar (2012).

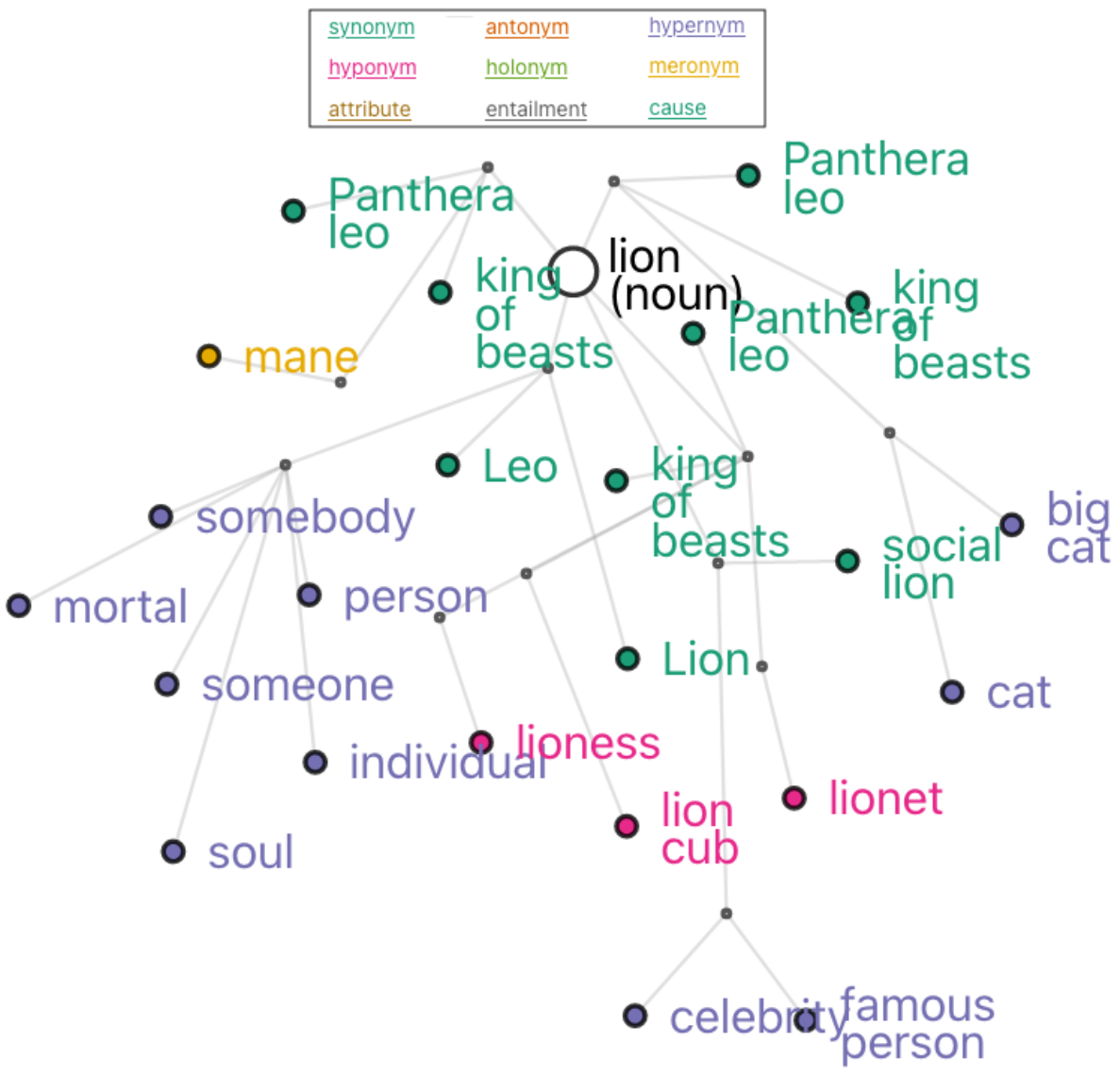

**Figure 2.5:** The lexical network of 'lion' (noun) according to the Princeton WordNet 3.1 where its synsets (in green) and other semantic relations are specified (Visualized using Lexical Graph: https://github.com/aliiae/lexical-graph).

Furthermore, other aspects are included in the WordNet such as the lexical representation of affective knowledge as in WordNet-Affect (Strapparava et al., 2004), integrating subject fields as in WordNet Domains (Magnini and Cavaglia, 2000) and extending the Princeton WordNet with named entities (Toral and Monachini, 2008). WordNet has been one of the most important and ubiquitous resources in many NLP applications such as word sense disambiguation (Morato et al., 2004), information retrieval (Mandala et al., 1998) and semantic similarity detection (Varelas et al., 2005). Many surveys have addressed various aspects of WordNets as in Bond and Paik (2012); Lin and Sandkuhl (2008); Petrolito and Bond (2014).

The current version of the Princeton WordNet (version 3.0) contains 206,941 word-sense pairs and 117,659 synsets. Among the senses, 33.8% are polysemous (79,450) with an average polysemy of 1.24 for nouns and 2.17 for verbs[11]. To further update the Princeton WordNet under an open source paradigm, McCrae et al. (2020) created the Open English WordNet recently. Despite the richness of WordNet in terms of semantic relations, there have been challenges regarding its applications within NLP applications where detection of polysemy is required. There have been many studies that focus on the description and detection of various types of polysemy, particularly regular polysemy (Barque and Chaumartin, 2009), systematic polysemy (Buitelaar, 1998; Peters and Peters, 2000) and also, clustering of word senses based on metonymy, and diathesis alternation (Peters et al., 1998).

11 According to https://wordnet.princeton.edu/documentation/wnstats7wn

### 2.3.2 FrameNet

FrameNet (Baker et al., 1998; Fillmore et al., 2004; Fillmore, 2008) is a resource containing words and their meanings, known as lexical units, provided according to the frame semantics which lays out "schematic representations of the conceptual structures and patterns of beliefs, practices, institutions, images, etc. that provide a foundation for meaningful interaction in a given speech community." (Fillmore et al., 2003, p. 235). The frame is a conceptualization of knowledge in such a way that each frame can be evoked by specific words, i.e. lemmata, based on their meaning. Figure 2.6 illustrates the frame of 'cogitation' (noun) and the relations of this frame with other frames, i.e. frame-to-frame relations, such as 'assessing' and 'mental activity'. In addition to such semantic relations, FrameNet provides an analysis of words based on their syntactic properties in actual utterances taken from large corpora. For instance, the frame 'cogitation' is evoked by 'consideration', 'thought' and 'thoughts' in the following phrases[12]:

(a) your application was submitted for *consideration* by the committee.

(b) the teacher gave some *thought* to a career change.

(c) your *thoughts* about art are not relevant.

Therefore, one can argue that 'thoughts' and 'consideration' in the aforementioned examples are related as they both evoke the same frame of 'cogitation'. Unlike WordNet which provides polysemous items based on their syntactic properties, FrameNet represents polysemy by associating different frames to a word. Additionally, FrameNet provides other properties in both semantic and syntactic levels such as valences, arguments and complements, which specifies the structure based on which the word, particularly verbs, can be used.

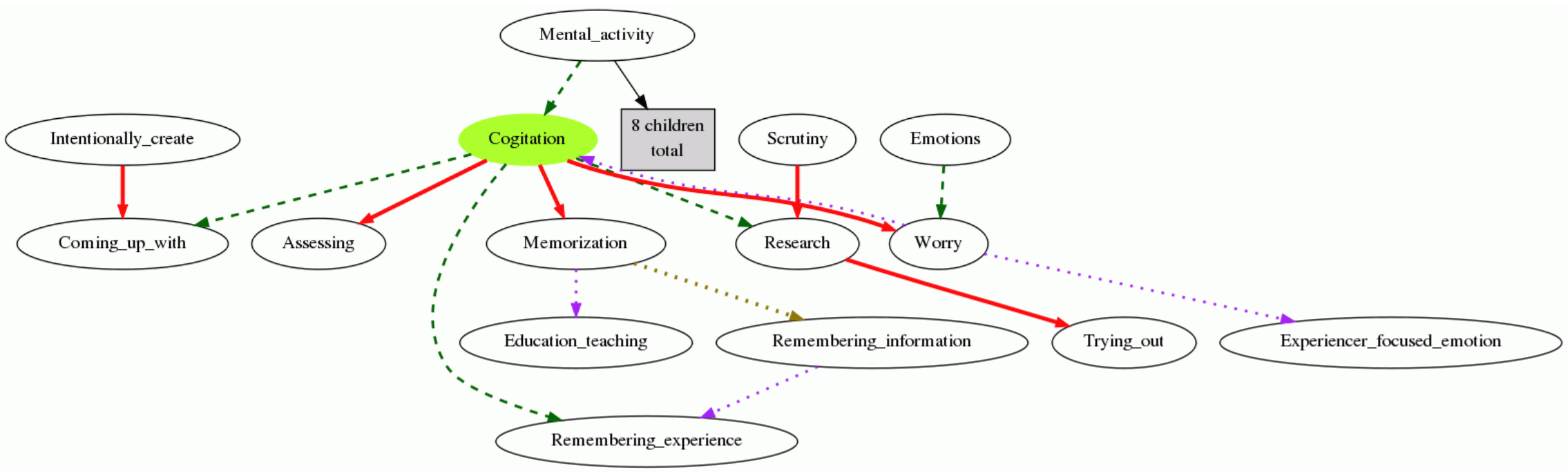

**Figure 2.6:** The frame of 'cogitation' (noun) in green along with its relations with other frames, such as 'memorization' according to the Berkeley FrameNet (Visualized using FrameGrapher: https://framenet.icsi.berkeley.edu/fndrupal/FrameGrapher).

12 From https://framenet2.icsi.berkeley.edu/fnReports/data/frameIndex.xml?frame=Cogitation

Unlike WordNet which does not rely on examples extracted from corpus, FrameNet provides annotated phrases based on corpora for each frame pattern (Boas, 2005). This enables FrameNet to represent richer "syntagmatic information about the combinatorial possibilities of each lexical unit" (Baker and Fellbaum, 2009), despite the considerably smaller size of FrameNet which contains approximately 1,000 manually defined frames.

### 2.3.3 JeuxDeMots

JeuxDeMots[13] (Lafourcade, 2007) is a gamified platform to create a lexical network by collecting terms from players containing over 404 million relations, such as *associated* for association, *pos* for part-of-speech, *isa* for predicate, *hypo* for hyponyms and *has_part* for meronyms, between over five million terms in French. The relationships within the network are labeled semantically, oriented and weighted, with possibly negative relationships denoting false associations. New relations between terms can be extracted by automatic induction and it has been enriched and aligned with external resources (Tchechmedjiev et al., 2017; Plu et al., 2018).

## 2.4 EXPLANATORY COMBINATORIAL DICTIONARY

The Explanatory Combinatorial Dictionary (ECD) (Mel'čuk, 2006) is a type of dictionary that is compiled based on the Meaning-Text Theory. The Meaning-Text Theory (MTT) (Mel'čuk, 1973, 1981) aims to formalize the correspondence between meaning and text by using lexical functions and categorizing lexical items according to lexical, syntactic and semantic features. As such, lexical functions can be related to lexical units, i.e. a lexeme, a morpheme or multi-word expression, to indicate the types of phrases that a given word can form with other words. The following table, based on Lareau et al. (2012), provides the lexical functions associated with 'attention':

| ATTENTION [*of* X *to* Y] | |
|---|---|
| `Magn` | close/whole/complete/undivided ~ |
| $\texttt{Func}_2$ | X's ~ is on Y |
| $\texttt{nonFunc}_0$ | X's ~ wanders |
| $\texttt{Oper}_{12}$ | X gives his/pays ~ to Y |
| $\texttt{Oper}_2$ | Y attracts/receives/enjoys X's ~ |
| $\texttt{Oper}_2\texttt{+Magn}_{quant-X}$ | Y is the center of ~ (of many Xs) |
| $\texttt{IncepOper}_{12}$ | X turns his ~ to Y |
| $\texttt{IncepOper}_2$ | Y gets X's ~ |
| $\texttt{ContOper}_2$ | Y holds/keeps X's ~ |
| $\texttt{CausFunc}_2$ | Z draws/calls/brings X's ~ to Y |
| $\texttt{LiquFunc}_2$ | Z diverts/distracts/draws X's ~ from Y |

13 JeuxDeMots ('*jeux de mots*' lit. games of words): http://www.jeuxdemots.org

Lexical functions of MTT aim to represent paradigmatic relations such as synonymy as in `Syn(wealthy)=rich`, antonymy as in `Anti(dark)=bright` and meronymy as in `Mult(bee)=swarm`, and syntagmatic links such as `Magn` (Lat. *magnus*, great, big) as intensifier in `Magn(rain)=heavily` and `IncepOper` (Lat. *incipere*, to begin and *operari*, operate) showing the act of starting to do, make, or have as in `IncepOper1(victory) = achieve, gain, score, win`. There are originally 27 paradigmatic and 37 syntagmatic lexical functions (Mel'čuk, 1996). Therefore, an ECD does not focus on a limited number of semantic properties but a wider range of lexical data, such as synsets in WordNet or frames in FrameNet (Lux-Pogodalla and Polguère, 2011). The various paradigmatic and syntagmatic links are shown in the lexical unit of 'attention'. In addition, not only the definition of attention as a lexeme is provided, but also collocations such as 'complete attention' or 'whole attention', and idioms such as 'pay attention' and 'enjoy attention' are definable. This demonstrates the flexibility and strength of the theoretical stance of MTT.

| | |
|---|---|
| IMPROVE, verb | |
| IMPROVE-I.1A | *X improves* ≡ 'The value of the quality of X becomes higher'<br>[*The weather suddenly improved; The system will improve over time*] |
| IMPROVE-I.1B | *X improves Y* ≡ 'X causes-1 that Y improves-I.1A'<br>[*The most recent changes drastically improved the system*] |
| IMPROVE-I.2 | *X improves* ≡ 'The health of a sick person X improves-I.1A'<br>[*Jim is steadily improving*] |
| IMPROVE-I.3 | *X improves at Y* ≡ 'X's execution of Y improves-I.1A, which is caused1 by X's having practiced or practicing Y'<br>[*Jim is steadily improving at algebra*] |
| IMPROVE-II | *X improves Y by Z-ing* ≡ 'X voluntarily causes-2 that the market value of a piece of real estate Y becomes higher by doing Z-ing to Y'<br>[*Jim improved his house by installing indoor plumbing*] |
| IMPROVE-III | *X improves upon Y* ≡ 'X creates a new Y' by improving-I.1B Y'<br>[*Jim has drastically improved upon Patrick's translation*] |

Table 2.2: Lexical units defined in the vocable of 'improve' based on ECD (Mel'čuk, 2006, p. 19)

The descriptive approach of Explanatory and Combinatorial Lexicography has chiefly led to two types of lexical resources: dictionaries such as the *Dictionnaire explicatif et combinatoire du français contemporain* (Combinatorial and Explanatory Dictionary of Contemporary French) (Mel'čuk and Arbatchewsky-Jumarie, 1999) and lexical databases such as the DiCo (Polguere, 2000). Unlike dictionaries where lexical entries are lemmas, an ECD is organized based on meaning by presenting lexical units as lexical entries. Therefore, lexical entries in an ECD do not represent the same level of polysemy as in a traditional dictionary. In fact, it is possible to group multiple lexical entries together in an ECD as a vocable as long as they show an intersection in meaning, or are roughly synonyms. The relationship between such lexical units in the same vocable is referred to as *co-polysemy* by Polguère (2018). In this sense, a vocable corresponds more or less to the micro-structure of a dictionary.

The example in Table 2.2 further clarifies this by showing six lexical units with unique identifiers, e.g. I.1A, which are grouped in the vocable of 'improve' based on Mel'čuk (2006, p. 19). In this vocable, the semantic closeness of the lexical units is categorized based on Roman and Arabic numerals and letters, in order of closeness. So, IMPROVE-I.1A and IMPROVE-I.1B, for example, are semantically more related than to IMPROVE-II. This enumeration is analogous to sense numbers in a traditional dictionary. In addition, examples are provided in brackets.

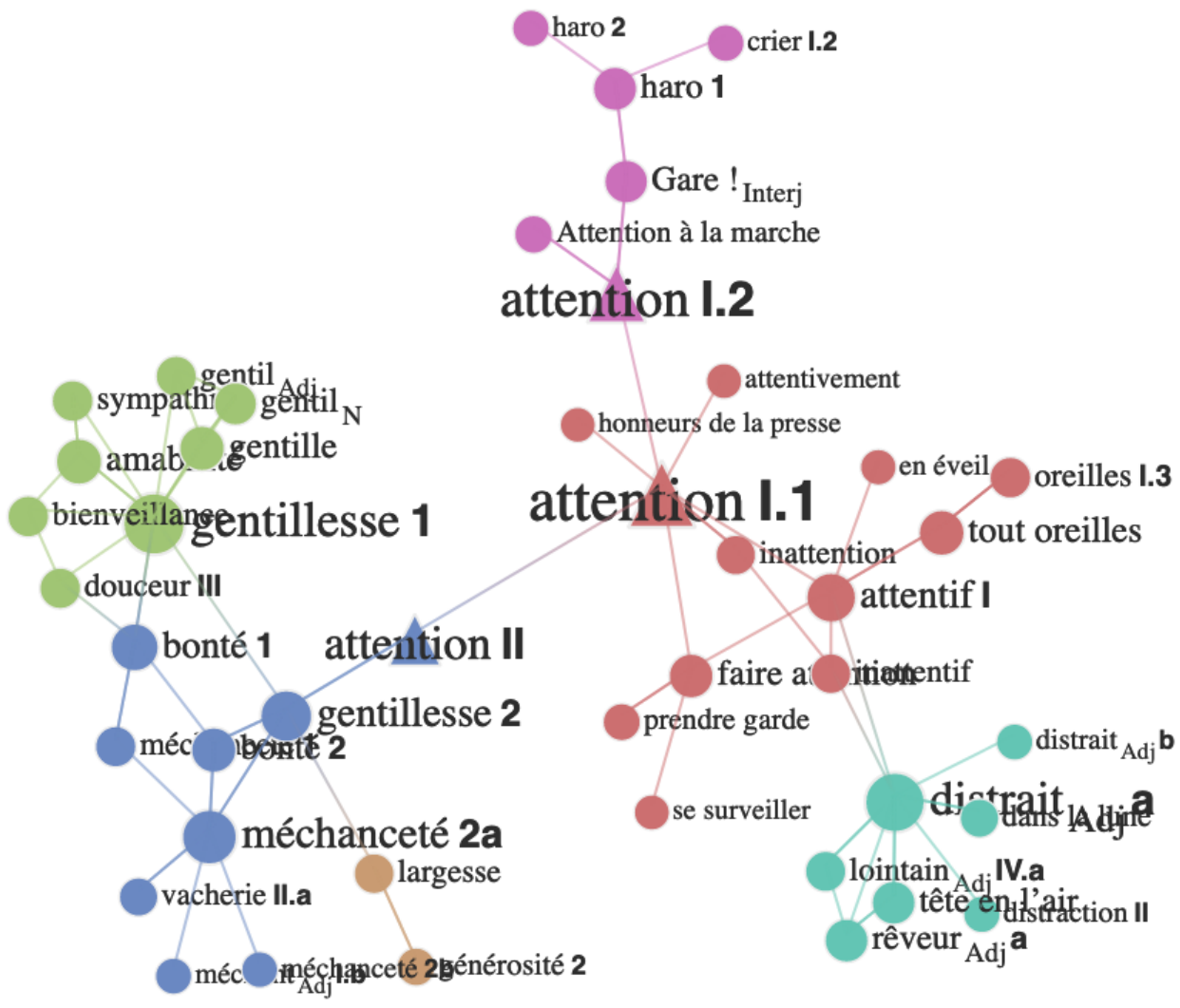

Figure 2.7: The network of '*attention*' (noun, En. attention) in the French Lexical Network (visualized using Spiderlex: https://spiderlex.atilf.fr/fr/q/*attention***)

A resource of interest that puts forward the theoretical and descriptive principles of ECD is the French Lexical Network (FLN, or "*Réseau Lexical du Français*") (Polguère, 2014). FLN is in fact a lexical system where the vertices of the graph are the lexical units, i.e. senses not synsets as in WordNet, in the language and the edges denote the paradigmatic and syntagmatic links. These are standardized by means of the system of lexical functions in MTT. Akin to the Generative Lexicon, FLN aims to make lexical databases compatible with the computational processing of linguistic data by bringing both MTT and ECD frameworks together. Figure 2.7 illustrates the lexical network of the vocable '*attention*' in French in FLN where the coloring of nodes reflect various senses of the lexeme which currently amount to six. ATTENTION I.1 is the basic lexical unit of the vocable which is associated to ATTENTION I, ATTENTION I.2, ATTENTION II, AMABILITÉ (kindness), DISTRACTION II and GÉNÉROSITÉ 2 (generosity). The internal edges within each cluster, i.e. nodes with identical colors, are tagged with lexical functions, such as `Syn`(ATTENTION I.2)= *Gare !* (watch out) or `Ant`(INATTENTION)= *attention*. Furthermore, the length of the edges represents the measure of proximity in such a way that nodes associated with shorter edges are more semantically close.

In addition to the linguistic application of both ECD and MTT being applied to other languages and even other types of resources such as FrameNet (Bouveret and Fillmore, 2008; Coyne and Rambow, 2009) and data such as terminologies (L'Homme, 2007) and collocations (Lareau et al., 2012), there have been efforts in a few language technology applications such as machine translation (Sasha, 2008), lexical and semantic disambiguation (Kolesnikova, 2020) and syntactic analysis (Mille et al., 2012) to leverage this conceptualization to assist in NLP tasks.

## 2.5 GENERATIVE LEXICON

One of the major limitations in sense inventories, as in dictionaries, is the restrictive number of senses, oftentimes enumerated, that are provided for a word. This approach, which is referred to as 'static' or 'frozen' by Pustejovsky (2006, p. 4), limits the number of senses for a word to a fixed number and also impedes finer distinctions between word senses and descriptions. As such, a dictionary user is coerced into context enforcing and matching the closest available sense. On the other hand, the traditional approaches to word-sense disambiguation (WSD) would also rely heavily on the limited number of senses listed for a word. More importantly, providing word senses as an exhaustive list reduces the possibility to further generalize or extend existing senses in a formal way.

As a solution, the Generative Lexicon (Pustejovsky, 1995) presents a lexical semantics theory that addresses the problem of polysemy and sense representation and attempts to provide compositional semantics for the contextual modulations in language. In other words, given a finite set of means, a generative lexicon allows creating and extending an indefinite number of senses by means of a rich and expressive vocabulary that can be used as data structure associated with the lexical encoding of semantic information. Initially motivated by compositional semantics and inspired by Aristotelian foundations for defining concepts based on *genus-differentia* (see Section 2.2.2), the core idea of a generative lexicon revolves around the inference between words and lays the theoretical foundation for the computational processing of word meaning in a theoretical manner.

For a lexicon to be generative, Pustejovsky (1995) proposes the following four levels:

1. **Lexical typing structure** which identifies how a lexical structure is related to other ones within the lexicon shaping the general world knowledge. For instance, knowing the structure of 'wood' (noun) being a MATERIAL that inherits the properties of its super-class ARTIFACT and then PHYSICAL-OBJECT, the given word 'rubber' can be related to the same structure.
2. **Argument structure** which specifies the type and number of argument structures, i.e. the relationship between words and functions, and their syntactic realizations (Grimshaw, 1990). For instance, the verb 'build' would have three arguments: subject and two objects, as in "He builds the box with wood". This can be shown as follows:

$$\begin{bmatrix} \text{BUILD}_{verb} & \\ \text{ARGSTR} & \begin{bmatrix} \text{ARG}_1 & \textbf{animate_individual} \\ \text{ARG}_2 & \textbf{artifact} \\ \text{D-ARG}_1 & \textbf{material} \end{bmatrix} \end{bmatrix}$$

where D-ARG1 refers to 'wooden' as a modifier in the direct object and defined as a default argument (Pustejovsky, 1995, p. 66).

3. **Event structure** which refers to the event type of a verb or phrase. Three primitive types are defined, that is *state*, *process* and *transition*. For example, the interaction between people denotes a transition event, while the state of things refers to a state event.
4. **Qualia structure** which is the most important component of a word sense, composed of the following four qualia roles:
   - The formal role which identifies an entity and is roughly similar to a genus, e.g. *a car is a physical object*.
   - The constitutive role which identifies the components of the entity, e.g. *a car is a physical object that has a body, a chassis and an engine*.
   - The agentive role which provides information about the origin of the entity, e.g. *a car is a physical object that has a body, a chassis and an engine that is an artifact and is constructed by an entity* Y.
   - The telic role which refers to the purpose of the entity, e.g. *a car is a physical object that has a body, a chassis and an engine that is an artifact and is constructed by an entity* Y *and an individual, like* Y, *can drive it. Driving is a process (P).*

   The qualia structure of the example 'car' can be summarized as follows:

$$\begin{bmatrix} \text{CAR}(x) & \\ \text{FORMAL} & \text{PHYSICAL_OBJECT}(x) \\ \text{CONST} & \left\{ \text{BODY, CHASSIS, ENGINE, ...} \right\} \\ \text{AGENTIVE} & \text{ARTIFACT}(x), \text{CONSTRUCT}(y, x) \\ \text{TELIC} & \text{DRIVE}(p, y, x) \end{bmatrix}$$

The Generative Lexicon theory has received some attention in both language technology and formal semantics as in Vikner and Jensen (2002). The Brandeis Semantic Ontology project (Pustejovsky et al., 2006), integrating Generative Lexicon event structures into VerbNet (Brown et al., 2018), a toolkit to create Generative Lexicon resources (Henry and Bassac, 2008) and the application of the Generative Lexicon in non-literal linguistic language resolution (Bergler, 2013) and the Semantic Web (Toral et al., 2007) are a few of such efforts. Even though the Generative Lexicon initially comes with the promise of enhancing the exploitation of information thanks to compositions, the idea seems to have yet to become operational given that there are not many resources or applications in NLP that rely on a generative lexicon. Moreover, questions have been raised regarding the level of compositionality and lexical generativity of the Generative Lexicon theory (Fodor and Lepore, 2000).

## 2.6 NATURAL SEMANTIC METALANGUAGE

One of the main characteristics of the previously mentioned LSRs is the theories based on which lexical units are defined in relation with other ones, as in 'book' vs. 'publishing house'. This process can be quite language-dependent given that essentially languages do not have the same vocabulary. One idea would be to define lexical units relative to a set of predefined, small yet universal meanings. This is the core idea of the Natural Semantic Metalanguage.

The Natural Semantic Metalanguage (NSM) approach (Wierzbicka, 1996) aims to describe complex meanings using simpler semantic primitives, called *primes*, such as `I`, `BIG` and `NOW`. Semantic primes are a set of basic meanings which are, arguably, universal to all languages and irreducible to smaller semantic cores (Goddard, 2008; Goddard and Wierzbicka, 2014). These primes are used to represent more complex linguistic concepts. To do so, a conceptual syntax is followed in which various primes can be combined together to form NSM clauses such as substantives and predicates, e.g. "`SOMETHING HAPPENS TO SOMEONE/SOMETHING`" to define undergoer frames. "Reductive paraphrase" is the name given to this approach of semantic analysis, and the results are known as "explications". Since only primes are used in a reductive paraphrase, there are no terms, neologisms or abbreviations and also, no obfuscated definitions such as "juridical (adjective): pertaining to a judge or to jurisprudence" (Princeton WordNet 3.1). The following example defines 'amazement' in the NSM (taken from (Wierzbicka, 1992, p. 549)) :

| |
|---|
| amazement<br>X feels something<br>sometimes a person thinks something like this:<br>something happened now<br>I didn't know before now: this can happen<br>if I thought about this I would have said: this cannot happen<br>because of this, this person feels something<br>X feels like this |

Moreover, it is easier to develop cross-translatable semantics that is not essentially Anglocentric, using NSM. In addition to the primes which consist of 65 based on significant efforts to empirical investigation (Goddard and Wierzbicka, 2013), it is possible to create new semantic cores depending on the language. Table 2.3 provides the list of primes in English defined according to (Goddard and Wierzbicka, 2002). It is important to note that semantic primes are conceptually defined and not lexically. That means that, although there might be many meanings for a semantic prime, such as 'move', in a dictionary, the basic meaning of the words are meant to define a prime. Therefore, lexical ambiguity should not change the logical meaning of the word.

| | |
|---|---|
| Substantives | I, YOU, SOMEONE, PEOPLE, SOMETHING, THING, BODY |
| Relational substantives | KIND, PART |
| Determiners | THIS, THE SAME, OTHER ELSE |
| Quantifiers | ONE, TWO, SOME, ALL, MUCH, MANY, LITTLE, FEW |
| Evaluaters | GOOD, BAD |
| Descriptors | BIG, SMALL |
| Mental predicates | THINK, KNOW, WANT, FEEL, SEE, HEAR |
| Speech | SAY, WORDS, TRUE |
| Actions, events, movement | DO, HAPPEN, MOVE |
| Location, existence, specification | BE (SOMEWHERE),THERE IS, BE (SOMEONE/SOMETHING) |
| Possession | (IS) MINE |
| Life and death | LIVE, DIE |
| Time | WHEN TIME, NOW, BEFORE, AFTER, A LONG TIME, A SHORT TIME, FOR SOME TIME, MOMENT |
| Space | WHERE PLACE, HERE, ABOVE, BELOW, FAR, NEAR, SIDE, INSIDE, TOUCH |
| Logical concepts | NOT, MAYBE, CAN, BECAUSE, IF |
| Intensifier, augmenter | VERY, MORE |
| Similarity | LIKE AS WAY |

Table 2.3: A list of semantic primes grouped into related categories according to Goddard and Wierzbicka (2002)

NSM also comes with the notion of semantic *molecules*. Semantic molecules refer to non-primitive semantic units that can be used within the semantic structure of another more complex word. For instance, 'bird' in the clause "a KIND of bird", as the definition of the word 'robin', refers to a semantic molecule that may be used in the explication of 'eagle' as well. This way, a taxonomic structure is created according to explications which allows inducing semantic relationships between words. On the other hand, explications provided to some words can reveal other types of semantic structures; for example, explications for 'fork', 'spoon' and 'plate' include the semantic molecule 'eat', where the semantic relationship corresponds to the intended purpose or function (Goddard, 2010).

Even though the NSM approach has been adapted to other languages such as Italian (Maher, 2002) and Spanish (Verdín-Armenta and Díaz-Rodríguez, 2017) and has been demonstrated as a flexible approach to tackle polysemy (Goddard, 2000), some faults in the underlying assumptions of this approach, particularly concerning the universality and untranslatability of culture-specific concepts (Blumczyński, 2013), have been raised. Moreover, the integration of the NSM approach in NLP applications seems to be rather unexplored, with very few studies such as Zamblera (2010) and Stock (2008).

## 2.7 TERMINOLOGIES

Terminology is a set of specialized terms relating to a field of activity, like the Unified Medical Language System (Bodenreider, 2004) describing the terminology for medicine. Terminology also designates the task of identifying, analyzing and, if necessary, coining neologisms for a given technique such as 'cyber-attack', in concrete situations, in such a way as to meet the needs for expression of the notions and concepts of a domain. Terminology (or terminography) applies to specialized languages, analogous to lexicography that relates to general language (Alberts, 2001; Costa, 2013). The alphabetical list of technical terms in a particular domain of knowledge is also referred to as a *glossary*.

Similar to dictionaries, terminologies provide linguistic descriptions such as morphosyntactic categories and definitions. However, this is of secondary importance as the focus of terminology is on the organization of knowledge and concept modeling in lieu of the linguistic characteristics of terms. Therefore, semantic relations, such as meronymy as in 'car' vs. 'chassis' and hyponymy as in 'eagle' vs 'bird', are widely used in terminologies that are rather taxonomic and more in line with ontological relationships. This being said, recent studies, as L'Homme (2020) states, tend to utilize lexical semantics in terminologies to better understand the nature of words and their relationships.

As with language resources, terminologies have been used in many language technology applications such as entity labeling (Ly et al., 2018) and recommender systems (Demner-Fushman et al., 2009). Moreover, terminologies are important resources to increase coverage of other language resources, like dictionaries, for various other tasks such as machine translation. The semantic enrichment of terminological resources using formalization of ontological information has also received attention (Speranza et al., 2020; Lamé et al., 2019). In addition, the automatic creation of controlled terminologies (Sarica et al., 2020) and term extraction (Aubin and Hamon, 2006) are other tasks in the same field.

## 2.8 ONTOLOGICAL RESOURCES

An ontology provides a formal and explicit specification of how the vocabulary in a specific domain in a language is used with the actual occurrence of a specific situation. According to Declerck et al. (2006), "*the main goal of ontologies is to formalize domain knowledge for ensuring a more compact description of it and a more efficient access to it.*". To this end, a set of representational primitives are used to model a domain using classes and relationships between them. In this regard, many LSRs can have a few ontological features and can be categorized based on their "ontological precision", i.e. the precision of conveying the intended meaning of a situation based on the conceptualization in an ontology, as defined by Guarino (2006). This way, catalogs, glossaries, taxonomies, thesauri and axiomatic theories are ontological resources ordered from the least ontological precision (catalogs) to the most precise one (axiomatic theories). The parameter that varies the ontological precision of a resource within this range is the formalization that provides structure and constrained meaning. According to Jarrar (2021), "ontologies are strictly formal, specified in some logical language with formal semantics" making it possible to reason over the concepts and entities and carry out a logical analysis. This is also the case of WordNet which is sometimes categorized as an ontological resource (Miller and Fellbaum, 2007). Finally, there are many interpretations for an ontology as a philosophical discipline and as a conceptual and formal system described in Guarino and Giaretta (1995). In this context, the latter is meant.

Nevertheless, language resources are primarily created at the language level and do not necessarily provide references to non-lexicalized ontological categories. For instance, '*marrke*' (noun, [mɑ:rkæ]) appears in a Kurdish lexicon to denote 'the egg that is put in a coop to encourage chickens to lay eggs', while such a concept is not lexicalized in the English lexicon but can be formalized in an ontology. Furthermore, many semantic relations available in a lexicon may not be similarly represented in an ontology; for instance, near-synonyms such as `hungry-peckish` in a lexicon appear with an overlapping word sense while in an ontology, they are defined in a different hierarchy which is not essentially compatible with that of a lexicon. The lack of such distinctions along with the presence of semantic relations such as semantic categorizations that are not required in an ontology increases the gap between these two resources. On the impact of linguistic categorizations on this gap, Hirst (2009) states that:

> "Often, linguistic categorizations are not even a reliable reflection of the world. For example, many languages distinguish in their syntax between objects that are discrete and those that are not: countable and mass nouns. This is also an important distinction for many ontologies; but one should not look in the lexicon to find the ontological data, for in practice, the actual linguistic categorization is rather arbitrary and not a very accurate or consistent reflection of discreteness and non-discreteness in the world. For example, in English, spaghetti is a mass noun, but noodle is countable."

Since the late 1990s, significant progress has been made in ontology engineering and creating standards for ontologies, with standards such as Resource Description Framework (RDF), RDF Schema (Brickley, 2004) and Web Ontology Language (OWL) (Antoniou and Van Harmelen, 2004). In the same vein, given the potential of ontologies for representation and modeling of linguistic data and their semantic interoperability, there has been much interest in creating resources to bring together the ontological features of ontologies and lexical properties of lexicons, with many initiating studies such as Knight and Luk (1994); Dahlgren (1995); Buitelaar (1998); Farrar et al. (2002); Farrar and Langendoen (2003). One of the important efforts in this realm is LexInfo–linguistic grounding of ontologies (Buitelaar et al., 2009) which combines necessary design aspects of previous models such as LingInfo (Buitelaar et al., 2006) and LexOnto (Cimiano et al., 2007) to associate linguistic information and computational lexica with ontologies. Other prominent ontologies in linguistics are SUMO–Suggested Upper Merged Ontology (Niles and Pease, 2001), MCR–Meaning Multilingual Central Repository Atserias et al. (2004), Olia–ontologies of linguistic annotation (Chiarcos and Sukhareva, 2015) and Lemon–Lexicon Model for Ontologies (McCrae et al., 2011).

### 2.8.1 Linguistic Linked Data

The studies on representing linguistic data as ontologies have paved the way to a new interdisciplinary branch called the Linguistic Linked Data (LLD) (Cimiano et al., 2020c). The concept of the Web of Linked Data, which makes RDF data available using the HyperText Transfer Protocol (HTTP), has gained traction along with the Web of data, also known as the Semantic Web, particularly in the NLP community as a standard for linguistic resource creation. According to the official definition of W3C[14],

> "Linked Data lies at the heart of what Semantic Web is all about: large scale integration of, and reasoning on, data on the Web. Almost all applications listed in, say collection of Semantic Web Case Studies and Use Cases are essentially based on the accessibility of, and integration of Linked Data at various level of complexities."

Moreover, the unique potential that the Semantic Web and Linked Data offer for electronic lexicography enables interoperability across lexical resources by leveraging printed or unstructured linguistic data to machine-readable semantic formats. To this end, linguistic linked open data aims to promote the creation, maintenance and

14 https://www.w3.org/standards/semanticweb/data

publication of language and linguistic data according to the principles of linked data (Berners-Lee, 2006) and openly. Figure 2.8 shows the resources which are available in linked data according to the Linguistic Linked Open Data Cloud.

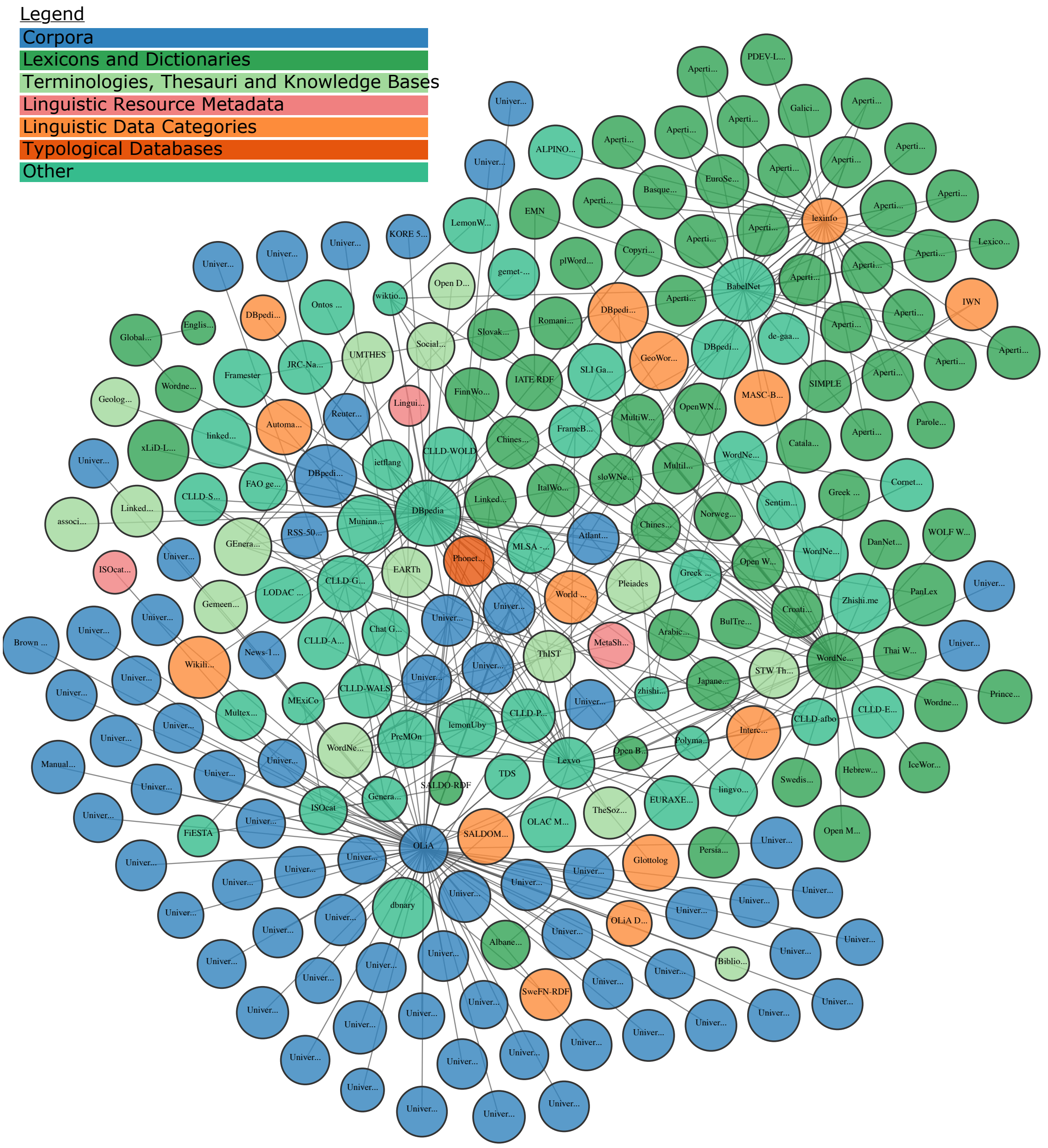

**Figure 2.8:** The Linguistic Linked Open Data Cloud (From https://linguistic-lod.org)

Over the recent years, the usage of LLD as a mechanism to create and publish has gained much wider attention and recognition in the field. This is thanks to the interoperability and accessibility that LLD offers and further tendency towards more open-access standards for community-based projects. One particular example that distinctly focuses on inter-separability and accessibility of data is Ontolex-Lemon. Building upon a few data models, especially LexInfo (Cimiano et al., 2011), LIR (Montiel-Ponsoda et al., 2008) and LMF (Francopoulo et al., 2006), Ontolex-Lemon provides an RDF native data model based on the Lemon (McCrae et al., 2011) that realizes the representation of linguistic data on the Semantic Web. This model provides comprehensive linguistic descriptions such as information related to the representation of morphological and syntactic properties of lexical entries.

Furthermore, the current trend of applying digital technologies to disciplines of the humanities, also known as Digital Humanities, has created a wider range of applications in which LLD can be used, in addition to more traditional fields as computational linguistics and NLP (Khan et al., 2021).

## 2.9 KNOWLEDGE GRAPHS

A knowledge graph is a representation of knowledge about a domain in a machine-readable form. The representation of knowledge in the form of a graph is a key element for the efficient and contextual retrieval of rich information and knowledge. A knowledge graph is made up of three components: an ontology defining a data model in the RDF, schemata or controlled vocabularies that can be linked and the resources covered by the graph (Ehrlinger and Wöß, 2016). Knowledge graphs were initially introduced by Google as information boxes in response to the search queries and, are a specification of knowledge base. The latter may refer to a wider range of knowledge resources, whether structured and unstructured information. In addition to Google's original knowledge graph, there are many other services and platforms which are referred to as knowledge graphs.[15].

Knowledge graphs are critical to many domains. They contain large amounts of information, used in applications as diverse as search, question-answering systems, and conversational agents. They are the backbone of linked open data, helping connect entities from different datasets. Finally, they create rich knowledge engineering ecosystems, making significant, empirical contributions to our understanding of knowledge representation, engineering, and practices. Knowledge graphs have been widely used in many knowledge-aware applications, in general in information retrieval (Reinanda et al., 2020) and artificial intelligence (Nicholson and Greene, 2020) and more specifically in NLP to exploit and extend lexical, ontological and terminological resources for tasks such as language representation learning (Logan IV et al., 2019), question answering (Mohammed et al., 2018), relation extraction (Ristoski et al., 2020), terminology extraction (Speranza et al., 2019) and language resource enrichment (Martín-Chozas et al., 2020).

Analogous to other language resources, there are two approaches to create, maintain and publish knowledge graphs: open knowledge graphs and enterprise knowl-

15 For further information regarding the definitions of knowledge base and knowledge graph and their differences according to the literature, see (Hogan et al., 2021, p. 111) edge graphs. Widely-known examples of open knowledge graphs are Wikidata[16], DBpedia[17] (Bizer et al., 2009), GeoNames[18] and YAGO (Suchanek et al., 2007) which are accessible openly. Even though these knowledge graphs are not identical in architecture, technology and purpose, they try to promote the creation and publication of data in an open manner with the contribution of various communities online. On the other hand, enterprise knowledge graphs are restricted to a company and are not essentially openly accessible. Prominent examples of such knowledge graphs are search engines such as Microsoft's Bing[19], Microsoft's Knowledge Mining API[20] and Google's search engine. In the same vein, there are knowledge graphs with the particular objective of representing and storing linguistic data such as ConceptNet (Speer et al., 2017) and Wikidata's Lexicographical Data[21] which is built on Wiktionary and is compatible with Ontolex-Lemon.

Since ontologies are one of the building components of knowledge graphs, it is possible to reason over the data using inference rules. This can be carried out thanks to the RDF data model implemented in each knowledge graph. RDF is a framework for describing resources on the Web which was initially designed to represent metadata. However, nowadays RDF is the foundational data model for Semantic Web and knowledge graphs. In addition, RDF along with other technologies such as SPARQL, OWL, and SKOS empower Semantic Web and Linked Data. RDF expressions are in the form of subject–predicate–object expressions, i.e. `subject predicate object` known as a semantic triple. A triple is the basic principle to define knowledge in a data model where the subject should be a Uniform Resource Identifier (URI) or a blank node, the predicate is a URI and the object can be a URI, a literal or a blank node. Unlike traditional databases where data has to adhere to a fixed schema, there is no prescribed schema for RDF documents. This is the reason that RDF is called semi-structured. On the other hand, an RDF document includes schema information and can be described without additional information. Therefore, an RDF data model is self-describing too. The following snippet in RDF from DBpedia[22] contains triplets describing France and Paris:

```
dbr:France dbo:countryCode "+33" .
dbr:France dbo:demonym  "French"@en .
dbr:France dbo:governmentType  dbr:Unitary_state .
dbr:France dbo:anthem dbr:La_Marseillaise .
dbr:France dbo:capital dbr:Paris .
dbr:Paris dbo:populationTotal "2229621"^^xsd:integer .
```

16 https://www.wikidata.org
17 https://www.dbpedia.org
18 https://www.geonames.org
19 https://www.bing.com
20 https://www.microsoft.com/en-us/research/project/knowledge-mining-api
21 https://www.wikidata.org/wiki/Wikidata:Lexicographical_data
22 https://dbpedia.org/page/France

where dbr and dbo are prefixes referring to DBpedia's resource[23] and DBpedia's ontology[24], respectively. This way, dbr:France is equivalent to https://dbpedia.org/page/France. These triples define France as an entity with predicates such as capital, country code and anthem. Each triple is structured according to the data model used in the knowledge graph which is defined by the four principals of RDF Schema, i.e., class hierarchy, property hierarchy, domain and range of properties. Figure 2.9 shows a simple schema of the ontology of a place as a class derived from Thing (the highest superclass), then Country being a sub-class of Place, County being a sub-class of Country and City being a sub-class of County. The schema also shows our individuals as France, Ile-de-France and Paris.

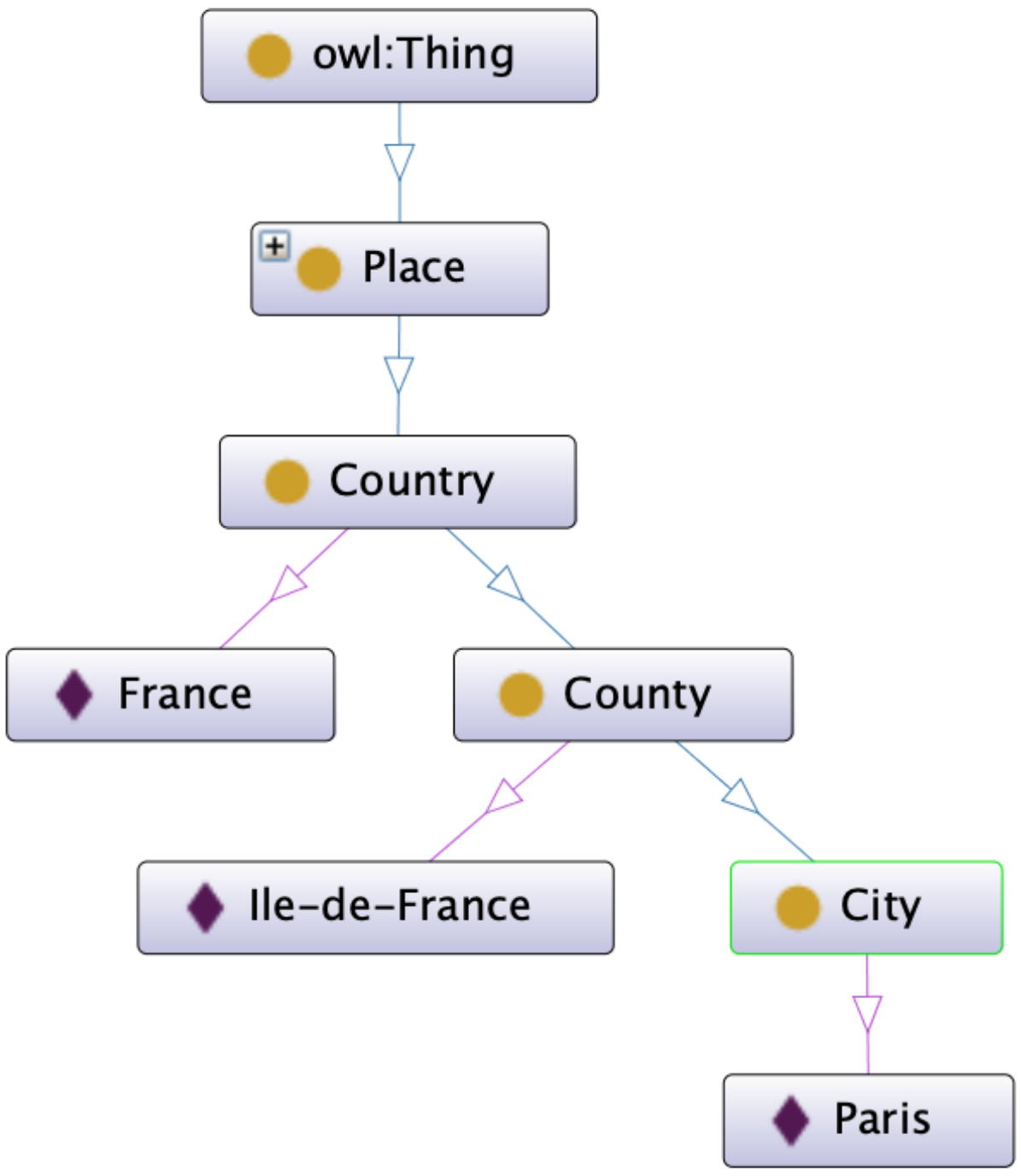

Figure 2.9: A simple ontology for Place, showing France, Ile-de-France and Paris as instances. This is created using Protégé (https://protege.stanford.edu).

It should be noted that the name of the components of an RDF triple, e.g. dbr:France, does not necessarily correspond to the label or the alias of an entity but to the resource that describes that entity. Unlike DBpedia, in some knowledge graphs such as Wikidata, entities are also defined by unique identifiers. For instance, the same entity France is defined as Q142 at https://www.wikidata.org/wiki/Q142 on Wikidata. This being said, URI names, labels and descriptive data can sometimes be considered as linguistic information.

With the advances of neural network-based techniques and the growing size of knowledge graphs, there has been an increasing interest in learning knowledge graph

23 http://dbpedia.org/resource
24 http://dbpedia.org/ontology

representations and creating graph embeddings (Wang et al., 2017). To do so, a low-dimensional distributional representation of data in a knowledge graph is created based on the rich semantic information of entities and relations (Ji et al., 2021). This embedded distributional representation is then used in various other tasks, such as semantic parsing (Heck and Huang, 2014), chatbots (Yoo and Jeong, 2020), machine translation (Lu et al., 2018) and explainability in artificial intelligence (Xian et al., 2019). Given the significant size of the structured data in knowledge graphs, they have been proved beneficial for incorporating commonsense knowledge (Chang et al., 2021) and increasing factual information in text-based applications (Zhu et al., 2020). Furthermore, many knowledge graphs come with query services, also known as SPARQL endpoint, that can facilitate their integration into NLP applications. For instance, the following SPARQL query retrieves lexemes describing 'book' (noun) in different languages via Wikidata's SPARQL endpoint[25]:

```
SELECT ?lemma ?languageLabel WHERE {
  ?l a ontolex:LexicalEntry ;
       ontolex:sense ?sense ;
       dct:language ?language ;
       wikibase:lemma ?lemma.
  ?sense wdt:P5137 wd:Q571 .

  SERVICE wikibase:label {
    bd:serviceParam wikibase:language "en".
  }
}
```

## 2.10 LANGUAGE MODELS

The fundamental idea of language modeling is based on the sequential nature of language production, that is the procedure of transforming meaning into speech or more importantly for our case, into text (Garrett, 1989). This topic has been widely studied in the context of psycholinguistics under the connectionist models of language production (Dell et al., 1999). Similarly, many aspects of a language can be studied through statistical language modeling where information such as dependency between words, lexical variations and structural features, e.g. syntax and semantics, and pattern associations and various meanings of words based on context, e.g. polysemy, are modeled and analyzed based on Bayesian inference. In other words, the ultimate goal is to estimate the probability of a given word $w_i$ or character, depending

25 https://query.wikidata.org

on the level of modeling, given the $n$ preceding and succeeding surrounding words or characters, i.e. the context, as follows:

$$P(w_i | w_{i-n} ... w_{i-1}, w_{i+1} ... w_{i+n})$$

Pre-trained language models are an important kind of language resource which are trained on large corpora and represent relationships between words by means of vector space models. Pre-trained word representations, also known as word embeddings, such as WORD2VEC (Mikolov et al., 2013a) and GLOVE (Pennington et al., 2014), have been heralded as major breakthroughs in NLP. Instead of representing words as atomic units, word embeddings learn a distributional representation of words from large language data. As such, they are widely used in NLP pipelines and have been demonstrated to be efficient in many tasks such as part-of-speech tagging, machine translation and sentiment analysis (Naseem et al., 2021), word-sense disambiguation (Loureiro et al., 2020) and word-sense induction (Amrami and Goldberg, 2019).

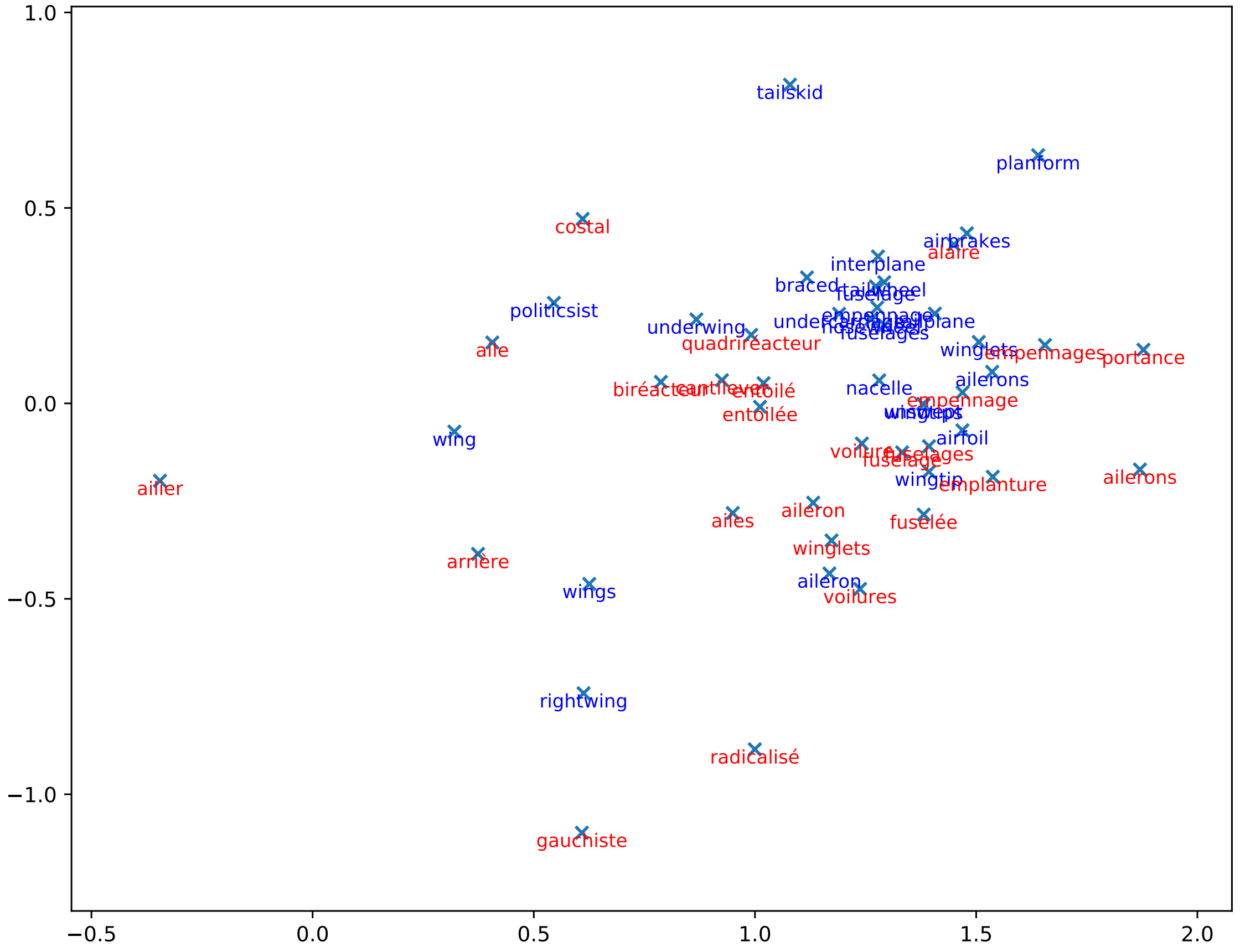

**Figure 2.10:** Closest words to the vector space of the word 'wing' (noun) in the MUSE embeddings of English along with the nearest words in the mapped vector space in French

It should be noted that language models differ from word embeddings in the sense that the latter create a distributed representation of a word and capture a more limited range of information, mainly based on word associations and without taking context into account. To remedy this, contextualized word embeddings have been proposed to incorporate context into word embeddings (Ethayarajh, 2019). In these techniques, a finer distinction can be drawn between word senses by capturing further information such as syntactic and semantic features (Mikolov et al., 2013b). There have been tremendous efforts in this field, from n-gram and tree-based models (Bahl et al., 1989) to more robust neural-based models (Bengio et al., 2003), particularly ELMO (Peters et al., 2018a), Bidirectional Encoder Representations from Transformer (BERT) (Devlin et al., 2018) and Generative Pre-trained Transformer 3 (GPT-3) (Brown et al., 2020).

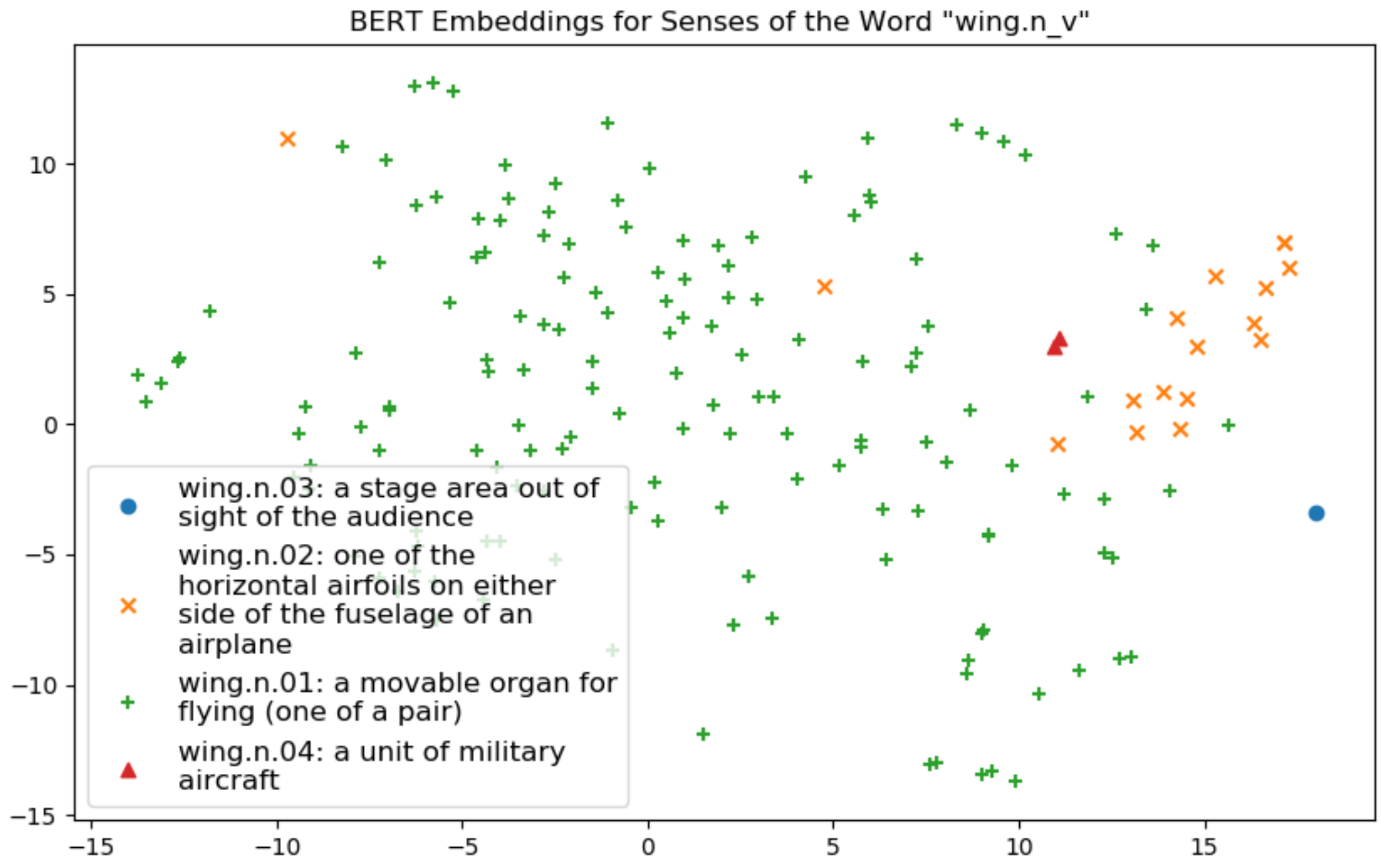

**Figure 2.11:** Distinction between senses of ‘wing’ (noun) using BERT

In order to demonstrate the ability of language models to deal with polysemy, we carry out a few experiments. Figure 2.10 illustrates words nearest to the vector space of the word ‘wing’ (noun) in the MUSE embeddings[26] using the t-Distributed Stochastic Neighbor Embedding (t-SNE) plot (Van der Maaten and Hinton, 2008). Considering the lemmata of these words, e.g. ‘wing’ and ‘winglet’ respectively for ‘wings’ and ‘winglets’, this plot indicates that embeddings can retrieve some of the semantically related words such as ‘rightwing’ with a political connotation, ‘underwing’ as the hindwing of an insect and ‘empennage’ as the tail in aeronautics. In addition, the mapping of the monolingual word spaces of English and French results an approximate alignment of words in the two languages, with applications in un-

26 https://github.com/facebookresearch/MUSE

supervised bilingual lexicon induction (Conneau et al., 2017). Interestingly, *gauchiste* ('leftwing') and *radicalisé* (radicalized) are among the words related to the political sense of 'wing'. Similarly, Figure 2.11 depicts various senses of 'wing' (noun) in BERT where our four targeted senses are retrieved.

Various studies have further demonstrated that word senses can be retrieved based on language models (Athiwaratkun et al., 2018) or achieve state-of-the-art results when used along with other LSRs (Levine et al., 2020). This being said, there is a sprawling literature on how word and contextual embeddings can be analyzed (Arora et al., 2020; Ji et al., 2022), evaluated (Nayak et al., 2016; Basta et al., 2019) and adapted to new domains and tasks (Merchant et al., 2020; Ueda et al., 2021). The latter is also known as 'fine-tuning'.

The status of language models as language resources does not seem to be agreed upon. Certainly, language models do not fit into the conventional definitions of LSRs, chiefly due to the lack of a structural representation of meaning. However, the fascinating ability of language models to capture contextual and semantic information using unlabeled corpora is advancing the frontier in many NLP applications. As such, among the various lexical semantic theories, distributional semantics seem to provide cutting-edge solutions in language technology thanks to the multilingualism of the Web and the availability of large corpora for many languages that do not necessarily need human annotation.

## 2.11 CONCLUSION

In this chapter, we shed light on various types of lexical semantic resources. Some of the major concepts in lexical semantics are first described, especially polysemy, that have motivated different theories to represent word senses and shaped differences between many resources. After discussing the difference between a lexicon and a dictionary, we proceed with a thorough description of dictionaries based on their content and structure. Furthermore, we present a few resources as network-based as they employ common representation of data, for instance as networks or lexical graphs. We also discuss how the structure of traditional electronic dictionaries is deemed inefficient for the computational processing of natural language; therefore, a set of new resources are previously created with the primary goal of compatibility with technology. Among these resources, Explanatory Combinatorial Dictionary (Mel'čuk, 2006), Generative Lexicon (Pustejovsky, 1995) and Natural Semantic Metalanguage (Wierzbicka, 1996) are presented. In the last part of the chapter a description on terminologies, ontological resources and most importantly, knowledge graphs and language models was provided. The latter two resources have led to much progress in language technology recently thanks to their coverage, convenient representations using embeddings and easy integration within applications.

Given the heterogeneity of lexical and semantic data and differences in structure and representation, important but challenging questions are raised regarding the alignment and inter-operability of lexical semantic resources. As such, the focus of the next chapter is on providing a systematic review of methodologies for, and approaches to, working with LSRs – and particularly on aligning them. We will also go through the major current tasks in NLP where LSRs play a central role.

# 3 SYSTEMATIC LITERATURE REVIEW

## 3.1 INTRODUCTION

In the previous chapter, a background was provided on the different types of language resources, particularly lexical semantic resources. Given the diversity in types of resources as in dictionaries vs. wordnets, in coverage as in dictionaries vs. terminologies, in languages as in multilingual vs. monolingual resources and in creation type as in expert-made vs. community-created resources, an important question is naturally raised on how to model, create, enrich and publish language resources. Although there are many approaches to distinguish between these tasks, broadly referred to as the life cycle of language resources (Rehm, 2016; Mattern, 2022), we categorize the task related to managing language resources in these four components, as illustrated in Figure 3.1.

The components of Figure 3.1 are defined as follows: the modeling phase focuses on the way that lexicalized data are modeled, as in Ontolex-Lemon (McCrae et al., 2017b) vs. conventional dictionaries, in the creation phase, the curation method is determined, such as open collaborative, expert-made and semi-automatic methods, the enrichment phase concerns tasks to maintain the resource and that increase the inter-operability and coverage of the resource, and finally, the publishing step regards platforms, such as catalogs, and methods based on which a resource is published and made accessible. The two last steps of enrichment and publishing are like a cycle that a resource can visit many times.

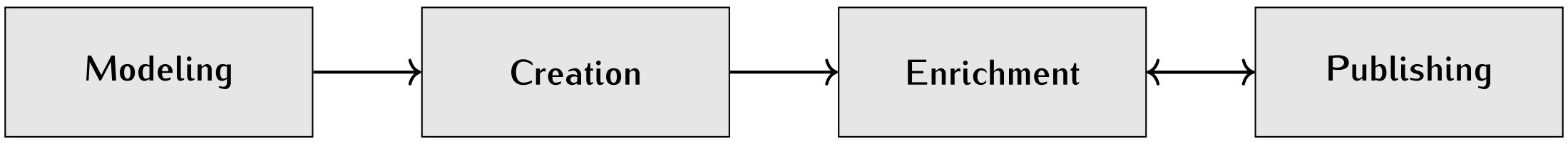

**Figure 3.1:** A general description of the life cycle of a lexical semantic resource

Going beyond the characteristics of language resources, we shed light on the life cycle of language resources in this chapter. In line with the focus of this thesis on the alignment of lexicographical content, we delve more into the enrichment component, mostly from a monolingual perspective. To this end, the previous studies that explicitly address the alignment task or implicitly solve a problem related to the alignment task are reviewed. The alignment task consists of finding similar entities with different structures, as in ontologies, or content, as in sense definitions in dictionaries.

As the topic of this thesis, the alignment of word senses and definitions aims to find potentially alignable senses and definitions in two resources, and also, predict the type of semantic relations that exist between the senses. This task can be implicitly addressed in tasks such as textual and non-textual similarity detection and semantic relation induction which are relevant as underlying problems.

Since the focus in this thesis is on monolingual resources, it is necessary to highlight the distinction between monolinguality, multilinguality and cross-linguality, terms that are widely used in the literature. A monolingual resource contains information in one language, regardless of the historical period that the language is documented. A multilingual resource is characterized as one containing information in more than one language. Multilinguality is thus the presence of content in at least two languages for a resource, such as a multilingual website or code-mixed corpora. The characterization of an approach or tool as "multilingual" can indicate that it is language-independent, i.e. can be applied to or used with more than one language. In contrast, cross-linguality is typically applicable to tasks or operations in which items expressed in one language are to be associated with items expressed in another language. A cross-lingual resource is characterized by the presence of links or equivalences between data in different languages that allow navigation from information in one language to information in another language. For example, cross-lingual information retrieval focuses on retrieving relevant documents that are written in a language other than the language in which the query is written. Other examples in NLP are cross-lingual word embeddings (Ruder et al., 2019), Universal Dependencies[1] and multilingual ontology mapping (Spohr et al., 2011).

Furthermore, this chapter is extended to a summary of the utility of language resources in general, and dictionaries in particular, in language technology. To this end, some of the most important current applications of various language resources in computational linguistics and NLP are introduced. In addition, we focus on the current standards and techniques for the creation, maintenance, enrichment and publication of language resources. Motivated by the current advances in the semantic web and linguistic linked open data, the most recent developments in these fields are summarized as well.

## 3.2 CREATION AND MODELING

In the creation and modeling step, a resource is to be implemented according to an encoding scheme. According to the Text Encoding Initiative (TEI) (TEI, 2020) guidelines regarding dictionaries[2], which can be used for a wide range of computational lexica, there are three views on modeling dictionary data, among many possible ones:

---

1 https://universaldependencies.org
2 http://web.uvic.ca/lancenrd/martin/guidelines/tei_DI.html

(i) a typographic view that focuses on the two-dimensional printed page, including information about the line and page breaks and other features of layout, (ii) an editorial view that concerns the one-dimensional sequence of tokens which can be seen as the input to the typesetting process and (iii) a lexical view which includes the underlying information represented in a dictionary without concern for its exact textual form. In a print dictionary, the editorial and typographic views are addressed in an orderly way after the lexical view that may require extracting information from a database. TEI provides a rich structure for modeling various types of resources and can be extended using other modules such as meta-data. Among the structured formats for lexicographical data, TEI is also the most widely used one. The following example shows an example for an entry in a Kurdish dictionary. The entry in the print dictionary is shown in Figure 3.2.

```
<entry>
    <form>
        <orth>bend</orth>
        <pron>bænd</pron>
    </form>
    <gramGrp>
        <pos>n</pos>
        <gen>f</gen>
    </gramGrp>
    <sense n="1">
        <cit type="translation" xml:lang="en">
            <quote>bond</quote>
        </cit>
        <cit type="example">
            <quote>divê em êdî li benda sibehê ranewestin.</quote>
            <cit type="translation" xml:lang="en">
                <quote>we shouldn't stand around waiting for tomorrow.</quote>
            </cit>
        </cit>
    </sense>
</entry>
```

Given the increasing importance of findable, accessible, interoperable and reusable data, known as the FAIR principle (Wilkinson et al., 2016; Khan et al., 2021), the semantic web technologies are ideal for lexicographic data in comparison to XML-based or less semantic models such as TEI (Tchechmedjiev, 2016; Klimek and Brümmer, 2015). In this vein, one of the most important models for lexicographical data is Ontolex-Lemon (McCrae et al., 2017b). This model, as briefly described in Section 2.8.1, is RDF-native and built in a modular way. Bosque-Gil et al. (2017) extend Ontolex-Lemon by adding a module for lexicography called *lexicog*. This module represents information, structures and annotations that are commonly found in lexicography. The *lexicographic resource* class in this module is used to represent the original printed entry structures. In addition, it is possible to add usage examples using the `UsageExample` class.

**bend** *f* bond; **li ~a** for the sake of, chained to, waiting for: *divê em êdî li benda sibehê ranewestin* we shouldn't stand around waiting for tomorrow; ~ **kirin** *v.t.* to fetter, arrest; **man di ~a** to wait for

```
1  :lexicon a lime:Lexicon;
2    lime:language
         <www.lexvo.org/page/iso639-3/kmr> ;
3    lime:entry :lex_bend .
4
5  :lex_bend a ontolex:LexicalEntry,
     ontolex:Word ;
6    ontolex:canonicalForm :form_bend ;
7    rdfs:label "bend"@kmr-latn .
8    lexinfo:partOfSpeech lexinfo:noun ;
9    lexinfo:gender lexinfo:feminine ;
10   ontolex:sense :bend_n_sense ;
11 :form_bend a ontolex:Form ;
12   dct:language
         <www.lexvo.org/page/iso639-3/kmr> ;
13   ontolex:writtenRep "bend"@kmr-latn ;
14   lexinfo:number lexinfo:singular ;
15 :bend_n_sense a ontolex:LexicalSense ;
16   lexicog:usageExample :bend_n_sense_ex .
17 :en_bond a ontolex:LexicalEntry ;
18   dct:language
         <http://lexvo.org/id/iso639-1/en> ;
19   rdfs:label "bond"@en ;
20   ontolex:sense :en_bond_sense .
21 :trans a vartrans:Translation ;
22  vartrans:source :bend_n_sense ;
23  vartrans:target :en_bond_sense .
24 :bend_n_sense_ex a lexicog:UsageExample;
25   rdf:value "divê em êdî li benda sibehê
         ranewestin."@kmr-latn .
26   rdf:value "we shouldn't stand around
         waiting for tomorrow."@en .
```

**Figure 3.2:** The conversion of an example entry from a Kurdish-English dictionary into RDF Turtle based on the OntoLex-Lemon model (to the right)

Figure 3.2 shows an example for the entry '*bend*' ('bond' (noun)) in a print dictionary versus the transformation in RDF Turtle in Ontolex-Lemon. In this entry, the following core modules of Ontolex-Lemon are used:

- Linguistic Metadata module–*lime* that allows us to describe metadata at the level of the lexicon-ontology interface with information such as lexical entries and language (lines 1 to 3 in Figure 3.2).
- Syntax and Semantics–*synsem* allows us to describe syntactic behavior. Syntactic frames are used to relate a lexical entry to one of its various syntactic roles, such as the canonical form of the word *bend* described in lines 5 to 7 in Figure 3.2).

- *lexinfo* (Cimiano et al., 2011) for describing relevant linguistic categories and properties, particularly part-of-speech, gender and number (lines 9 to 15 in Figure 3.2).
- Variation and Translation–*vartrans* is used to describe relations between lexical entries, particularly translations as shown in lines 17 to 19 in Figure 3.2.

It is worth emphasizing the extent to which data can be finely represented in this model; for instance, lexicalized data, i.e. literals in RDF, can be specified by language tags according to ISO 639-3[3] as well as scripts such as `arab` for Arabic and `latn` for Latin. Furthermore, the micro-structure of a print dictionary is not necessarily identical in terms of information when converted into Ontolex-Lemon as it requires additional processing. For instance, compound forms '∼ *kirin*' (to arrest, to fetter) and '*man di ∼a*' (to wait for), where ∼ refers to the lemma, are respectively replaced by '*bend kirin*' and '*man di benda*' and defined as separate entries.

The modeling phase requires an extensive study on the aspect of the lexicographical data, such as morphology (Klimek et al., 2019), phonology (Moran and McCloy, 2019) and syntax (Corcoglioniti et al., 2016). Some of the other data models and vocabularies developed for lexicographic data are the lexical markup framework (Francopoulo et al., 2006), ELEXIS Data Model (Tiberius et al.), TELIX (Rubiera et al., 2012), MLR model (Spohr, 2012), Lexfom (Fonseca et al., 2016), among others. Bosque-Gil et al. (2018) provides a survey on such models.

Similarly, some of the the platforms for creating language dictionaries are LexO (Bellandi, 2021), FLEx (Butler and Van Volkinburg, 2007), Léacslann (Měchura, 2012), SooSL[4], Lexique Pro (Guérin and Lacrampe, 2007) and ELEXIFIER[5]. On the other hand, Wiktionary[6], Language Forge[7], LEO[8] and Reverso[9], focus not only on dictionary creation, but also making the creation process accessible to a community.

## 3.3 ENRICHMENT

Automatic and semi-automatic enrichment of language resources has received much attention thanks to the advances in language technology and also, the expansion of multilingualism and the plethora of electronic resources available. Inter-connected and inter-operable resources not only improve word, knowledge and domain coverage and enhance multilinguality, it has been also shown that it is beneficial to improve language technology applications such as machine translation (Zhao et al., 2020) and

3 https://iso639-3.sil.org/code_tables/639/
4 https://www.soosl.net
5 https://elexifier.elex.is
6 https://www.wiktionary.org
7 https://languageforge.org
8 https://www.leo.org
9 https://www.reverso.net

also, create centralized repositories of data (Gurevych et al., 2012). However, linking concepts and words across resources is challenging, especially due to the complexity of the structure of LSRs which generally contain heterogeneous and multi-lingual data.

In this section, among the many tasks to enrich language resources, we focus on the tasks most related to the main topic of the thesis. The tasks are semantic similarity detection and language resource alignment in its broad scope. We will also summarize the previous works in translation inference.

### 3.3.1 Semantic Similarity Detection

The objective of text similarity detection techniques is to estimate the level of similarity of two text documents or strings in an automatic way. Given the usage of this task in many applications, from a basic string-based text finder in text editors to more sophisticated semantic-based problems such as word sense disambiguation (Prior and Geffet, 2003), question answering (Jin et al., 2019), paraphrase identification (Mohamed and Oussalah, 2020) and natural language inference (Lan and Xu, 2018), semantic similarity detection has received much attention in NLP and has been instrumental. Many surveys are available on the same topic, as in Vijaymeena and Kavitha (2016); Sunilkumar and Shaji (2019); Chandrasekaran and Mago (2021).

This task has been widely studied in two major branches: lexical similarity detection and semantic similarity detection. While the first task aims to evaluate the similarity of two strings at the string level, for instance 'book' and 'books' are similar lexically, semantic-based approaches take the meaning of the words into account as well, as in 'king' and 'queen'. To define what is meant by semantic or conceptual similarity, there are different objectives such as association, relatedness or topical similarity, that can be indicative of the relationships. Furthermore, similarity can be calculated at various levels, such as character-level, word-level and sentence-level. As a basic method, the similarity at a higher level, i.e. sentence, can be calculated based on the similarity at a lower level, i.e. word, meaning that words within two sentences are compared to determine the level of similarity. In embedding-based approaches, the similarity of sentences can be estimated by merging the vector values of the composing words and averaging them. This being said, there are other techniques based on convolutional neural networks (Yao et al., 2018), semantic networks (Li et al., 2006) and using syntactic and semantic features (Quan et al., 2019), to mention but a few.

Beyond string-based similarity detection approaches, as discussed in Cohen et al. (2003), this section provides various methods that have been previously proposed for similarity detection. In addition to these approaches, there are hybrid approaches that rely on the information extracted from various sources, as in Zhu and Iglesias (2016), Cai et al. (2018), Li et al. (2003) and Mohamed and Oussalah (2020).

### *Corpus-based approaches*

In this approach, the similarity is calculated based on the information extracted from large corpora. The assumption of a word being known by "the company it keeps" is not only foundational to corpus-based approaches in similarity detection, but also to many other advances in distributional semantics, especially word embeddings (Almeida and Xexéo, 2019). Previous studies have proposed co-occurrence extraction (Kolb, 2008), point-wise mutual information (Bouma, 2009; Islam and Inkpen, 2006), latent semantic analysis (Hofmann, 1999; Islam and Hoque, 2010), hyperspace analogue to language (Azzopardi et al., 2005), normalized Google distance (Cilibrasi and Vitanyi, 2007), explicit semantic analysis (Egozi et al., 2011) and dependency-based models (Agirre et al., 2009).

### *LSR-based approaches*

These approaches incorporate knowledge about words, such as lexical and semantic relations, from LSRs to estimate the similarity. WordNet (Miller, 1995) as a rich semantic network, for instance, has been widely used in the similarity task (Meng et al., 2013). Various measures have been proposed to use the taxonomic and lexical information of WordNet and dictionaries. Leacock and Chodorow (1998) introduce a similarity measure according to the path length between two words in WordNet. Lesk (1986) proposes an algorithm to calculate the similarity based on the overlap of word definitions in dictionaries, with primary application in word sense disambiguation. Wu and Palmer (1994) uses depth of concepts in the WordNet to measure the structural relations, i.e. how they are related in the hierarchy, for similarity detection. Kubis (2015) creates a framework for similarity detection based on WordNets and various other resources such as Wikipedia. And, Jiang et al. (2015) employs a feature-based technique using definitions of word extracted from Wikipedia.

### *Embeddings-based approaches*

With the emergence of various frameworks in deep learning and the availability of large corpora for many languages, there have been many advances in utilizing distributional semantics, particularly embeddings, for semantic similarity detection. Akin to corpus-based approaches, the surrounding words of a given word within a text are taken into account to create the embeddings of that specific word. By comparing the embeddings of two words, which are in fact two vectors with the same dimension, the level of similarity can be estimated by measuring the distance between the vectors. As it was discussed in Section 2.10, word embeddings were previously struggling to differentiate between different senses of a word in context, a problem known as meaning conflation deficiency (Camacho-Collados and Pilehvar, 2020). This has been addressed, to some extent, in contextual embeddings.

Creating embeddings has been carried out at various levels such as word embeddings (Pennington et al., 2014), sense embeddings (Iacobacci et al., 2015), dictionary embeddings (Tissier et al., 2017), topic embeddings (Liu et al., 2015), context embeddings (Melamud et al., 2016), tweet embeddings (Vosoughi et al., 2016), meta-embeddings (Kiela et al., 2018) and sentence embeddings (Gao et al., 2021). Although these embeddings are intended to perform differently and more effectively based on the application, they follow the same principle of learning vector representations for words.

### *Datasets*

Despite the efforts in making machines understand words and sentences and capture meaning from documents, estimating semantic similarity is a challenging task. A solution would be to create datasets based on domain, application and language, initially for evaluation purposes but ultimately, for training deep learning models and fine-tuning pre-trained models. There are many datasets to train and evaluate semantic similarity for different tasks. One of the most important initiatives in this field is the SemEval semantic textual similarity (STS) shared tasks (Agirre et al., 2013, 2016b; Cer et al., 2017). STS shared tasks are organized based on a framework where a pair of sentences are scored (Baudiš et al., 2016). Many datasets[10] have been created for STS in English and based on various sources such as news, captions and forum. In the STS benchmark, the semantic similarity of independent pairs of texts, typically short sentences, is determined and the similarity is rated as a number between 0 to 5 to each pair denoting the level of similarity or entailment. In SemEval-2014 (Jurgens et al., 2014), the cross-level semantic similarity was targeted where the similarity is to be estimated at four levels as paragraph to sentence, sentence to phrase, phrase to word and word to sense. The similarity score is categorically rated as very similar, somewhat similar, somewhat related but not similar, slightly related and unrelated. Furthermore, there are other benchmarks focusing on specific techniques, such as the evaluation of distributional semantic models as in (Marelli et al., 2014; Hill et al., 2015), or specific linguistic properties as in Verb-3500 (Gerz et al., 2016).

In the same vein, there are a few datasets for evaluation of semantic similarity in monolingual setups such as for Russian (Panchenko et al., 2018), Finnish (Venekoski and Vankka, 2017) and Turkish (Ercan and Yıldız, 2018). This being said, there are not many datasets for lesser-resourced languages with significant coverage. In a multilingual context, Vulić et al. (2020a) introduce Multi-SimLex – a suite of 66 cross-lingual semantic similarity data sets of 12 languages. Barzegar et al. (2018) create datasets for 11 languages for semantic similarity and relatedness detection. In the context of lexicography, (Li et al., 2006) carries out a relevant study to create an evaluation dataset for semantic similarity using word definitions extracted from a dictionary and containing 65 noun pairs.

10 Available at http://ixa2.si.ehu.eus/stswiki/index.php/STSbenchmark

### 3.3.2 Language Resource Alignment

According to Gurevych et al. (2016), there are three main linking approaches applicable to language resources in general, and electronic lexicography in particular: ontology matching, schema matching and graph matching. As the basis of our analysis of the previous work, we focus on these approaches as well as follows:

**Ontology matching** which aims to find semantically-related entities across ontologies. As described in Section 2.8, many language resources have ontological features for which ontology matching can be beneficial. This has been previously the case for aligning WordNet (Lin and Sandkuhl, 2008). In monolingual ontology matching, entities in the source and target ontologies are compared by their labels which are in a single language. In multilingual ontology matching, this comparison is carried out between at least two languages. On the other hand, in cross-lingual ontology matching, the process is done by translating labels between languages (Fu et al., 2009). Previously, many approaches have been proposed, particularly in the context of the Ontology Alignment Evaluation Initiative[11]. The matching is carried out in two levels: terminological matching which focuses on the lexical comparison of ontologies and the semantic similarity of lexicalized data, and structural matching which takes the conceptual properties along with the structural information into account. SAMBO (Lambrix and Tan, 2006), OAANN (Huang et al., 2008) and Logmap (Jiménez-Ruiz and Cuenca Grau, 2011), are some of the many systems proposed for matching. Surveys on the recent advances in this field are provided by Shvaiko and Euzenat (2011); Khoudja et al. (2018); Ochieng and Kyanda (2018).

**Graph matching** refers to a set of techniques that rely on structural properties of the data where lexicalized data are considered as the nodes of a graph and the semantic relation between them are the edges. This way, a range of graph matching problems that have been historically of interest in mathematics, such as the PageRank algorithm (Xing and Ghorbani, 2004) and random walk (Lawler and Limic, 2010), are applicable to the problem. Furthermore, the current advances in representing graphs using neural networks, also known as graph neural networks, have paved the way for more robust systems that are currently the state-of-the-art (Ling et al., 2022).

**Database schema matching** which is similar to the previous two approaches in the sense that the structural properties of linguistic data as in relational databases are used for the matching problem. This approach is less of use given the emerging widespread usage of semantic technologies, like linked data, which are not dependent on relational data anymore.

11 http://oaei.ontologymatching.org

It should be noted that multilingualism is an aspect that can exist in all the approaches. Furthermore, the usage of a particular method for dictionary alignment depends on the heterogeneity of the data caused by the structure or data model that the two resources employ. For instance, aligning two dictionaries, one in Ontolex-Lemon and the other in LMF, represents challenges different from aligning both of them in the same form. As such, there are many hybrid techniques for the same problem.

Table 3.1 summarizes some of the previous contributions in resource alignment based on approach, level (monolingual, cross-lingual), type of resource as described in Chapter 2 and type of linking (structural, conceptual, lexical). Regarding approaches, "formalism" refers to any symbolic or axiomatic solution for matching resources that has been introduced in the literature.

| Paper | Approach | Level | Type of resource | Type of linking |
|---|---|---|---|---|
| (Caselli et al., 2014) | semantic similarity | monolingual | lexical | lexical |
| (Bennett and Fellbaum, 2006; Kwong, 1998) | formalisms | monolingual | lexical | ontological |
| (Niles and Pease, 2003) | formalisms | monolingual | lexical | ontological |
| (Sánchez-Rada and Iglesias, 2016) | formalisms | monolingual | ontological | structural |
| (Diosan et al., 2008) | machine learning | monolingual | lexical | lexical |
| (Subirats and Sato, 2004) | corpus | monolingual | lexical | lexical |
| (Bond and Foster, 2013) | formalisms | multilingual | lexical | structural |
| (Caracciolo et al., 2012; Cimiano et al., 2020b; Moussallem et al., 2018) | string similarity | multilingual | lexical | ontological |
| (Lesnikova, 2013; Lesnikova et al., 2016) | machine translation | multilingual | lexical | structural |
| (Gracia, 2015; Gracia et al., 2018) | formalisms | multilingual | lexical | structural |
| (Spohr et al., 2011) | machine learning | multilingual | ontological | structural |
| (Damova et al., 2013) | formalisms | multilingual | lexical/ontological | structural |
| (Charles et al., 2018) | formalisms | multilingual | knowledge graph | ontological |
| (Biemann et al., 2018) | distributional semantics | multilingual | lexical | lexical/ontological |
| (Chen et al., 2016c) | graph neural networks | cross-lingual | knowledge graph | structural |
| (Schuster et al., 2019) | mapping of word spaces | cross-lingual | contextual embeddings | lexical |

Table 3.1: A summary of the previous studies in aligning resources

### 3.3.3 Translation Inference

One particular task that can facilitate the usage of dictionaries across languages and further add to the content value is translation inference, also known as bilingual lexicon induction. In this section, we give an overview of methods to acquire multilingual lexicons.

The translation inference task aims to generate a new dictionary by inducing new translations from the existing ones. For instance, the translation pair '*pomme*' (French) $\rightarrow$ '*sêw*' (Kurdish) can be induced from '*pomme*' (French) $\rightarrow$ 'apple', 'apple' $\rightarrow$ '*sêw*' (Kurdish). In this context, the Translation Inference Across Dictionaries (TIAD) shared task aims to incentivize researchers to propose new techniques and approaches to translation inference in an unsupervised way. Table 3.2 summarizes the previously proposed techniques in the TIAD shared tasks.

Among the previous techniques, pivot-based (Torregrosa et al., 2019), cycle-based (Donandt et al., 2017b) and one time inverse consultation approaches (Lanau-Coronas and Gracia, 2020) are applied. On the other hand, external resources are used in an unsupervised way to train multi-way machine translation models and cross-lingual word embedding mappings. Although these techniques align translations without being trained on parallel corpora, they face challenges in retrieving part-of-speech tags and lemmatizing various word forms (Arcan et al., 2019a).

Tanaka and Umemura (1994b) generate a bilingual dictionary using the structure of the source dictionaries. They introduced the inverse consultation approach which measures the semantic distance between two words based on the number of their common words in the pivot language. Using this method, Schafer and Yarowsky (2002) created an English-Gujarati lexicon using Hindi as the pivot. Similarly, Tsunakawa et al. (2009) used English as an intermediate language to create a Chinese-Japanese lexicon. The IC method was extended by taking more lexical and semantic information into account Kaji and Aizono (1996). For instance, Bond and Ogura (2008) used part-of-speech information and semantic classes to produce a Japanese-Malay dictionary with English as the pivot. Sjöbergh (2005) created a Japanese-Swedish dictionary by linking words based on the sense definitions, whereas Kaji et al. (2008) constructed a Japanese-Chinese dictionary using a pivot English lexicon and co-occurrences information for more accurate alignment.

The high dependency of the inverse consultation method on one language as a pivot has been shown to create limited translations with ambiguity and low recall Shezaf and Rappoport (2010); Saralegi et al. (2011). One way to remedy this is to use multiple pivot languages with additional resources. Paik et al. (2001) generated a Korean-Japanese dictionary using English and Chinese pivot languages and an English thesaurus. Soderland et al. (2009) described the automatic generation of a multilingual resource, called PanDictionary. In this work, the authors used probabilistic inference over the translation graph. The construction of the dictionary consisted of extracting knowledge from existing dictionaries and combining the obtained knowledge into a single resource. István and Shoichi (2009) took advantage of the semantic structure of WordNet as the pivot language for creating a new lexicon for less-resourced languages. Mann and Yarowsky (2001) used string distance to create bilingual lexicons based on transduction models of cognates, as languages belonging to a specific language family usually share many cognates.

Furthermore, Paik et al. (2004) examines the impact of the translation direction in merging dictionaries. Mausam et al. (2009) describe the automatic generation of a multilingual resource, called PanDictionary. In the proposed work, the authors use probabilistic inference over the translation graph. The construction of the dictionary consists of large-scale information extraction over the Web, i.e. extracting knowledge from existing dictionaries, and combining the obtained knowledge into a single resource. The final step consists of performing automated reasoning over the graph. Tanaka and Umemura (1994a) introduce a bilingual dictionary generation using the Princeton WordNet of the pivot language to build a new bilingual dictionary. Instead of focusing on the lexical overlap, the authors calculate the semantic lexical overlap of the source-to-pivot and target-to-pivot translations.

| Year | Target dictionaries | Paper | Approach | External resources |
|---|---|---|---|---|
| 2017 | German-Portuguese Danish-Spanish Dutch-French | (Alper, 2017b) | graph analysis | - |
| | | (Proisl et al., 2017b) | graph analysis and collocation-based models | Europarl corpus |
| | | (Donandt et al., 2017b) | Support Vector Machine using features based on the translation graph and string similarity | - |
| 2019 | English French Portuguese | (Arcan et al., 2019a) | multi-way neural machine translation | corpora of languages from the same family and Wiktionary |
| | | (Torregrosa et al., 2019) | graph analysis and neural machine translation | Directorate General for Translation corpus (Steinberger et al., 2013) |
| | | (Garcia et al., 2019) | pivot-based and cross-lingual word embeddings | monolingual corpora |
| | | (Donandt and Chiarcos, 2019) | multi-lingual word embedding | pretrained embedding model |
| | | (McCrae, 2019) | unsupervised document embedding using Orthonormal Explicit Topic Analysis | Wikipedia corpora |
| 2020 | English French Portuguese | (McCrae and Arcan, 2020) | unsupervised multi-way neural machine translation and unsupervised document embedding | Directorate General for Translation corpus (Steinberger et al., 2013) |
| | | (Chiarcos et al., 2020) | propagation of concepts over a graph of interconnected dictionaries using WordNet synsets and lexical entries as concepts | WordNet |
| | | (Lanau-Coronas and Gracia, 2020) | graph analysis and cross-lingual word embeddings | monolingual corpora of Common Crawl and Wikipedia |
| | | (Dranca, 2020) | graph analysis relying on paths, synonyms, similarities and cardinality in the translation graph | - |

**Table 3.2:** An overview of the approaches proposed in the previous TIAD shared tasks

## 3.4 PUBLICATION AND STORAGE

Given the increasing number of language and linguistic resources, providing sustainable mechanisms for access and discovery of such resources is a challenging task. Once a resource is released many issues are to be tackled, particularly sustainability, both technical and organizational, availability and findability, selection and qualification for long-term archiving, and legal issues (Georg et al., 2010). These factors have led to many initiatives to create catalogs, platforms, portals, and various linguistic resource metadata for linguistic data management.

Although metadata schemata may be represented with many common features, such as name, language and license of the resource, they are usually different in coverage, features such as labels and datatypes, approach as collaborative or centralized curation, format, e.g. XML or RDF, and standards. Recent schemata further emphasize the findable, accessible, interoperable and reusable factors as well.

One of the well-known sources of publishing and maintaining language resources is META-SHARE (Piperidis, 2012). META-SHARE is a platform to document repositories of language data and tools from the production stage up to their usage. The schema used in this network is implemented in XML and XML Schema Definition. In addition to this, there are many other services for the same objective, such as SPLICR (Rehm et al., 2008), the European Language Resources Association[12], the Linguistic Data Consortium[13], the Language Grid[14], Open Language Archives Community[15] and CLARIN Virtual Language Observatory[16]. Moreover, there are several others which are created by crowd-sourcing and community efforts, such as LRE Map[17], EUDAT[18], OpenAire[19], PHOIBLE (Moran and McCloy, 2019) and Datahub[20]. Metadata has received fair attention in the semantic web fields, as well. Notable examples are the metadata vocabularies such as the DCAT vocabularyMaali et al. (2014) and the LInguistic MEtadata module of Ontolex-Lemon–*lime* (Fiorelli et al., 2015).

One of the main problems in dealing with these repositories and catalogs is the lack of harmonization among them (Cimiano et al., 2020a). To this end, there have been many initiatives such as the ISOcat data category registry (ISOcat DCR) (Kemps-Snijders et al., 2008), the General Ontology of Linguistic Description (Farrar and Langendoen, 2010) and Ontologies of Linguistic Annotation–OLiA (Chiarcos and Sukhareva, 2015). Motivated to ensure more qualitative metadata and increasing homogeneity, McCrae et al. (2015) develop an ontology in OWL to represent the

12 www.elra.info
13 https://www.ldc.upenn.edu
14 https://langrid.org
15 http://www.language-archives.org
16 https://www.clarin.eu/content/virtual-language-observatory-vlo
17 http://lremap.elra.info
18 https://www.eudat.eu
19 https://www.openaire.eu
20 https://datahub.io

metadata schemes. The data model is also available through a platform called Yuzu[21] (McCrae, 2016) which enables running SPARQL queries on the aggregated datasets. Similarly, the data model is represented in the Linguistic Metadata Hub – Linghub[22] (McCrae and Cimiano, 2015) and the IULA LOD catalogue[23] as well.

## 3.5 DICTIONARIES IN NLP APPLICATIONS

Dictionaries have been historically an essential part of NLP applications such as spell checkers and morphological analyzers (Ahmadi, 2021), word sense disambiguation (Krovetz and Croft, 1989), part-of-speech tagging (Täckström et al., 2013), syntactic analysis (Gross, 1984) and co-reference resolution (Sleeman and Finin, 2013).

In this section, some of the most relevant tasks to the topic of the thesis are briefly discussed. These tasks facilitate the alignment tasks in one way or another.

### 3.5.1 Word Sense Disambiguation

Word Sense Disambiguation (WSD) is the task of identifying the meaning of a word in the context. For instance, in the bank model described in Section 2.2.2, WSD detects 'bank' in the sentence 'paid into a bank account' as a financial institution while 'bank' in 'the bank of the river Corrib' refers to a riverbank. This task has been extensively studied with many approaches ranging from knowledge bases (Scozzafava et al., 2020; Blevins and Zettlemoyer, 2020) to unsupervised methods (Navigli and Lapata, 2009; Raganato et al., 2017; Ustalov et al., 2018), with the former methods or hybrid ones outperforming the latter methods. Bevilacqua et al. (2021) provide a recent survey on the current trends of this historical task in NLP.

WSD is related to WSA as both of them focus on identifying the meaning of polysemous words; however, the reliance of WSD on the context makes it quite different from WSA, as defined in this thesis, which only takes lexical and structural properties of words in language resources into account, regardless of the context in which the word appears. This being said, merging WSA and WSD by incorporating word sense contexts based on corpora in the alignment task is a compelling task that can be addressed.

### 3.5.2 Semantic Role Labeling

Semantic role labeling aims to identify and label arguments and role-bearing constituents in a text. The primary goal of this task is to find the semantic relations

21 https://github.com/jmccrae/yuzu
22 https://github.com/liderproject/linghub
23 http://lod.iula.upf.edu

between a predicate and its associated participants and properties (Màrquez et al., 2008). Such relations are predefined based on the predicates, as in 'whisper' in the sentence 'the girl whispered to the boy' where 'the girl' is the agent and 'the boy' is the recipient.

Semantic role labeling is important to the alignment task as identifying semantic roles in sense definitions could be beneficial to determine if two senses should be linked. Employing semantic role labeling, however, has not received much attention for WSA. One notable example is proposed by (Silva et al., 2016) where a rule-based framework is presented to extract semantic roles from definitions relying on syntactic and semantic features. These roles are to be labeled based on a genus-differentia pattern which is further defined at a finer level as shown in Figure 3.3.

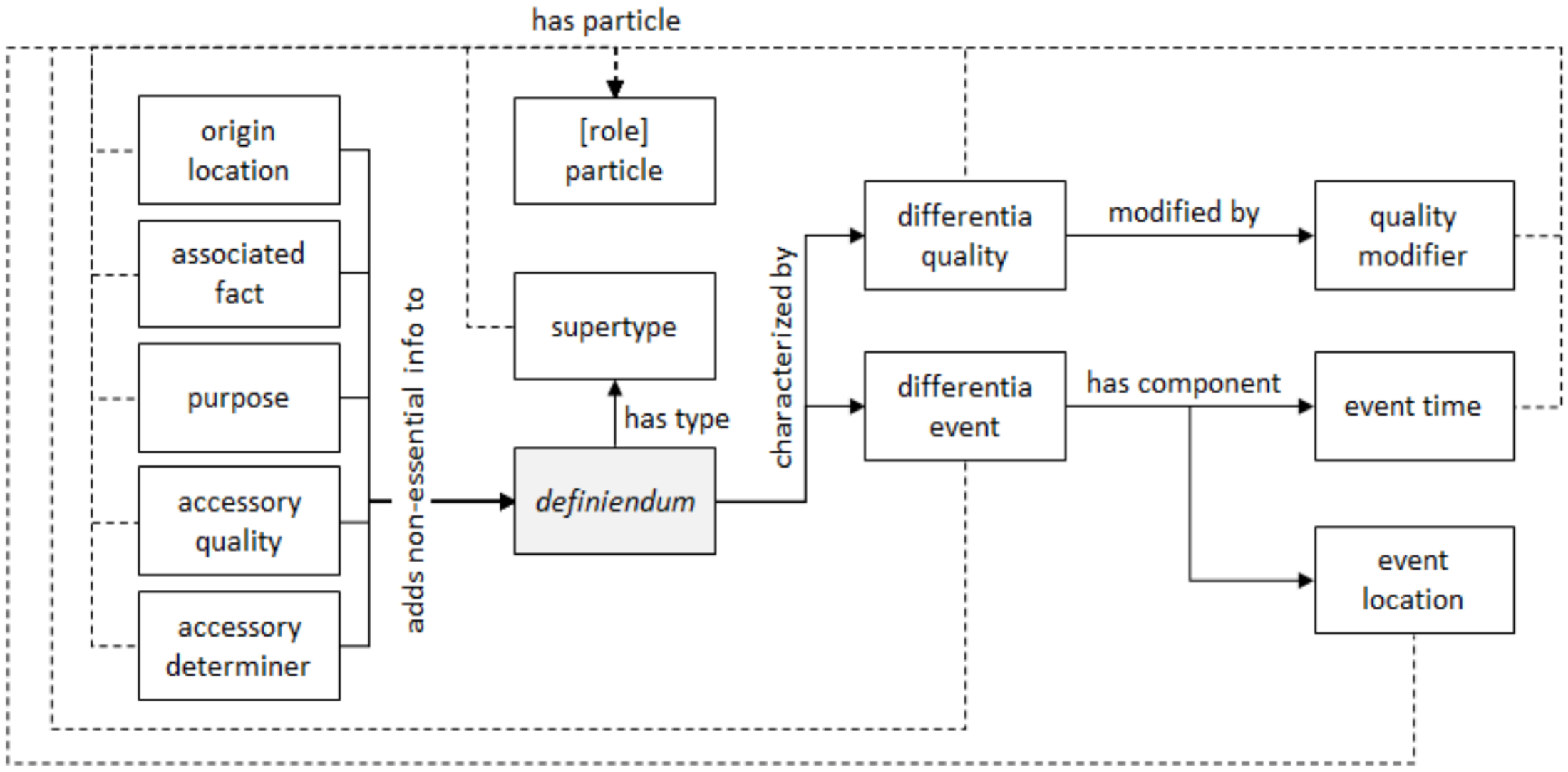

Figure 3.3: A conceptual model proposed by Silva et al. (2016) to extract semantic roles from sense definitions. Dashes lines refer to relationships between particles, such as a phrasal verb complement, in the definition.

### 3.5.3 Reverse Dictionary

Another interesting avenue of research is *reverse dictionary* or *concept finder* which given a definition, an associated concept is returned (Zock and Bilac, 2004). In an attempt to fill the gap of representation of lexical and phrasal semantics, Hill et al. (2016) propose a neural language model technique to map between arbitrary-length phrases and fixed-length continuous-valued word vectors using word definitions. This task is of interest, particularly as it provides a mechanism to explain the behavior and the type of information that a neural model would learn in the embedding process. In other words, how a sense can be embedded differently or similarly through the word and the definition knowing that both the word and the definition are conceptually and semantically referring to the same meaning.

## 3.6 WHAT IS MISSING?

In this chapter, some of the most related tasks to word sense alignment are presented. We started with an introduction into the life cycle of language resources at a broad scope as follows: modeling, creation, enrichment and publication. Even though the modeling based on which a resource is created may remain the same, a resource can undergo various changes and be published in different versions. Given the focus of the thesis, it made sense to focus on the enrichment step more than the other ones. To this end, the most relevant tasks to word sense alignment are briefly discussed. These tasks are semantic similarity detection, resource alignment and semantic resource induction. Finally, the chapter ends with a summary of the applications of LSRs in NLP such as word sense disambiguation and translation inference.

As the survey indicates, there are two major limitations in the literature: (i) there is no multilingual benchmark for the evaluation for word sense alignment, (ii) despite the previous tasks in language resource alignment, there is not much focus on lexicographical data specifically. These hinder the progress in exploring, integrating and exploitation of dictionaries in language technology, linked data and various applications in NLP.

# 4 LEVERAGING THE GRAPH STRUCTURE OF LEXICOGRAPHICAL RESOURCES

## 4.1 INTRODUCTION

As described in the previous chapters, some of the LSRs represent lexical items in a structural and conceptualized way. This structure can be leveraged for many tasks, particularly word sense alignment, by using graph-based methods. Employing graphs has been of interest to many tasks in NLP (Nastase et al., 2015), such as fake news detection (Hamid et al., 2020), document summarization (Wang et al., 2020), entity relation extraction (Fei et al., 2021) and named-entity recognition (Gui et al., 2019).

In this chapter, we shed light on the structure of lexicographical resources as a network where lexical items are represented as nodes and the relation between words or definitions are represented as edges. Such relations can denote translations, as in bilingual dictionaries, semantic relations such as synonymy and antonymy or alignment of senses across resources. In the same vein, we provide a few techniques to leverage the graph structure of resources. To do so, a brief overview of graph-based approaches in the alignment of lexicographical resources is provided in Section 4.2. In Section 4.3, we evaluate lexicographical networks of bilingual dictionaries based on a few graph measuring metrics, such as average degree, density and clustering coefficient. Following this, in Section 4.4, we introduce an optimization algorithm, called the weighted bipartite b-matching, for alignment scenarios to model restrictions over the number of links that can be established among different nodes, i.e. entries. This algorithm shows that assigning weights to an alignment problem can improve the performance of a linking system, despite its complexity in tuning the optimal parameters. We also introduced a few string-based similarity measures to estimate the similarity of words or definitions.

Moreover, we explore the usage of graphs in translation inference in Section 4.5. Relying on the paths and cycles between a source word in a language and a target word in another language, we show that the graph structure can be explored to generate new translation pairs for two languages for which a bilingual dictionary is not available. This task is of also of interest to create new bilingual dictionaries in an unsupervised way for under-resourced languages.

At the end of the chapter, we discuss some of the limitations of graph-based methods, particularly those due to missing information in the data that requires exploring and incorporating external resources to improve the performance of linking. Our analysis based on the limitations motivates the curation of a standard benchmark for evaluating the tasks of word sense alignment which will be discussed in the following section.

## 4.2 RELATED WORK

Representing and formalizing language and linguistic data as a graph has been a compelling and ubiquitous idea for a long time. Graphs represent mechanisms to extract further information regarding a specific node based on topological properties. Some of the graph analysis techniques applied to NLP tasks are random walk (Ethayarajh, 2018; Arora et al., 2016; Hughes and Ramage, 2007), PageRank (Engström, 2016; Pershina et al., 2015) and label propagation (Speriosu et al., 2011; Lin et al., 2013). Thanks to the advances in neural networks, graph neural networks have been demonstrated to effectively capture structural information and therefore, achieve state-of-the-art results in many NLP tasks (Schlichtkrull et al., 2020; Wang et al., 2020). Surveys on NLP with graph-based techniques and graph neural networks are respectively provided in Nastase et al. (2015) and Wu et al. (2021).

Graph-based approaches have been widely used for the WSA task. Matuschek and Gurevych (2013) propose a graph-based approach, called Dijkstra-WSA, for aligning lexical semantic resources, namely WordNet, OmegaWiki, Wiktionary and Wikipedia. In this approach, senses are represented as the nodes of a graph where the edges represent the semantic relation between them. Assuming that monosemous lemmata have a more specific meaning and are therefore less ambiguous to match, a semantic relation is created among the senses of such lemmata when they appear in a sense of a polysemous lemma. Using Dijkstra's shortest path algorithm along with semantic similarity scores and without requiring any external data or corpora, a set of possible sense matches are retrieved.

More related to the topic of this thesis, graph analysis has been addressed as possible solutions for the WSA task (Matuschek, 2015, p. 67) and translation inference. In the translation inference task, graph analysis techniques rely on the analysis of translation graphs to determine a possible connection between two words. Over the past few years, there has been an increasing usage of graph-based algorithms such as random walk and graph sampling techniques for multilingual dictionary generation (Andrieu et al., 2003; Villegas et al., 2016). Proisl et al. (2017a) proposed a system for generating translation candidates using a graph-based approach with a weighting scheme and a collocation-based model based on the parallel corpus Europarl. In contrast, Alper (2017a) focused on finding cycles of translations in the graph. By find-ing cycles of translations in the graph of all lexical entries with translations treated as undirected edges, the proposed approach was able to infer translations with reasonable accuracy. Donandt et al. (2017a) used supervised machine learning to predict high-quality candidate translation pairs. They trained a support vector machine (SVM) for classifying valid or invalid translation candidates. For this, they used several features, e.g. frequency of source word in a dictionary or minimum/maximum path length. Furthermore, string similarity leveraging, i.e. Levenshtein distance, was also taken into consideration.

In addition to using textual similarity methods, a number of non-textual methods can be used that are useful for linking dictionaries, introduced by McCrae and Cillessen (2021). This method is graph-based similarity, which relies on there being a graph relating the senses of an entry and so is primarily used in the case of WordNet linking. NAISC implements the FastPPR method (Lofgren et al., 2014) to find graph similarity.

In the case of WordNet linking, graph similarity cannot be naively applied as there are not generally links between the graphs of the two WordNets, instead, it is possible to rely on the *hapax legomenon* links, i.e. links that are created when there is only one sense for the lemma in both dictionaries. These links allow us to create a graph between the two graphs as shown in Figure 4.1.

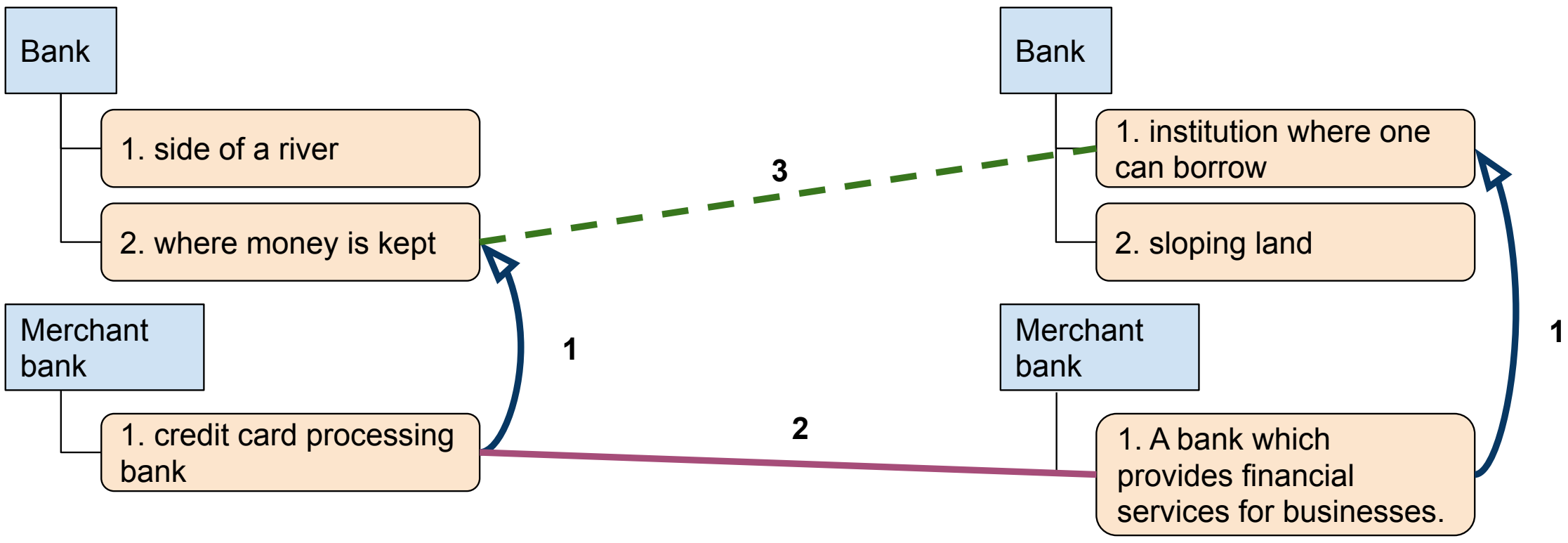

**Figure 4.1:** An example of the use of non-textual features for linking. Here the two senses of bank are distinguished by the hypernym links (1) and an inferred *hapax legomenon* link (2), so that the correct sense (3) can be selected.

McCrae and Cillessen (2021) explored this method in the context of linking English WordNet (McCrae et al., 2019a) with Wikidata[1], where the NAISC system is used to find equivalent senses of WordNet synsets and entities in the Wikidata database. This study suggest that 67,569 (55.3%) of WordNet's synsets have a matching lemma in Wikidata, of which 16,452 (19.5%) counted as *hapax legomenon* links. Therefore, the accuracy of the *hapax legomenon* links was directly evaluated and consequently, it was found that accuracy, when applying some simple filters, was 96.1% based on an evaluation of two annotators, who had a Cohen's kappa agreement of 81.4%.

1 https://www.wikidata.org

On the other hand, using the non-textual methods along with simple textual methods, similar to the ones described in Section 6.4, the Naisc system could achieve an accuracy of 65-66% in predicting links between WordNet and Wikidata. Divided by the prediction scores, those links predicted with a confidence of less than 60% by the system were all incorrect (0.0% accuracy), those with a 60-80% accuracy were correct 23/39 times (59.0% accuracy) and those with a greater than 80% confidence were correct 42/49 times (85.7% accuracy), indicating that the system's confidence was a good predictor of the accuracy of links. Given the effectiveness of such non-textual linking methods in a few kinds of dictionary linking tasks, especially with large-scale knowledge graphs such as Wikidata, they are also included in Naisc.

## 4.3 LEXICOGRAPHICAL NETWORK

In this section, we analyze lexicographical networks based on basic graph notions. We define a *lexicographical network* as a network of two disjoint sets of vocabulary which are interconnected based on a sense relation. Analyzing the structure of such networks provide further information that may be of help in using alignment algorithms based on link prediction methods. Figure 4.2 illustrates a set of entries of a bilingual English-French dictionary and their lexicographical network schema.

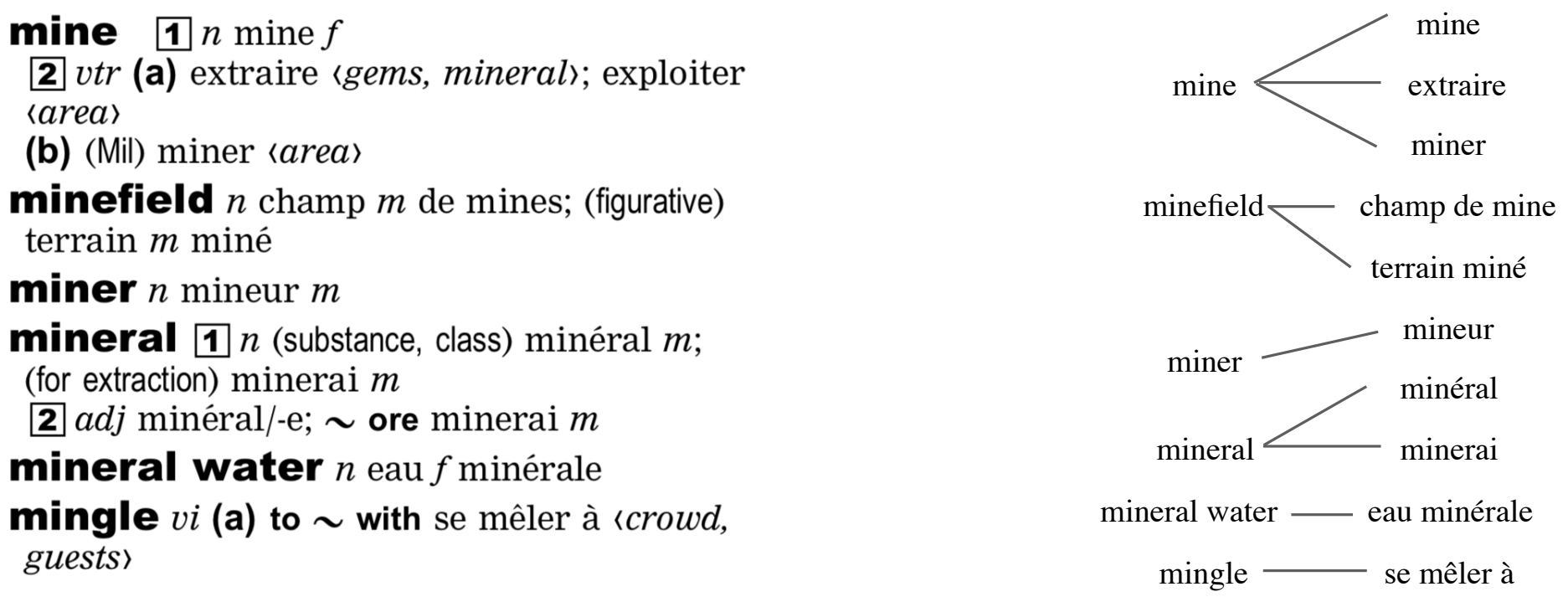

Figure 4.2: A set of dictionary entries (left) and the equivalent lexicographical network (right).

(Gurevych et al., 2016) categorizes linking approaches applicable to lexical data as ontology alignment, schema matching and graph matching. Despite the dependency of the two former ones on lexical semantic similarity, graph matching also relies on structural properties. Therefore, the analysis of lexicographical networks elucidates the employment of such methods.

### 4.3.1 Analysis of lexicographical networks

We assume that a graph $G = ((U, V), E, W)$ is unweighted, undirected, and bipartite. In other words, $U$ and $V$ are disjoint sets of vertices and the edge set $E \subseteq U \times V$ contains only edges between vertices in $U$ (source entries) and vertices in $V$ (target entries)[2]. We use similar notions to Latapy et al. (2008) to define basic bipartite statistics.

Given the bipartite graph $G$, we denote the number of right and left nodes by $n_U = |U|$ and $n_V = |V|$. We also denote the number of links in the graph by $m = |E|$. The average degree of each set of vertices is defined as $k_U = \frac{m}{n_U}$ and $k_V = \frac{m}{n_V}$. Therefore, the average degree of the whole graph $G' = (U \cup V, E, W)$ can be calculated as $k = \frac{2m}{n_U + n_V} = \frac{n_U \times k_U + n_V \times k_V}{n_U + n_V}$. Finally, we define the number of existing links divided by the number of possible links as the bipartite density $\delta(G) = \frac{m}{n_U \times n_V}$.

In order to capture a notion of overlap, we also define *clustering coefficient* which measures the probability that two nodes are linked based on the common neighbors. (Borgatti and Everett, 1997) define the clustering coefficient in bipartite graphs as the following:

$$cc(u) = \frac{\sum_{v \in N(N(u))} cc(u, v)}{|N(N(u))|} \tag{4.1}$$

where $cc(u, v)$ measures the overlap between neighbourhoods of $u$ and $v$ and $N(u)$ refers to the neighbours of $u$. If there is no common neighbours between $u$ and $v$, then $cc(u, v) = 0$. If they have the same common neighbours, $cc(u, v) = 1$. Therefore, cc(u, v) is defined as:

$$cc(u, v) = \frac{N(u) \cap N(v)}{N(u) \cup N(v)} \tag{4.2}$$

Finally, we define the average clustering coefficient in $U$ (or in $V$) as the average of $cc(u)$ (or $cc(v)$) over the whole number of nodes:

$$cc(U) = \frac{\sum_{u \in U} cc(u)}{|U|} \tag{4.3}$$

### 4.3.2 Experiments

We analyze the lexicographical network of the 10 largest multilingual dictionaries freely-accessible on FreeDict[3], Apertium dictionaries[4], English WordNet-Wiktionary mapping dataset (McCrae, 2018) and Matuschek and Gurevych (2014)'s English Wik-

2 This assumption may not be always correct as in a real-world dictionary an entry can refer to another entry in the same set, for instance, using *see* or *cf.* keywords.
3 https://freedict.org/
4 Data of the 1st edition of 2018

tionary and Wikipedia sense-aligned dataset. The evaluation results of each network are shown in Table 4.1 and Table 4.2.

Regarding Table 4.1, although the sizes of the dictionaries are not identical, their feature values seem to be uniformly varying in a specific range. The average degree $k$ changes in the range of [1, 2] indicating one-to-many relations between source entries and target entries. A higher degree in each side of the network, i.e., $k_U$ and $k_V$, shows a higher number of edges connected to the nodes. Norwegian Nynorsk-Norwegian Bokmål and Dutch-English present the lowest and the highest average degrees respectively. This range of degree is expected as in a dictionary, entries are mostly linked to other words. In most of the cases, there is a remarkable difference between the clustering coefficients of $U$ and $V$. $cc_U$ tending to zero suggests the scarcity of entries with common neighbors in $U$. On the other hand, the clustering coefficient in $V$, $cc_V$, indicates a higher number of common neighbors. This metric is particularly interesting as it may be used as a heuristic in link discovery algorithms.

On the other hand, Table 4.2 provides the results of the same metrics on a different set of dictionaries and datasets. Except in a few datasets such as English-Spanish (EN-ES) and English-Galician (EN-GL), there is no significant number of polysemous items. This indicates that there is a one-to-one relation between almost all dictionary entries. More interestingly, the bipartite density, i.e. $\delta$, varies across the datasets showing that some datasets are providing lexical entries less or more richly linked. For instance, McCrae (2018)'s WordNet-Wikipedia dataset (WN-WP) provides a denser network in comparison to Matuschek and Gurevych (2014)'s WordNet-Wiktionary dataset (WN-WKN) which contains less links.

| Language pairs | $n_U$ | $n_V$ | $m$ | $k_U$ | $k_V$ | $k$ | $\delta$ | $cc_U$ | $cc_V$ |
|---|---|---|---|---|---|---|---|---|---|
| German-English | 81540 | 92982 | 123490 | 1.51 | 1.32 | 1.41 | 1.62e-05 | 2.86e-23 | 0.0046 |
| English-Arabic | 87424 | 56410 | 89028 | 1.01 | 1.57 | 1.23 | 1.80e-05 | 0.0 | 0.0001 |
| Dutch-English | 22747 | 15424 | 45151 | 1.98 | 2.92 | 2.36 | 1.28e-4 | 7.57e-14 | 0.2694 |
| Kurdish-German | 10562 | 6374 | 10562 | 1.0 | 1.65 | 1.24 | 1.56e-4 | 0.0 | 0.0012 |
| English-Hindi | 22907 | 49534 | 55635 | 2.42 | 1.12 | 1.53 | 4.90e-05 | 2.09e-20 | 0.0001 |
| Japanese-French | 13233 | 17869 | 27692 | 2.09 | 1.54 | 1.78 | 1.17e-4 | 0.0 | 0.0 |
| Breton-French | 23109 | 29141 | 42730 | 1.84 | 1.46 | 1.63 | 6.34e-05 | 6.44e-29 | 0.0168 |
| Hungarian-English | 139935 | 89679 | 254734 | 1.82 | 2.84 | 2.21 | 2.02e-05 | 1.54e-78 | 0.0143 |
| Icelandic-English | 8416 | 6405 | 8416 | 1.0 | 1.31 | 1.13 | 1.56e-4 | 1.32e-05 | 0.0344 |
| Norwegian Nynorsk -Norwegian Bokmål | 63509 | 62103 | 63509 | 1.0 | 1.02 | 1.01 | 1.61e-05 | 7.87e-06 | 0.9559 |

**Table 4.1:** Evaluation of lexicographical networks based on basic graph notions

| Dataset | $n_U$ | $n_V$ | $m$ | $k_U$ | $k_V$ | $k$ | $\delta$ |
|---|---|---|---|---|---|---|---|
| OC-ES | 14566 | 14566 | 14566 | 1.0 | 1.0 | 1.0 | 6.87e-05 |
| ES-RO | 17320 | 17320 | 17320 | 1.0 | 1.0 | 1.0 | 5.77e-05 |
| ES-GL | 8990 | 8990 | 8990 | 1.0 | 1.0 | 1.0 | 10.11e-5 |
| EO-EN | 31482 | 31483 | 31485 | $\approx$ 1.0 | $\approx$ 1.0 | $\approx$ 1.0 | 3.17e-05 |
| PT-GL | 10148 | 10148 | 10148 | 1.0 | 1.0 | 1.0 | 9.85e-05 |
| FR-ES | 21478 | 21478 | 21478 | 1.0 | 1.0 | 1.0 | 4.65e-05 |
| ES-PT | 12060 | 12059 | 12060 | 1.0 | $\approx$ 1.0 | $\approx$ 1.0 | 8.29e-05 |
| EU-ES | 11883 | 11882 | 11883 | 1.0 | $\approx$ 1.0 | $\approx$ 1.0 | 8.41e-05 |
| ES-AST | 36111 | 36113 | 36114 | $\approx$ 1.0 | $\approx$ 1.0 | $\approx$ 1.0 | 2.76e-05 |
| EN-CA | 31861 | 30538 | 34879 | $\approx$ 1.1 | $\approx$ 1.1 | $\approx$ 1.1 | 3.58e-05 |
| OC-CA | 15985 | 15985 | 15985 | 1.0 | 1.0 | 1.0 | 6.25e-05 |
| EO-FR | 35799 | 35803 | 35803 | $\approx$ 1.0 | 1.0 | $\approx$ 1.0 | 2.79e-05 |
| CA-IT | 7871 | 7875 | 7875 | $\approx$ 1.0 | 1.0 | $\approx$ 1.0 | 10.27e-5 |
| EO-CA | 19967 | 19966 | 19967 | 1.0 | $\approx$ 1.0 | $\approx$ 1.0 | 5.00e-05 |
| ES-AN | 3111 | 3111 | 3111 | 1.0 | 1.0 | 1.0 | 30.21e-5 |
| ES-CA | 29991 | 29821 | 32918 | $\approx$ 1.1 | $\approx$ 1.1 | $\approx$ 1.1 | 3.68e-05 |
| FR-CA | 6550 | 6550 | 6550 | 1.0 | 1.0 | 1.0 | 10.52e-5 |
| PT-CA | 7117 | 7117 | 7117 | 1.0 | 1.0 | 1.0 | 10.40e-5 |
| EU-EN | 13231 | 13230 | 13231 | 1.0 | $\approx$ 1.0 | $\approx$ 1.0 | 7.55e-05 |
| EO-ES | 17230 | 17232 | 17232 | $\approx$ 1.0 | 1.0 | $\approx$ 1.0 | 5.80e-05 |
| EN-GL | 17913 | 18273 | 21863 | $\approx$ 1.2 | $\approx$ 1.1 | $\approx$ 1.2 | 6.67e-05 |
| EN-ES | 23900 | 23789 | 27220 | $\approx$ 1.1 | $\approx$ 1.1 | $\approx$ 1.1 | 4.78e-05 |
| WN-WKN (Matuschek and Gurevych, 2014) | 63851 | 63708 | 72747 | 1.14 | 1.14 | 1.14 | 1.8e-05 |
| WN-WP (McCrae, 2018) | 7667 | 7523 | 7687 | $\approx$1.0 | 1.02 | 1.01 | 10.33e-5 |

**Table 4.2:** Basic statistics of sense-aligned resources. The first rows describe the Apertium dictionaries (2018) with ISO 639-1 language codes.

## 4.4 WEIGHTED BIPARTITE b-MATCHING

Considering the task of word-sense alignment, linking is a task that cannot only be achieved by looking at pairs of definitions by themselves but instead a holistic approach looks at all the links being generated and considers whether this leads to a good overall linking. It is clear that mapping multiple senses to the same senses or generating many more or fewer links than the number of senses is not ideal. Therefore, we look at a method for solving the problem of sense linking holistically. To do so, we present a similarity-based approach for WSA in English WordNet and Wiktionary with a focus on the polysemous items. Our approach, as illustrated in Figure 4.3, relies on semantic and textual similarity using features such as string distance metrics and word embeddings, and a graph matching algorithm. Transforming the alignment problem into a bipartite graph matching enables us to apply graph matching algorithms, in particular, weighted bipartite b-matching (WBbM).

Two-mode networks contain two sets of units, such as authors and papers, which are connected based on their relation, for instance, citation. Such networks have been already analyzed for various applications, such as actors-movies network (Newman et al., 2001), authoring networks (Newman, 2001), occurrence networks (i Cancho and Solé, 2001) and peer-to-peer exchange networks (Le Fessant et al., 2004). As another application, we are interested in applying the WBbM algorithm to the two-mode net-work of word-senses in two different resources. WBbM is one of the widely studied classical problems in combinatorial optimization for modeling data management applications, e-commerce and resource allocation systems (Ahmed et al., 2017; Chen et al., 2016a,b).

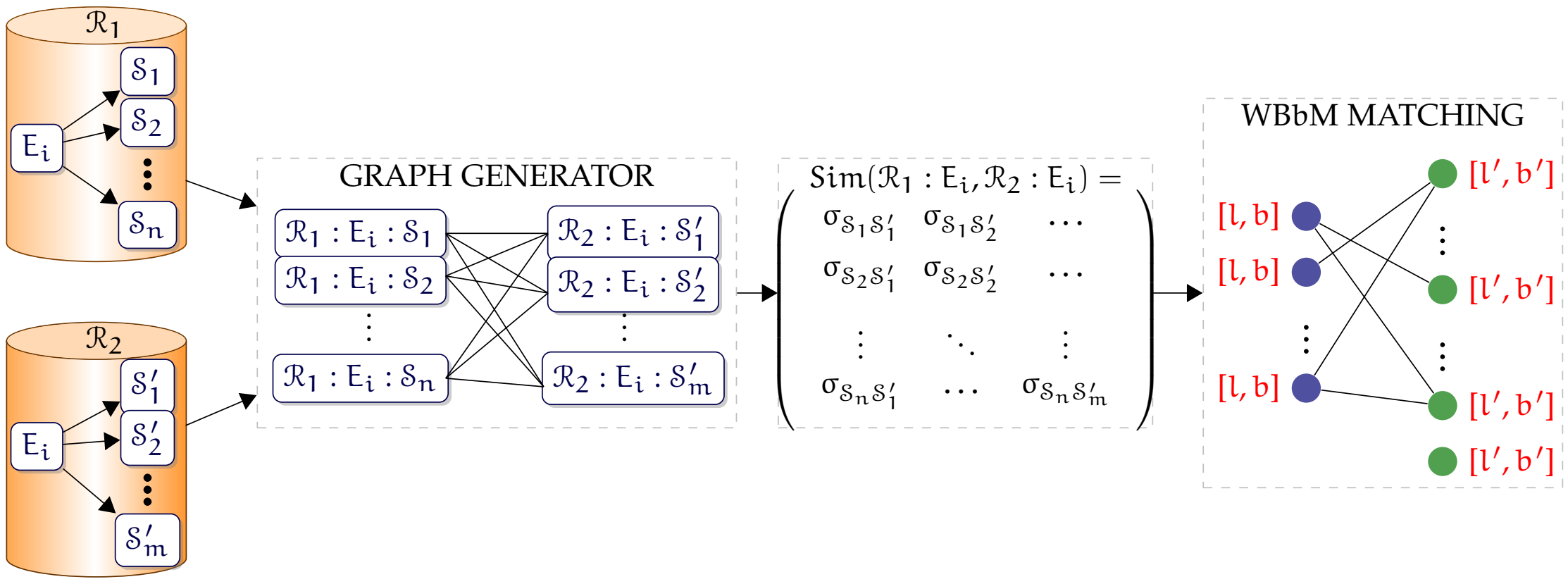

**Figure 4.3:** Schema of the sense alignment system with a focus on the graph matching component. It should be noted that in resources $R_1$ and $R_2$, $E_i$ refers to entry $i$ with senses shown as $S$ and $S'$. $\sigma$ denotes the similarity function.

### 4.4.1 String-based Methods

In order to estimate the similarity of two words or definitions, i.e. the function $sim$ in Figure 4.3, we use string-based measures. Such measures rely on the resemblance of textual information in two texts regardless of lexical or semantic variations; for instance, 'was' and 'is' in such measures are not detected to be similar due to the morphological differences even though both are word forms of the same lemma 'to be'. Some of the well-known methods of this type use edit distance measures and character-level comparisons. Gomaa et al. (2013) and Yu et al. (2016) provide surveys on such methods.

Let $A$ and $B$ be the set of words belonging to two senses or definitions in a dictionary. We calculate their similarity using the following methods:

**LONGEST COMMON SUBSEQUENCE** The longest subsequence of words (characters) that match between the two strings as a ratio to the average length between the two strings.

**LONGEST COMMON PREFIX/SUFFIX** The longest subsequence of words (characters) from the start/end of each string, as a ratio to the average length.

**$n$-GRAM** The number of matching subsequences of words (characters) of length $n$ between the two strings as a ratio to the average maximum number of $n$-grams that could match (e.g. length of string minus $n$ plus one)

**JACCARD/DICE/CONTAINMENT** The match between the words of the two definitions using the Jaccard and Dice coefficients.

$$\text{Jaccard} = \frac{|A \cap B|}{|A \cup B|} \tag{4.4}$$

$$\text{Dice} = \frac{2|AB|}{|A| + |B|} \tag{4.5}$$

$$\text{Containment} = \frac{|A \cap B|}{\min(|A|, |B|)} \tag{4.6}$$

**SMOOTHED JACCARD** This metric is an improved formulation of the Jaccard coefficient that makes the optimization possible and can be adjusted using the parameter $\alpha$ to distinguish matches on shorter texts (McCrae et al., 2017a). It is defined as follows:

$$J_\sigma(A, B) = \frac{\sigma(|A \cap B|)}{\sigma(|A|) + \sigma(|B|) - \sigma(|A| \cup |B|)} \tag{4.7}$$

where $\sigma(x) = 1 – exp(–\alpha x)$ and it tends to Jaccard as $\alpha \to 0$.

**SENTENCE LENGTH RATIO (SLR)** The ratio of the length of the sentences as

$$SLR(A, B) = 1 - \frac{min(|A|, |B|)}{max(|A|, |B|)} \tag{4.8}$$

**JARO-WINKLER, LEVENSHTEIN** The Jaro-Winkler (Jaro, 1989) and Levenshtein (Levenshtein et al., 1966) distances, respectively $d_{Jaro}$ and $lev(a, b)$, are two standard string similarity functions that are defined as follows:

$$d_{Jaro} = \begin{cases} 0 & m = 0 \\ \frac{1}{3}\left(\frac{m}{|A|} + \frac{m}{|B|} + \frac{m-t}{m}\right) & m \neq 0 \end{cases} \tag{4.9}$$

$$lev_{A,B}(i, j) = \begin{cases} max(i, j) & \text{if } min(i, j) = 0 \\ \min \begin{cases} lev_{A,B}(i-1, j)) + 1 \\ lev_{A,B}(i, j-1) + 1 \\ lev_{A,B}(i-1, j-1) + 1_{(A_i \neq B_j)} \end{cases} & \text{otherwise.} \end{cases} \tag{4.10}$$

where $m$ and $t$ are respectively the number of matching characters and half the number of transpositions, $i$ and $j$ are the terminal character position of strings $A$ and $B$ and $1_{(A_i \neq B_j)}$ is the indicator function equal to $0$ when $A_i \neq B_j$ and equal to 1 otherwise. We use the Apache Commons Text[5] implementations for these two functions.

**MONGE-ELKAN** This is defined as follows where $sim$ is a word similarity function (using either Jaro-Winkler or Levenshtein):

$$ME(A, B) = \frac{1}{|A|} \sum_{i=1}^{|A|} \max_{j=1,\ldots t} sim(A_i, B_j) \tag{4.11}$$

**AVERAGE WORD LENGTH RATIO** The ratio of the average word length in each sentence normalized to the range [0,1] as for SLR.

**NEGATION** Whether either both sentences contain negation words or both don't (1 if true, 0 if false).

**NUMBER** If both sentences contain numbers do these numbers match (1 if all numbers match).

**BAG-OF-WORD** This is a basic vector space model where the vocabulary of the language is mapped to hashed values. This way, a set of words can be transformed into a vector of integers, from which the bag term comes, regardless of the position where a specific word appears. Having the two bags of words corresponding to the definitions, it is then to calculate their similarity by applying a metric such as Jaccard.

It should be noted that the aforementioned measures are all symmetric, meaning that the similarity function returns the same value if $A$ is compared to $B$ or $B$ is compared to $A$.

### 4.4.2 The WBbM algorithm

WBbM is a variation of the weighted bipartite matching (WBM), also known as the assignment problem. In the assignment problem (Kuhn, 1955), the optimal matching only contains one-to-one matchings with the highest sum of weights. This bijective mapping restriction is not realistic in the case of lexical resources where an entry may be linked to more than one entry. Therefore, WBbM aims at providing a more diversified matching where a node may be connected to a certain number of nodes. Formally, given $G = ((U, V), E)$ with weights $W$ and vertex-labelling functions $L : U \cup V \to \mathbb{N}$ and $B : U \cup V \to \mathbb{N}$, WBbM finds a subgraph $H = ((U, V), E')$ which maximizes $\sum_{e \in E'} W(e)$ having $u \in [L(u), B(u)]$ and $v \in [L(v), B(v)]$. In other words, the number of edges that can be connected to a node is determined by the lower

5 https://commons.apache.org/proper/commons-text/

and upper bound functions L and B, respectively. Algorithm 4.1 presents the WBbM algorithm with a greedy approach where an edge is selected under the condition that adding such an edge does not violate the condition over the lower and the upper bounds, i.e. L and B.

**Algorithm 4.1: Greedy WBbM**

**Input:** $G = ((U, V), E, W)$, lower bound L and upper bound B
**Output:** $H = ((U, V)), E', W)$ satisfying bound constraints with a greedily-maximized score $\sum_{e \in E'} W(e)$

$E' = \emptyset$
Sort E by descending W(e)
**for** $e$ **to** E **do**
    **if** $H = ((U, V)), E' \cup \{e\}, W)$ *does not violate B* **then**
        $E' = E' \cup \{e\}$
    **end**
**end**
**if** $H = ((U, V)), E', W)$ *does not violate L* **then**
    **return** $H = ((U, V)), E', W)$
**end**
**else**
    Print "matching impossible"
**end**

We evaluate the performance of our approach on aligning sense definitions in WordNet and Wiktionary using an aligned resource presented by Meyer and Gurevych (2011). Given an identical entry in English WordNet and Wiktionary with the same part-of-speech tags, we first convert the senses to a bipartite graph where each side of the graph represents the senses belonging to one resource. Then, we extract the similarity scores between those senses using a similarity function. The similarity function is a classifier trained using support vector machine[6] based on the similarity metrics mentioned in Section 4.4.1.

Once the semantic similarity scores are extracted and assigned as the weight of the edges of our bipartite graph, the senses graph are matched by the WBbM algorithm. This process is illustrated in Figure 4.3 where senses of entry $E_i$ in resource $\mathcal{R}_1$, $\{S_1, S_2, ..., S_n\}$, are aligned with the senses of the same entry in $\mathcal{R}_2$, $\{S'_1, S'_2, ..., S'_n\}$. The lower and upper bounds of the right side and left side of the graph, respectively $[l, b]$ and $[l', b']$, are the parameters to be tuned.

6 LibSVM: https://github.com/cjlin1/libsvm

### 4.4.3 Experiments

In order to evaluate the performance of our alignment approach, we calculated macro precision $P_{macro}$, macro recall $R_{macro}$, average F-measure $F_{avg}$ and average accuracy $A_{avg}$ as follows:

$$P = \frac{TP}{TP + FP} \tag{4.12}$$

$$P_{macro} = \frac{1}{|E|} \sum_{i=1}^{|E|} \frac{TP_i}{TP_i + FP_i} \tag{4.13}$$

$$R = \frac{TP}{TP + FN} \tag{4.14}$$

$$R_{macro} = \frac{1}{|E|} \sum_{i=1}^{|E|} \frac{TP_i}{TP_i + FN_i} \tag{4.15}$$

$$F = 2 \times \frac{P \times R}{P + R} \tag{4.16}$$

$$F_{avg} = \frac{1}{|E|} \sum_{i=1}^{|E|} F_i \tag{4.17}$$

$$A_{avg} = \frac{1}{|E|} \sum_{i=1}^{|E|} \frac{TP_i + TN_i}{TP_i + TN_i + FP_i + FN_i} \tag{4.18}$$

where $E$ refers to the set of entries, TP, TN, FN and FP respectively refer to true positive, true negative, false negative and false positive.

Table 4.3 provides the evaluation results using the WBbM algorithm with different combinations of the matching bounds over the left side (WordNet senses) and the right side (Wiktionary senses) of the alignment graph. We observe that a higher upper bound increases the recall. On the other hand, setting the lower bound to 1 provides higher precision, while parameters with a lower bound of 0, e.g. [0, 3], lack precision. Note that [0, 1] parameter performs similarly as a bijective mapping algorithm such as the assignment problem where a node can be only matched to one node.

| Left bound, right bound | $P_{macro}$ | $R_{macro}$ | $F_{avg}$ | $A_{avg}$ |
|---|---|---|---|---|
| [0, 1], [0, 1] | **81.86** | 61.83 | 68.51 | 69.48 |
| [0, 2], [0, 1] | 78.13 | 70.74 | 73.28 | 76.57 |
| [0, 3], [0, 1] | 77.88 | 71.38 | 73.59 | 77.13 |
| [1, 2], [1, 2] | 81.21 | 74.17 | 76.59 | 79.49 |
| [1, 3], [1, 3] | 81.26 | 75.02 | 77.12 | 80.14 |
| [1, 5], [0, 1] | 81.25 | **75.25** | **77.28** | **80.33** |
| [1, 5], [1, 2] | 81.25 | 75.23 | 77.26 | 80.32 |

**Table 4.3:** WBbM algorithm performance on alignment of WordNet and Wiktionary

## 4.5 TRANSLATION INFERENCE

In addition to WSA, translation inference can also utilize a graph-based method to generate new translation pairs. In this vein, two methods are proposed based on graph traversal heuristic: cycle-based and path-based translation inference. We apply these approaches to the data of the Translation Inference Across Dictionaries 2019 Shared Task which focused on the automatic generation of dictionary entries in order to enrich knowledge graphs with multilingual knowledge. The task focused on the generation of French-English, Portuguese-English and French-Portuguese dictionaries. We use the Apertium RDF dataset[7] (Forcada et al., 2011) which is illustrated in Figure 4.4. It is worth mentioning that no external resource is used where a direct link between any of those target languages exists, with the exception being parallel data between the targeted languages. The datasets contain 44 languages and 53 language pairs, with a total number of 1,540,996 translations between 1,750,917 lexical entries.

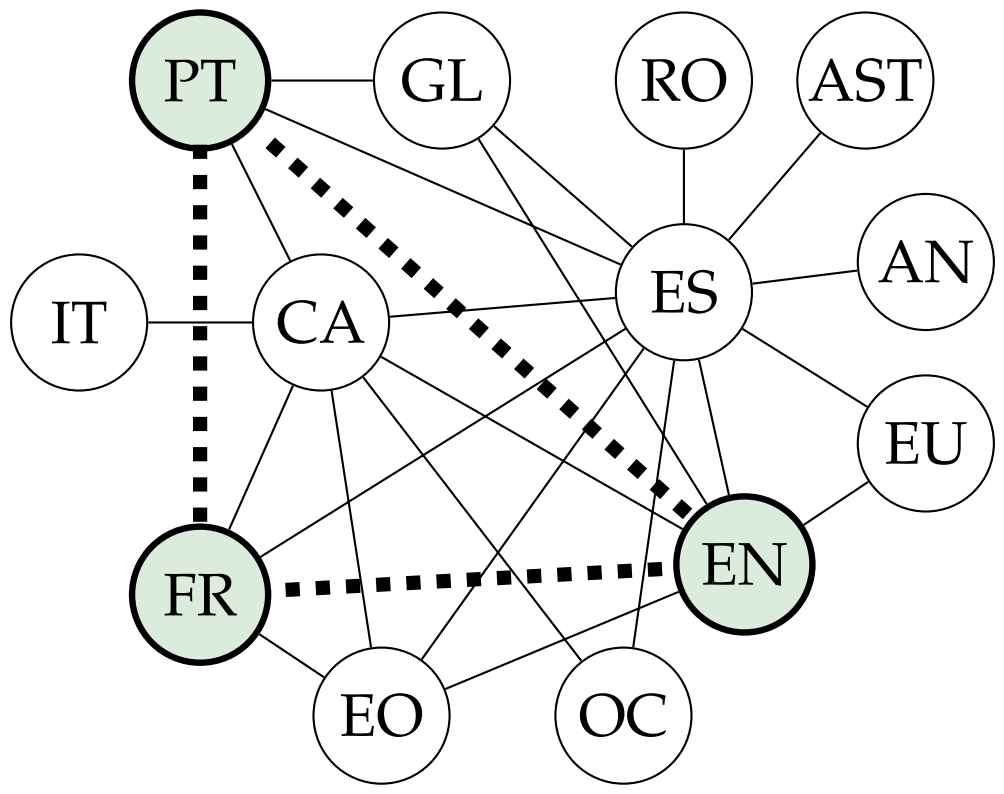

**Figure 4.4:** Apertium RDF graph, with the target languages shown in green and bilingual dictionaries to be generated indicated in dashed lines.

7 linguistic.linkeddata.es/apertium/, version `cfc5995c5369ddb158cd266fcb8d4b74ed8dbdd0`.

### 4.5.1 Cycle-based approach

A heuristic is devised that focuses on producing high precision entries, even though the recall might suffer. In this approach, loops of length four are searched in the Apertium dictionaries in order to discover new translations. The model builds a graph with all the bilingual word-POS pairs in all the Apertium dictionaries that can be used as a pivot between Portuguese, French and English; those are language pairs that appear as an intermediate node between the source language and the target language in Figure 4.4, like Spanish (ES) or Catalan (CA). Whenever a cycle of length 4 in the graph is retrieved, all the nodes in the cycle are assumed to be translations, thus are connected. All discovered edges for the respective language pair are used to generate a new dictionary. Figure 4.5 shows an example of discovered translations for the word 'antique' (adjective).

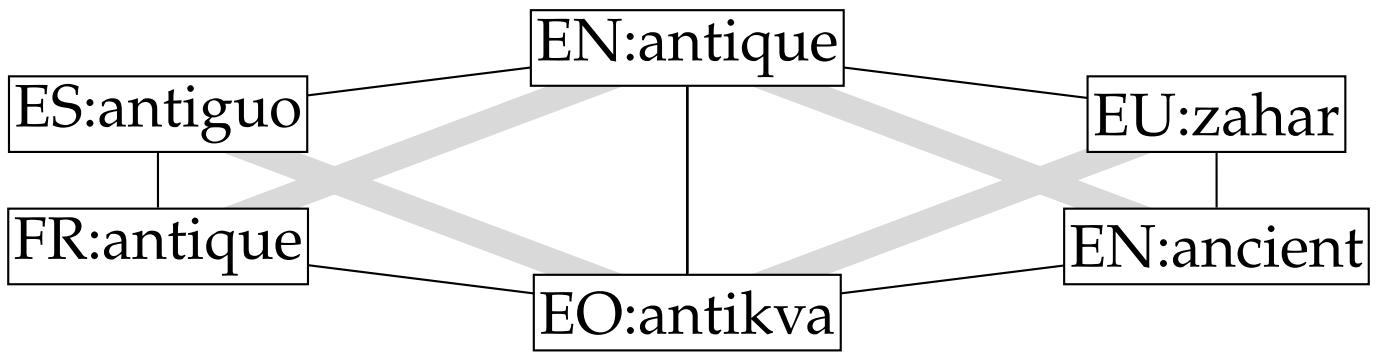

**Figure 4.5:** Cycles found in the dictionary (solid lines) and inferred translations (transparent lines). Some of the lines identify possible same-language synonyms (e.g. *ancient* and *antique* in English), while others identify newly discovered possible translations (e.g. *antiguo* in Spanish and *antikva* in Esperanto).

### 4.5.2 Path-based approach

Similar to the cycle-based method, we use another heuristic technique that aims to create translation candidates by traversing the paths between the source and the target languages in the Apertium language graph. The candidate translations $T$ are weighted with respect to the path length and the frequency. In this section, *language graph* refers to the Apertium dictionary graph (Figure 4.4) and *translation graph* refers to a graph where vertices represent a word and edges represent the translations in other languages. Figure 4.6 illustrates the translation graph of the word spring as a noun in English based on the following language path:
English→Basque→Spanish→French→Esperanto→Catalan→Portuguese.

The basic idea behind pivot-oriented translation inference is the transitivity assumption of translations. If $w_p$, a pivot word in the dictionary $\mathcal{D}_p$, has the translation equivalents $w_i$ and $w_j$ in dictionaries $\mathcal{D}_{p \to 1}$ and $\mathcal{D}_{p \to 2}$ respectively, then $w_i$ and $w_j$ may be equivalents in the $\mathcal{D}_{1 \to 2}$ dictionary. Although the pivot-oriented approach can mostly create accurate translations for monosemous words (depending on the lexicon completeness), this oversimplifies for polysemous words leading to incorrect translations (Saralegi et al., 2011).

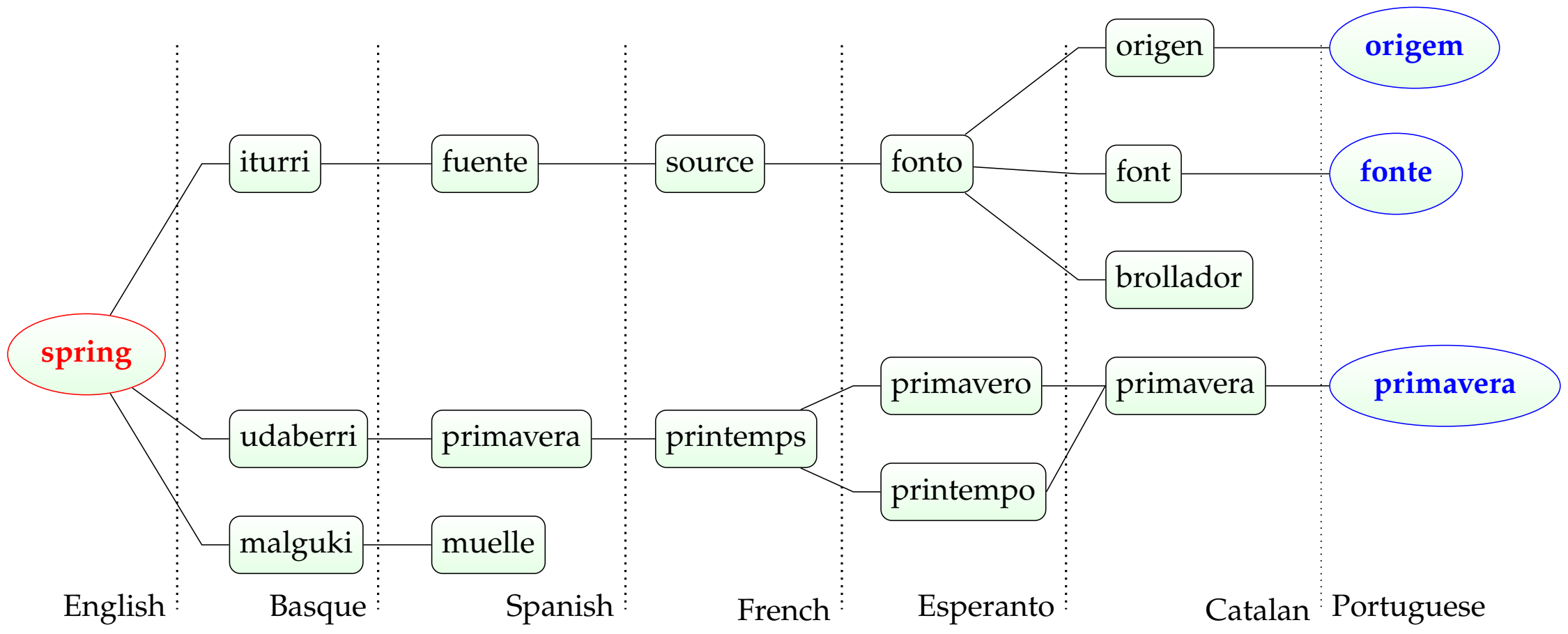

**Figure 4.6:** Translation graph of *spring* (noun) (in red) resulting in Portuguese translations (in blue) using the pivot languages.

For the current task, we have considered all the simple paths, i.e. paths without any repeating vertex, starting from and ending with the goal languages of the shared task, of which there are 92, 67 and 58 simple paths between Portuguese→French, French→English and English→Portuguese, respectively. As the language graph is undirected, the paths between the vertices are identical in each direction. For instance, the dictionary paths from English to Portuguese are the same as those from Portuguese to English.

In order to describe the likelihood of a translation being correct, we introduce the following weighting factor:

$$w_t = frequency(t) \times \alpha^l \tag{4.19}$$

where $w_t$ is the weight of the translation candidate $t \in T$, $frequency(t)$ is the number of times translation $t$ is reached, $\alpha \in (0, 1)$ is a penalization constant and $l$ is the length of the path leading to $w_t$. $\alpha^l$ penalizes the translation candidates in a way that paths of lower length and higher frequency get a lower weight. On the other hand, a longer translation path results in a lower weight factor. For instance, in the translation graph in Figure 4.6, the frequency of the words *primavera*, *font* and *origem* is respectively 2, 1 and 1 and the length of their translation path is 7. For the current task, we set $\alpha = 0.5$ and have included the part-of-speech tags in the inference.

Finally, the weights are normalized such that $\sum_{t \in T} w_t = 1$. It is worth noting that this method requires optimization, in the sense of rewriting codes, as it is computationally expensive due to the high number of node combinations between a source and a target word.

### 4.5.3 Experiments

Table 4.4 shows the size of the discovered dictionaries using both the cycle and path approaches. Unlike the cycle strategy which creates symmetric dictionaries, the path strategy creates different translation pairs based on translation direction, i.e. traversing nodes starting from English to French yields different translation pairs than starting from French to English.

| | EN-FR | FR-EN | EN-PT | PT-EN | FR-PT | PT-FR |
|---|---|---|---|---|---|---|
| Cycle | 7041 | | 142 | | 100 | |
| Path | 25 594 | 26 492 | 16 273 | 22 195 | 19 079 | 27 678 |

**Table 4.4:** Sizes of the extracted dictionaries in the cycle-based and path-based approaches.

Moreover, the performance of each method are evaluated based on the manually compiled pairs of *K Dictionaries* as a gold standard. The evaluation is carried out according to precision, recall, f-measure and coverage where the latter refers to the number of entries in the source language for which a translation is generated. The evaluation results are provided in Table 4.5. In the generated translation, a confidence score is created as well. The impact of this score, normalized in [0-1] is evaluated using a varying threshold.

| | Baseline (OTIC) | | | | Cycle-based | | | | Path-based | | | |
|---|---|---|---|---|---|---|---|---|---|---|---|---|
| T | P | R | F1 | C | P | R | F1 | C | P | R | F1 | C |
| 0 | 0.63 | 0.27 | **0.38** | **0.46** | 0.75 | 0.07 | 0.11 | 0.13 | 0.26 | **0.28** | 0.26 | 0.45 |
| 0.1 | 0.63 | 0.27 | **0.38** | **0.46** | 0.75 | 0.07 | 0.11 | 0.13 | 0.4 | 0.26 | 0.31 | 0.45 |
| 0.2 | 0.63 | 0.27 | **0.38** | **0.46** | 0.75 | 0.07 | 0.11 | 0.13 | 0.51 | 0.24 | 0.32 | 0.44 |
| 0.3 | 0.63 | 0.27 | **0.38** | **0.46** | 0.75 | 0.07 | 0.11 | 0.13 | 0.59 | 0.21 | 0.31 | 0.4 |
| 0.4 | 0.64 | 0.27 | **0.38** | **0.46** | 0.75 | 0.07 | 0.11 | 0.13 | 0.65 | 0.19 | 0.29 | 0.36 |
| 0.5 | 0.64 | 0.26 | 0.37 | 0.45 | 0.75 | 0.07 | 0.11 | 0.13 | 0.68 | 0.17 | 0.26 | 0.33 |
| 0.6 | 0.66 | 0.24 | 0.35 | 0.43 | 0.75 | 0.07 | 0.11 | 0.13 | 0.75 | 0.14 | 0.23 | 0.27 |
| 0.7 | 0.71 | 0.19 | 0.29 | 0.34 | 0.75 | 0.07 | 0.11 | 0.13 | **0.76** | 0.12 | 0.21 | 0.24 |
| 0.8 | 0.71 | 0.19 | 0.29 | 0.34 | 0.75 | 0.07 | 0.11 | 0.13 | **0.76** | 0.11 | 0.19 | 0.22 |
| 0.9 | 0.71 | 0.18 | 0.29 | 0.33 | 0.75 | 0.07 | 0.11 | 0.13 | **0.76** | 0.1 | 0.18 | 0.2 |
| 1 | 0.71 | 0.18 | 0.29 | 0.33 | 0.75 | 0.07 | 0.11 | 0.13 | 0.75 | 0.09 | 0.16 | 0.18 |

**Table 4.5:** Results of the evaluation with precision (P), recall (R), F-measure (F1), and coverage (C) for all the different thresholds (T).

The evaluation results indicate that the path-based approach performs better than the cycle-based one. Although both techniques generate translation pairs with higher precision, they have a lower f-measure than the baseline system which is based on the One Time Inverse Consultation (OTIC) method (Tanaka and Umemura, 1994b). This being said, none of the techniques outperforms the coverage of the baseline.

## 4.6 CONCLUSION AND CONTRIBUTIONS

In this chapter, we revisited WordNet-Wiktionary alignment task and proposed an approach that divides the alignment problem into two sub-tasks of retrieving similarity of sense pair based on textual and semantic similarity and aligning in a holistic way using graph-based WBbM algorithm. We demonstrated that this approach is efficient for aligning resources in comparison to the baseline results thanks to the flexibility of the matching algorithm. However, tuning the parameters of the matching algorithm needs further investigations of the resource and is not following a rule. Additional studies propose further customization of the WBbM algorithm by considering conflicts between potential links (Chen et al., 2016b), focusing on a smaller group of target nodes (Chen et al., 2016a) and increasing diversity (Ahmed et al., 2017). These can be applied to the same linking setup that we described in this chapter. In addition, our experiments can be extended to more resources in the future, such as the OAEI ontology alignment datasets[8] and WordNet-Wikipedia mapping dataset (McCrae, 2018).

On the other hand, we focused on the translation inference task and demonstrated that graph-based methods can also be used for generating new translation pairs in an unsupervised way. To this end, we proposed cycle-based and path-based methods where the path length, i.e. number of edges between two nodes, determines the probability that two words are translations. One major limitation of graph-based methods is the limited coverage of connectivity between certain translations. Figure 4.7 illustrates some of the translations that can be retrieved for the word 'chaotic' (adjective) in the Apertium translation graph (Goel et al., 2021) where the Portuguese translation '*caótico*' ('chaotic') is not retrievable by traversing intermediate nodes. As future work, this can be addressed using external resources and more state-of-the-art techniques such as graph neural networks and unsupervised cross-lingual word embeddings (**Sina Ahmadi** et al., 2021b).

The latest version of the TIAD 2021 data[9] include more bilingual dictionaries in other languages, such as Maltese (MT) and Arabic (AR). Although the translation graph contains more languages, some of them are isolated nodes that can be pruned, such as MT-AR in Figure 4.8. Therefore, semi-supervised graph methods, in particular those that focus on label propagation as described in (Talukdar and Crammer, 2009), would also be another avenue of research. In this scenario, a seed dictionary, i.e. a dictionary that contains translation pairs as highlighted in Figure 4.8 to the right, is used to create a probability distribution over the unseen nodes that can be potential translations for a given word. In the same vein, identifying isomorphisms in lexicographical networks in two dictionaries may be beneficial to find potentially identical lexical items in two resources.

8 http://oaei.ontologymatching.org/
9 https://tiad2021.unizar.es/task.html

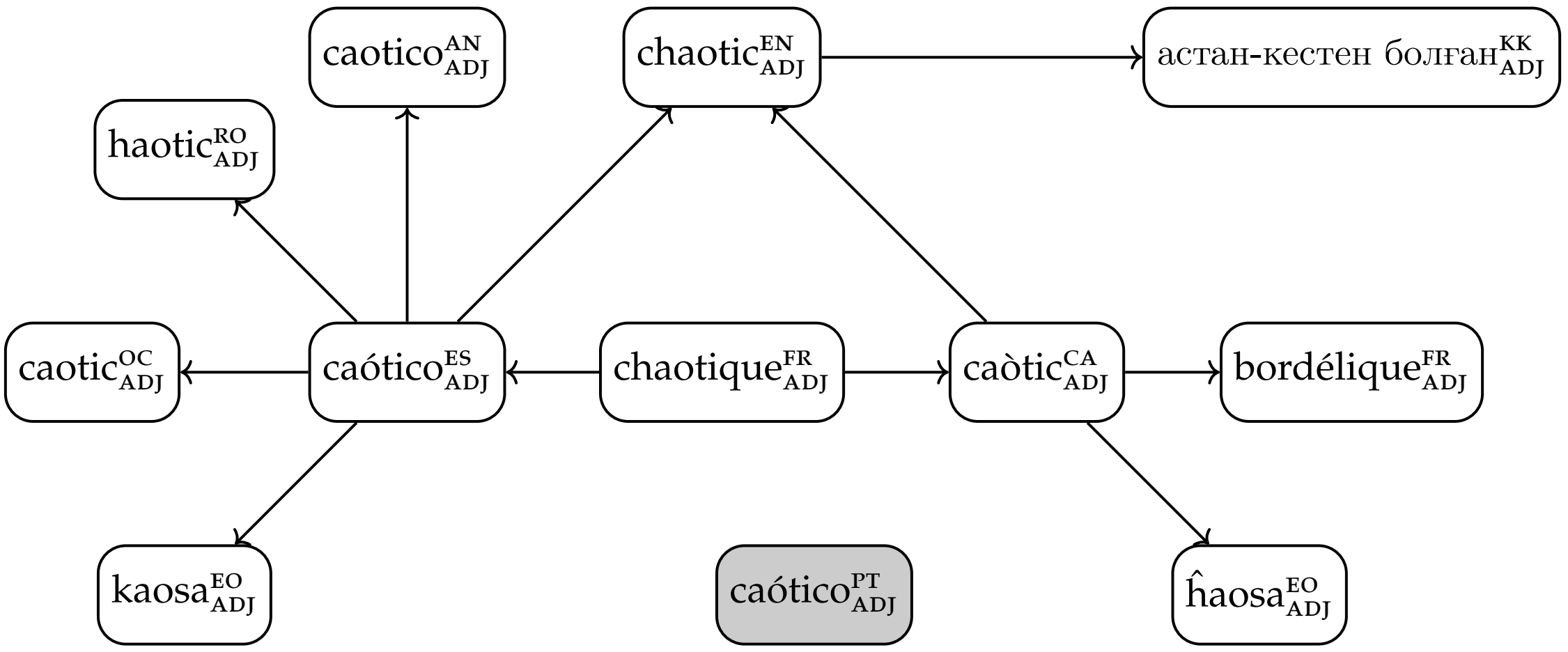

**Figure 4.7:** Paths starting from '*chaotique*' (adjective in French) in the Apertium translation data. Language codes and part-of-speech tags are respectively provided in subscript and superscript.

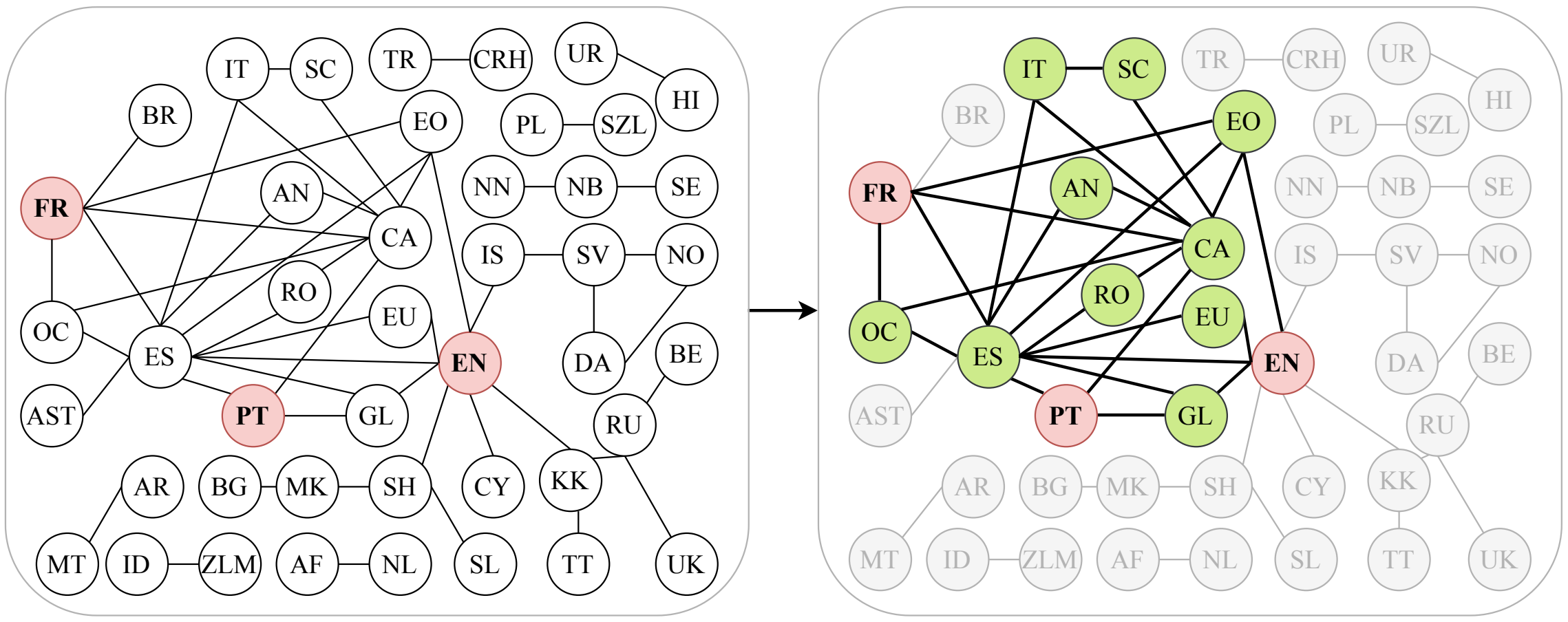

**Figure 4.8:** To the left, Apertium RDF graph (version 2) where nodes refer to languages and edges the availability of translations. Nodes unconnected to the target languages French (FR), English (EN) and Portuguese (PT) are pruned.

# 5 A BENCHMARK FOR MONOLINGUAL WORD SENSE ALIGNMENT

In the previous chapter, we introduced graph-based techniques for the alignment of lexicographical resources. As concluded, there is a lack of annotated datasets for the task of word sense alignment for evaluation purposes. This chapter addresses the task of word sense alignment and describes the manual annotation of various expert-made dictionaries for 15 languages.

## 5.1 INTRODUCTION

Different dictionaries and related resources such as Wordnets and encyclopedia have significant differences in structure and heterogeneity in content, which makes aligning information across resources and languages a challenging task. Word sense alignment (WSA) is a more specific task of linking dictionary content at sense level which has been proved to be beneficial in various NLP tasks, such as word-sense disambiguation (Navigli and Ponzetto, 2012a), semantic role labeling (Palmer, 2009) and information extraction (Moro et al., 2013). In a wider scope, combining LSRs can enhance domain coverage in terms of the number of lexical items and types of lexical semantic information (Gurevych et al., 2012).

Given the current progress of artificial intelligence and the usage of data to train neural networks, annotated data with specific features play a crucial role in tackling data-driven challenges, particularly in NLP. In recent years, a few efforts have been made to create *gold-standard* dataset, i.e., a dataset of instances used for learning and fitting parameters, for aligning senses across monolingual resources including collaboratively-curated ones such as Wiktionary[1], and expert-made ones such as WordNet. However, the previous work is limited to a handful of languages and much of it is not on the core vocabulary of the language, but instead on named entities and specialist terminology. Moreover, despite the huge endeavor of lexicographers to compile dictionaries, proper lexicographic data are rarely openly accessible to researchers. In addition, many of the resources are quite small and the extent to which the mapping is reliable is unclear.

1 https://www.wiktionary.org

Lexicographical resources mainly vary in structure. WordNet (Miller, 1995), for instance, is an LSR similar to a dictionary that provides lexical items with definitions and examples, but also conceptual, semantic and lexical relations such as meronymy, hypernymy, and hyponymy. Encyclopedias, on the other hand, have more descriptive content in comparison to other resources but limited semantic relationships between entries. In addition to structure, lexicographical resources vary in content as well. Even though the headwords in two monolingual dictionaries may be identical, ignoring spelling variations, the more descriptive part of the content that is sense definition, also known as gloss, can vary quite differently based on the lexicographer's preference, editorial decisions and more importantly, the paradigm followed to define a concept (as discussed in Section 2.2.2). In the following examples, the various sense definitions of 'spring' (noun) in a few monolingual English resources are provided:

SPRING (noun) in Collins

1. spring is the season between winter and summer when the weather becomes warmer and plants start to grow again.
2. a spring is a spiral of wire which returns to its original shape after it is pressed or pulled.
3. a spring is a place where water comes up through the ground. It is also the water that comes from that place.

SPRING (noun) in Longman

1. the season between winter and summer when leaves and flowers appear
2. (a) something, usually a twisted piece of metal, that will return to its previous shape after it has been pressed down

   (b) the ability of a chair, bed, etc to return to its normal shape after being pressed down
3. a place where water comes up naturally from the ground
4. a sudden quick movement or jump in a particular direction

SPRING (noun) in the Princeton WordNet:

1. the season of growth
2. a metal elastic device that returns to its shape or position when pushed or pulled or pressed
3. a natural flow of ground water
4. a point at which water issues forth
5. the elasticity of something that can be stretched and returns to its original length
6. a light, self-propelled movement upwards or forwards

SPRING (noun) in MacMillan

1. the season of the year between winter and summer
   (a) happening in spring, or relating to spring
2. water that flows up from under the ground and forms a small stream or pool
3. a long thin piece of metal in the shape of a coil that quickly gets its original shape again after you stop stretching it
   (a) the ability of something to get its original shape again after you stop stretching it
4. a quick jump forward or up

Differences in structure as in sense-subsense differentiation and content are obvious in these examples. For instance, the definition of 'spring' as a water source varies between Collins[2] (sense # 3), Longman[3] (sense #3) and MacMillan[4] (sense #2). In the first dictionary, the emphasis is on the water that comes up through the ground, while in the second, the 'natural' aspect of a water source is highlighted too and in the third one, the result of the water forming a small pool is mentioned as well. Nevertheless, such specifications of a word sense in definition may seem less relevant when it comes to metaphorical senses, as in '*therefore you will joyously draw water, from the springs of salvation*' where 'spring' is considered as a source but not essentially that of water. Similarly, 'spring' as a season is defined specifically as the season between winter and summer in MacMillan, while the same sense is defined as 'the season of growth' without further specifications.

Therefore, the manual alignment of word sense involves searching for matching senses within dictionary entries of different lexical resources and linking them, which poses significant challenges. The lexicographic criteria are not always entirely consistent within individual dictionaries and even less so across different projects where different options may have been assumed in terms of structure and especially wording techniques of lexicographic glosses. Aligning senses across languages poses an extra challenge due to culture-specific differences in conceptualization and vocabulary. To remedy the nuances of meanings in definitions, we propose a framework where the alignment task is extended to semantic relations, namely exact, broader, narrower and related. These relations are according to Simple Knowledge Organization System (SKOS) (Miles and Bechhofer, 2009) and enable a fine-grained alignment at the sense level.

In this chapter, we present a set of datasets for the task of WSA containing manually-annotated monolingual resources in 15 languages. The annotation is carried out at sense level where semantic relations between definition pairs are also se-

2 https://www.collinsdictionary.com
3 https://www.ldoceonline.com
4 https://www.macmillandictionary.com

lected in the two resources by native lexicographers. Given the lexicographic context of this study, we have tried to provide lexicographic data from expert-made dictionaries. We believe that our datasets will pave the way for further developments in exploring statistical and neural methods, as well as for evaluation purposes. It will also fill the existing gap of lack of bench-marking for the WSA task.

The rest of the chapter is organized as follows: we first describe the previous work in Section 5.2. After having described our methodology in Section 5.3, we further elaborate on the challenges of sense annotation based on the experiences of the annotators in Section 5.4. We evaluate the datasets in Section 5.5 and finally, conclude the the chapter in Section 5.6.

## 5.2 RELATED WORK

Aligning senses across lexical resources has been attempted in several lexicographical milieus over recent years. Such resources mainly include open-source dictionaries, WordNet and collaboratively-curated resources, such as Wiktionary. The latter has been shown to be reliable resources to construct accurate sense classifiers (Dandala et al., 2013).

There has been a significant body of research in aligning English resources, particularly, the Princeton WordNet with Wikipedia as in Ruiz-Casado et al. (2005); Ponzetto and Navigli (2010); Niemann and Gurevych (2011); McCrae (2018)), with the Longman Dictionary of Contemporary English and with Roget's thesaurus as in Kwong (1998), with Wiktionary[5] as in Meyer and Gurevych (2011) and with the Oxford Dictionary of English as in Navigli (2006). Meyer and Gurevych (2011) also present a manually-annotated dataset for WSA between the English WordNet and Wiktionary.

On the other hand, there are a fewer number of manually aligned monolingual resources in other languages. For instance, there have been considerable efforts in aligning lexical semantic resources in German, particularly, the GermaNet–the German Wordnet (Hamp and Feldweg, 1997) with the German Wiktionary (Henrich et al., 2011), with the German Wikipedia (Henrich et al., 2012) and with the Digital Dictionary of the German Language (*Digitales Wörterbuch der Deutschen Sprache* (Klein and Geyken, 2010)) (Henrich et al., 2014) where 470 lemmas with 1,517 senses are aligned. Gurevych et al. (2012) present UKB–a large-scale lexical semantic resource containing pairwise sense alignments between a subset of nine resources in English and German which are mapped to a uniform representation. Although the datasets contain many resources in German and English, namely Wiktionary, Wikipedia and OmegaWiki[6], they lack multilingualism and expert-made data.

5 https://www.wiktionary.org/
6 http://www.omegawiki.org

For Danish, aligning senses across modern lexical resources has been carried out in several projects in recent years, and a next natural step is to link these to historical Danish dictionaries (Pedersen et al., 2018). Pedersen et al. (2009) describe the semi-automatic compilation of a WordNet for Danish, *DanNet*, based on a monolingual dictionary, the Danish Dictionary (*Den Danske Ordbog (DDO)*). Later, the semantic links between these two resources facilitated the compilation of a comprehensive thesaurus (*Den Danske Begrebsordbog*) (Nimb et al., 2014). The semantic links between thesaurus and dictionary made it possible to combine verb groups and dictionary valency information, used as input for the compilation of the Danish FrameNet Lexicon (Nimb, 2018). Furthermore, they constitute the basis for the automatically integrated information on related words in the Danish dictionary, on the fly for each dictionary sense (Nimb et al., 2018). Similarly, Simov et al. (2019) report the manual mapping of the Bulgarian Word-Net BTB-WN with the Bulgarian Wikipedia.

Aligning word senses and the resolution of lexical ambiguity, a task also known as word sense disambiguation, are two tightly related tasks. For instance, Dandala et al. (2013) provides a Wikipedia-based sense tagged corpus generated for four languages and describes explorations in word sense disambiguation for sense alignment. Additionally, a few other LSRs have been aligned, such as VerbNet with WordNet by Kipper et al. (2006), VerbNet with FrameNet by Palmer (2009) and finally, FrameNet with Wiktionary by Hartmann and Gurevych (2013).

Given the amount of effort required to construct and maintain expert-made resources, various solutions have been proposed to automatically link and merge existing LSRs at different levels. LSRs being very diverse in domain coverage (Meyer, 2010; Burgun and Bodenreider, 2001), previous works have focused on methods to increase domain coverage, enrich sense representations and decrease sense granularity Miller (2016). Miller and Gurevych (2014) describe a technique for constructing an $n$-way alignment of LSRs and applied it to the production of a three-way alignment of the English WordNet, Wikipedia and Wiktionary. Niemann and Gurevych (2011) propose a threshold-based Personalized PageRank method for extracting a set of Wikipedia articles as alignment candidates and automatically aligning them with WordNet synsets. This method yields a sense inventory of higher coverage in comparison to taxonomy mapping techniques where Wikipedia categories are aligned to WordNet synsets Ponzetto and Navigli (2009). Matuschek and Gurevych (2013) present the Dijkstra-WSA algorithm as a graph-based approach and a machine learning approach where features such as sense distances and gloss similarities are used for the task of WSA Matuschek and Gurevych (2014). It should be noted that all of these approaches produce results that are of lower reliability than gold standard datasets such as the ones presented in this paper.

In a recent study, Yao et al. (2021) present a methodology to align dictionary content where lexicographical data is used to train a neural network model for the automatic alignment of word senses. However, one major obstacle in such studies is the copyright of the data. Expert-made dictionaries are oftentimes not openly available. We believe that this major limitation can be addressed by our benchmark.

## 5.3 METHODOLOGY

The main goal of the current task in this chapter is to provide semantic relationships between two sets of senses for the same lemmas in two monolingual dictionaries. As an example, Figure 5.1 illustrates the senses for the entry “clog" (verb) in the English WordNet Miller (1995) (to the left in the third column) and the Webster’s Dictionary 1913 Webster and Slater (1828) (to the right in the sixth column). For further clarification, we provide a few case studies in Section 5.4.

| Headword (POS) | R1-IDs | R1 senses | Semantic relation | Sense match | R2 senses | R2-IDs |
|---|---|---|---|---|---|---|
| clog (verb) | | | | | | |
| | clog.v.02 | dance a clog dance | | | to become clogged; to become loaded or encumbered, as with extraneous matter. | |
| | clog.v.03 | impede the motion of, as with a chain or a burden | | | to encumber or load, especially with something that impedes motion; to hamper. | |
| | clog.v.01 | become or cause to become obstructed | | | to coalesce or adhere; to unite in a mass. | |
| | clog.v.06 | fill to excess so that function is impaired | | | to obstruct so as to hinder motion in or through; to choke up; . | |
| | clog.v.04 | impede with a clog or as if with a clog | | | to burden; to trammel; to embarrass; to perplex. | |
| | clog.v.05 | coalesce or unite in a mass | | | | |

**Figure 5.1:** Sense provided for *clog* (verb) in the English WordNet (R1) and the Webster Dictionary (R2). Drop-down lists are created dynamically for semantic relationship annotation.

Given the range of languages that were initially targeted, we invited all partners and observers of the ELEXIS project to join this task. According to the annotation guidelines, we asked participants to provide at least one expert-made resource for their language of interest. The annotation guidelines were finalized after a pilot study where various issues regarding the annotation framework and difficulties regarding polysemous items were raised and addressed.

The actual annotation was implemented by means of dynamic spreadsheets that provide a simple but effective manner to complete the annotation. This also had the added advantage that the annotation task could be easily completed from any device. In order to collect the data that was required for the annotation, each of the participating institutes provided their data in some form. We asked them, where possible, to organize their two dictionaries either in OntoLex-Lemon (Cimiano et al., 2016), TEI-Lexo (Romary and Tasovac, 2018) or by following a simple TSV (tab-separated values) or Excel format providing the following data:

- An entry identifier, that locates the entry in the resource

- A sense identifier marking the sense in the resource, for example, the sense number
- The lemma of the entry. Spelling variations and orthographies should be normalized, as in the case of the old and modern Danish orthographies ("*kjø* → *kø*").
- The part-of-speech of the entry
- The sense text, including the definition. If the senses are provided in a hierarchical form to represent semantically-related concepts, they should be unified to bring all the senses along with subsenses at the same level.

In order to facilitate the task of annotation, we convert the initial data into spreadsheets similar to Figure 5.1. These spreadsheets provided an easy mapping and had the following columns:

- The headword and part of speech (given in parentheses after the headword);
- The sense text (definition) in the first resource;
- An interactive drop-down to specify one of the 5 semantic relations (see below) from the sense in the first resource;
- The sense text (abbreviated) in a drop-down list from the second resource, which the first resource is matched to;
- The full sense text of the second resource.

The fifth column played no technical role in the annotation, but was provided for reference, however as it was formatted with text wrapping on, it allowed the annotators to see the full definition of the second resource. In general, we arranged the spreadsheets such that there were more senses for the first resource. In cases where the number of senses between the two resources was roughly equal, we created two spreadsheets based on which of the two datasets had more senses for those entries. In other cases, such as the English WordNet-Webster mapping where one resource (in this case WordNet) has many more senses, we used this as the first resource. Even still, there were some cases where the resource with more senses may contain a sense that corresponds to multiple senses in the second resource and in this case, the annotators were instructed to simply use the "Insert Row Below" feature of the spreadsheet, which also duplicated the drop-down lists.

### 5.3.1 Semantic Relationships

One of the challenges is that sense granularity between two dictionaries is rarely such that we would expect one-to-one mapping between the senses of an entry. In this respect, we followed a simple approach such as that in SKOS (Miles and Bechhofer, 2009) providing different kinds of linking predicates, which are described in Table 5.1. While it is certainly not easy to decide which relationship is to be used, we found that this methodology was broadly effective and we believe will simplify the development of machine-learning-based classifiers for sense alignment prediction.

| | |
|---|---|
| `exact` | senses are the same, for example the definitions are simply paraphrases |
| `broader` | senses in the first dictionary completely covers the meaning of the sense in the second dictionary and is applicable to further meanings |
| `narrower` | senses in the first dictionary is entirely covered by the sense of the second dictionary, which is applicable to further meanings |
| `related` | There are cases when the senses may be equal but the definitions in both dictionaries differ in key aspects |
| `none` | There is no match for this sense |

**Table 5.1:** Semantic relationships according to SKOS used for WSA task

SKOS is a data model designed for knowledge organization systems such as representation of thesauri, classification schemes, taxonomies and other types of controlled vocabularies or documentary languages (Miles and Bechhofer, 2009). SKOS created based as an RDF data model, its main objective is to allow the easy publication of structured vocabularies for their use within the framework of the Semantic Web. SKOS supports relationships that are commonly found in thesauri as hierarchical such as broader and narrower, associative as related and equivalence. Figure 5.2 illustrates an example where the concept of 'economic cooperation' is schematized using SKOS's core vocabulary.

As we will later see in the case studies (Section 5.4), distinguishing the type of semantic relation, particularly where it concerns exact and related, is deemed the most challenging part of the project by annotators.

The 'related' relation is chiefly used when there are differences in ontological type between the two definitions, e.g. the property of 'being able to sleep', a sense of the noun '*søvn*' ('sleep') in Danish is considered to be 'related' to 'the state of sleeping' sense in another resource. Often, such differences in ontological type across the two dictionaries are due to regular polysemy, as in act/result, semiotic artifact/content, animal/food, organization/building (Pustejovsky, 1995). Two dictionaries will often differ in their descriptions in cases of regular polysemy, focusing on either one or the other sense leaving one of them under-specified, or describing both of them. For instance, while the noun '*afsked*' 'farewell' in Danish describes the act of saying farewell, the corresponding sense definition in the other Danish resource focuses on the result, namely the phrase 'farewell', therefore the senses are only related and not exact matches.

Moreover, 'related' has also been used when the ontological type is, in fact, the same for the two senses, but where other parts of the definitions differ slightly, e.g. in the case of the noun *bamse* ('bear, teddy bear') in Danish where 'fat, clumsy person, especially a child' is considered to be a related sense to the other sense of the same lemma 'fat, good-natured person'.

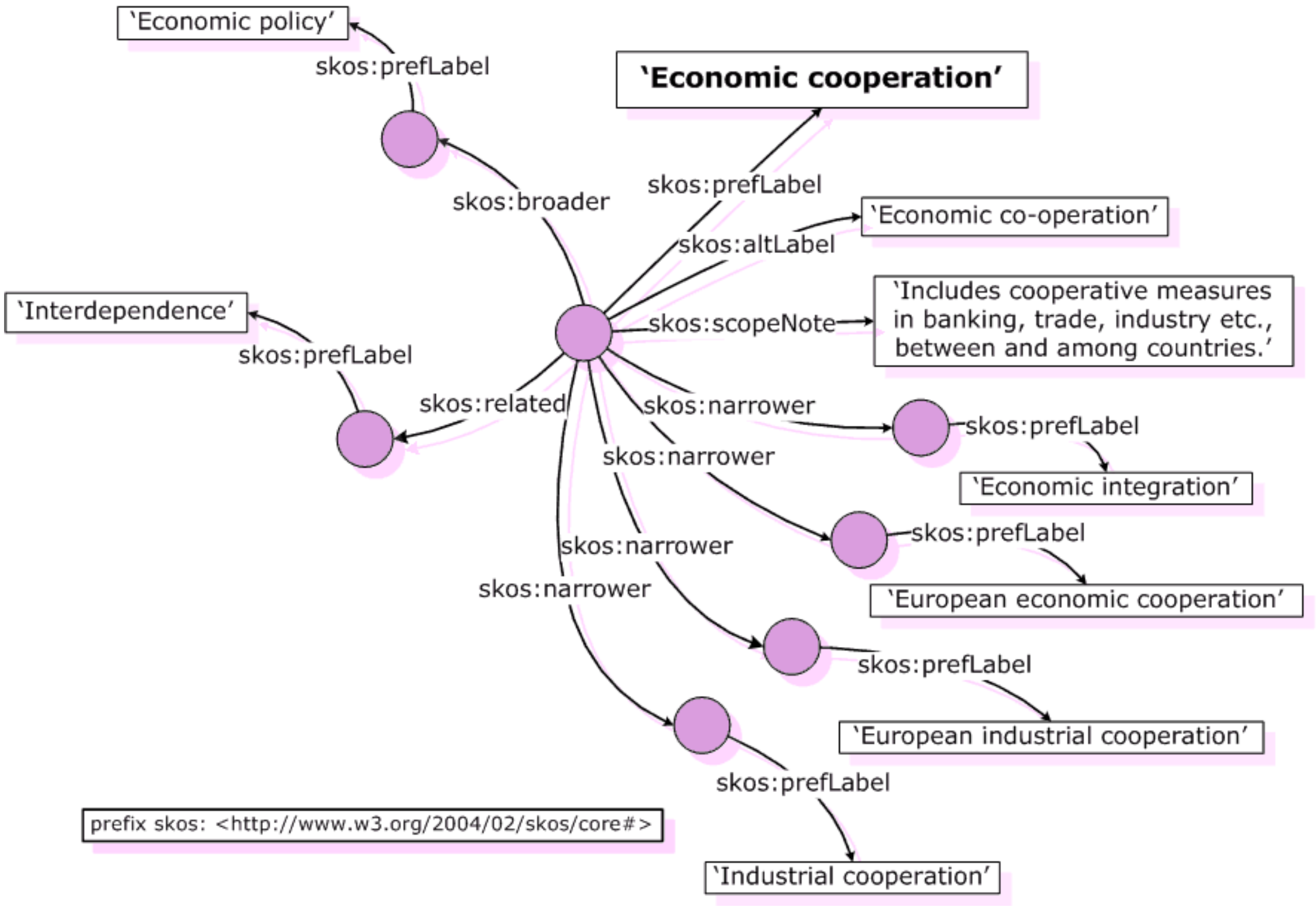

**Figure 5.2:** The concept scheme of 'economic cooperation' as an RDF graph in the SKOS core vocabulary (source: https://www.w3.org/TR/2005/WD-swbp-skos-core-guide-20051102/)

Regarding the broader and narrower relations, there are two indications in sense definitions that are of help to distinguish these two relations: i) using of specific words with a broader or narrower sense as in taxonomy and ii) providing further details in one of the sense definitions, especially when the two definitions initially overlap considerably. For instance, in the case of the Danish noun '*værge*' (guardian) defined as 'a guardian of anything or anybody' in one of the dictionaries, a broader relation is selected in comparison to 'a guardian in legal context' (i.e. a guardian for a child not yet legally competent or for an incapacitated adult) due to the specifications provided in the latter gloss. Being the inverse relation of broader, a narrower relation reflects the more fine-grained semantic specifications in the first sense in comparison to the second one, as in the Danish adjective *spids* ('sharp') where one of the dictionaries defines two separate senses, one about sound and another one about the smell, while the other dictionary merges the two senses into one as 'pungent in an unpleasant way (about smell, taste or sound)'.

### 5.3.2 Data Selection

The selection of the initial set of lemmas and senses to be aligned is guided by the following criteria:

- The lemmas should represent all open class words, namely nouns, verbs, adjectives and adverbs.
- Another criterion was that the lemmas should represent different degrees of polysemy, i.e. both highly polysemous lemmas, as well as monosemous ones, should be included.
- The lemmas in the two resources have the same part-of-speech tags. Spelling variations are normalized to a unique variation.

### 5.3.3 Dictionaries used in the creation of the dataset

For alignment, we used the following dictionaries:

**BASQUE** The Basque Wordnet (MCR 3.0) and the Basque Monolingual Dictionary "*Euskal Hiztegia*" (copyright by the author, Ibon Sarasola) were linked.

**BULGARIAN** The BulTreeBank Wordnet (BTB-WN) (Osenova and Simov, 2017) and the Bulgarian Wiktionary[7] were used.

**DANISH** We used the *Ordbog over det danske Sprog* (ODS)[8] (Dahlerup, 1918), a historical dictionary covering 188,000 lemmas in Danish from 1700-1950, and *Den Danske Ordbog* (DDO) (Farø et al., 2003) a dictionary of modern Danish covering Danish from 1950 till today. One additional criterion in data selection was that at least one of the senses in DDO should be linked to a base or core concept in the Princeton WordNet via the Danish WordNet (Pedersen et al., 2019). This resulted in 4,500 DDO lemmas (of 97,500 in the dictionary). The lemma intersection (86%) with ODS was selected for our task.

**DUTCH** We used the *Woordenboek der Nederlandsche Taal* (Dictionary of the Dutch Language, WNT) [9] and the *Algemeen Nederlands Woordenboek* (Dictionary of Contemporary Dutch, ANW)[10]. The Dutch lemmas were selected based on the Danish lemma list due to the close relationship between the two languages, facilitated by the information on the English equivalents from the Princeton WordNet.

**ENGLISH (KD)** We used the Password and Global dictionary series provided by K Dictionaries through Lexicala[11]. However, due to copyright limitations we are not able to expand this benchmark.

7 https://bg.wiktionary.org
8 https://ordnet.dk/ods_en
9 http://gtb.ivdnt.org/search
10 http://anw.ivdnt.org/search
11 https://www.lexicala.com/

**ENGLISH (NUIG)** We developed a second English dataset using Princeton WordNet (Miller, 1995) (Fellbaum, 2010) and the public domain version of Webster's dictionary from 1913[12].

**ESTONIAN** We used the EKS Dictionary of Estonian and the PSV Basic Estonian Dictionary (Kallas et al., 2014).

**GERMAN** We used the German versions of OmegaWiki[13] and Wiktionary[14].

**HUNGARIAN** We linked the Explanatory Dictionary of Hungarian (1959-1962)[15] containing 60,000 entries and, the Comprehensive Dictionary of Hungarian[16] containing 110,000 entries since 2006. Both are typical academic dictionaries.

**IRISH** We used the Wiktionary data[17] and *An Foclóir Beag* (Dónaill and Maoileoin, 1991, 'The Little Dictionary'), the only two monolingual dictionaries available for this language.

**ITALIAN** We used ItalWordNet (Roventini et al., 2000) and SIMPLE (Lenci et al., 2000).

**SERBIAN** We used the Serbian WordNet (Krstev et al., 2004; Stanković et al., 2018) and the *Rečnik Matice srpske I-VI: Rečnik srpskohrvatskog književnog jezika* (Dictionary of the Serbo-Croatian Literary Language).

**SLOVENE (JSI)** Slovene WordNet (Erjavec and Fiser, 2006) and Slovene Lexical Database (Gantar and Krek, 2011) were used.

**SLOVENE (ISJFR)** eSSKJ–Dictionary of the Slovenian Standard Language (3rd edition) (Gliha Komac et al., 2016) and the *Kostelski slovar* (Gregorič, 2014) were aligned.

**SPANISH** The *Diccionario de la lengua española* (2011 edition) (RAE, 2001) was linked with the entries in the Spanish Wiktionary[18] (backup dump of late August 2019) sharing the same lemmas.

**PORTUGUESE** *Dicionário da Língua Portuguesa Contemporânea* (DLPC, (Academia das Ciências de Lisboa, 2001)) and *Dicionário Aberto* (DA)[19] were used.

**RUSSIAN** Ozhegov and Shvedova's "The Dictionary of the Russian Language" (Ozhegov and Shvedova, 1992) and the Dictionary of the Russian Language edited by A.P. Evgenyeva, or Maliy Akademicheskiy Slovar (Short Academic Dictionary) (Evgenyeva, 1999, MAS) were used.

12 https://www.websters1913.com
13 http://www.omegawiki.org
14 https://de.wiktionary.org
15 http://mek.oszk.hu/adatbazis/magyar-nyelv-ertelmezo-szotara
16 http://nagyszotar.nytud.hu
17 https://ga.wiktionary.org
18 https://es.wiktionary.org
19 https://dicionario-aberto.net

### 5.3.4 Dataset Structure

Listing 1 presents the structure of the datasets in JSON format. External keys such as `meta_ID` and `external_ID` will enable future lexicographers to integrate the annotations in external resources. Given that some of the semantic relationships, such as `narrower` and `broader`, are not symmetric, `sense_source` and `sense_target` are important classes in determining the semantic relationship correctly.

```
{
    "lemma": "splenetic",
    "POS_tag": "adjective",
    "gender": "",
    "meta_ID": "",
    "resource_1_senses": [
        {
            "#text": "of or relating to the spleen",
            "external_ID": "splenic.a.01"},
        {
            "#text": "very irritable",
            "external_ID": "bristly.s.01"}
    ],
    "resource_2_senses": [
        {
            "#text": "affected with spleen; malicious; spiteful; peevish; fretful.",
            "external_ID": ""}
    ],
    "alignment": [
        {
            "sense_source": "very irritable",
            "sense_target": "affected with spleen; malicious; spiteful; peevish; fretful.",
            "semantic_relationship": "exact"}
    ]
}
```

**Listing 1:** An example of the structure of senses and their alignments in the datasets

Furthermore, the datasets are available in the resource description framework (RDF) and TSV at https://github.com/elexis-eu/mwsa.

## 5.4 CASE STUDIES

In this section, some of the dictionaries used in this task are described. Furthermore, some of the challenges based on the qualitative experience of the annotation teams are reported.

### 5.4.1 Danish

The *Den Danske Ordbog*[20] ('The Danish Dictionary', DDO) and *Ordbog over det danske Sprog* ('Dictionary of the Danish Language', ODS) are two monolingual dictionaries of the Danish language. While DDO covers senses of more than 100,000 Danish

20 https://ordnet.dk/ddo

lemmas from 1955 until today, ODS is a retro-digitized historical dictionary covering 220,000 Danish lemmas from 1700 until 1955. Both of the resources are considered to be key lexical resources of the Danish language. The annotation of these resources was carried out by Danish partners of the ELEXIS project at the Society for Danish Language and Literature[21] and the Centre for Language Technology[22].

Despite the significant overlaps of entries in the two resources, there are distinctive and unique features particular to any of one of them. Broadly, there are the following major differences between ODS and DDO:

### *Sense structure*

Senses in both dictionaries are provided in a hierarchy where semantically related concepts are provided as subsenses to a main sense. However, the sense granularity and the exact distinctions drawn between both main senses and subsenses of a lemma differ across the two monolingual dictionaries. The order of main senses and subsenses in ODS is based on etymology while in DDO it is based on corpus frequency. DDO establishes only main senses proved by concrete textual examples before the closely-related senses are listed in the form of subsenses. These might represent either a broader, a narrower, or a figurative use to a higher degree of the main sense, and also have to be manifested in concrete examples in the language. On the contrary, ODS operates with main senses in the structure which are rather a kind of heading or very broad *summing up* sense description for a series of subsenses to be listed, which are then the only ones to be manifested in concrete language. Consequently, two main senses in ODS along with their subsenses are semantically related to the same one sense in DDO. Therefore, ODS splits in more senses than DDO does.

When it comes to sense granularity, the two dictionaries also differ in other ways. For instance, ODS often merges content that DDO would instead express as several senses in the structure. This is carried out by using formulations like '*også om*' ('also about'), '*dels .., dels*' ('both .. and') and '*også uegentl.*' ('also figurative') in one single sense definition in ODS. Furthermore, limitations regarding material might have influenced the sense granularity of the two dictionaries. For instance, DDO editors were often encouraged to rather lump the senses of the less frequent lemmas due to the limited space in the printed edition; this particularly influences cases of regular polysemy.

### *Definition content*

The time span between the edition of the first volumes of ODS and the most recent edition of lemmas in DDO is 100 years. This leads to many differences in lexicographic description style making differences even more salient. The definition style of ODS is compact, aiming at presenting as many details as possible in one phrase.

21 https://dsl.dk
22 https://cst.ku.dk

The editors of DDO focused, instead, on the communicative qualities of a definition. Where ODS uses many parentheses, additional words and phrases and a deep syntactic structure with many attributives and subordinate phrases in order to try to cover all aspects of a sense, DDO focuses on the prototypical aspects and prefers a more flat syntactic structure. In addition, when DDO makes use of supplementary explanations, these are easily identifiable as they are initiated by a semicolon in the definition text, or being placed in two separate XML-fields, one for connotative and the other for encyclopedic information. Table 5.2 shows an example in which the definition of the same sense in DDO and ODS is provided.

| *lukke* ('to close') | | |
|---|---|---|
| ODS | DA | *trække, lægge, skyde hen for (over) en aabning, saaledes at denne spærres, udfyldes, tilstoppes; især m. h. t. et dertil beregnet og anbragt (i aabningen passende) spærremiddel, fx. klap, lem, dør; m. h. t. dør olgn. ogs. undertiden: (trække til og) laase ell. stænge* |
| | EN | 'pull, place, shoot over (over) an opening so that it is blocked, filled out, clogged; in particular w.r.t. a specially designed and arranged (in the aperture) blocking means, e.g. a clap, limb, door; w.r.t. doors or the like also sometimes: (pull and) lock or close' |
| DDO | DA | *bevæge noget dertil indrettet hen foran eller hen over en åbning så den spærres* |
| | EN | 'move something to the front of or across an opening to lock it' |

**Table 5.2:** Definition of the same sense of the verb *lukke* ('to close') in DDO and ODS

Similarly, when it comes to the content of the definition, many differences are either due to the time span between the edition of the two dictionaries or simply to the lexicographer's individual preference in each case. For instance, in the definition of the lemma *standpunkt* (noun, 'view') in Table 5.3, there is no word at all in common between the two definitions, even though they convey the same meaning:

Furthermore, Table 5.4 illustrates that definitions might also focus on different aspects of word meaning, i.e. different qualia roles (Pustejovsky, 1995, p. 76) (See Section 2.5). In ODS, *honning* ('honey') is described by focusing on how it is *produced*, i.e. the agentive role: "factors involved in its origin or bringing it about", while in DDO it is defined mainly by focusing on how it is *used*, the telic role: "its purpose and function" having as a consequence that the resulting definitions become different. That said, there are many parallel definitions in the two dictionaries, both with respect to the syntactic style and lexical choice with some differences in morphology. Table 5.5 shows one of such cases. It is important to note that there are some cases in ODS where meta-information in the form of precise sense references, such as numbers, are provided for words in the definition text itself. This mostly concerns some morpho-logical forms and semantic relations, such as synonyms. Moreover, some senses in ODS are only provided with examples and no definitions.

| *standpunkt* ('view') | | |
|---|---|---|
| ODS | DA | *om en persons åndelige stade som forudsætning ell. baggrund for hans anskuelser, synsmåde ell. handlemåde; synspunkt; ogs. om den anskuelse, hvortil man er kommet, det grundsyn, man anlægger på noget, ell. (i videre anv.) om stadium ell. trin i en persons åndelige ell. sociale udvikling ell. i en sags, et forholds udvikling olgn.* |
| | EN | 'about a person's spiritual state as a prerequisite or background to his views, mode of view or mode of action; point of view. about the view to which one has come, the basic view that one is applying to something, or (further use) about the stage or step of a person's spiritual or social development or the development of a case, a relationship, etc.' |
| DDO | DA | *opfattelse af og holdning til et bestemt spørgsmål el. anliggende* |
| | EN | 'perception of and attitude to a particular issue or matter' |

**Table 5.3:** Definition of the same sense of the noun *standpunkt* ('view') in DDO and ODS

| *honning* ('honey') | | |
|---|---|---|
| ODS | DA | *plantesaft, der er opsuget af bier, omdannet i deres tarmkanal og atter gylpet op* |
| | EN | 'sap/plant juice which is soaked up by bees, transformed in their intestinal tracts and regurgitated' |
| DDO | DA | *sød klæbrig masse som bier danner af blomsters nektar, og som fx spises på brød eller bruges som ingrediens i mad* |
| | EN | 'sweet sticky mass that bees form from the nectar of flowers and which for example is eaten on bread or used as an ingredient in food' |

**Table 5.4:** Definition of the same sense of the noun *standpunkt* ('view') in DDO and ODS

Finally, there are some divergences in the orthography of the two dictionaries due to a Danish language spelling reform in 1948 where for example the letters 'aa' were replaced by a new letter 'å'. For example in Figure 5.2, the spelling of '*aabning*' and '*laase*' are respectively changed into '*åbning*' and '*låse*'. Many abbreviations are also spelled differently in the two dictionaries– for example, '*p. gr. af*' (standing for '*på grund af*' meaning 'because/due to' in ODS is replaced with '*pga.*' in DDO.

### *Data structure*

Both DDO and ODS are available in XML. In the XML structure, ODS has been enriched using semi-automatic methods with links at the lemma level to a number of other historical dictionaries covering Danish before 1700 (Pedersen et al., 2009). On the other hand, many other lexical resources for modern Danish are all linked with DDO, not only at the lemma level, but also at the sense level. This means that DDO shares sense ID numbers with not only *Den Danske Begrebsordbog*–a Danish thesaurus (Nimb et al., 2014) but also with the Danish WordNet DanNet and the Danish FrameNet lexicon (Nimb, 2018).

| *sikre* ('to secure') | | |
|---|---|---|
| ODS | DA | ***beskytte*** *en ell. noget* ***mod angreb****, skade,* ***overlast****, forstyrrelse olgn. v. hj. af* ***forebyggende foranstaltninger*** |
| | EN | 'protect somebody or something from attack, injury, nuisance, disruption, etc. using preventative measures' |
| DDO | DA | ***beskytte mod angreb****,* ***overlast****, forringelser e.l. vha.* ***forebyggende foranstaltninger*** |
| | EN | 'protect against attack, nuisance, deterioration etc. using preventive measures' |

**Table 5.5:** Definition of the same sense of the noun *sikre* ('to secure') in DDO and ODS with overlaps

Given the disparities between the content and the structure of the two resources, the XML structures are not fully compatible. In fact, the two dictionaries highly differ when it comes to the number of markups in the XML structure. While DDO has been initially created and edited in a fine-grained XML structure with isolated content-named elements, e.g. one for the definition, another for the citation, etc., constituting the perfect basis for the later online edition, ODS has been retro-digitized based on the printed version in order to be published online and has yet to be fully transformed into a well-defined XML structure. The definition text from ODS is often initiated by different types of meta-information such as frequency, chronology, domain, as well as usage, which is not part of the DDO definition text. Furthermore, meta-information can even be part of the definition text itself, as described above.

#### *Manual annotation*

Regarding the linking process between the senses of the two dictionaries, all senses and sub-senses within the sense hierarchy are brought together at the same level. This facilitated the annotation task as all possibilities could be visually taken into account easily. However, such a relaxation over the hierarchy may result in semantically less-representative senses.

In the manual annotation process, senses of lemmas of the same etymology were annotated. Such lemmas were picked out randomly among a selection of core concept lemmas, already having been identified in DDO, constituting a total of 4,646 DDO lemmas of which at least one sense constitutes the Danish equivalent of one of the 5000 base concept synsets in Princeton Wordnet (Pedersen et al., 2019). Approximately 75% of these DDO core concept lemmas are polysemous, and even though they only constitute 5% of the total number of lemmas in the dictionary, they cover more than 20% of its senses. The lemma selection thereby represents a high degree of polysemy which makes it highly suitable for our task. The DDO core concept lemmas cover both nouns, verbs, adjectives and adverbs, and 86% of them have a lemma match in ODS, confirming that even though the DDO core concept lemmas were selected via an English selection, they are in fact central lemmas also in the Danish language. We excluded senses from fixed expressions in our dataset. Table 5.6 summarizes the sense statistics of the annotated data set.

| pyramide (sb.)-19036640 | | | |
|---|---|---|---|
| 1) (massivt) bygningsværk af sten med firkantet grundflade og trekante | exact | 4- | 1-todimensional figur der har form som en trekant med s |
| 2) (mat.) legeme, hvis grundflade er en polygon, og hvis trekantede sid | exact | 3- | 2-bygning el. konstruktion med form som et sådant grav |
| 3) hvad der har form af en pyramide ell. kegle; ogs. i videre anv., om hv | exact | 2- | 3-rumlig geometrisk figur der fremkommer ved at der fra |
| 3. 1 ) om (del af et) bygningsværk (tårn, spir olgn.); nu især (jf. Pyramid | narrow | 2- | 4-egyptisk gravmonument, ofte af meget store dimensio |
| 3. 2 ) (især gart.) om træer (sjældnere andre planter). Pyramideaster, d | narrow | 2- | |
| 3. 3 ) om (lille) pyramideformet ting ell. figur; fks. til havepynt: små Pyra | narrow | 2- | |
| 3. 4 ) opstabling, opstilling af ting, der har form som en pyramide, tilspic | narrow | 2- | |
| 3. 5 ) om (del af) møbel (hylde, opsats olgn.), der tilspidses opefter; spe | | | |
| 3. 6 ) (fagl.) krystalform bestående af to mod hinanden vendte pyramide | | | |
| 3. 7 ) (anat.) om forsk. fremspring olgn. d. s. s. Nyrepyramide . Anat.(18 | | | |
| 3. 8 ) om (konkylier med stærkt opsvulmet nederste vinding af) forsk. f | | | |

**Figure 5.3:** The sense definition of the noun *pyramide* ('pyramid') in ODS (column 1 to the left) and DDO (column 4 to the right) in the annotation process.

| Resource | Nouns | Verbs | Adjectives | Adverbs | Other | All |
|---|---|---|---|---|---|---|
| ODS | 2,176 (282,040) | 983 (119,163) | 436 (60,599) | 0 (0) | 0 (0) | 3595 (461,802) |
| DDO | 1,036 (12,326) | 383 (4,045) | 248 (2,228) | 0 (0) | 0 (0) | 1667 (18,599) |

**Table 5.6:** The statistics of the annotated data for Danish. The numbers in parentheses refer to the overall number of the tokens in the senses.

In the manual annotation task, the hierarchical sense structure in ODS including the main sense as well as subsense numbers is visible to the annotator while the DDO senses are presented in a random linear order with no information on the original sense numbers and hierarchical relations between senses. This facilitated the manual linking process since cases of potentially very different hierarchies in the two dictionaries did not disturb the picture. Figure 5.3 shows the annotation environment where sense definitions of the noun *pyramide* ('pyramid') in ODS (column 1 to the left) and DDO (column 4 to the right) is carried out. The aligning values such as 'exact' and sense numbers, e.g. '4' are annotated in columns 2 and 3. Sense numbers in DDO are ad hoc, and the order does not correspond to the one in the dictionary.

The selection of semantic relations is carried out as follows (examples are translated from Danish):

- **none**: There is no match for this ODS sense in DDO
- **exact**: The sense in ODS corresponds to the sense in DDO, for example, the definitions are simply paraphrases, as seen in the examples in Figure 5.5, or they describe the same concept in rather different ways, as seen in the examples in Figures 5.2, 5.3 and 5.4. Senses are also considered to be exact matches in cases where the only difference is due to the modernization of the society. For instance, the ODS sense of the noun '*passager*' ('passenger') 'person traveling with mail coach etc.' was considered an exact match to the DDO sense 'person traveling with private or public means of transportation'.

- **broader**: The sense in ODS completely covers the meaning of the sense in DDO but is also applicable to further meanings. For instance, the ODS sense of the noun '*værge*' ('guardian'): 'a guardian of anything or anybody' is a broader sense of the DDO sense restricted to 'a guardian in legal context' (i.e. a guardian for a child not yet legally competent or for an incapacitated adult).

- **narrower**: The sense in ODS is entirely covered by the sense of DDO and is also applicable to further meanings. In ODS the adjective '*spids*' ('sharp') has, for example, two specific senses, one about a sound and another one about a smell, where DDO covers both senses in one definition: 'pungent in an unpleasant way (about smell, taste or sound)'. Therefore, both ODS senses are considered to be narrower than the merged DDO sense.

- **related**: Annotators found this relationship to be the vaguest one. There are cases when the senses may be related even though the definitions in ODS and DDO differ in *key aspects*. For example, the property of 'being able to sleep', a sense of the noun *søvn* ('sleep') in ODS is considered 'related' to 'the state of sleeping' sense in DDO, however not identical. The noun '*bamse*' ('teddy bear') in ODS described as a 'fat, clumsy person, especially a child', is defined as a 'fat, good-natured person' in DDO. As such, these two senses are also considered to be related. Also, cases of regular polysemy are considered to be 'related' matches. For instance, ODS has only one sense for the noun '*ambassade*' ('embassy'), namely the organization sense, while DDO has two: the organization sense as well as the building sense. While the organization sense is an exact match to the sense in ODS, the building sense is considered to be only 'related' to it.

### 5.4.2 Italian

Regarding Italian, two Italian language lexical resources were selected: ItalWordNet and SIMPLE. The annotation task was carried out by ELEXIS partners at ILC-CNR[23].

ItalWordNet (Roventini et al., 2002) is a lexical semantic network for Italian which is part of the WordNet family (Miller, 1995). As such, it is organized around the notion of a synset of word senses and the network structure based on lexical semantic relations which hold between senses across synsets. The 50,000 Italian synsets contained in ItalWordNet are linked to the Princeton Wordnet. On the other hand,

23 http://www.ilc.cnr.it

SIMPLE constitutes the semantic level of a quadripartite Italian lexicon: its structure is inspired by the Generative Lexicon theory (Pustejovsky, 1995) and in particular the notion of qualia structure which is used to organize the semantic units which constitute the basic structures representing word senses. SIMPLE contains 20,000 semantic units and we used the definitions of such semantic units for the task. Both lexicons share a set of common base concepts that provided the basis of a previous semi-automatic mapping of the two lexicons on the basis of their respective ontological organizations (Roventini et al., 2007; Roventini and Ruimy, 2008). Although this mapping did not make the five-fold distinction of semantic relations, i.e., exact, narrower, broader, related, and none, it did constitute a useful starting point and a basis for comparison for the task.

The teams that had originally compiled ItalWordNet and SIMPLE shared many members in common and thus, the definitions for corresponding senses across the two lexicons are sometimes very similar or differ solely on the basis of an additional clause. This made it easy to determine, in many cases, if two senses were 'exact' matches or if one was 'broader' or 'narrower' than the other by just comparing strings. The applicability of the 'related' category was less clear than the others but the annotator made use of it in cases where two senses referred to different concepts which did not match but were semantically related, as well as in cases of metaphoric senses in which one sense refers to the concrete and the other to the metaphorical meaning.

It was also reported that the annotators found the most challenging aspect of the task to lie in the necessity of having to choose the type of matching relationship from out of the five options available. This choice was not always an intuitive one and the procedure often called for careful analysis in order to achieve as objective an assessment of the case under consideration as possible. The annotators also found it useful to consult other lexical resources, in particular two online versions of the well known *Treccani*[24] and *Garzanti*[25] reference dictionaries.

### 5.4.3 Portuguese

For the Portuguese language, two dictionaries are used: the *Dicionário da Língua Portuguesa Contemporânea* (Dictionary of Contemporary Portuguese Language, DLPC) (Academia das Ciências de Lisboa, 2001) which the first complete edition of a Portuguese Academy dictionary containing around 70,000 entries, and the *Dicionário Aberto* (DA) (Simões and Farinha, 2010), a retro-digitized Portuguese language dictionary based on the *Novo Dicionário da Língua Portuguesa* ('New dictionary of the Portuguese language') which is in the public domain (the 1913 edition). Over a few years, the dictionary has expanded by including more entries. After the complete transcription, the dictionary was subject to automatic orthography update and was

24 http://www.treccani.it
25 https://www.garzantilinguistica.it

used for different experiments regarding NLP tasks, such as the automatic extraction of information for the creation of Wordnets or ontologies (Gonçalo Oliveira, 2018; Oliveira and Gomes, 2014). The current version of DA contains 128,521 entries. Although the number of entries seems high, it is necessary to bear in mind that this resource registers orthographic variants of the same entry. Concerning formats, both Portuguese language resources are available in printed editions and XML versions using a slightly customized version of the P5 schema of the Text Encoding Initiative (TEI) (Simões et al., 2016).

#### *Sense structure*

DLPC's micro-structure is more complex than the DA's having more structured and hierarchical information. Both dictionaries follow lexicographic conventions such as bold type in headwords. Nevertheless, comparing the sample of entries, we may observe certain typographic differences as in initial lowercase entries in DLPC in comparison to the capitalized entries of DA. Furthermore, only DLPC provides full pronunciation information. The DLPC etymological information figures after the grammatical properties of the lexical item while, in DA such information, appears at the end of the entry. While DLPC indicates the part-of-speech and gender, DA displays the gender in the case of nouns. This is a common lexicographic practice where masculine is marked as *m.* and therefore, explicitly specified that the entry is a noun.

One of the main features of DLPC is the organization of entries. Not only etymological homonyms are treated as independent entries, but also homonyms of the same etymological family belonging to different part-of-speech classes are differentiated by numeric superscripts to the right of the lemma in order to distinguish the respective entries. For instance, *perfurador* can function as an adjective meaning 'perforating', or a noun meaning 'punch' so is defined into two entries.

#### *Definition content*

Senses are enumerated in DLPC providing more organized and fine-grained information while in DA, senses are specified by newlines. This is the result of the lack of metadata added to the dictionary during the retro-digitization and transcription process. Nevertheless, the dictionary has the basic microstructure annotated including grammatical information, definitions, quotations, usage examples and etymological information. DLPC is more comprehensive regarding semantic information such as synonyms (preceded by ≈), examples (shown in italics), cross-reference to lexical units that preferentially co-occur (shown by the symbol +), usage labeling, among other relevant features.

One of the main differences between sense definitions in these resources is due to the time span between the two dictionaries: DLPC was published in 2001 and the DA in 1913. The Portuguese lexicon and language have undergone many transformations over that period of time, particularly the Portuguese spelling reform, e.g. '*periphrástico*' vs '*perifrástico*' ('periphrastic') respectively in the old and the modern orthographies, and semantic changes of many lexical items such as '*computador*' ('computer') which is not defined as an electronic device in DA. Moreover, new terms have entered general vocabulary such as '*futebol*' ('football') which is not included in the DA.

### *Manual annotation*

A set of lemmas are selected in both dictionaries in a random way and alphabetically. The randomness of the selection was conditioned by the fact that the same headwords along with their part-of-speech tags, such as '*banco*' ('bank') and '*tripeiro*' ('tripe seller and native of Porto'), appear in both dictionaries. The selection of entries took into account some points previously defined in the project, namely that words belonging to an open-class part-of-speech tag should be represented and also, monosemous and polysemous lemmas should appear.

There are several cases where there are exact relations. Of such cases are entries for which only one sense is defined. For instance, sense 1 of DLPC in Figure 5.4 shows the sense of the word '*banderilla*' in the bullfighting domain which can possibly correspond to the only sense of DA. On the other hand, sense 2 related to the bookbinding domain only appears in the DLPC and not DA. Nevertheless, the definition of identical senses is not similar in textual terms as they are described differently. DLPC also uses domain labels while in the DA, there is no label. In other cases, the correspondence of senses is evident but the lexicographic criteria adopted differ as shown in Figure 5.5. The structure of these lexicographic articles is different as DLPC has two entries for *tripeiro* ('*tripeiro*$^1$' ('tripeman') and *tripeiro*$^2$, 'of or pertaining to the city of Oporto') as an adjective and a noun, i.e. part-of-speech homonyms. On the other hand, DA has only one entry and gender information. Between *tripeiro*$^2$ (DLPC) and *tripeiro* (DA), there is an exact match in the first sense which is an obsolete one. However, the technique of writing the gloss differs: '*Pessoa que vende tripas*' ('Person who sells tripes') in DLPC versus '*Vendedor de tripas*' ('Tripe seller') in DA. Given that sense definitions are not enumerated in DA, each paragraph is considered an independent sense. This way, the sense definition starting with '*pop.*' (popular) in DLPC can be aligned with the one starting with '*Deprec.*' (depreciative) in DA.

One of the challenging alignment cases happens when a sense definition may correspond to more than one sense definition in the other resource. For instance, '*praia*' ('beach') entry in the sense of '*Beira-mar*' ('seaside') (Figure 5.6); in DA, the senses '*Beira-mar*' ('seaside') and '*Região, banhada pelo mar; litoral; margem*' ('Region, bathed by the sea; coast') correspond to sense 2 of DLPC: '*Zona banhada pelo mar; zona balnear*' ('Zone bathed by the sea; bathing area'). In such cases, the annotator's evaluation of the definition determines which relation is more likely to be selected.

**bandarilha** [bɐ̃dɐríʎɐ]. *s. f.* (Do cast. *banderilla*). **1.** *Taurom.* Haste munida de ponta de metal penetrante, enfeitada com uma bandeira ou com fitas de papel de cores e que se espeta no cachaço dos touros, durante a corrida. ≃ FARPA, FERRO. *A elegância com que espetou o par de bandarilhas no touro pôs a praça de pé. «Abrem-se então as portas e a manada entra, esta que será toureada hoje consoante os preceitos inteiros da arte, passada à capa, espetada de bandarilhas, castigada de varas»* (SARAMAGO, *Levantado do Chão*, p. 165). *Cravar, espetar as +s; um par de +s; tércio de +s.* **bandarilhas a quarteio,** variedade de farpas em que o toureiro faz um quarto de volta ao espetá-la no touro. **bandarilhas a recorte,** movimento que consiste em colocar os ferros no touro no momento em que o toureiro evita a marrada. **2.** *Encad.* Tira de papel que se cola na margem de um original ou prova, quando as emendas não cabem nas margens.

Entrada

Bandarilha

f.
Farpa, enfeitada com bandeiras ou fitas, e destinada a cravar-se no cachaço dos toiros, quando se correm.
(Por *bandeirilha*, cast. *banderilla*)

Figure 5.4: '*bandarilha*' '(banderilla') in DLPC (above) and DA (below)

**tripeiro**[1] [tripɐ́jru]. *adj. m.* e *f.* (De *tripa* + suf. *-eiro*). *Pop.* O m. que *portuense*[1].
**tripeiro**[2] [tripɐ́jru]. *s. m.* e *f.* (De *tripa* + suf. *-eiro*). **1.** Pessoa que vende tripas. **2.** *Pop.* O m. que *portuense*[2].

Entrada

Tripeiro

m.
Vendedor de tripas.
Aquele que se sustenta de tripas.
*Deprec.*
Habitante do Porto.
(De *tripa*)

Figure 5.5: '*tripeiro*' ('tripe seller and native of Porto') in DLPC (above) and DA (below)

**praia** [prájɐ]. *s. f.* (Do lat. tardio *plagia*, talvez do gr. πλάγιος 'oblíquo'). **1.** Faixa arenosa do litoral marítimo, de fraca inclinação, muito utilizada por banhistas nas zonas de veraneio ou em estâncias de turismo. *«e a débil pegada que o meu obscuro pé imprimiu nas praias do Mindelo há-de ficar gravada na história»* (GARRETT, *Discursos*, p. 121). **casa**+ **de praia. colchão**+ **de praia. voleibol**+ **de praia. 2.** Zona banhada pelo mar; zona balnear. ≃ BEIRA-MAR, COSTA, LITORAL. *Passaram as férias na praia.*

Entrada

Praia

f.
Orla de terra, geralmente coberta de areia, confinando com o mar.
Beiramar.
Região, banhada pelo mar; litoral; margem.
Pl. *Marn.*
Depósito geral das águas que alimentam a salina, e que também se chama loiças, (cp. *loiça*).
(Do lat. *plaga*)

Figure 5.6: '*praia*' ('beach') in DLPC (above) and DA (below)

# 5.5 EVALUATION

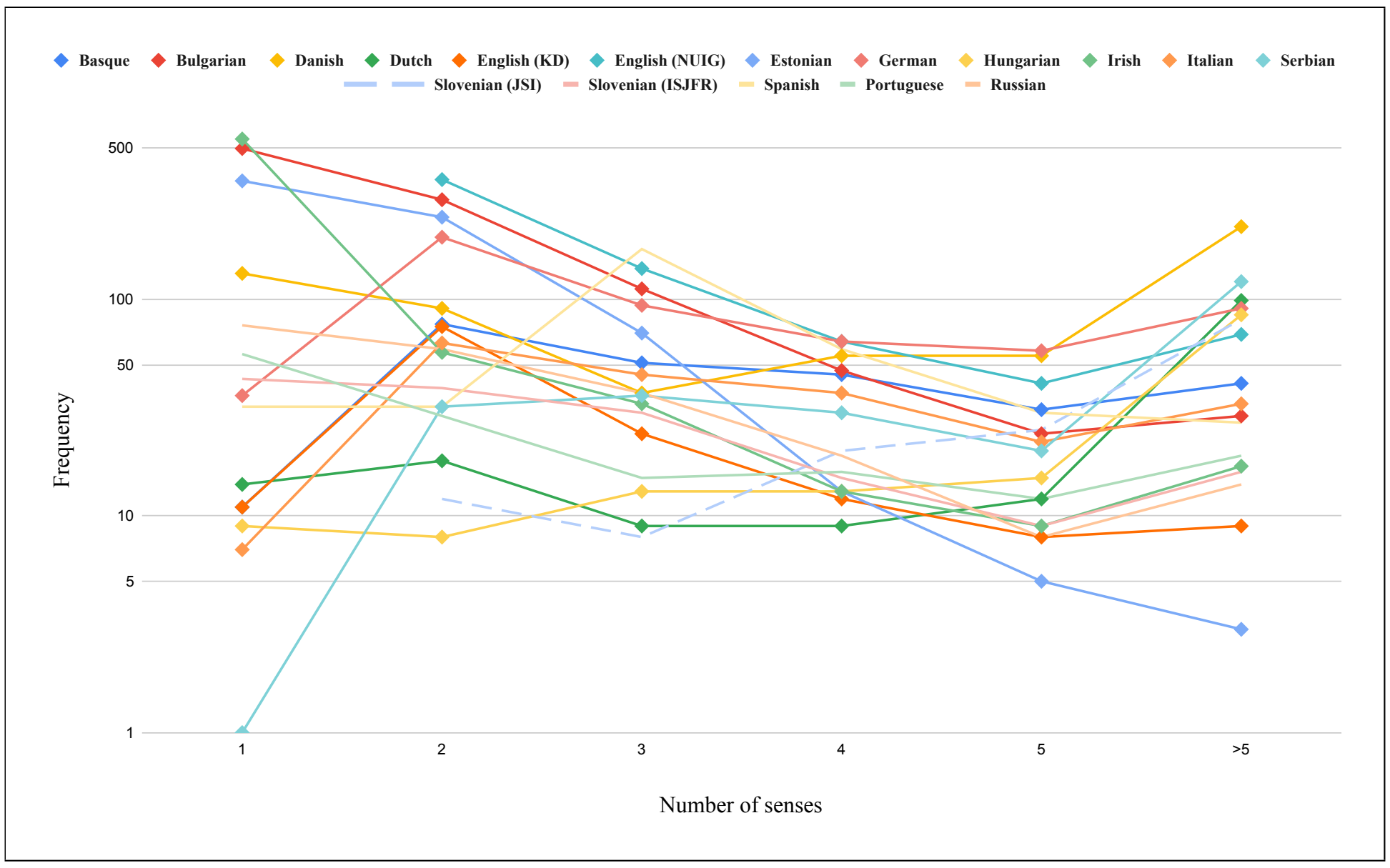

**Figure 5.7:** Frequency of the number of senses in the datasets per language in the left resource

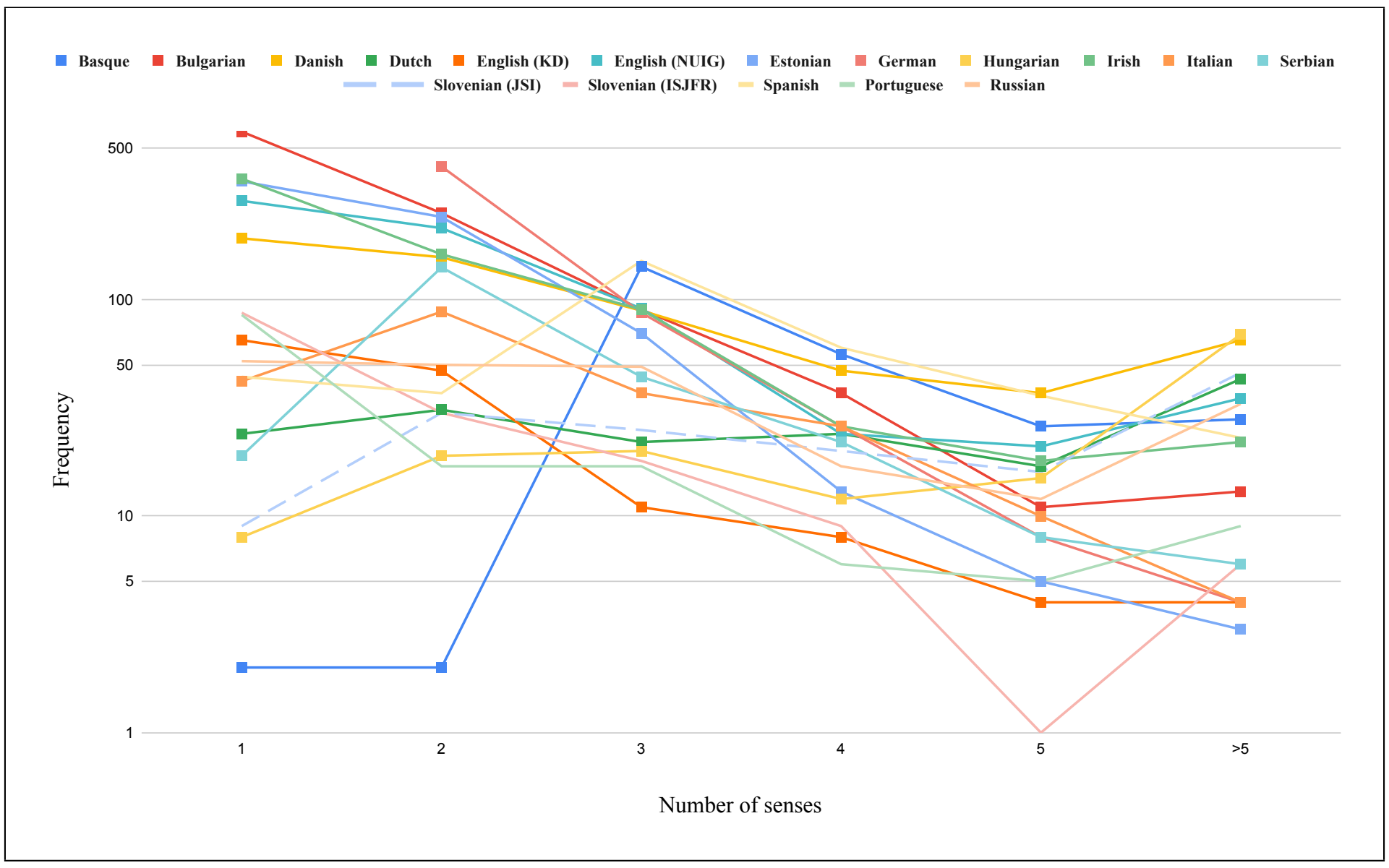

**Figure 5.8:** Frequency of the number of senses in the datasets per language in the right resource

We performed an intrinsic evaluation on our datasets by computing a number of resource statistics on the senses. Table 5.7 provides resource statistics based on part-of-speech tags and languages. As most of the lemmas available in the resources belong to open classes, namely nouns, verbs, adjectives and adverbs, we carried out our experiments with respect to those part-of-speech tags. Moreover, there are few languages, such as German, Italian and Serbian, for which only a certain number of the part-of-speech tags are available. As a unique case, the Hungarian entries are aligned at lemma-level without taking the POS tags into account; the POS tags are provided within the senses, upon the lexicographers' request.

Moreover, the distribution of the frequency of the number of senses of the left resource (source senses) and the right resource (target senses) are respectively presented in Figures 5.7 and 5.8, where we show for each resource how many entries had 1, 2, 3, 4, 5 or more senses.

### 5.5.1 Sense Granularity

The granularity of senses is a determining factor in applying automatic approaches for semantic similarity evaluation. Sense granularity does not follow an identical pattern across resources and languages. The type of the resource, the preference of the lexicographer and the historical period of the resource edition are some of the factors on how senses are shaped.

Figure 5.9 illustrates the number of tokens in the first and second resources of the languages provided in our datasets. The bigger the difference between the number of tokens, the more the difference between the definitions and length of senses. As an additional analysis, we divide the number of space-separated tokens in one of the resources by the other resource. Although most of the resources have a ratio of [1, 2] which indicates a relatively similar granularity of senses in the two resources, Danish and English (NUIG) represent higher ratios. In the case of Danish, a correlation score of 24.8 demonstrates a huge difference in how senses are expanded in the resources. This can be justified by the fact that ODS as a historical resource provides many senses which are no longer used in the language. In addition, the structure of the resource is in a way that further details are provided at sense-level rather than separately.

### 5.5.2 Sense Alignments

One of the main challenges in aligning senses is due to the structure of the senses. A resource that provides senses in a hierarchy based on main senses and their subsenses represents semantically context-dependent senses in comparison to one in which senses are *semantically independent*, which are stand-alone senses not influenced by the hierarchy. On the other hand, senses may contain descriptions beyond the definition, such as usage examples and idioms.

| Language | Resource | Nouns | Verbs | Adjectives | Adverbs | Other | All |
|---|---|---|---|---|---|---|---|
| Basque | Basque Wordnet | 929 (6836) | 0 (0) | 0 (0) | 0 (0) | 0 (0) | 929 (6836) |
| | *Euskal Hiztegia* | 971 (7754) | 0 (0) | 0 (0) | 0 (0) | 0 (0) | 971 (7754) |
| Bulgarian | BTB-WN | 1394 (15649) | 175 (1698) | 305 (3187) | 50 (338) | 0 (0) | 1924 (20872) |
| | Bulgarian Wiktionary | 1273 (12883) | 164 (1107) | 194 (1418) | 39 (306) | 0 (0) | 1670 (15714) |
| Danish | *Ordbog over det danske Sprog* | 2176 (282040) | 983 (119163) | 436 (60599) | 0 (0) | 0 (0) | 3595 (461802) |
| | *Den Danske Ordbog* | 1036 (12326) | 383 (4045) | 248 (2228) | 0 (0) | 0 (0) | 1667 (18599) |
| Dutch | *Woordenboek der Nederlandsche Taal* | 1459 (28979) | 405 (5185) | 527 (7878) | 106 (2662) | 0 (0) | 2497 (44704) |
| | *Algemeen Nederlands Woordenboek* | 497 (8443) | 140 (1542) | 109 (1393) | 13 (172) | 0 (0) | 759 (11550) |
| English (KD) | Global | 92 (532) | 107 (617) | 80 (457) | 57 (257) | 61 (283) | 397 (2146) |
| | Password | 66 (536) | 72 (417) | 62 (324) | 33 (177) | 46 (188) | 279 (1642) |
| English (NUIG) | *Webster* | 1131 (11606) | 741 (4622) | 373 (2585) | 45 (269) | 0 (0) | 2290 (19082) |
| | *Princeton WordNet* | 730 (12166) | 496 (6980) | 249 (2892) | 24 (207) | 0 (0) | 1499 (22245) |
| Estonian | Dictionary of Estonian (EKS) | 543 (4012) | 273 (1598) | 151 (747) | 98 (451) | 78 (370) | 1143 (7178) |
| | Estonian Basic Dictionary (PSV) | 543 (4492) | 273 (1983) | 151 (1097) | 98 (596) | 79 (468) | 1144 (8636) |
| German | German Wiktionary | 2026 (15160) | 0 (0) | 0 (0) | 0 (0) | 0 (0) | 2026 (15160) |
| | German OmegaWiki | 1266 (14354) | 0 (0) | 0 (0) | 0 (0) | 0 (0) | 1266 (14354) |
| Hungarian | Comprehensive | | | | | | 1355 (14654) |
| | Explanatory | | | | | | 1038 (10934) |
| Irish | *An Foclóir Beag* | 891 (8053) | 11 (95) | 55 (267) | 10 (56) | 36 (171) | 1003 (8642) |
| | Irish Wiktionary | 1209 (6696) | 8 (45) | 61 (181) | 10 (41) | 36 (109) | 1324 (7072) |
| Italian | ItalWordNet | 408 (3128) | 352 (2411) | 0 (0) | 0 (0) | 0 (0) | 760 (5539) |
| | SIMPLE | 290 (1990) | 218 (1240) | 0 (0) | 0 (0) | 0 (0) | 508 (3230) |
| Serbian | Serbian WordNet | 691 (5864) | 985 (6522) | 92 (713) | 0 (0) | 0 (0) | 1768 (13099) |
| | Dictionary of Serbo-Croatian Literary Language | 289 (2360) | 281 (1527) | 29 (215) | 0 (0) | 0 (0) | 599 (4102) |
| Slovenian (JSI) | Slovene WordNet | 409 (1106) | 303 (901) | 237 (733) | 44 (133) | 0 (0) | 993 (2873) |
| | Slovene Lexical Database | 284 (2237) | 191 (1047) | 220 (1486) | 29 (102) | 0 (0) | 724 (4872) |
| Slovenian (ISJFR) | Standard Slovenian Dictionary (eSSKJ) | 229 (2060) | 109 (911) | 76 (620) | 0 (0) | 60 (588) | 474 (4179) |
| | *Kostelski slovar* | 151 (1050) | 61 (308) | 45 (257) | 0 (0) | 38 (263) | 295 (1878) |
| Spanish | *Diccionario de la lengua española* | 617 (7986) | 225 (2426) | 305 (3269) | 26 (161) | 24 (250) | 1197 (14092) |
| | Spanish Wiktionary | 602 (6421) | 227 (2045) | 294 (2825) | 25 (129) | 22 (123) | 1170 (11543) |
| Portuguese | *Dicionário da Língua Portuguesa Contemporânea* | 285 (4060) | 58 (686) | 110 (1287) | 9 (143) | 1 (9) | 463 (6185) |
| | *Dicionário Aberto* | 199 (1521) | 53 (203) | 67 (372) | 3 (15) | 1 (5) | 323 (2116) |
| Russian | *Ozhegov-Shvedova* | 258 (2038) | 109 (615) | 101 (533) | 15 (77) | 44 (368) | 527 (3631) |
| | Dictionary of the Russian Language (MAS) | 310 (2811) | 173 (1338) | 190 (1219) | 20 (114) | 71 (1010) | 764 (6492) |

**Table 5.7:** Statistics of the datasets. This table shows the number of senses in the resources (the number of the words in the definitions are provided in parentheses).

To evaluate the distribution of the alignments with respect to the senses, we assume that each entry is a *lexicographic network* (**Sina Ahmadi** et al., 2018), i.e., a graph where the nodes and edges are the senses and the alignments, respectively. Given a set of aligned senses, we denote the number of senses in resource 1 and resource 2 by $n_1$ and $n_2$, respectively. We also denote the number of alignments in each entry by $m$. Therefore, the average degree of senses in each resource is defined as $k_1 = \frac{m}{n_1}$ and $k_2 = \frac{m}{n_2}$. Similarly, the average degree of the whole dataset can be calculated as follows:

$$k = \frac{2 \times m}{n_1 + n_2} = \frac{n_1 \times k_1 + n_2 \times k_2}{n_1 + n_2} \tag{5.1}$$

Finally, we define the number of existing alignments divided by the number of possible alignments as the density $\delta = \frac{m}{n_1 \times n_2}$.

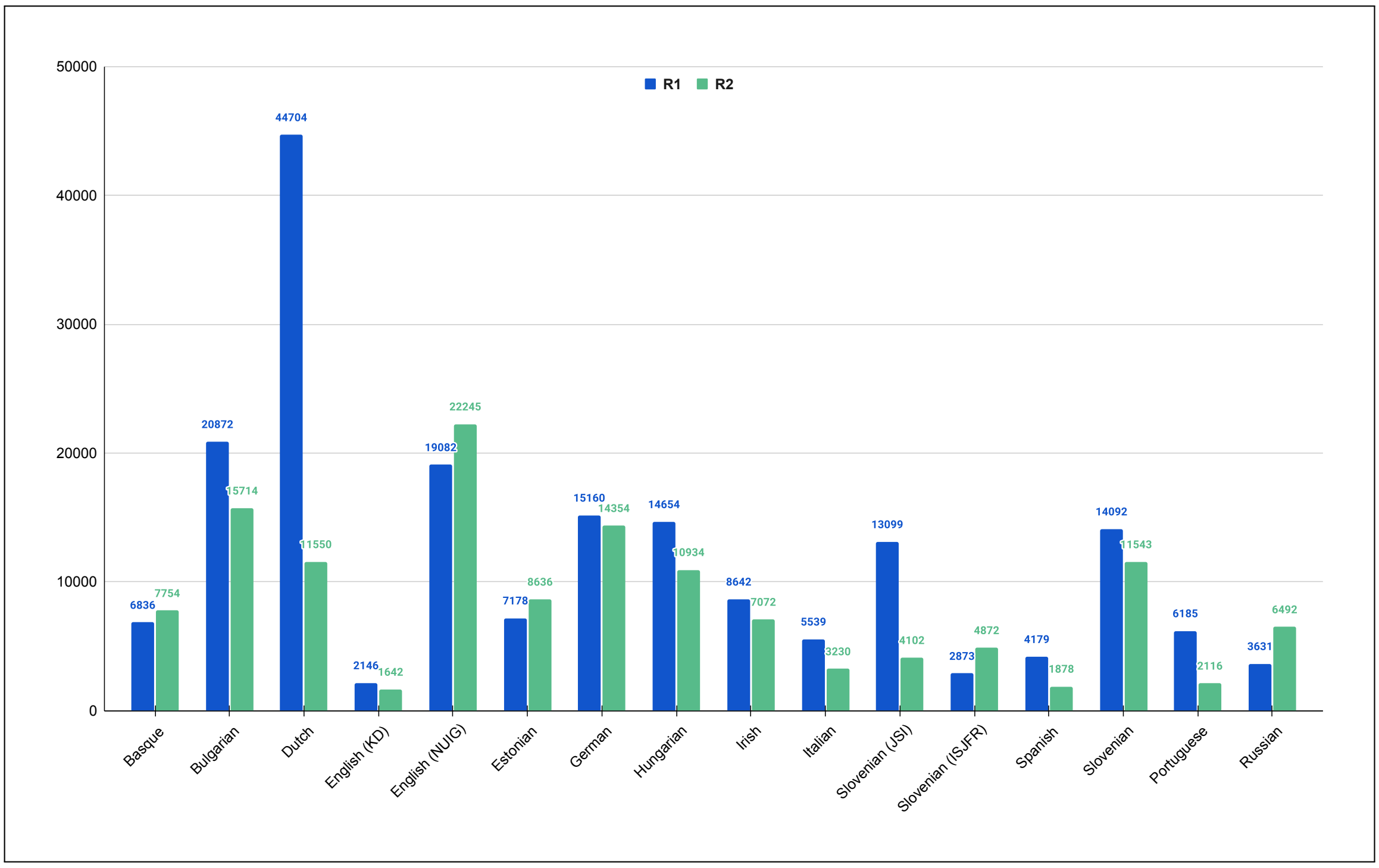

**Figure 5.9:** Number of tokens per dataset. R1 and R2 respectively correspond to the resources in Table 5.7 from up to down. Danish is removed due to its larger size.

Table 5.8 represents the results of our evaluations on the aligned senses. The degree indicates the distribution of the alignments with respect to the senses. For instance, a degree of 1.182 ($k_1$) in the case of Russian shows that every sense is at least aligned with another one. On the other hand, a low degree of 0.250 ($k_1$) in the case of Dutch indicates the sparsity of alignments over the senses. Moreover, density $\delta$ provides an insight into how alignments are distributed over the combination of all senses. In other words, a higher density represents a higher probability that two senses are aligned in the two resources. Estonian and German resources, for example, have the highest density among the resources.

| Language | Semantic relationship | | | | | $k_1$ | $k_2$ | $k$ | $\delta \times 10^3$ |
|---|---|---|---|---|---|---|---|---|---|
| | exact | narrower | broader | related | all | | | | |
| Basque | 399 | 138 | 94 | 184 | 815 | 0.877 | 0.839 | 0.858 | 0.9035 |
| Bulgarian | 958 | 274 | 254 | 492 | 1978 | 1.028 | 1.184 | 1.101 | 0.6156 |
| Danish | 1103 | 316 | 189 | 36 | 1644 | 0.457 | 0.986 | 0.625 | 0.0001 |
| Dutch | 489 | 30 | 64 | 42 | 625 | 0.250 | 0.823 | 0.384 | 0.3298 |
| English (KD) | 107 | 78 | 28 | 88 | 301 | 0.758 | 1.079 | 0.891 | 2.7175 |
| English (NUIG) | 885 | 339 | 42 | 67 | 1333 | 0.582 | 0.889 | 0.704 | 0.3883 |
| Estonian | 1025 | 61 | 54 | 4 | 1144 | 1.001 | 1.000 | 1.000 | 0.8749 |
| German | 354 | 311 | 426 | 126 | 1217 | 0.601 | 0.961 | 0.739 | 0.4745 |
| Hungarian | 465 | 214 | 227 | 43 | 949 | 0.700 | 0.914 | 0.793 | 0.6747 |
| Irish | 731 | 45 | 67 | 132 | 975 | 0.972 | 0.736 | 0.838 | 0.7342 |
| Italian | 327 | 132 | 44 | 89 | 592 | 0.779 | 1.165 | 0.934 | 1.5334 |
| Serbian | 325 | 47 | 73 | 146 | 591 | 0.334 | 0.987 | 0.499 | 0.5581 |
| Slovenian (JSI) | 306 | 183 | 169 | 54 | 712 | 0.717 | 0.983 | 0.829 | 0.9904 |
| Slovenian (ISJFR) | 110 | 88 | 10 | 39 | 247 | 0.521 | 0.837 | 0.642 | 1.7664 |
| Spanish | 867 | 185 | 114 | 93 | 1259 | 1.052 | 1.076 | 1.064 | 0.8990 |
| Portuguese | 207 | 38 | 2 | 28 | 275 | 0.594 | 0.851 | 0.700 | 1.8389 |
| Russian | 363 | 15 | 159 | 86 | 623 | 1.182 | 0.815 | 0.965 | 1.5473 |

**Table 5.8:** A description of the semantic relationship alignments using basic graph measures

### 5.5.3 Inter-annotator Agreement

Krippendorff's Alpha is a reliability coefficient to measure the agreement of two or more annotators. Reliability coefficients are an important metric to determine the difficulty of a task to be carried out by a human, and also, the agreement or potential disagreement that occur between annotators. There are mainly two families of reliability coefficients known as Kappa and Alpha. Different from the Kappa ($\kappa$) family of reliability coefficients, such as Fleiss's $k$, where the disagreements are treated equally important, Alpha ($\alpha$) coefficients are designed to incorporate a weight function that sets the level of disagreement based on each pair of labels. This is useful for cases where taking a decision regarding an annotation is not equally easy or difficult with respect to a specific class. Krippendorff's Alpha is recommended as the standard reliability measure in content analysis (Hayes and Krippendorff, 2007) and has been widely used in many annotation tasks in NLP (Artstein, 2017).

We selected Krippendorff's Alpha as the number of the annotators for some of the datasets varies between two and three, and Krippendorff's Alpha is able to handle various numbers of annotators. Moreover, Krippendorff's Alpha is able to calculate reliability at any measurement level, i.e. nominal, ordinal, interval, ratio. In addition, we were initially interested to explore the agreement score based on various weight functions. This was another reason to select this measure. However, the weight function was finally selected as the identity function where all annotations are considered equally significant.

According to Krippendorff (2011), $\alpha$ is calculated as follows:

$$\alpha = 1 - \frac{D_o}{D_e} \tag{5.2}$$

where $D_o$ is the observed disagreement among values and is defined as follows:

$$D_o = \frac{1}{n} \sum_{c \in R} \sum_{k \in R} o_{ck} \delta^2_{ck} \tag{5.3}$$

and $D_e$ is the expected disagreement which is attributable by chance:

$$D_e = \frac{1}{n(n-1)} \sum_{c \in R} \sum_{k \in R} n_c . n_k \delta^2_{ck} \tag{5.4}$$

where $\delta$ is the metric function of difference, $n$ is the number of annotators, $n(n-1)$ is the number of pairs of annotators whose annotations are being compared, $R$ is the set of all possible annotations that an annotator can select and finally, $o$ is the matrix of observed coincidences containing frequencies. As our annotations are nominal, $\delta^2_{ck}$ is 1 in the case of agreement and 0 otherwise. These parameters are calculated based on the coincidence matrices which are created according to the annotations for each pair of sense definitions.

Equation 5.2 denotes that when the agreement among annotators is the highest, observed disagreement, i.e. $D_o = 0$ and therefore, $\alpha = 1$. This indicates perfect reliability. On the other hand, when the annotations are carried out as if they were produced by chance, observed disagreement and the expected disagreement by chance are equal, i.e. $D_o = D_e$, which indicates the lack of reliability. This results in $\alpha = 0$ and denotes that there is no point of agreement between the annotators. It is worth noting that $\alpha$ varies between the range of 0 and 1, $0 \leqslant \alpha \leqslant 1$; any score out of the extremes, indicates systematic disagreement and exceeds any agreement produced by chance, i.e. $\alpha = 0$.

While the linking for most of the languages was only developed by a single annotator, we collected multiple annotations for four languages which enabled us to evaluate the alignment agreement over the same senses. Given the conditions of our annotation problem, we used Krippendorff's alpha reliability for calculating the inter-annotator agreement (IAA) where we considered each possible sense pair as an item for the agreement. Thus, if a pair of senses was not chosen by any of the annotators, they are considered to agree that the link between this is `none`. Table 5.9 presents the IAA in a 5-class model, that is the five semantic relations namely exact, narrower, broader, related and none; the latter refers to the cases where no relation is specified between a pair of senses. Moreover, we provide a 2-class model where all types of semantic relationships, namely exact, broader, narrower and related, are merged and compared against none as the other class. The weight function for all classes is considered uniform. Regarding the number of senses, 561, 4979, 185 and 270 senses were annotated by more than one annotator for English, German, Irish and Danish, respectively, which made it possible to calculate IAA.

Regarding the English (KD) resources, an internal evaluation of the annotated data with two annotators shows an agreement for 76% of the annotators.

| | Agreement (5-class) | Agreement (2-class) |
|---|---|---|
| Irish (3) | 0.83 | 0.99 |
| English (NUIG) (3) | 0.43 | 0.73 |
| Danish (2) | 0.95 | 0.92 |
| German (2) | 0.71 | 0.58 |

**Table 5.9:** Inter-annotator agreement using Krippendorff's alpha. Number of annotators provided in parentheses.

The IAA results in Table 5.9 indicate that the alignment task is not equally difficult for all resources and languages, even though there is a considerable agreement between annotators, generally speaking. The highest IAA score (0.99) belongs to the 2-class problem for Irish where the possible correspondence of a sense definition pair is determined by the annotators, regardless of the semantic relation. On the other hand, the 5-class problem for the English (NUIG) data is the lowest (0.43).

After a qualitative analysis of the English data, we believe that determining a semantic relation between sense definitions is subjective to the annotators' understanding of the sense. For instance, Table 5.10 provides a comparison between the annotations of sense definitions in Princeton English WordNet and Webster's Dictionary 1913 for the entry VULGAR (adjective); Although the two annotators agree upon one annotation pair, they are in disagreement in three other cases. Regarding the sense of 'vulgar' as 'lacking refinement', annotator 1 seems to show more indulgence in such a way that the overall meaning of the definition in resource 1 ($R1_C$) is compared with that of resource 2 ($R2_C$) ignoring semantic nuances due to additional words such as 'low' or 'good taste' in $R2_C$. On the other hand, annotator 2 considers the additional meanings, provided after a semicolon (;) in resource 2, to determine the semantic relation between the two sense definitions and therefore, decided to annotate it as narrower.

The same phenomenon can be seen in Table 5.11 for the Irish entry CAORA ('sheep', noun) where, even though the two annotators have detected the correspondence of the senses similarly, one selected narrower as the relation and the other one, exact. One obvious reason is the differences of the definition of 'sheep' in the two resources: the first resource specifies two meanings as 'domestic sheep' (*Ovis aries*) in sense $R1_A$ and 'sheep' ('*Ovis*') as a genus in sense $R1_B$ while the second resource only defines it as a genus in sense $R2_A$. Moreover, as Table 5.9 illustrates, English glosses are more descriptive being provided with synonyms or near-synonyms after semi-colons while the Irish ones are more concise. This also explains how the length of glosses has been disadvantageous to determine a more widely-agreed semantic relation between sense pairs in English.

Finally, in cases where the annotation was carried out by more than one annotator, a third annotator analyzed the annotations and resolved cases of conflict to produce a final dataset.

| VULGAR (adjective) | |
|---|---|
| **Sense definitions in resource 1, the Princeton English WordNet ($R_1$)** | |
| $R_{1A}$ | of or associated with the great masses of people |
| $R_{1B}$ | being or characteristic of or appropriate to everyday language |
| $R_{1C}$ | lacking refinement or cultivation or taste |
| $R_{1D}$ | conspicuously and tastelessly indecent |
| **Sense definitions in resource 2, Webster's Dictionary 1913 ($R_2$)** | |
| $R_{2A}$ | of or pertaining to the mass, or multitude, of people; common; general; ordinary; public; hence, in general use; vernacular. |
| $R_{2B}$ | belonging or relating to the common people, as distinguished from the cultivated or educated; pertaining to common life; plebeian; not select or distinguished; hence, sometimes, of little or no value. |
| $R_{2C}$ | hence, lacking cultivation or refinement; rustic; boorish; also, offensive to good taste or refined feelings; low; coarse; mean; base; . |
| **Annotations** | |
| Annotator 1 | $R_{1A}$ exact $R_{2A}$ |
| | $R_{1C}$ exact $R_{2C}$ |
| Annotator 2 | $R_{1A}$ exact $R_{2A}$ |
| | $R_{1B}$ narrower $R_{2A}$ |
| | $R_{1C}$ narrower $R_{2C}$ |
| | $R_{1D}$ narrower $R_{2C}$ |

**Table 5.10:** A comparison of annotations for VULGAR (adjective) in English by the two annotators

| CAORA ('sheep', noun) | |
|---|---|
| **Sense definitions in resource 1, the Irish Wiktionary ($R_1$)** | |
| $R_{1A}$ | *ainmhí leathmhór féaránach athchogantach cuideachtúil a mbíonn adharca air go minic. Gné Ovis aries. Tá a lán saghas de tugtha chun tíreachais ar fud an domhain mar gheall ar a lomra, a mbainne agus a bhfeoil.* |
| | a semi-large, ruminant, grazing, domestic animal usually with horns. Of the species Ovis aries. A large number of varieties have been domesticated for their fleece, milk and meat around the world. |
| $R_{1B}$ | *ainmhíthe den ghéineas Ovis, agus géinis gaolta leis, a bhfuil dlúthbhaint acu leis na gabhair.* |
| | animals of the genus Ovis and related genera, which are closely related to goats. |
| **Sense definitions in resource 2, *An Foclóir Beag* ($R_2$)** | |
| $R_{2A}$ | *ainmhí athchogantach de chineál an ghabhair a thugann olann agus feoil dúinn* |
| | a ruminant animal similar to a goat that gives us wool and meat. |
| **Annotations** | |
| Annotator 1 | $R_{1A}$ narrower $R_{2A}$ |
| | $R_{1B}$ narrower $R_{2A}$ |
| Annotator 2 | $R_{1A}$ exact $R_{2A}$ |
| | $R_{1B}$ exact $R_{2A}$ |

**Table 5.11:** A comparison of annotations for CAORA ('sheep', noun) in Irish by the two annotators

## 5.6 CONCLUSION AND CONTRIBUTIONS

In this chapter, we presented a set of 17 datasets for the task of monolingual word sense alignment covering 15 languages. This dataset innovates on previous datasets by focusing on general vocabulary, which is much harder to link than the focus of previous works, such as named entities. In addition to the collaboratively-curated resources such as Wiktionary, many expert-made resources are used in our datasets for the task. Given the difficulty of determining the overlap of sense definitions in various resources and languages, we created a framework where the word sense alignment task is carried out using 5 categories of semantic links based on SKOS (Miles and Bechhofer, 2009), namely exact, broader, narrower, related and not related, i.e. none. Given the significant size of the datasets, we believe that they can be beneficial not only for evaluation purposes but also for training new statistical and neural models for various tasks such as word sense alignment, semantic relationship detection, paraphrasing and semantic entailment, to mention but a few. Various challenges related to this task are described in a few case studies in Danish, Italian and Portuguese.

The inter-annotator agreement scores indicate that the task of word sense alignment is not equally challenging for all resources and all languages. For instance, there is roughly 50% agreement among the annotators of the English resources. This reveals that determining a semantic relation is indeed a challenging task even for expert lexicographers. Regardless of how the sense granularity and coverage in two resources are identical, the textual information provided in sense definitions creates further nuances in meaning which are not easily distinguishable. We provide a few examples for 'spring' (noun) as a season where further specifications in a definition, as in 'the season between winter and summer', denotes a delicately different meaning in comparison to another definition, as in 'the season of growth'.

In the following chapter, we focus on the automatic alignment of sense definitions. A few techniques are proposed where the textual, lexical and semantic information are taken into account. The performance of various methods is evaluated for the tasks of sense alignment and semantic relationship detection using these datasets. Moreover, we explore language-independent techniques to facilitate monolingual lexical data linking and increase the interoperability of monolingual dictionaries.

# 6 MONOLINGUAL WORD SENSE ALIGNMENT

In the previous chapter, we discussed the manual annotation of sense definitions in 15 languages to create a benchmark for the evaluation of word sense alignment (WSA). In this chapter, we delve more into the automatic alignment of sense definitions by introducing a tool called Naisc in which a set of techniques for semantic similarity detection is implemented, ranging from string-based similarity detection to more recent techniques using contextual embeddings, and semantic induction techniques. We will discuss how NLP techniques can be applied in the task of aligning the senses of identical lemmas in two dictionaries, a task which otherwise would be a time-consuming and difficult challenge due to the high number of senses and the heterogeneity of data, in both structure and content.

## 6.1 INTRODUCTION

Senses and definitions are important components of dictionaries where dictionary entries, i.e. lemmata, are described in plain language. Therefore, unlike other properties such as references, comparisons (*cf.*), synonyms and antonyms, sense definitions are unique in the sense that they are more descriptive but also highly contextualized. Moreover, unlike lemmata which remain identical among resources in the same language, except in spelling variations, senses can undergo tremendous changes based on the choice of the editor, lexicographer and publication period, to mention but a few factors. Ignoring the differences in dictionary structures and formats such as XML, LMF (Francopoulo et al., 2006) and Ontolex-Lemon (McCrae et al., 2017b), there are different lexicographic and logical ways for describing senses in a dictionary (Solomonick, 1996). As an example, Table 6.1 provides the senses available for ENTIRE (adjective) in various lexical resources where the predominant sense of 'whole' or 'complete' is provided in all resources. However, all resources do not equally cover specific domains such as 'botany' and 'mathematics'. Therefore, there are differences in the number of provided senses, as in Macmillan in which one sense is provided while the Oxford Dictionary provides five.

1 https://wordnet.princeton.edu
2 https://www.websters1913.com
3 https://en.wiktionary.org

| ENTIRE (adjective) | |
|---|---|
| WordNet[1] | 1- (of leaves or petals) having a smooth edge; not broken up into teeth or lobes<br>2- constituting the full quantity or extent; complete<br>3- constituting the undiminished entirety; lacking nothing essential especially not damaged<br>4- (used of domestic animals) sexually competent |
| Webster[2] | 1- complete in all parts; undivided; undiminished; whole; full and perfect; not deficient<br>2- without mixture or alloy of anything; unqualified; morally whole; pure; faithful<br>3- not gelded; – said of a horse<br>4- internal; interior. |
| Wiktionary[3] | 1- (sometimes postpositive) Whole; complete.<br>2- (botany) Having a smooth margin without any indentation.<br>3- (botany) Consisting of a single piece, as a corolla.<br>4- (complex analysis, of a complex function) Complex-differentiable on all of $\mathbb{C}$.<br>5- (of a male animal) Not gelded.<br>6- morally whole; pure; sheer |
| Macmillan[4] | 1- used for emphasizing that you mean all or every part of something |
| Longman[5] | 12- used when you want to emphasize that you mean all of a group, period of time, amount etc |
| Oxford[6] | 1- [attributive] with no part left out; whole.<br>2- Without qualification or reservations; absolute.<br>3- Not broken, damaged, or decayed.<br>4- (of a male horse) not castrated.<br>5- Botany (of a leaf) without indentations or division into leaflets. |
| Cambridge[7] | 1- whole or complete, with nothing lacking, or continuous, without interruption |

**Table 6.1:** Senses of ENTIRE (adjective) in various monolingual English dictionaries

Our focus in this chapter is on the monolingual word senses alignment (MWSA) that aims to find potential links and identify the semantic relations between senses or sense definitions within the micro-structure of two identical lemmas belonging to the same part-of-speech category. Given the setup that we described in the previous chapter, aligning sense definitions is meaningful when the alignment task is completed with semantic relations, such as exact, narrower, broader, related and none, to capture nuances in meaning. Indeed, this goes beyond the task of establishing a link between sense definitions which are potentially referring to the same meanings. Therefore, MWSA is also the task of inferring any possible relationships between senses stored in two dictionaries.

4 https://www.macmillandictionary.com
5 https://www.ldoceonline.com
6 https://www.lexico.com
7 https://dictionary.cambridge.org

From a technical point of view, we address the MWSA task based on two sub-tasks as follows:

- **Semantic similarity detection**: scoring in a numerical way the extent to which two senses or definitions are similar, based on their textual and non-textual properties, and
- **Semantic relation induction**: finding the type of semantic relations that may exist between the two sense definitions. This can be defined formally as the following function:

$$rel = sem(p, s_i, s_j) \tag{6.1}$$

  where $p$ is the part-of-speech of the lemma, $s_i$ and $s_j$ are senses belonging to the same lexemes in two monolingual resources and $rel$ is a semantic relation, namely exact, broader, narrower, related and none. Our goal is to predict a semantic relation, i.e. $rel$, given a pair of senses.

Furthermore, we define the MWSA task at three different levels as follows:

- **Binary classification** to predict if two senses or definitions can possibly be aligned together. In other terms, the 'exact', 'broader', 'narrower' and 'related" links are merged into a single positive class, i.e. $rel \in 0, 1$ in Equation 6.1. This is motivated by the fact that many applications do not care about the specific type of link and that detecting the presence of the link is a harder task from predicting the type of the link.
- **SKOS classification** to predict a semantic relation among `exact`, `broader`, `narrower` and `related` semantic relationships.
- **SKOS+none classification** to predict a semantic relation among `exact`, `broader`, `narrower`, `related` and `none`.

It is important to note that the alignment is carried out at the micro-structure level of each entry; this includes senses, whether defined or not, and sense definitions, i.e. glosses. The latter may contain further information such as semantically-related words which are usually specified with a semi-colon. If such information are systematically structured differently in the micro-structure, i.e. the dictionary defined a different XML tag without grouping them with the glosses, they are not taken into account in the alignment task.

Table 6.2 provides an example of the entry 'entire' described in Table 6.1; this table, provides the scores of various similarity measures carried out on three sense definitions of 'entire', where Definition 1 ($D_1$) and Definition 2 ($D_2$) are semantically closer together, both referring to the sense of 'entire' as being complete in parts, and Definition 3 ($D_3$) denotes a slightly different function of the word used for emphasis. The similarity measures are calculated per each pair of alignment, namely $D_1$-$D_2$, $D_1$-$D_3$ and $D_2$-$D_3$. In this example, three similarity types are provided: string-based, LSR-based and embeddings-based. There are two major differences between these methods, among others. First, the calculated scores are not normalized and therefore, cannot be compared. For instance, the longest common substring method provides the distance as a score while Word2vec provides a float in the range of [0, 1]. And second, they calculate the similarity score based on various features which do not capture the same information essentially, as in edit distance versus vector representations. This leads to different scores where $D_1$ and $D_2$ are considered to be more similar (0.11), as in the method based on WordNet, or $D_1$ and $D_3$ are more similar (0.423) using sentence transformers.

| ENTIRE (adjective) | | | | | |
|---|---|---|---|---|---|
| Senses | Definition 1 ($D_1$) | constituting the full quantity or extent | | | |
| | Definition 2 ($D_2$) | complete in all parts | | | |
| | Definition 3 ($D_3$) | used for emphasizing that you mean all or every part of something | | | |
| similarity | similarity type | method | scores | | |
| | | | $D_1$ - $D_2$ | $D_1$ - $D_3$ | $D_2$ - $D_3$ |
| | String-based | Longest Common Substring | 3 | 6 | 6 |
| | | Levenshtein | 28 | 47 | 49 |
| | | Jaro-Winkler | 0.58 | 0.592 | 0.56 |
| | | Longest Common Subsequence | 37 | 61 | 54 |
| | | four-gram | 0.74 | 0.76 | 0.81 |
| | LSR-based | Levenshtein + lemmas | 6 | 11 | 10 |
| | | WordNet (Pawar and Mago, 2018) | 0.11 | 0 | 0 |
| | Embeddings | Word2vec + Cosine | 0.13 | 0.11 | 0.10 |
| | | Sentence transformers (Reimers and Gurevych, 2019) | 0.261 | 0.293 | 0.423 |

**Table 6.2:** A comparison of three definitions of 'entire' (adjective) based on various similarity detection methods. Definition 1 and 2 refer to the same sense, while Definition 3 is different.

Of the benefits of WSA, it should be noted that it will facilitate the integration of various resources and the creation of inter-linked language resources. Moreover, aligning word senses across monolingual lexicographic resources increases domain coverage and enables integration and incorporation of data. Considering the literature, various aspect of WSA in general, and MWSA in particular, have been matters of research previously. However, a few of previous papers address the alignment task as a specific task on its own. To remedy this and to bring various semantic similarity methods together for the task of WSA, a framework called NAISC (meaning 'links' in Irish and pronounced 'nashk') is built. The idea of NAISC as a framework for data alignment was initially proposed by McCrae et al. (2017a) and McCrae and Buitelaar (2018) chiefly for ontology alignment. However, given that the tool aims to combine various semantic similarity measures at the structural and textual levels, it was further developed in the context of the ELEXIS project and this thesis. Therefore, in this chapter, the architecture of this system is also described as part of the ELEXIS in-frastructure, which covers all parts of the lexicographic process including dictionary drafting. In addition, the contributions of the thesis which are also integrated into the tool, are presented.

The rest of this chapter is organized as follows:

- In Section 6.2, we go through the previous work in aligning word senses or definitions.
- In Section 6.3, the architecture of NAISC is described.
- Section 6.4 describes some of the major metrics to estimate the similarity of two sense definitions, respectively based on textual and non-textual information extracted from them. We first look at some basic methodologies and then implement more advanced methods that use deep learning models.
- Finally, Section 6.6 brings the described methods together by carrying out various experiments. In this experiments, a baseline system is created along with three other systems based on classification, a knowledge graph and deep learning.

We believe that the combination of these tools provides a highly flexible implementation that can link senses between a wide variety of input dictionaries and we demonstrate how linking can be done as part of the ELEXIS toolchain. The outcomes of our developments are beneficial to e-lexicography and the integration and maintenance of lexicographical resources, opening up for new ways of presenting word information to the users of the resources.

## 6.2 RELATED WORK

The alignment of lexical resources has been previously of interest both to create resources and propose alignment approaches. In this section, we only focus on WSA techniques in the related literature.

As discussed in Chapter 2, there have been many studies where graph-based approaches have been used for the WSA task. To recap, we can mention the work of Matuschek and Gurevych (2013) who propose a graph-based approach, called Dijkstra-WSA, for aligning lexical semantic resources, namely Wordnet, OmegaWiki, Wiktionary and Wikipedia. In the same vein, in Chapter 4, the alignment task was modeled as a bipartite-graph where an optimal alignment solution is selected among the combination of possible sense matches in two resources. Although this algorithm performs competitively with the Dijkstra-WSA technique on the same datasets, no viable solution is provided regarding the tuning of the matching algorithm, as discussed in Chapter 2 in detail. Similarly, other studies such as Nancy and Véronis (1990); Pantel and Pennacchiotti (2008); Meyer and Gurevych (2010); Pilehvar and Navigli (2014) focus on linking senses without considering semantic relationships or sense definitions. Aligning resources have been also of interest to create a sense inventory, as described by Pedersen et al. (2018), as a common reference for a language.

Beyond aligning lexical resources, there has been much effort in inducing semantic relationships, particularly within more generic fields such as taxonomy extraction (Bordea et al., 2015), hypernym discovery (Camacho-Collados et al., 2018) and semantic textual similarity (Agirre et al., 2016b). Although in these tasks the focus is on the relationship within words, there are a few works exploring how to induce semantic relationships between definitions. Heidenreich and Williams (2019) introduce an algorithm using a directed acyclic graph to construct a Wordnet based on the Wiktionary data and enriched with the synonym and antonym relationships. Using the semantic relationship annotations provided in Wiktionary, the method induces a semantic hierarchy by identifying a subset within each sense that can relate two lemmas together. In addition to graph-based methods, there are various other closely-related fields, such as word sense disambiguation (Maru et al., 2019) and sense embeddings (Iacobacci et al., 2015), which can potentially contribute to the task of WSA. However, we could not find any previous work exploring those approaches.

One major limitation of the previous work is with respect to the nature of the data used for the WSA task. Expert-made resources, such as the Oxford English Dictionary are not as widely available as collaboratively-curated ones like Wiktionary due to copyright restrictions. Therefore, in the previous work, experiments are carried out either on copyrighted material that cannot be shared or collaboratively-curated resources for specific languages with limited applicability of the proposed methods (as explained in more detail in Chapter 3). As such, the creation of the benchmark, that was explained in the previous chapter, containing a set of 17 datasets of monolingual dictionaries in 15 languages and annotated by language experts with five semantic relationships according to SKOS (Miles and Bechhofer, 2009), namely, broader, narrower, related, exact and none, would provide a common ground for future endeavors in the field. Further, our techniques being open-source and implemented all in one tool called Naisc should make it easier for user to align lexicographical data more easily.

In this context, we also organized a shared task at the GLOBALEX workshop in 2020[8] where a new baseline is developed that covers 15 languages and we invited participants to address the same task (Kernerman et al., 2020a). The result of the participating systems are reported in the Section 6.7.

## 6.3 NAISC ARCHITECTURE

The Naisc architecture is depicted in Figure 6.1. The architecture of Naisc was originally designed by McCrae and Buitelaar (2018) for linking any RDF datasets and this

8 https://competitions.codalab.org/competitions/22163

can be applied to the MWSA task by converting the dictionaries into an RDF format such as OntoLex (McCrae et al., 2017b; Cimiano et al., 2016). The process of linking is broken down into a number of steps that are described as follows:

- **Blocking**: The blocking step finds the set of pairs that are possible candidates for linking. For more general linking tasks and for the multilingual linking task, this may trade off some accuracy for computational efficiency. However, for the MWSA task we only link on matching headwords so the blocking task has a single implementation that simply finds matching headwords and outputs every sense pair between these two entries.
  Signature: (Dataset, Dataset) $\Rightarrow$ (Sense, Sense)*
- **Lens**: The lens examines the data around the sense pair to be linked and extracts text that can be compared for similarity. Clearly, the most important lens for this task extracts the senses' definitions. However, other information such as examples can also be extracted here.
  Signature: (Sense, Sense) $\Rightarrow$ (Text, Text)
- **Text features**: The text features extract a set of similarity judgments about the texts extracted with the lenses and are described in more detail in the following section.
  Signature: (Text, Text) $\Rightarrow \mathbb{R}^*$
- **Graph features**: Graph (or non-textual) features do not rely on the text in the dataset but instead look at other features. They are described in more detail later in this chapter.
  Signature: (Sense, Sense) $\Rightarrow \mathbb{R}^*$
- **Scorer**: From a set of features extracted either from the text or from other graph elements, a score must be estimated for each of the sense pairs. This can be done in either a supervised or unsupervised manner and we implement standard methods for supervised classification such as support vector machines (SVMs) and unsupervised classification using voting.
  Signature: $\mathbb{R}^* \Rightarrow [0,1]^*$ - *Output corresponds to a probability distribution over the relation classes*
- **Matcher and Constraint**: There are normally some constraints that are useful to enforce on the matching and these are applied by the matcher
  Signature: (Sense, Sense, $[0,1]^*$)* $\Rightarrow$ (Sense, Sense)* - *Output is a subset of the input*

Naisc is implemented in Java and is openly available at https://github.com/insight-centre/naisc. The configuration of each run can be specified by giving a JSON description of the components that can be used. For example, this is a default configuration for the MWSA task (presented using YAML syntax):

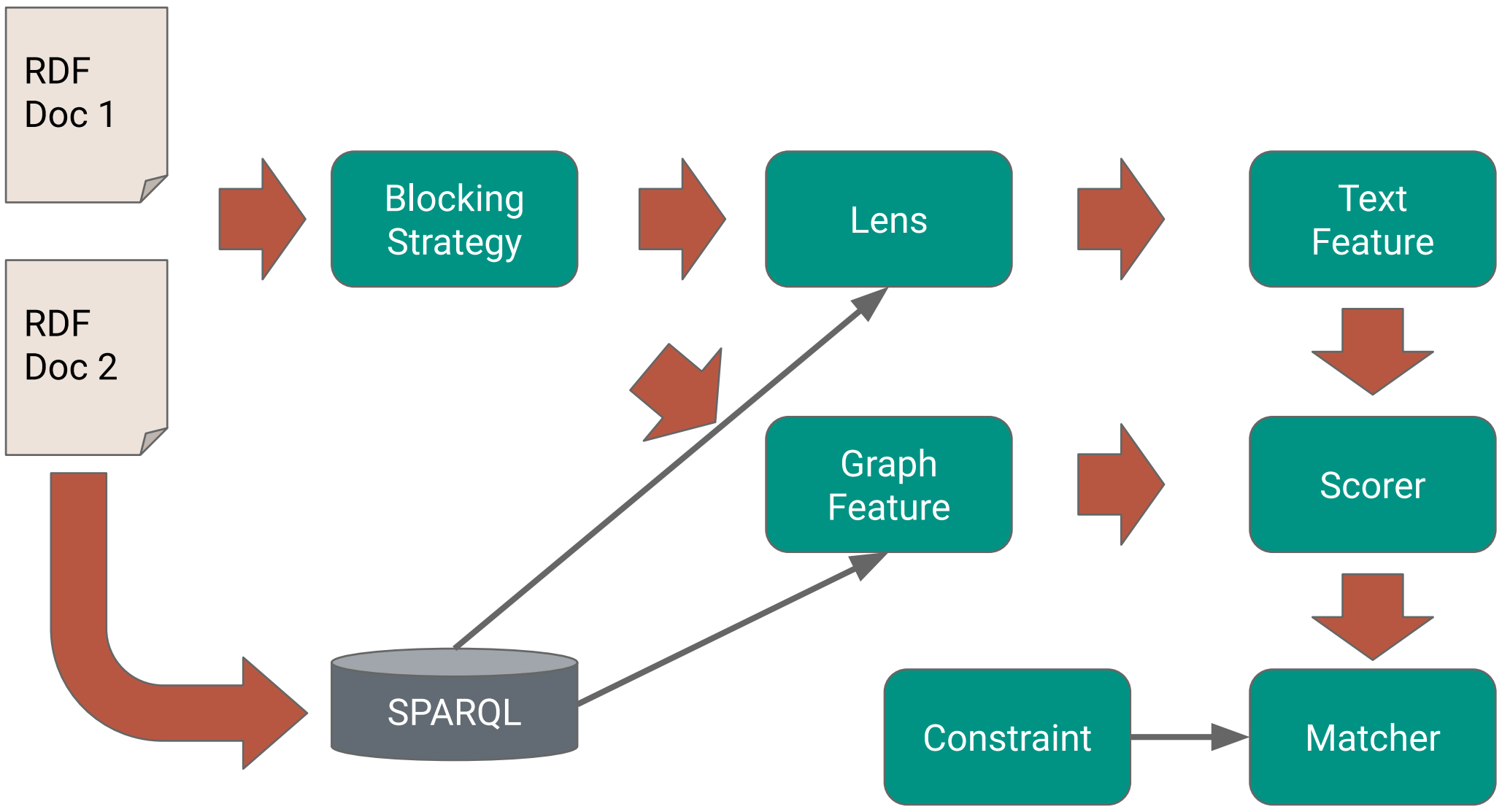

Figure 6.1: The Architecture of the Naisc system for sense linking. Single thin arrows refer to the usage of SPARQL by a component.

```
blocking:
  name: blocking.OntoLex
lenses:
- name: lens.Label
  property:
  - http://www.w3.org/2004/02/skos/core#definition
  id: label
textFeatures:
- name: feature.BasicString
  wordWeights: models/idf
  ngramWeights: models/ngidf
  labelChar: true
- name: feature.WordEmbeddings
  embeddingPath: models/glove.6B.100d.txt
scorers:
- name: scorer.LibSVM
  modelFile: models/default.libsvm
matcher:
  name: matcher.BeamSearch
  constraint:
    name: constraint.Taxonomic
description: The default setting for processing two OntoLex dictionaries
```

This configuration assumes that the dictionary is in the OntoLex format for blocking and processes it. Then, it extracts the definitions using the 'Label' lens and applies various text similarity features which are described in the following sections. The scores for each property type are calculated using LibSVM (Chang and Lin, 2011) and finally the overall linking is calculated using the taxonomic constraints, which will be defined later in this chapter.

One of the main objectives of the development of Naisc and our techniques is the multilingual aspect of the alignment problem. Even though our focus is on monolingual sense and definition alignment, it is possible to use the tool for any other language by implementing language-independent approaches or proper configurations in the tools, particularly for word embeddings and language-specific tasks such as removal of stop-words or acronym finder. As an additional feature, it is also possible to annotate data semi-automatically and save annotations for training purposes. The tool is fully documented at https://uld.pages.insight-centre.org/naisc. A screenshot of the interface of the tool is provided in Figure 6.2.

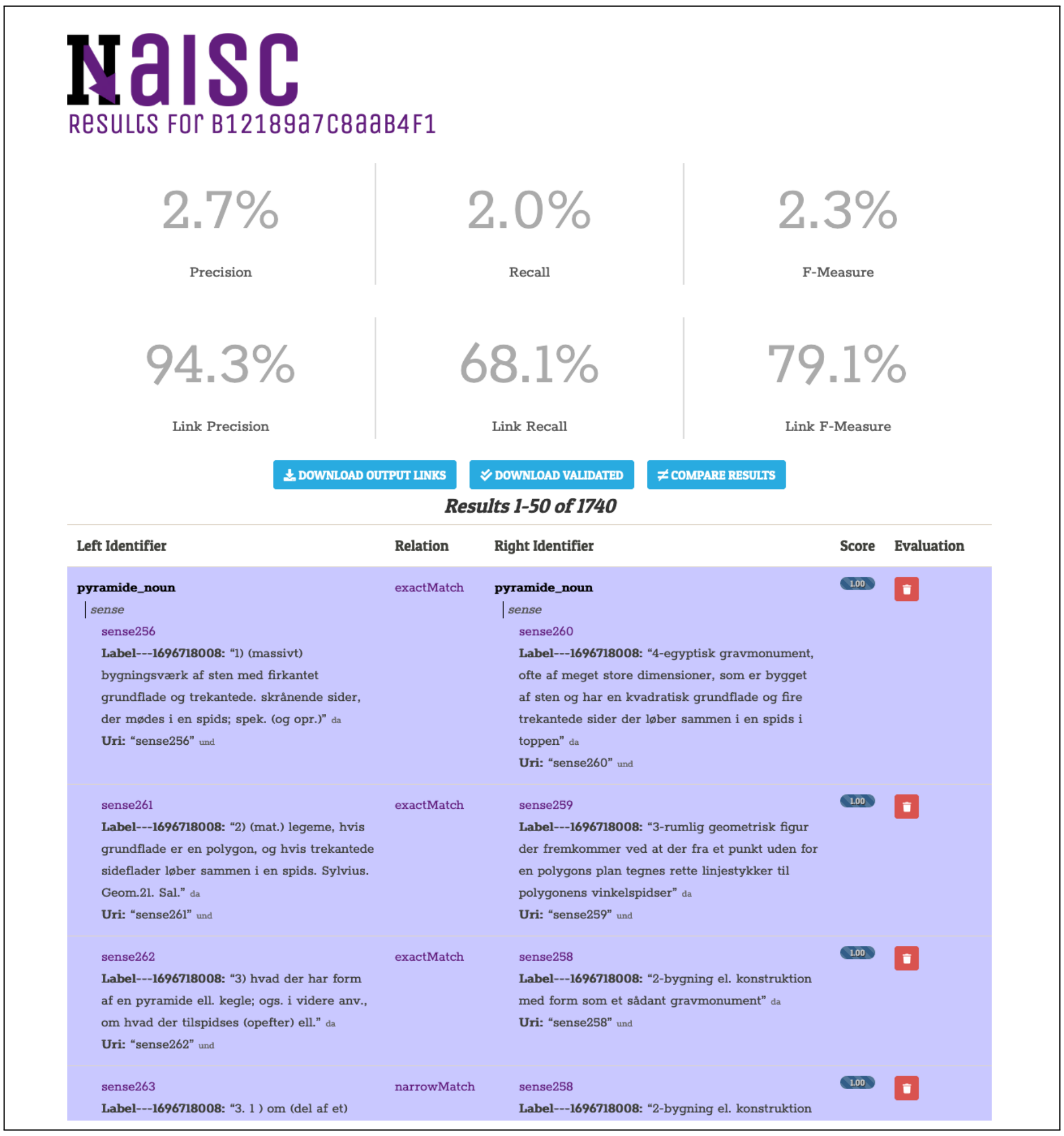

Figure 6.2: A screenshot of Naisc – the automated linking tool

## 6.4 TEXTUAL SIMILARITY METHODS

In this section, we discuss a few approaches that rely on the textual data rather than the structure of the alignment problem to calculate the similarity of senses and definitions. These approaches are categorized as string-based, resource-based, and embeddings-based. Throughout this section, we assume that $A$ and $B$ are the set of words belonging to two senses or definitions in a dictionary, and try to define various similarity functions to estimate their similarity.

### 6.4.1 String-based Methods

Detecting similarity between of textual data, i.e. data of string type, using string-based measures are widely used not only in general usage of computer but also in NLP (Islam and Inkpen, 2008). Within the Naisc framework, we implement string-based basic methods using frequency and surface forms of the strings to compute features. These methods were introduced in Section 4.4.1 of Chapter 4. Most of these methods can work on words or on characters.

### 6.4.2 Beyond String Similarity

Thanks to the increasing size of openly-available language resources and the wide availability of such resources for many languages, it is possible to incorporate lexical and semantic resources in the semantic similarity detection task. Such metrics are also provided in Naisc as follows:

**DICTIONARY** In this method, a dictionary containing words and their synonyms is used to detect whether two terms are synonyms.

**WORDNET** WordNet has been extensively used to detect semantic similarity thanks to various relationships defined in it, such as hypernymy and synonymy (Pedersen et al., 2004; Meng et al., 2013). Figure 6.3 illustrates the synonyms defined for 'entire' (adjective) in the Princeton English WordNet. Based on this structure, it is expected to score the similarity of definitions that contain synonyms higher than those not having.

In the same vein, WordNets are used in the target language of the user as specified in the configuration file to measure the relatedness of two terms in the definitions. To do so, the similarity of words are calculated in Naisc based on the overlap of synonymous and closely-related words according to WordNet. In addition, methods based on shortest path (Lin and Sandkuhl, 2008; Wu and Palmer, 1994) and context (Leacock and Chodorow, 1998) are implemented.

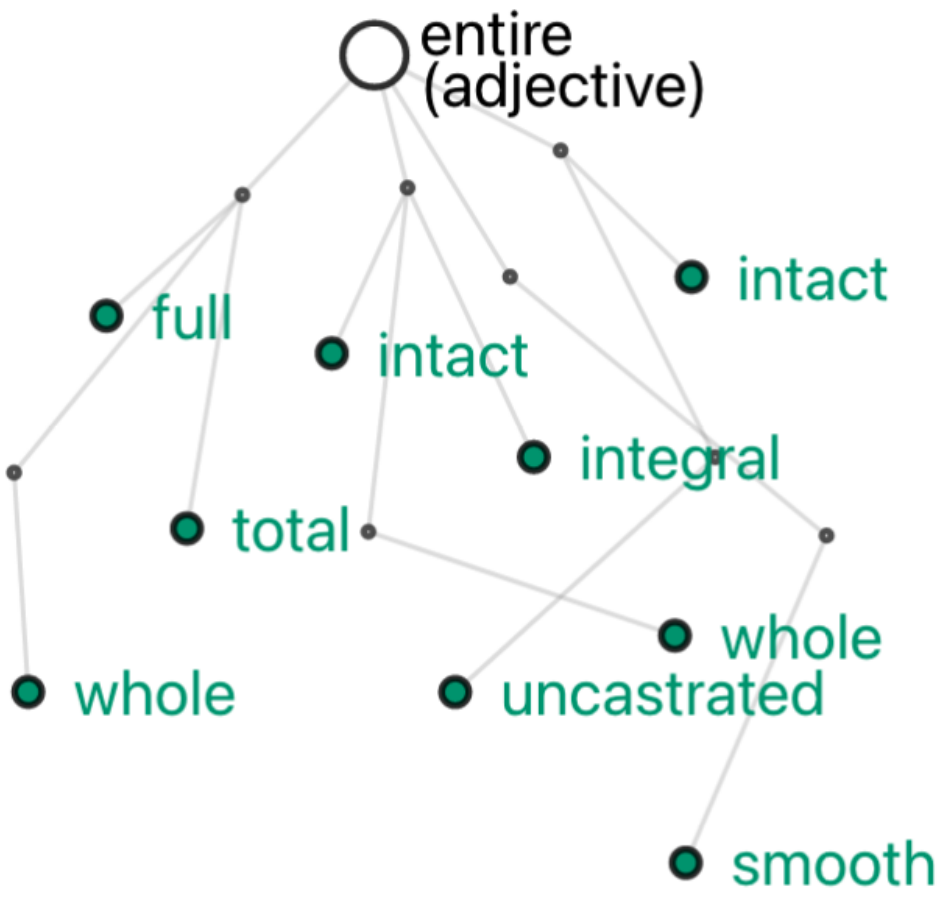

Figure 6.3: Structure of 'entire' (adjective) in the Princeton English WordNet. Green nodes denote synonyms.

**CONCEPTNET** ConceptNet (Speer et al., 2012, 2017) is a knowledge graph that brings various lexical resources together and provides a large network of monolingual and multilingual data in natural language. Words in a given language are used as labels of the nodes of a graph among which semantic relations are extracted and weighted as edges. The weight is calculated according to the number of occurrences in the resources and can have a negative value. Figure 6.4 provides a few monolingual relations associated to 'entire' (adjective) where 'whole' is confidently presented as a related concept to 'entire' with weight value 4.2 and 'department' has the lowest score of being an antonym of 'entire'. The resource based on which the relation is extracted or inferred is provided as well.

**EMBEDDINGS** Unlike similarity measures that rely on manually-curated lexical and semantic resources such as WordNet and dictionaries, embeddings only require to be trained on large corpora. As embeddings are widely accessible for many languages, including low-resourced languages, it is essential to integrate measure that rely on embeddings in NAISC. Therefore, the following approach is implemented based on Global Vectors for Word Representation (GloVe) vectors (Pennington et al., 2014) where the word embeddings for each word is calculated in the two definitions and then compare pairwise the words of each definition. Given $V_A$ and $V_B$, respectively as the corresponding vector representations of sense definition $A$ and sense definition $B$, the similarity between vectors is calculated using the cosine similarity as follows:

$$\cos(\theta) = \frac{V_A . V_B}{|V_A||V_B|} \tag{6.2}$$

where $\theta$ is the angle between two vectors projected in a multi-dimensional plane.

| Source word | Relation / Weight | Target word | Resource |
|---|---|---|---|
| full (a, wn) | — Synonym → Weight: 2.0 | entire (a, wn) | Open Multilingual WordNet |
| intact (a, wn) | — Synonym → Weight: 2.0 | entire (a, wn) | Open Multilingual WordNet |
| entire (a, wn) | — SimilarTo → Weight: 2.0 | smooth (a, botany) | Open Multilingual WordNet |
| uncastrated (a, wn) | — SimilarTo → Weight: 2.0 | entire (a, wn) | Open Multilingual WordNet |
| entire | — RelatedTo → Weight: 4.2 | whole | Verbosity players |
| entire | — RelatedTo → Weight: 1.56 | whole thing | Verbosity players |
| entire | — RelatedTo → Weight: 1.28 | everything | Verbosity players |
| entire | — Antonym → Weight: 0.27 | nothing | Verbosity players |
| entire | — DistinctFrom → Weight: 0.19 | part | Verbosity players |
| department | — Antonym → Weight: 0.18 | entire | Verbosity players |
| entire (a) | — HasContext → Weight: 1.0 | complex analysis | English Wiktionary |
| entire (a) | — HasContext → Weight: 1.0 | male animal | English Wiktionary |

**Figure 6.4:** A few monolingual relations associated to 'entire' (adjective) in ConceptNet (https://conceptnet.io/c/en/entire). Underlined items refer to concepts.

In addition, the tool is extendable to various other embedding methods in the configuration, such as Word2vec (Mikolov et al., 2013a), fastText (Grave et al., 2018), GloVe (Pennington et al., 2014) and BERT (Devlin et al., 2018). Given the key role of the fine-tuned pre-trained neural network language models in achieving state-of-the-art results in many NLP applications, we further expand on it in the following.

**DEEP LEARNING** : Currently, deep learning methods are widely used for word and sentence similarity detection. In the implementations, two methods are used based on the recurrent neural network and transformers.

**Recurrent Neural Networks (RNNs)**: RNNs (Medsker and Jain, 2001) are a set of neural networks for processing sequential data and modeling long-distance dependencies. Given that many tasks in NLP are compatible with sequential modeling of data, RNNs currently form the backbone of many tools, particularly in machine translation and sentence embedding (Agirre et al., 2016a). The initial models are further optimized, particularly using gate mechanism such a long short-term memory (LSTM) (Hochreiter and Schmidhuber, 1997) and the attention mechanism (Vaswani et al., 2017) to tackle vanishing gradient in long sequences (Ahmadi, 2018).

**Transformers**: Another notable technique is based on transformers. A transformer is a type of neural network architecture that consist of self-attention mechanism using an encoder-decoder structure (Devlin et al., 2018). In an encoder-decoder network, the restriction over the fixed length of sentences which was previously faced in neural network based methods, such as RNNs, is addressed (Pascanu et al., 2013). As such, a variable length sequence is encoded into a state with a fixed shape and then mapped into a variable length sequence, from where such a network is called an encoder-decoder.

Bidirectional Encoder Representations from Transformers (BERT) (Devlin et al., 2018) is one of the well-known transformer-based methods that is trained on unlabeled data, such as Wikipedia-based corpora, using masked language modeling and next sentence prediction for language modeling. By jointly conditioning on both left and right context of a given word, which is masked in the training phase, the model learns dependencies that occur at the lexical, semantic and syntactic levels. BERT has achieved state-of-the-art results in many tasks by outperforming various other models leveraging RNNs or convolutional neural networks (CNNs), both in terms of evaluation score and training time. From an empirical point of view, it is not fully clear yet how and on what tasks contextual embeddings, in general, and BERT, in particular, perform, making researchers wonder about *bertology* (Rogers et al., 2020) as a new field of interest in NLP.

As the results of the systems submitted to the shared task organized for the same task of alignment indicate (see Section 6.7), participants used BERT and Robustly optimized BERT pre-training approach (Liu et al., 2019) as well. This is done by using the Hugging Face transformers library which provides the API for fine-tuning of transformer models (Wolf et al., 2019). To do so, a definition is fed into the embeddings network, also referred to as transformer model, and a 768 dimensional dense vector space is returned as the representation of the definition. Following this, the pooling operation is applied on the contextualized word embeddings where all contextualized word embeddings are simply averaged. Finally, the similarity of two given definitions is calculated by computing the cosine similarity of their vectors. More documentation regarding BERT is available at https://huggingface.co/sentence-transformers.

### 6.4.3 Word Alignment

In addition to the string-based techniques and those based on resources and embeddings, aligning two given definitions at word level can be used to estimate similarity as well. In this approach, similarity matrix $S_{N \times M}$ is created based on the lengths of definitions A and B, respectively as N and M. If two words in the two definitions are to be aligned, their value is set to 1, otherwise 0. This way, the value $S_{ij}$ in the matrix corresponds to the alignment of words $i$ in definition A and $j$ in definition B. In order to extract a single similarity score based on the matrix, the following methods are used.

Firstly, the precision of mapping is defined according to McCrae and Buitelaar (2018) by calculating the number of words that are aligned with a similarity greater than one half:

$$f = \frac{1}{N}\left|\left\{i \mid \exists j : \frac{s_{ij}}{\sum_{j'} s_{ij'}} > 0.5\right\}\right| \quad \text{Forward precision} \quad (6.3)$$

$$b = \frac{1}{M}\left|\left\{j \mid \exists i : \frac{s_{ij}}{\sum_{i'} s_{i'j}} > 0.5\right\}\right| \quad \text{Backward precision} \quad (6.4)$$

The forward and backward precision measures can be combined using a harmonic mean, as defined by Sultan et al. (2014):

$$m = \frac{2fb}{f+b} \quad \text{Harmonized Alignment Mean} \quad (6.5)$$

Another feature that can be used is the normalized and parallel column and row maximums using norms (p-Norm) and parallel maxima (p-Max). p-Max takes one or more vectors or matrices as f and b as arguments and returns a single vector for the parallel maxima of the vectors where the first element of the result is the maximum of the first elements of all the arguments, the second element of the result is the maximum of the second elements of all the arguments and so on (Crawley, 2012, p. 45). This will be larger if we tend to have one value in the similarity matrix that is much larger and, smaller if all values are approximately equal.

$$c_{b,p} = \frac{1}{M}\left(\sum_j max_i\left(\frac{s_{ij}}{\sum_{i')} s_{i'j}}\right)^p\right)^{\frac{1}{p}} \quad \text{Column Mean p-Max} \quad (6.6)$$

$$col_{f,p} = \frac{1}{N}\sum_i\left(\sum_j s_{ij}^p\right)^{\frac{1}{p}} \quad \text{Column Mean p-Norm} \quad (6.7)$$

To measure the quality of an alignment and normalizing some metrics by dividing by matrix size to ensure uniform output, the sparsity metric introduced by Hurley and Rickard (2009) is additionally used. Following McCrae and Buitelaar (2018)'s findings, the Gaussian Entropy Diversity is also used as the sparsity metric as follows:

$$H_G = \frac{1}{MN} \sum_i \sum_j -\log(s_{ij}^2) \tag{6.8}$$

It is important to note that this method can be applied to definitions of any length, and thus, they can produce similar scores regardless of the length of the input strings.

### 6.4.4 Monolingual Alignment

In addition to word alignment in the two definitions, a simplified version of the monolingual alignment technique described by Sultan et al. (2014) and McCrae and Buitelaar (2018) is also implemented. This technique aims to discover and align semantic units in a given pair of sentences or phrases, in our case sense definitions, to detect semantic similarity. To do so, first, identical word sequences are identified and linked in the two definitions. Then, named entities, such as proper names or name of organizations, are linked using Stanford Named Entity Recognizer (Finkel et al., 2005) and, content and stop words are aligned in the two definitions. Finally, the similarity of any pair of words in the definition pair is calculated as follows:

$$sim(w_i, w_j) = \omega \times wordSim(w_i, w_j) + (1 - \omega) \times contextSim(w_i, w_j) \tag{6.9}$$

where

$$contextSim(w_i, w_j) = \frac{1}{|W_i||W_j|} \sum_{m \in W_i} \sum_{n \in W_j} wordSim(W_{i,m}, W_{j,n}) \tag{6.10}$$

where $W_i$ and $W_j$ are a fixed window of words respectively around $w_i$ and $w_j$, $wordSim$ is a word similarity function, similar to the previously mentioned ones in Section 4.4.1, returning a value in the range of [0, 1], and $\omega$ is the weight of word similarity relative to contextual similarity. It is worth mentioning that the alignment measures mentioned for the similarity task refers to the alignment of words within sentences, akin to alignment in machine translation (Fraser and Marcu, 2007), and is different from our general task of alignment in dictionaries.

Once the various measures of these techniques, namely monolingual alignment described here and word alignment described in Section 6.4.3, are calculated, an additional step is taken to calculate a single similarity score. To do so, a range of methods had been implemented in Naisc McCrae et al. (2017a) which are referred to as 'Scorer' in Figure 6.1. Throughout our experiments, LibSVM (Chang and Lin, 2011) is used which trains an SVM model based on the features extracted from the training data.

## 6.5 SEMANTIC RELATION INDUCTION

Another approach to detect the type of semantic relation that exists between a pair of senses is by incorporating a knowledge graph and using textual features. We define a few features which use the lengths of senses along with their textual and semantic similarities. In addition, we incorporate word-level semantic relationships to determine the type of relation that two senses may possibly have. To this end, we use ConceptNet (Speer et al., 2016), an openly-available and multilingual semantic network with relational knowledge from various other resources, such as Wiktionary and WordNet (Miller, 1995). As a supervised method, we rely on the data instances extracted from the MWSA benchmark to train models. A similar approach has been previously proposed for aligning bilingual with monolingual dictionaries Saurí et al. (2019). Our approach is illustrated in Figure 6.5 where data instances are enriched using the knowledge base; the following sections describe the components of this approach.

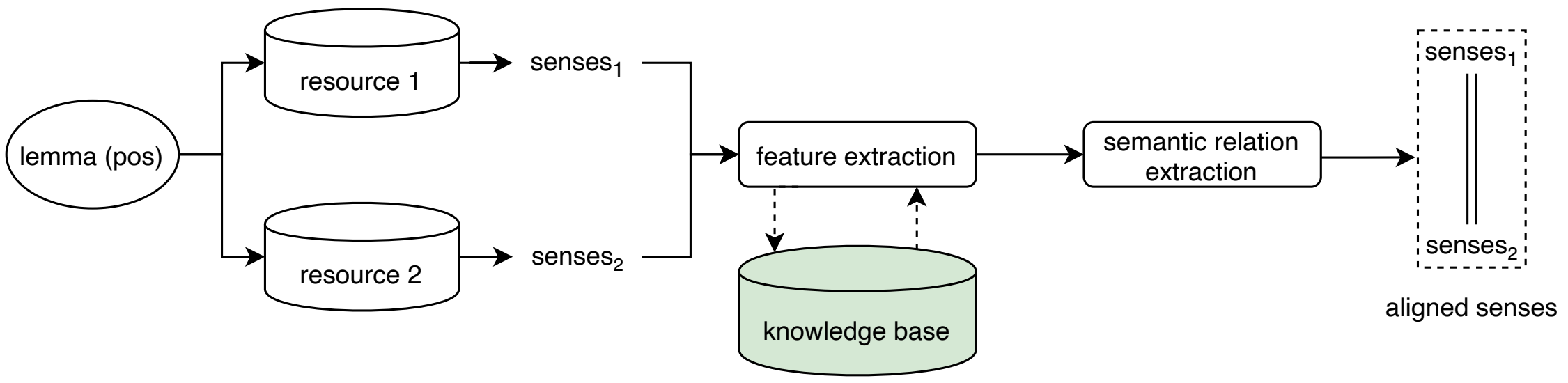

Figure 6.5: Our approach where features are extracted from word senses and external semantic resources

### 6.5.1 Data Extraction

In this step, sense instances are extracted from the MWSA datasets that were described in the previous chapter, as $t = (p, s_i, s_j, r_{i_j})$. This instance is interpreted as sense $s_i$ has relation $r_{i_j}$ with sense $s_j$. Therefore, the order of appearance is important to correctly determine the relationship. It should also be noted that both senses belong to the same lemma with the part-of-speech p. Table 6.3 provides the basic statistics of the senses and their semantic relationships in various languages, extracted for training purposes from the MWSA benchmark. "# Entries" and "# SKOS" refer to the number of entries and senses with a relationship within SKOS. In addi-tion, the senses within the two resources which belong to the same lemma but are not annotated with a SKOS relationship are included with a none relationship.

Given the class imbalance where senses with a none relationship are more frequent than the others, a data augmentation technique is carried out based on the symmetric property of the semantic relationships. By changing the order of the senses, also known as relation direction, in each data instance, a new instance can be created by semantically reversing the relationship. For instance, for the entry 'observer' (noun), if the sense definition 'an expert who observes and comments on something' in WordNet is annotated with a narrower relation with respect to 'an annotator' in Webster, a new instance is created where 'an annotator' has a broader relation in comparison to 'an expert who observes and comments on something'. In other words, for each $t = (p, s_i, s_j, r_{ij})$ there is a $t' = (p, s_j, s_i, r'_{ij})$ where $r'_{ij}$ is the inverse of $r_{ij}$. Thus, exact and related as symmetric properties remain the same, however, the asymmetric property of the broader and narrower relationships yields narrower and broader, respectively. The overall number of instances as the result of this data augmentation along with SKOS and none relations is provided in the column # All in Table 6.3. Finally, the percentage of relations that are none is specified in the last column.

| Language | # Entries | # SKOS | # SKOS + #none | # All | none (%) |
|---|---|---|---|---|---|
| Basque | 256 | 813 | 3661 | 4382 | 64.99% |
| Bulgarian | 1000 | 1976 | 3708 | 5656 | 30.62% |
| Danish | 587 | 1644 | 16520 | 18164 | 81.90% |
| Dutch | 161 | 622 | 20144 | 20766 | 94.01% |
| English[9] | 684 | 1682 | 9269 | 10951 | 69.28% |
| Estonian | 684 | 1142 | 2316 | 3426 | 34.27% |
| German | 537 | 1211 | 4975 | 6185 | 60.86% |
| Hungarian | 143 | 949 | 15774 | 16716 | 88.69% |
| Irish | 680 | 975 | 2816 | 3763 | 48.92% |
| Italian | 207 | 592 | 2173 | 2758 | 57.32% |
| Serbian | 301 | 736 | 5808 | 6542 | 77.53% |
| Slovenian | 152 | 244 | 1100 | 1343 | 63.74% |
| Spanish | 351 | 1071 | 4898 | 5919 | 64.66% |
| Portuguese | 147 | 275 | 2062 | 2337 | 76.47% |
| Russian | 213 | 483 | 3376 | 3845 | 75.24% |

**Table 6.3:** Number of data instances extracted from the MWSA benchmark for training purposes based on the type of semantic relation. # refers to the number.

Once the senses have been extracted, data instances are created using the features in Table 6.4. Feature 1 is a one-hot vector indicating the part-of-speech of the entry. Features 2 and 3 concern the length of senses and how they are different. Intuitively speaking, this regards the wordings used to describe two concepts and their semantic relationship. In features 2 to 3, this is respectively calculated with and without function words, i.e. words with little lexical meaning. One additional step is to query

9 English (NUIG) and English (KD) are merged as well as the Slovenian datasets.

ConceptNet to retrieve semantic relations between the content words in each sense pair. For instance, the two words "gelded" and "castrated" which appear in two different senses of 'entire' in Table 6.1 are synonyms and therefore, the whole senses can be possibly synonyms. In order to measure the reliability of the relationships, the weights of each relationship, also known as *assertions*, are summed up according to ConceptNet. Features 4 to 10 correspond to the weight of various relationships based on ConceptNet. Further, features 11 and 12 provide the semantic similarity of each sense pair using word embeddings, respectively with and without function words. For this purpose, GloVe (Pennington et al., 2014) or fastText[10] are used. Finally, features 13, 14 and 15 indicate the type of semantic relation based on the annotations in the MWSA benchmark. These are used as the target classes for our classification problem. It should be mentioned that the data instances are all standardized by scaling each feature to the range of [0-1].

| # | feature | definition | possible values |
|---|---|---|---|
| 1 | `POS_tag` | part of speech of the headword | a one-hot vector of {N, V, ADJ, ADV, OTHER} |
| 2 | `s_len_no_func_1/2` | number of space-separated tokens in $s_1$ and $s_2$ | $\mathbb{N}$ |
| 3 | `s_len_1/2` | number of space-separated tokens in $s_1$ and $s_2$ without function words | $\mathbb{N}$ |
| 4 | `hypernymy` | hypernymy score between tokens | sum of weights in ConceptNet |
| 5 | `hyponymy` | hyponymy score between tokens | sum of weights in ConceptNet |
| 6 | `relatedness` | relatedness score between tokens | sum of weights in ConceptNet |
| 7 | `synonymy` | synonymy score between tokens | sum of weights in ConceptNet |
| 8 | `antonymy` | antonymy score between tokens | sum of weights in ConceptNet |
| 9 | `meronymy` | meronymy score between tokens | sum of weights in ConceptNet |
| 10 | `similarity` | similarity score between tokens | sum of weights in ConceptNet |
| 11 | `sem_sim` | semantic similarity score between senses using word embeddings | averaging word vectors and cosine similarity [0-1] |
| 12 | `sem_sim_no_func` | semantic similarity score between senses without function words | averaging word vectors and cosine similarity excluding function words [0-1] |
| 13 | `sem_bin_rel` | target class | 1 for alignable, otherwise 0 |
| 14 | `sem_rel_with_none` | target class | {`exact, narrower, broader, related, none`} |
| 15 | `sem_rel` | target class | {`exact, narrower, broader, related`} |

Table 6.4: Manually extracted features for semantic classification of sense relationships (21 columns in total)

10 https://fasttext.cc

### 6.5.2 Feature Learning

Despite the differences in measuring various similarity score metrics, there could be a relation between some of the metrics (features) given their theoretical resemblance; for instance, similarity scores obtained from ConceptNet and embeddings could capture similarity between words being semantically related, such as synonyms. In such scenarios where various features are brought together for a specific task with a potential relation between them, defining and modeling the relationship between all the similarity measures is not feasible in a manual manner. Inferring how various similarity measures work in relation to each other is beneficial to evaluate the similarity between two definitions or senses more accurately. To this end, various representation learning methods have been developed. Therefore, in addition to the feature extraction and data augmentation, we proceed with a feature learning step.

Restricted Boltzmann machine (RBM) is a generative model representing a probability distribution given a set of observations (Fischer and Igel, 2012). Its main application in feature learning and dimensionality reduction has made it a popular solution in various classification and regression problems in machine learning, notably in the fields of speech recognition (Jaitly and Hinton, 2011; Zheng et al., 2013), image processing (Sheri et al., 2018; Teh and Hinton, 2001) and pattern recognition (Alani, 2017). It is also widely used in combination with various neural network methods such as deep belief networks (Hinton, 2009). In the task of similarity detection, instead of training our models using the similarity score values, an RBM model is first trained, which is unsupervised, and then the classifiers are trained using the latent features of the RBM model. These new features have binary values and can be configured and tuned depending on the performance of the models.

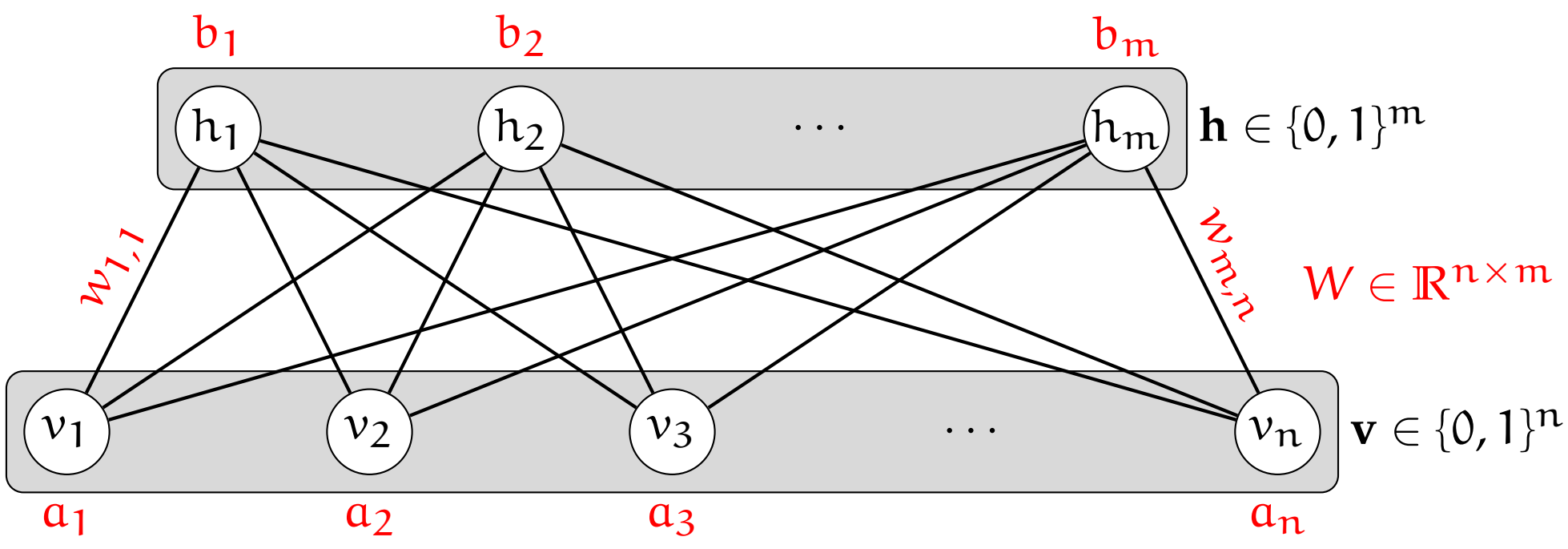

**Figure 6.6:** An example of restricted Boltzmann Machine with a $m$ latent units and $n$ visible units

An RBM, as illustrated in Figure 6.6, is composed of two layers: a visible one $v$ where the data instances are provided, and a latent one $h$ where a distribution is created by the model by retrieving dependencies within variables. While all the nodes in the visible layer are connected to the nodes in the latent layer, the nodes within the visible or latent layer are not interlinked; this restriction is the main difference of an RBM with a Boltzman machine (Montúfar, 2018). Once trained, an RBM model learns the underlying relation and distribution of the features in how the target classes are predicted. The features in the visible layer can be manually-created or based on the output of another classification problem. We follow the description of Hinton (2012) in implementing and using an RBM. In our case, we use similarity scores as visible units as they are observed; such scores are either binary or with a value between 0 and 1. In addition, the the values of the latent layers are binary. An RBM with such units is also known as Bernoulli Restricted Boltzmann Machine.

Given a finite set of features in the visible layer with states $v_i \in [0, 1]$ and bias values $a_i \in \mathbb{R}$, the energy of a joint configuration of the visible and latent units is defined as the following energy function (Hopfield, 1982):

$$E(v, h) = -\sum_{i \in \text{visible}} a_i v_i - \sum_{j \in \text{hidden}} b_j h_j - \sum_{i,j} v_i h_j w_{ij} \tag{6.11}$$

where $h_j$ is a hidden unit with state $h_j \in [0, 1]$ and bias value $b_j$ and $w_{ij} \in \mathbb{R}$ is a random variable as the weight between the visible unit $v_i$ and $h_j$. The RBM network assigns a probability to every connection between a visible node and a hidden one as follows:

$$p(v, h) = \frac{1}{Z} e^{E(v,h)} \tag{6.12}$$

where $Z$ refers to the sum of all possible combinations of visible and hidden units:

$$Z = \sum_{v,h} e^{E(v,h)} \tag{6.13}$$

And, the probability given to a visible unit is equivalent to the sum of the values of all the hidden units:

$$p(v) = \sum_{h} p(v, h) = \frac{1}{Z} \sum_{h} e^{-E(v,h)} \tag{6.14}$$

The objective of learning in this model is to maximize the probability that RBM assigns to the binary units in the training set, those being the similarity scores. Therefore, stochastic gradient ascent can be carried out by adjusting weights $W$, and biases such that the overall energy $E(v, h)$ decreases. To calculate this, a log-likelihood objective function is used and differentiated with respect to the weight as follows:

$$\frac{\partial \log p(v)}{\partial w_{ij}} = \langle v_i h_j \rangle_{data} - \langle v_i h_j \rangle_{model} \tag{6.15}$$

where brackets refer to the expectations under the specified distributions. This way, the following learning rule is defined which is optimized through the training phase:

$$\Delta w_{ij} = \eta(\langle v_i h_j \rangle_{data} - \langle v_i h_j \rangle_{model}) \tag{6.16}$$

where η refers to the learning rate. The optimization is carried out by calculating the overall energy of the model for a visible vector and then, update hidden vectors and visible vectors, accordingly. Given a visible unit, the binary state of each hidden unit j is set to 1 with the following probability (without calculating the bias):

$$p(h_j = 1|v) = \sigma(b_j + \sum_i v_i wij) \tag{6.17}$$

where σ is the logistic sigmoid function as non-linearity. Contrastive divergence learning (Carreira-Perpinan and Hinton, 2005) is usually used in the optimization phase to facilitate the convergence of the network.

## 6.6 EXPERIMENTS

In the previous sections, various textual and non-textual similarity detection methods were introduced to estimate how similar two definitions are. In addition, a few linking constraints were introduced to optimize the linking between sense pairs of the same lemma belonging to the same part-of-speech tags in two different monolingual dictionaries. Using these, a few experiments are carried out to evaluate various aspects of the alignment problem. A baseline system is first provided as the basis to create and compare more robust approaches based on classification, knowledge graph and deep learning. Given the number of datasets (17) in the evaluation benchmark, some of the experiments are exclusively limited to a few of the datasets. It should also be mentioned that the experiments do not include similarity scores calculated using deep learning, including BERT (described in Section 6.4.2).

Multiple metrics are used to evaluate the results of the systems. Firstly, accuracy measures the total number of links for which the correct class of relationship is predicted, in other words, the percentage of scores for which the predicted label matches the reference label. Secondly, recall, precision and F-Measure scores, defined based on Powers (2020). The systems are overall scored based on a macro-average of the accuracy, precision, recall and F-Measure, as defined in Chapter 4.

### 6.6.1 Baseline System

We create a simple baseline system where for each sense pair, the Jaccard similarity of the set of words within glosses is calculated, then the Hungarian algorithm is used to find the most likely unique assignment between these senses. The baseline only predicts the 'exact' and 'none' classes so it is expected that the results would be quite poor. The baseline is evaluated using accuracy, precision, recall and F-measure and the results along with the statistics of the datasets are provided in Table 6.5.

| Language | Accuracy | Precision | Recall | F-measure |
|---|---|---|---|---|
| Basque | 0.789 | 0.211 | 0.050 | 0.081 |
| Bulgarian | 0.728 | 0.250 | 0.011 | 0.020 |
| Danish | 0.817 | 0.300 | 0.023 | 0.043 |
| Dutch | 0.936 | 0 | 0 | 0 |
| English | 0.752 | 0 | 0 | 0 |
| Estonian | 0.482 | 0.545 | 0.093 | 0.159 |
| German | 0.7777 | 0 | 0 | 0 |
| Hungarian | 0.940 | 0.053 | 0.012 | 0.020 |
| Irish | 0.583 | 0.680 | 0.185 | 0.291 |
| Italian | 0.693 | 0 | 0 | 0 |
| Serbian | 0.853 | 0 | 0 | 0 |
| Slovene | 0.834 | 0.100 | 0.009 | 0.017 |
| Spanish | 0.678 | 0.255 | 0.127 | 0.170 |
| Portuguese | 0.921 | 0.083 | 0.024 | 0.037 |
| Russian | 0.754 | 0.438 | 0.179 | 0.255 |
| Average | 0.769 | 0.194 | 0.048 | 0.074 |

**Table 6.5:** The baseline system using Jaccard similarity metric and Hungarian algorithm for alignment.

Even though the baseline system is able to align definition pairs accurately in half of the cases, the fraction of relevant instances among the retrieved ones, i.e. precision, and the fraction of relevant instances that are retrieved, i.e. recall, is quite low. Consequently, the F-measure is low and even zero in some of the cases. Looking back at Table 6.3, we notice that, except for Bulgarian, Estonian and Irish, over 50% of all the relations in the datasets belong to 'none'. Therefore, there is no surprise that the baseline system can achieve such an accuracy. Furthermore, the alignment of the Hungarian data having the highest accuracy of 0.94 can be explained by the lack of distinction between part-of-speech tags in the datasets. This is an exceptional case as in other datasets, the alignment is carried out based on not only identical lemmas, but also identical part-of-speech tags. It should also be noted that for languages for which more than one dataset exists in the benchmark, the annotations are merged and treated as one dataset.

In the intrinsic evaluation of our benchmark in the previous chapter in Section 5.5, a few evaluations are carried out to describe the characteristics of the datasets, such as density and degree in Table 5.8, and the correlation between the number of tokens in the first and second resource in Figure 5.9. In addition to the imbalance of data, despite our efforts to create more data instances thanks to induction, the alignment task is difficult due to the difference in the granularity of senses and definitions. This can be demonstrated by comparing the number of tokens in the two dictionaries per dataset, respectively referred to as $R_1$ and $R_2$ in Figure 5.9.

In the rest of this section, we focus on the alignment problem with binary and SKOS relations, but also how the length of definitions can play a role in improving the results.

### 6.6.2 System 1: Classification and Feature Learning

In this system, we train various classification models using support vector machines (SVMs) based on different hyper-parameters, as implemented in Scikit-learn[11] (Pedregosa et al., 2011). After a preprocessing step, where the datasets are shuffled, normalized and scaled, we split them into train, test and validation sets with 80%, 10% and 10% proportions, respectively. The performance of the models is evaluated for all the languages with and without the feature learning step. In the latter case, instead of training our models using the data instances described in the previous section, we train the models using the latent features of an RBM model. These new features have binary values and can be configured and tuned depending on the performance of the models.

Table 6.6 presents the best performance of the models trained for each language. The experiments are carried out based on three classification objectives: (i) a binary classification, where the model classifies a pair of sense definitions to be aligned, i.e. the target class is 1, or not, i.e. the target class is 0, (ii) a SKOS classification where one of the SKOS relations, namely exact, narrower broader and related, is determined by the model on those sense pairs for which a relation exists in the gold-standard datasets, and finally (iii) a classification over all the possible relations, namely SKOS relations and none. The latter is referred to as All or RBM-All in the table. Our evaluation is carried out using the same metrics of the baseline in Table 6.5, but for all classification setups. Our optimal models that are reported in Table 6.6 were trained with 50 iterations, a learning rate within [0.05-0.2], a hidden unit number of the RBM model within the range of 400 and 600 and linear, radial basis and polynomial kernel functions for SVM. The tuning of the hyper-parameters was carried out using the exhaustive grid search in Scikit-learn[12]. The highest values of F-measure for each language and classification task are indicated in different colors: cyan for binary classification, violet for SKOS classification and green for the classification of all relations.

11 https://scikit-learn.org
12 https://scikit-learn.org/stable/modules/grid_search.html

Despite the high accuracy of the models for most languages, they do not perform equally efficiently for all languages in terms of precision and recall. Our classifiers outperform the baselines for all the relation prediction tasks in terms of accuracy and perform with high accuracy when trained for the binary classification and also given all data instances. This is thanks to the majority of sense pairs in two dictionaries having no semantic relation, i.e. their semantic relation is 'none'. However, there is a significant low performance when it comes to the classification of SKOS relationships. In other words, when a model is trained on the data instances that should be aligned as there are none with a 'none' relation, the F-measure is not exceeding 0.5349 (for Portuguese). While, the F-measure score for the binary and all classification tasks respectively reaches 0.8760 and 0.7890 (both for Dutch). It should be noted that F-measure combines both precision and recall, therefore, is a good metric to estimate the effectiveness of a model.

Regarding the performance of RBM in terms of accuracy and F-measure, we do not observe an improvement in the results of all classifiers and see that almost half of the highest scores are in fact thanks to using an RBM. However, the precision of the models which learn features with an RBM is higher in the majority of cases. As such, the initial hypothesis that using RBM would help improving the results by learning latent relations between features is not unquestionably sound. It has been demonstrated that models that are trained using log-likelihood objective functions, like RBM, are prone to fail to assign correct probability to instances for which a lower number of data are provided in the training phase (Fisher et al., 2018). In our case, a lower number of data instances with a SKOS relation, as Table 6.3 shows, means that the trained model tends to learn assigning relations with more frequency, such as 'exact' and 'none' rather than the less frequent ones, as 'broader', 'narrower' and 'related'.

Given the difficulty of determining the type of semantic relations, the lower performance of the systems with respect to the SKOS classification task is not surprising. Distinguishing certain types of relationships, such as related versus exact, is a challenging task even for an expert annotator. This has been previously discussed by Koskela (2014) regarding the systematic representation of senses and broader vs. narrower distinctions in lexicography. In our case, for instance, the relationship between two senses of ENTIRE in Table 6.1, "constituting the undiminished entirety" and "complete in all parts; undivided; undiminished; whole" is annotated as 'narrower' and exact by two different annotators.

Two examples of the alignment of 'tube' (verb) and 'glow' (verb) are respectively provided in Tables 6.7 and 6.8. In the case of 'tube', all systems, i.e. baseline, System 1 and BERT (Bajcetic and Yim, 2020), detect $R_{1A}$ and $R_{2A}$ as the senses to be aligned, and all detect a 'narrower' relation between them which is not the one that the gold-standard data shows (exact). On the other hand, 'glow' has three alignments in the gold-standard dataset, all with an 'exact' relation, while the systems, except BERT, detect only two senses to be aligned and none of them are identical with the reference annotations. It should be noted that only predictions with a similarity scores of over 0.1 are presented in the table and any predictions with a lower similarity score is considered to be 'none'. It is worth mentioning that BERT in Tables 6.7 and 6.8 refers to the shared task system submitted by Bajcetic and Yim (2020) (see Section 6.7).

Regarding the levels of difficulty of the semantic relations, the 'related' relation is the least-correctly predicted relation making it the most challenging one. Figure 6.7 shows different confusion matrices with a heatmap of the binary, SKOS and SKOS+none (all) classification problems. These are the results of the performance of the model trained on the English data and tested on a subset of the English data in MWSA. Therefore, there is a lower number of instances in the test sets of SKOS in comparison to SKOS+none and binary classification. In the binary classification in Figure 6.7a, we see that the model detects sense pairs with 'none' relations more frequently correctly. However, in 13% of cases (358 instances), the model incorrectly predicts no alignment and in 4% (125 instances) of cases, the model predicts that a sense pair should be aligned while the true label suggests otherwise. Similarly, in the SKOS relation in Figure 6.7b, the majority of the correct predictions have an 'exact' relation. However, the model tends to incorrectly predict other relations, namely 'broader', 'narrower' and 'related', as 'exact' as well. The same issue is observed with respect to the 'none' relation in the SKOS+none classification in Figure 6.7c. More results from other datasets are not provided given that it would take much more space.

On average, the SVM method without RBM performs better in terms of F-measure with 0.6845, 0.5459 and 0.3743 scores respectively for the binary, SKOS and SKOS+none classifications.

| Language | Metric | Binary | All | SKOS | RBM-Binary | RBM-All | RBM-SKOS |
|---|---|---|---|---|---|---|---|
| Basque | Accuracy | 0.7879 | 0.5847 | 0.4977 | 0.7037 | 0.5417 | 0.2885 |
| | Precision | 0.7140 | 0.5921 | 0.4365 | 0.6214 | 0.5908 | 0.2073 |
| | Recall | 0.7278 | 0.5845 | 0.4601 | 0.7493 | 0.5255 | 0.5087 |
| | F-measure | 0.7208 | 0.5883 | 0.4480 | 0.6794 | 0.5562 | 0.2946 |
| Bulgarian | Accuracy | 0.7060 | 0.6591 | 0.3405 | 0.7351 | 0.6338 | 0.3647 |
| | Precision | 0.6875 | 0.6479 | 0.3175 | 0.7746 | 0.3446 | 0.3685 |
| | Recall | 0.6932 | 0.6544 | 0.3183 | 0.7291 | 0.4987 | 0.2486 |
| | F-measure | 0.6903 | 0.6511 | 0.3179 | 0.7511 | 0.4076 | 0.2969 |
| Danish | Accuracy | 0.6647 | 0.3482 | 0.2787 | 0.7385 | 0.5008 | 0.2967 |
| | Precision | 0.7454 | 0.2370 | 0.3649 | 0.6059 | 0.6096 | 0.3047 |
| | Recall | 0.7551 | 0.6290 | 0.2287 | 0.5566 | 0.6692 | 0.7304 |
| | F-measure | 0.7502 | 0.3443 | 0.2812 | 0.5802 | 0.6380 | 0.4300 |
| Dutch | Accuracy | 0.8255 | 0.5999 | 0.2475 | 0.8390 | 0.5147 | 0.3634 |
| | Precision | 0.8697 | 0.7859 | 0.3138 | 0.5978 | 0.7782 | 0.3066 |
| | Recall | 0.8824 | 0.7922 | 0.3310 | 0.6733 | 0.3965 | 0.6603 |
| | F-measure | 0.8760 | 0.7890 | 0.3222 | 0.6333 | 0.5254 | 0.4188 |
| English | Accuracy | 0.8900 | 0.8100 | 0.4900 | 0.8016 | 0.6503 | 0.4857 |
| | Precision | 0.8235 | 0.7303 | 0.3931 | 0.6436 | 0.6367 | 0.5553 |
| | Recall | 0.8287 | 0.7641 | 0.4663 | 0.8213 | 0.7935 | 0.3451 |
| | F-measure | 0.8261 | 0.7468 | 0.4266 | 0.7217 | 0.7065 | 0.4257 |
| Estonian | Accuracy | 0.7898 | 0.5892 | 0.4611 | 0.7596 | 0.6275 | 0.4782 |
| | Precision | 0.7606 | 0.6883 | 0.4081 | 0.6353 | 0.6067 | 0.3663 |
| | Recall | 0.2076 | 0.5782 | 0.4402 | 0.2818 | 0.4935 | 0.2244 |
| | F-measure | 0.3262 | 0.6285 | 0.4235 | 0.3905 | 0.5443 | 0.2783 |
| German | Accuracy | 0.7314 | 0.6199 | 0.4958 | 0.7797 | 0.4323 | 0.4421 |
| | Precision | 0.7772 | 0.6474 | 0.4189 | 0.8044 | 0.6634 | 0.4099 |
| | Recall | 0.5441 | 0.5995 | 0.4373 | 0.2288 | 0.2792 | 0.4899 |
| | F-measure | 0.6401 | 0.6225 | 0.4279 | 0.3563 | 0.3930 | 0.4463 |
| Hungarian | Accuracy | 0.7965 | 0.5840 | 0.2295 | 0.8146 | 0.3627 | 0.1520 |
| | Precision | 0.4996 | 0.3014 | 0.2341 | 0.6850 | 0.5980 | 0.2658 |
| | Recall | 0.5447 | 0.3795 | 0.6808 | 0.5672 | 0.7385 | 0.2923 |
| | F-measure | 0.5212 | 0.3360 | 0.3485 | 0.6205 | 0.6609 | 0.2784 |
| Irish | Accuracy | 0.7500 | 0.5575 | 0.2627 | 0.7961 | 0.6084 | 0.2475 |
| | Precision | 0.8442 | 0.4658 | 0.3184 | 0.7903 | 0.4252 | 0.3025 |
| | Recall | 0.8446 | 0.3985 | 0.4615 | 0.5247 | 0.5465 | 0.2540 |
| | F-measure | 0.8444 | 0.4295 | 0.3768 | 0.6306 | 0.4783 | 0.2761 |
| Italian | Accuracy | 0.5908 | 0.5543 | 0.4448 | 0.7723 | 0.4626 | 0.4301 |
| | Precision | 0.5255 | 0.4298 | 0.2880 | 0.7569 | 0.4631 | 0.4056 |
| | Recall | 0.6647 | 0.5264 | 0.4216 | 0.4505 | 0.6867 | 0.3127 |
| | F-measure | 0.5869 | 0.4732 | 0.3422 | 0.5649 | 0.5532 | 0.3532 |
| Serbian | Accuracy | 0.8005 | 0.3253 | 0.2755 | 0.8235 | 0.4143 | 0.3296 |
| | Precision | 0.7678 | 0.4857 | 0.4306 | 0.7351 | 0.3770 | 0.2149 |
| | Recall | 0.6573 | 0.6940 | 0.2710 | 0.7746 | 0.4845 | 0.5553 |
| | F-measure | 0.7083 | 0.5715 | 0.3326 | 0.7543 | 0.4240 | 0.3099 |
| Slovenian | Accuracy | 0.8429 | 0.3613 | 0.2613 | 0.7893 | 0.3957 | 0.3163 |
| | Precision | 0.7308 | 0.2319 | 0.4698 | 0.7862 | 0.3859 | 0.2097 |
| | Recall | 0.8322 | 0.4507 | 0.2861 | 0.4164 | 0.2809 | 0.3302 |
| | F-measure | 0.7782 | 0.3062 | 0.3556 | 0.5445 | 0.3251 | 0.2565 |
| Spanish | Accuracy | 0.7379 | 0.5467 | 0.3028 | 0.8071 | 0.5438 | 0.5848 |
| | Precision | 0.7978 | 0.5507 | 0.3321 | 0.7940 | 0.4254 | 0.3957 |
| | Recall | 0.8037 | 0.5315 | 0.4004 | 0.6018 | 0.2068 | 0.3859 |
| | F-measure | 0.8007 | 0.5410 | 0.3631 | 0.6847 | 0.2783 | 0.3907 |
| Portuguese | Accuracy | 0.7131 | 0.6662 | 0.5171 | 0.7314 | 0.5569 | 0.4287 |
| | Precision | 0.4929 | 0.5823 | 0.5352 | 0.7772 | 0.6941 | 0.4045 |
| | Recall | 0.3747 | 0.7041 | 0.5347 | 0.5441 | 0.2232 | 0.3815 |
| | F-measure | 0.4257 | 0.6374 | 0.5349 | 0.6401 | 0.3378 | 0.3926 |
| Russian | Accuracy | 0.6088 | 0.5890 | 0.3775 | 0.7580 | 0.5976 | 0.3310 |
| | Precision | 0.7292 | 0.6383 | 0.2728 | 0.7338 | 0.7377 | 0.3271 |
| | Recall | 0.8221 | 0.4443 | 0.3674 | 0.6823 | 0.7039 | 0.4775 |
| | F-measure | 0.7729 | 0.5239 | 0.3131 | 0.7071 | 0.7204 | 0.3882 |
| **Average** | Accuracy | 0.7491 | 0.5597 | 0.3655 | 0.7766 | 0.5229 | 0.3693 |
| | Precision | 0.7177 | 0.5343 | 0.3689 | 0.7161 | 0.5558 | 0.3363 |
| | Recall | 0.6789 | 0.5821 | 0.4070 | 0.5735 | 0.5018 | 0.4131 |
| | F-measure | 0.6845 | 0.5459 | 0.3743 | 0.6173 | 0.5033 | 0.3491 |

Table 6.6: Results of the experiments on semantic induction on binary, SKOS and all relations, namely exact, related, broader, narrower and none, with and without an RBM. The highest F-measure scores in binary, all and SKOS classifications per language are respectively indicated in cyan, green and violet. Darker shades indicate higher values.

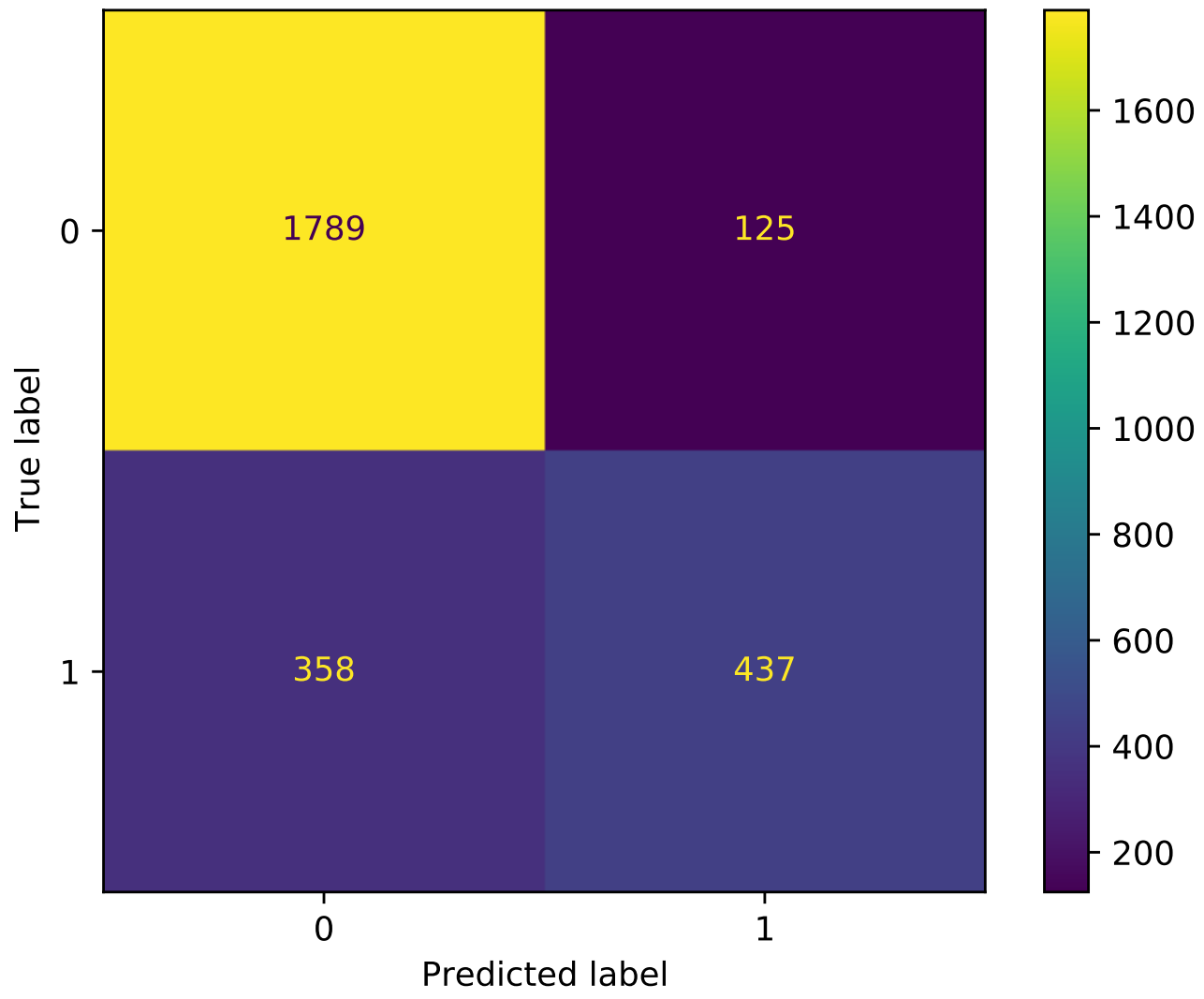

(a) binary classification

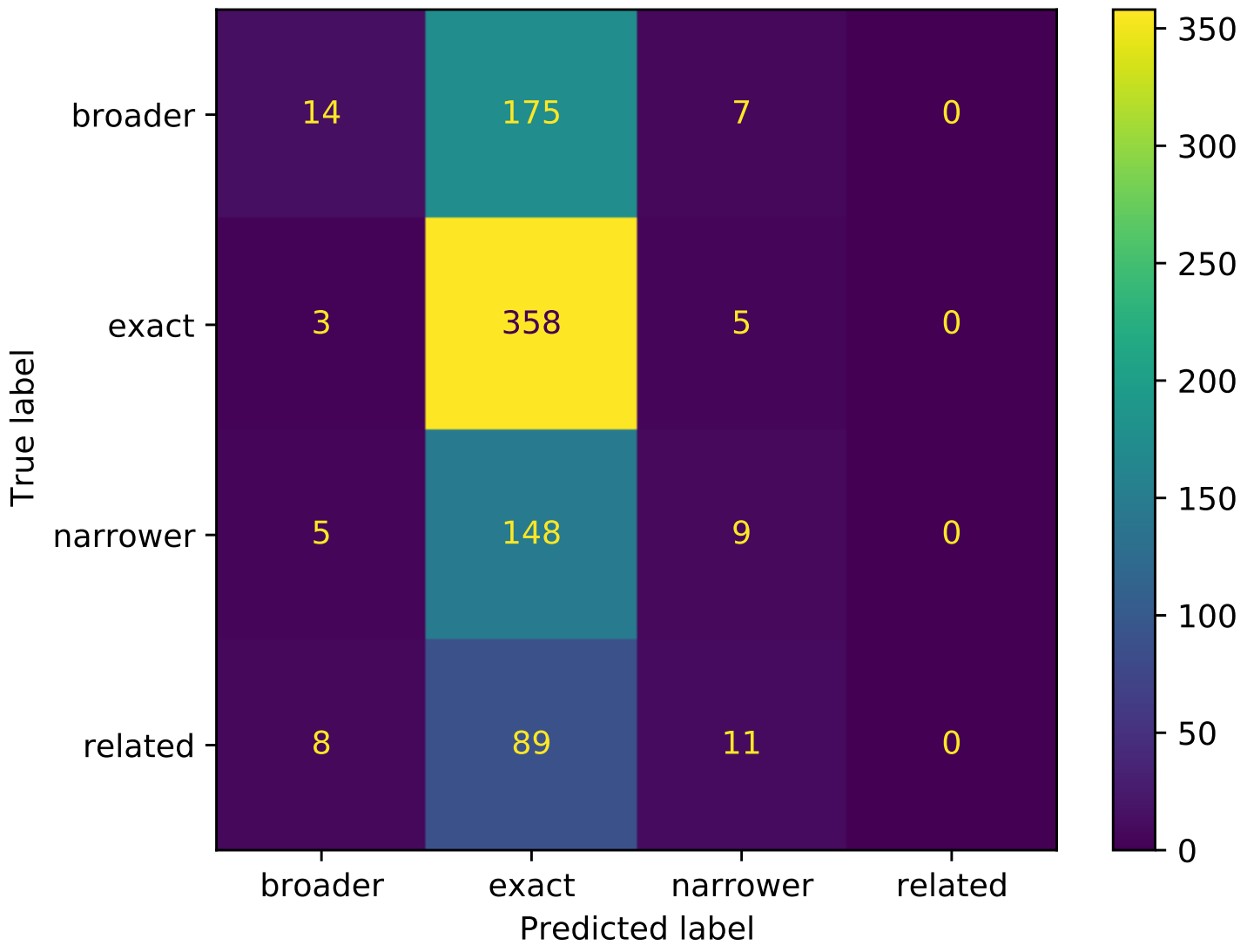

(b) SKOS classification

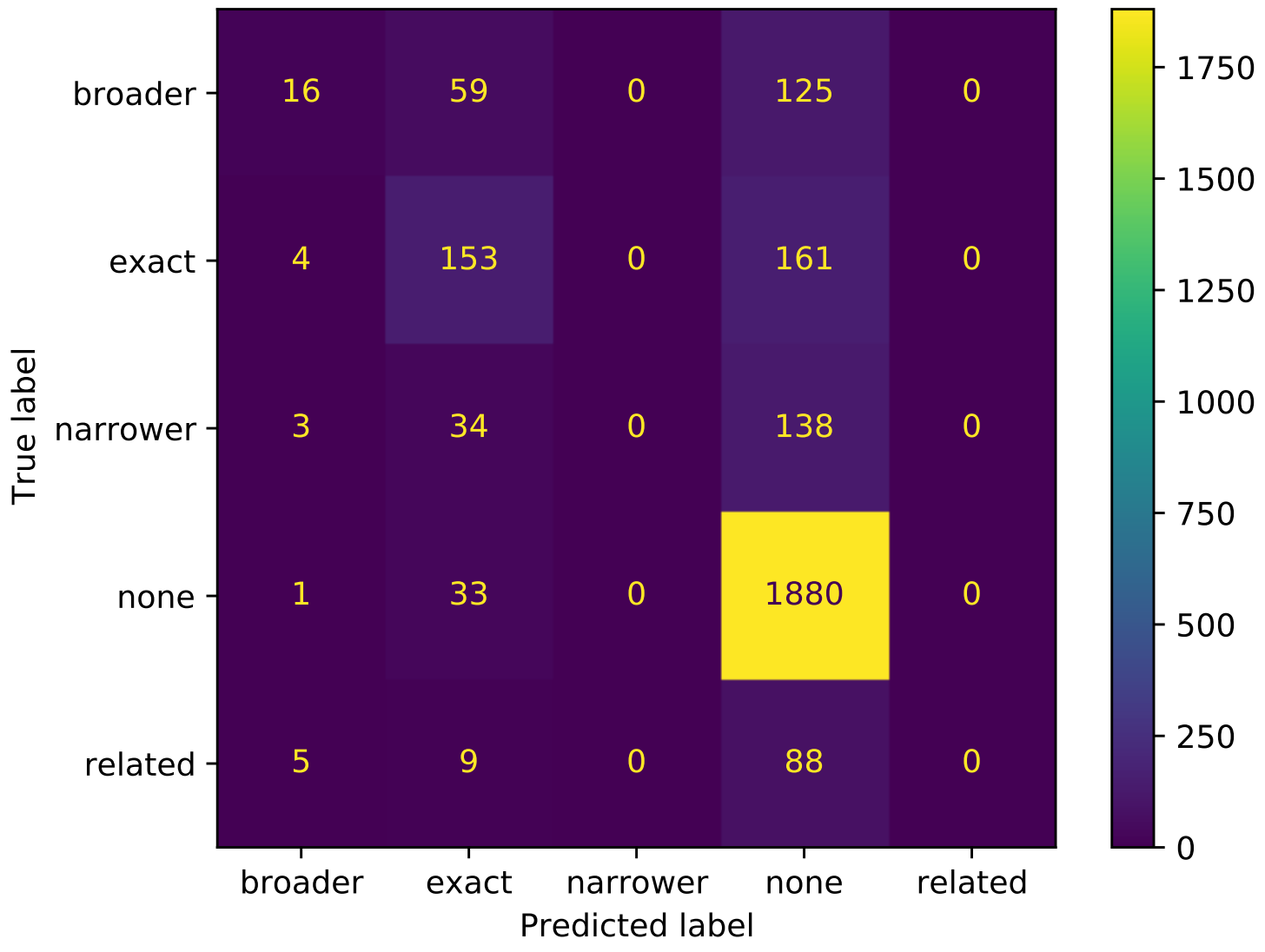

(c) SKOS+none classification

**Figure 6.7:** Confusion matrices of various classification models evaluated on the English test set

| TUBE (verb) | | |
|---|---|---|
| **Sense definitions in resource 1, the Princeton English WordNet ($R_1$)** | | |
| $R_{1A}$ | provide with a tube or insert a tube into | |
| $R_{1B}$ | ride or float on an inflated tube | |
| $R_{1C}$ | convey in a tube | |
| $R_{1D}$ | place or enclose in a tube | |
| **Sense definitions in resource 2, Webster's Dictionary 1913 ($R_2$)** | | |
| $R_{2A}$ | to furnish with a tube | |
| **Alignments** | | |
| Gold-standard | $R_{1A}$ exact $R_{2A}$ | |
| Baseline | $R_{1A}$ narrower $R_{2A}$ | (similarity score: 0.2793) |
| System 1 | $R_{1A}$ narrower $R_{2A}$ | (similarity score: 0.8013) |
| BERT | $R_{1A}$ narrower $R_{2A}$ | (similarity score: 0.93) |

**Table 6.7:** A comparison of the output of various systems for aligning sense definitions of TUBE (verb). BERT refers to the shared task system submitted by Bajcetic and Yim (2020).

| GLOW (verb) | | |
|---|---|---|
| **Sense definitions in resource 1, the Princeton English WordNet ($R_1$)** | | |
| $R_{1A}$ | shine intensely, as if with heat | |
| $R_{1B}$ | be exuberant or high-spirited | |
| $R_{1C}$ | experience a feeling of well-being or happiness, as from good health or an intense emotion | |
| $R_{1D}$ | have a complexion with a strong bright color, such as red or pink | |
| $R_{1E}$ | emit a steady even light without flames | |
| **Sense definitions in resource 2, Webster's Dictionary 1913 ($R_2$)** | | |
| $R_{2A}$ | to make hot; to flush. | |
| $R_{2B}$ | to shine with an intense or white heat; to give forth vivid light and heat; to be incandescent. | |
| $R_{2C}$ | to exhibit a strong, bright color; to be brilliant, as if with heat; to be bright or red with heat or animation, with blushes, etc. | |
| $R_{2D}$ | to feel hot; to have a burning sensation, as of the skin, from friction, exercise, etc.; to burn. | |
| $R_{2E}$ | to feel the heat of passion; to be animated, as by intense love, zeal, anger, etc. | |
| **Alignments** | | |
| Gold-standard | $R_{1A}$ exact $R_{2B}$ | |
| | $R_{1C}$ exact $R_{2D}$ | |
| | $R_{1E}$ exact $R_{2C}$ | |
| Baseline | $R_{1A}$ narrower $R_{2B}$ | (similarity score 0.1524) |
| | $R_{1C}$ narrower $R_{2E}$ | (similarity score: 0.1836) |
| System 1 | $R_{1A}$ narrower $R_{2B}$ | (similarity score 0.7455) |
| | $R_{1C}$ related $R_{2E}$ | (similarity score: 0.6429) |
| BERT | $R_{1A}$ none $R_{2A}$ | (similarity score: 0.952) |

**Table 6.8:** A comparison of the output of various systems for aligning sense definitions of GLOW (verb). BERT refers to the shared task system submitted by Bajcetic and Yim (2020).

### 6.6.3 System 2: Length-limited Alignment

In order to evaluate the impact of length of definitions on the alignment task, another study is carried out where models are trained based on a maximum number of tokens per definition. To this end, we focus on the Danish dataset which has the highest number of senses among all the datasets (see Table 5.8). In addition, the Danish dataset has a few other characteristics that makes it an interesting choice for experiments: (i) it has the highest number of tokens, making definitions wordier as Figure 5.9 shows, (ii) it aligns a historical and modern dictionary of Danish and (iii) both DDO and ODS represent senses in a hierarchy.

In this experiment, we particularly evaluate the performance of the following similarity detection methods to detect whether two sense definitions should be aligned or not, i.e. binary classification:

- String metrics as introduced in Section 4.4.1, namely longest common substring, length ratio, average word length ratio, Jaccard, Dice, and Containment and smoothed Jaccard.
- Similarity based on word embeddings trained on the Danish dictionaries *Ordbog over det danske Sprog* (ODS)[13] (Dahlerup, 1918) and *Den Danske Ordbog* (DDO) (Farø et al., 2003) as described in Section 6.4.2. To do so, we used Sørensen and Nimb (2018)'s word embeddings model trained using a corpus of approximately one billion running words of modern Danish. The model is trained using the GloVe model (Pennington et al., 2014) with 500 features, a window size of 5 and a minimum occurrence of 5 (any types below this threshold are discarded), and used the Skip-Gram version of the model.
- Automatic feature extraction in which we automatically extract useful features using the two previous methods of string similarity and word embeddings combined. Features are extracted using the training set in such a way that the performance of the extracted features is maximal among the whole combination of features. To do so, an SVM is used with cross-validation.

To do so, we summarize sense definitions as the ODS sense descriptions are often very detailed and syntactically complex and the borders between definition text, usage examples and idioms still remain to be fully identified in the XML structure. For the experiments in this section, in addition to the full original definition, we create three other datasets in such a way that the number of space-separated tokens is limited to only 15, 20 and 25 tokens.

Once the similarity scores are extracted based on the aforementioned methods, we automatically align senses in a greedy approach where, starting with no alignment, the sense pairs are ordered based on the similarity score and then aligned in such a way that a sense is linked to only one other sense in the other resource, i.e. ODS

13 https://ordnet.dk/ods_en

and DDO. Although this bijective constraint ignores polysemous senses, it yields a more diverse combination of sense matches. Table 6.9 provides the results of our experiments with respect to the number of tokens in ODS.

Although the precision of the models in automatically detecting the similarity of two senses varies in a close range of 50.3% (All-auto) and 66.7% (15-Word embeddings), there is a more significant difference between the recall of each dataset and so, in F-measure. The lowest recall appears in aligning DDO with ODS with its original senses. In other words, when senses with all the composing parts, such as usage examples and idioms, are aligned with DDO, all the three models can predict a link over 50% correctly. However, they only succeed in less than 10% of cases to retrieve relevant senses. Truncating senses from 25 tokens to 15 significantly improves both the precision and recall, proving our initial observation of the noisiness of senses in ODS. Figure 6.8 illustrates the correlation of senses sizes with F-measures in all the models.

| ODS sense size | Model | Precision | Recall | F-measure |
|---|---|---|---|---|
| 15 | String metrics | 0.653 | **0.481** | 0.554 |
| | Word embeddings | **0.667** | 0.48 | **0.558** |
| | Auto | 0.64 | 0.466 | 0.54 |
| 20 | String metrics | 0.615 | 0.443 | 0.515 |
| | Word embeddings | 0.647 | 0.467 | 0.543 |
| | Auto | 0.633 | 0.458 | 0.532 |
| 25 | String metrics | 0.575 | 0.219 | 0.317 |
| | Word embeddings | 0.559 | 0.212 | 0.308 |
| | Auto | 0.585 | 0.222 | 0.321 |
| All | String metrics | 0.547 | 0.098 | 0.167 |
| | Word embeddings | 0.507 | 0.097 | 0.163 |
| | Auto | 0.503 | 0.094 | 0.158 |

Table 6.9: The performance of our similarity detection models for automatic alignment of DDO and ODS within a specific limit of space-separated tokens (15, 20, 25 and all tokens)

The highest F-measure of 55.8% belongs to the ODS dataset with a maximum of 15 tokens and trained with the word embeddings model. In comparison to the baselines presented in Table 6.5 where an F-measure of 0.043 is reported, such an improvement is promising. This being said, the knowledge graph based approach described in the System 1 (see Section 6.6.2) outperforms this model with an F-measure of 0.7502. However, it is important to conclude based on the current experiment that the conciser definitions tend to facilitate the alignment task significantly.

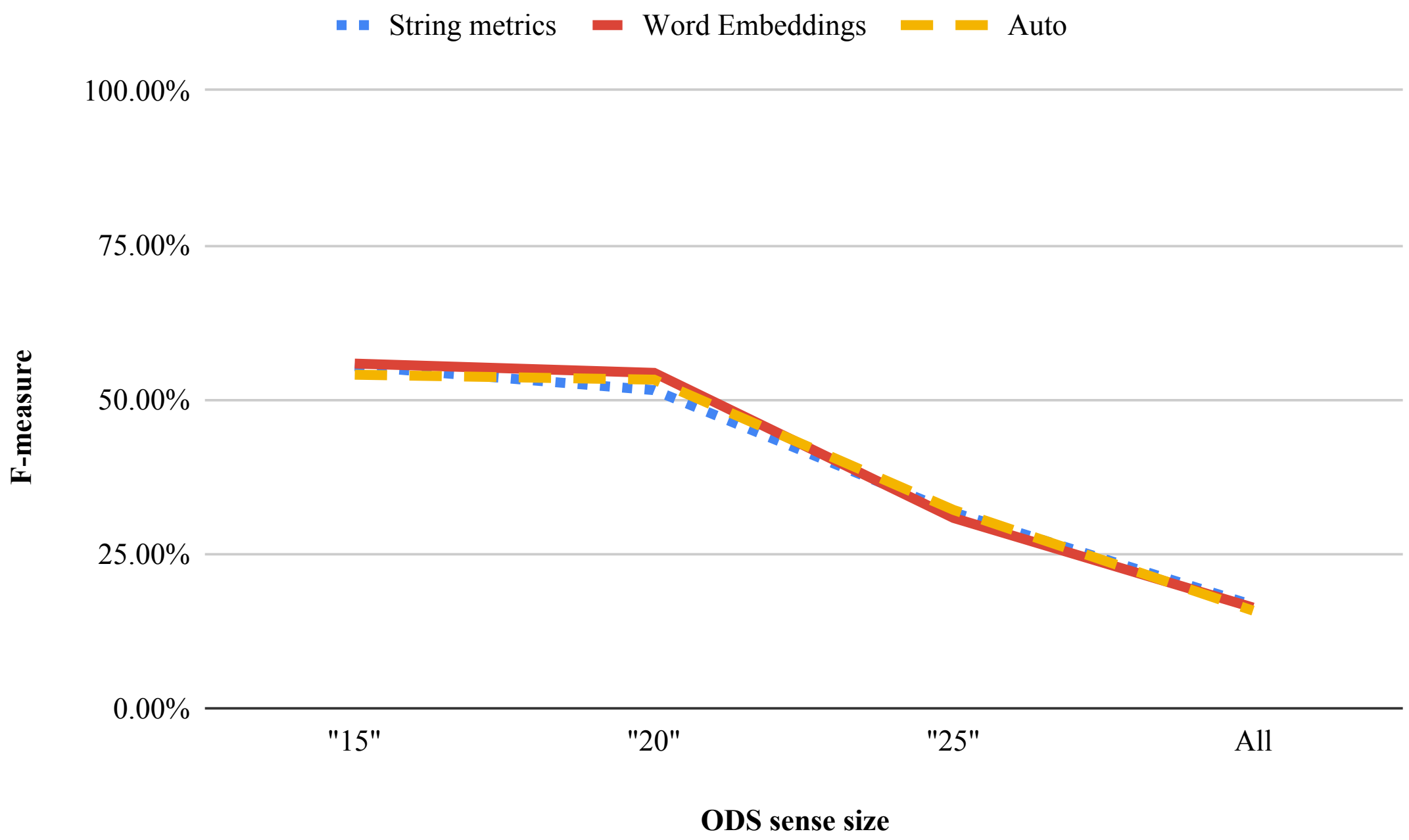

**Figure 6.8:** The correlation of sense sizes in ODS with F-measure using various methods

## 6.7 ELEXIS MONOLINGUAL WORD SENSE ALIGNMENT TASK

In order to motivate other researchers to address the MWSA task, we organized a shared task in the 3rd GLOBALEX Workshop at LREC 2020[14]. The task was organized using CodaLab[15] and three external teams participated, although not all teams participated for all languages. The evaluation of the shared task was carried out as described in this chapter, using accuracy, precision, recall and F-measure. The performance of the participating systems is compared with the baseline and our experiments in System 1 described in Section 6.6.2. The following approaches were suggested by the participants:

**RACAI** (Pais et al., 2020) The RACAI system viewed this task as a case of word-sense disambiguation, from this multiple features were extracted including scores based on the Lesk algorithm (Lesk, 1986) as well as features from BERT (Devlin et al., 2018) and other features, which were combined using a random forest (Ho, 1995).

**ACDH** (Bajcetic and Yim, 2020) This system defines the MWSA task as sentence pair classification task for which BERT can be fine-tuned since its use of self-attention mechanism (Vaswani et al., 2017) to encode concatenated text pairs effectively includes bidirectional cross attention between two definitions. A variety of features were combined in this approach including simple similarity methods such

14 https://globalex2020.globalex.link/globalex-workshop-lrec2020-about-globalex-lrec2020/
15 https://competitions.codalab.org/competitions/22163

as used in the baseline as well as similarities coming from ELMo (Peters et al., 2018b) and BERT (Devlin et al., 2018). These were also combined using a supervised learning framework, and different settings were used for each language.

**UNIOR** (Manna et al., 2020) This system used BERT as well as Siamese LSTMs (Mueller and Thyagarajan, 2016) improved with lexical semantic information related to the lemma's part-of-speech category.

The results of various systems along with System 1 are presented in Table 6.10 in terms of accuracy and F-measure. Given that the participants did not report the performance of their systems with respect to SKOS classification, we only compared results of the binary and SKOS+none (all) tasks. Even though all the evaluation results were not reported for all systems and for all languages, there are promising improvements in some of the systems, particularly ACDH and UNIOR in comparison to System 1. However, all systems can be said to have performed best on some of the tasks (even the baseline) and given that all systems used BERT, more research is needed into the best way to fine-tune BERT for this task.

| Language | Metric | Binary | | | | All | | | |
|---|---|---|---|---|---|---|---|---|---|
| | | System 1 | RACAI | ACDH | UNIOR | System 1 | RACAI | ACDH | UNIOR |
| Basque | Accuracy | **0.7879** | | | | **0.5847** | | 0.407 | |
| | F-measure | **0.7208** | | 0.342 | | **0.5883** | | | |
| Bulgarian | Accuracy | **0.7351** | | | | **0.6591** | | 0.395 | |
| | F-measure | **0.7511** | | 0.475 | | **0.6511** | | | |
| Danish | Accuracy | **0.7385** | | | | 0.5008 | | **0.522** | |
| | F-measure | **0.7502** | | 0.379 | | **0.638** | | | |
| Dutch | Accuracy | **0.839** | | | | 0.5999 | 0.798 | **0.94** | 0.931 |
| | F-measure | **0.876** | 0.48 | 0.35 | 0.145 | **0.789** | | | |
| English | Accuracy | **0.89** | | | | 0.81 | **0.944** | 0.766 | 0.759 |
| | F-measure | **0.8261** | 0.31 | 0.691 | 0.634 | **0.7468** | | | |
| Estonian | Accuracy | **0.7898** | | | | **0.6275** | | 0.565 | |
| | F-measure | 0.3905 | | **0.754** | | **0.6285** | | | |
| German | Accuracy | **0.7797** | | | | 0.6199 | | **0.798** | |
| | F-measure | 0.6401 | | **0.667** | | **0.6225** | | | |
| Hungarian | Accuracy | **0.8146** | | | | **0.584** | | | |
| | F-measure | **0.6205** | | | | **0.6609** | | | |
| Irish | Accuracy | **0.7961** | | | | **0.6084** | | 0.549 | |
| | F-measure | **0.8444** | | 0.739 | | **0.4783** | | | |
| Italian | Accuracy | **0.7723** | | | | 0.5543 | 0.761 | 0.537 | **0.766** |
| | F-measure | 0.5869 | 0.463 | 0.529 | **0.741** | **0.5532** | | | |
| Serbian | Accuracy | **0.8235** | | | | 0.4143 | | **0.599** | |
| | F-measure | **0.7543** | | 0.269 | | **0.5715** | | | |
| Slovenian | Accuracy | **0.8429** | | | | 0.3957 | | **0.442** | |
| | F-measure | **0.7782** | | 0.268 | | **0.3251** | | | |
| Spanish | Accuracy | **0.8071** | | | | 0.5467 | 0.786 | | **0.829** |
| | F-measure | 0.8007 | 0.661 | | **0.81** | **0.541** | | | |
| Portuguese | Accuracy | **0.7314** | | | | 0.6662 | | 0.87 | **0.933** |
| | F-measure | 0.6401 | | 0.441 | **0.641** | **0.6374** | | | |
| Russian | Accuracy | **0.758** | | | | 0.5976 | | **0.606** | |
| | F-measure | **0.7729** | | 0.512 | | **0.7204** | | | |

Table 6.10: The highest results reported by the participants of the MWSA shared task (2020) along with the highest results from System 1. The highest values are indicated in bold.

## 6.8 CONCLUSION AND CONTRIBUTIONS

In this chapter, the monolingual word sense alignment task is defined along with a pre-existing framework for solving this called NAISC. We looked at textual similarity metrics for which there are a large number of methods that are effective for estimating similarity; however, the task of distinguishing between exactly equivalent senses with broader/narrower senses is still a challenging task. We then looked at non-textual linking methods methods that are effective for a few kinds of dictionary linking tasks, especially with large-scale knowledge graphs such as Wikidata. Furthermore, we examined the constraints that can be used to find the best overall linking between senses and showed how these can be solved. Finally, a few experiments are carried out using the monolingual lexicographic datasets from 15 languages described in the previous chapter. We first develop a baseline system and then, model the alignment task as a classification problem and apply a knowledge graph. These experiments aim at classifying sense matches across dictionaries and also, prediction of the semantic relationship between two given senses, namely narrower, broader, exact and related.

A set of the experiments are based on manually-extracted features along with a representation learning technique, RBM, and semantic relationship detection. We demonstrate that the performance of classification methods varies based on the type of semantic relationships due to the nature of the task but performs better than the baseline. Despite the increasing popularity of deep learning methods in providing state-of-the-art results in various NLP fields, we believe that evaluating the performance of feature-engineered approaches is an initial and essential step to reflect the difficulties of the task. In the same vein, we organized a shared task on the same task inviting researchers to propose new ideas and insights to improve the current alignment results. The results indicate a better performance of the proposed approaches with respect to the baseline with differences between them, achieving strong performance with state-of-the-art methods such as BERT.

As we saw in another experiment, the length of sense definition plays an important role in the alignment task in such a way that conciser definitions tend to be aligned more easily. This being said, a more complete investigation is required to fully understand the characteristics of concise definitions, whether related to the linguistic features or the definition paradigm, that gives automatic approaches more effectiveness in addressing the alignment problem. We believe that the development of the benchmark and providing linking techniques within NAISC facilitates not only aligning lexicographical resources, but also makes it possible to evaluate various systems in the future as a part of the ELEXIS dictionary infrastructure.

One of the limitations that should be addressed is the evaluation of the models for alignment of senses based on part-of-speech tags to analyze the impact of grammatical roles on the alignment task and also, differences of definitions and senses in the task of semantic relation detection.

# 7 CONCLUSIONS

This thesis sheds light on the alignment of lexical semantic resources, notably dictionaries. The task of combining dictionaries from different sources is difficult, especially for the case of mapping the senses of entries, which often differ significantly in granularity and coverage. After providing a background on the language and linguistic resources that are of importance in natural language processing, such as wordnets, generative lexicons, knowledge graphs and language models, a review of the previous efforts in aligning language resources is presented in an essentially concise but systematic manner. The significance of the alignment task to increase the interoperability of resources, reduce the heterogeneity of linguistic data and facilitate the task of lexicographers and language experts for creating resources and documenting languages upholds the necessity of addressing this topic.

In this chapter, we provide the conclusions based on the contributions that were mainly presented in Chapter 4, Chapter 5 and Chapter 6. In addition to the contributions, some of the limitations of the current work are provided along with the future directions paving the way for potentially more substantial advances in this realm.

## 7.1 RESEARCH CONTRIBUTIONS

The main contributions of this thesis can be summarized as follows:

### 7.1.1 Benchmarking Word Sense Alignment — Chapter 5

One major limitation regarding previous work was with respect to the nature of the data used for the WSA task. Expert-made resources, such as the Oxford English Dictionary, require much effort to be created and therefore, are not available under an open-source license as collaboratively-curated ones like Wiktionary are due to copyright restrictions. On the other hand, the latter resources lack domain coverage and descriptive senses. To address this, we present a set of 17 datasets containing monolingual dictionaries in 15 languages, annotated with five semantic relations according to SKOS (Miles and Bechhofer, 2009), namely, exact, broader, narrower, related and none. The annotation was carried out by language experts and lexicographers within the group of ELEXIS volunteers and partners.

The main objectives of creating a benchmark for MWSA were the following:

(A) Collect lexicographical data from chiefly expert-made resources, especially dictionaries. In the collecting of the data, common practices of data management were also considered, particularly by specifying various types of data by their identifiers, making them more compatible with the semantic web standards and the Ontolex-Lemon ontology.

(B) Carry out an annotation task where the micro-structure of two entries with the same part-of-speech tags are aligned at sense level. The alignment is further completed with semantic annotations denoting the type of semantic relations that two senses via their definitions would have. This proved to be an effective way to capture information regarding sense nuances and potential biases that the descriptive and explanatory paradigms of sense defining would impose on the alignment task. Furthermore, the sense granularity between two dictionaries is rarely such that we would expect one-to-one mapping between the senses of an entry. In this respect, we followed a simple approach such as that in SKOS providing different kinds of linking predicates which is described as follows:

**EXACT** The sense are the same, for example, the definitions are simply paraphrases.

**BROADER** The sense in the first dictionary completely covers the meaning of the sense in the second dictionary and is applicable to further meanings.

**NARROWER** The sense in the first dictionary is entirely covered by the sense of the second dictionary, which is applicable to further meanings.

**RELATED** There are cases when the senses may be equal but the definitions in both dictionaries differ in key aspects.

**NONE** There is no match for these senses.

Given that some of the semantic relations, such as narrower and broader, are not symmetric, the order of the source and target senses is important in determining the semantic relation correctly. In addition, the senses within the two resources which belong to the same lemma but are not annotated with a SKOS relation, are considered with a *'none'* relation.

The annotation was implemented by means of dynamic spreadsheets that provided a simple but effective manner to complete the annotation. It should be noted that the senses and definitions of the headwords to be annotated were conditioned based on part-of-speech; this means that not only the headwords in two dictionaries should be identical, ignoring spelling variations that are systematically taken care of by being unified as we saw in the case of Danish in Section 5.3.1, but they should also be of the same morphosyntactic category.

Figure 7.1 shows an alignment example of the entry *domestication* (noun, feminine) in the *Trésor de la Langue Française Informatisé* (TLFi) and *Wiktionnaire* – the French

Wiktionary. The two major differences between these two resources, which occur oftentimes across other resources, is (i) the differences of distinction of senses and (ii) dissimilar and sometimes conceptually different definitions for the same senses. For instance, the sense of *domestication* as 'taming of animals' is mentioned as a separate sense to 'action of domesticating' in TLFi, while Wiktionnaire is limited to only one sense and that corresponds only to the 'action of domesticating' sense. On the other hand, 'subjugation, enslavement (of people, human groups, ideas).' is referring to a semantically different concept than that of 'the act of subjecting someone to housework.', which may be considered as having a broader or narrower relation. Therefore, definitions may not essentially convey the same meanings, even though they are approximately in the same context. This is due to many reasons, among which, editorial preferences, lexicographer's style, definition paradigms, lexical semantic theories and semantic shifts play an important role. Figure 7.1 also shows the alignment and annotation of semantic relations carried out by the annotators[1].

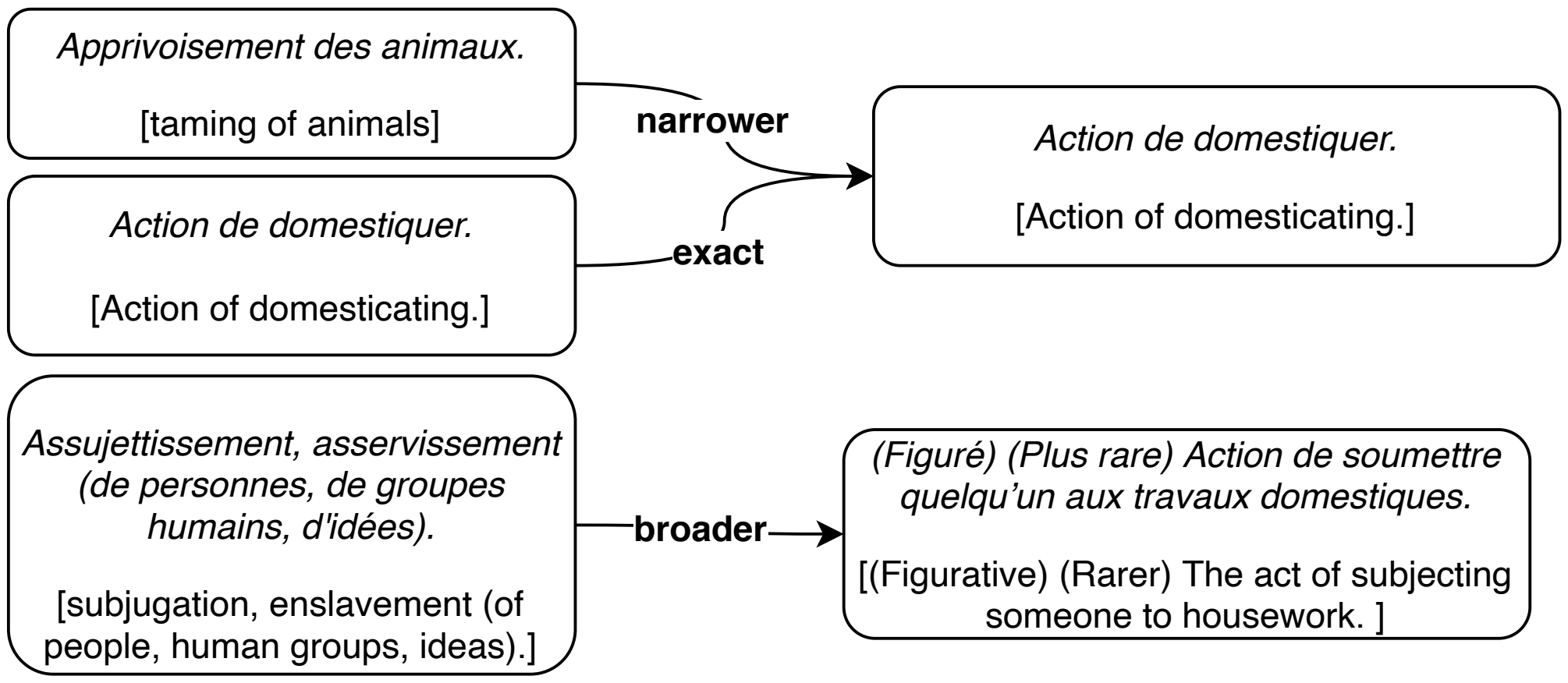

**Figure 7.1:** An alignment of sense definitions of *domestication* (noun, feminine) in the *Trésor de la Langue Française* (to the left) and *Wiktionnaire* (to the right). Translations are provided in brackets.

The inter-annotator agreement scores indicate that the task is not equally challenging for all resources and all languages. This indicates that it is certainly not easy to decide which relationship is to be used among word senses and how to align them. Regardless of the sense granularity and coverage in two resources, the definition of word senses, which often follows a paradigm as discussed in Section 2.2.2, creates further challenges and nuances in meaning that make the alignment task more difficult. However, this methodology was broadly effective and can simplify the development of machine-learning-based classifiers for sense alignment prediction.

The datasets are available in JSON, tabular and RDF formats and external keys such as `meta_ID` and `external_ID` enable future lexicographers to integrate the annotations in external resources. The benchmark can be used for training models,

1 The French dataset was not included in the experiments as it was created at the end of my Ph.D.

fine-tuning existing contextual embeddings like BERT (Devlin et al., 2018) and more importantly, evaluation purposes.

### 7.1.2 Alignment of dictionaries at the sense level Chapter 4 & 6

Following the curation of the benchmark, we propose solutions for the automatic alignment of word senses. The alignment techniques are based on a variety of methods ranging from basic string similarity features to finer representations using semantic similarity. Considering the literature, various components of the WSA task have been matters of research previously. However, a few of the previous papers address WSA as a specific task on its own in the context of dictionaries. In the same vein, our focus is on providing explainable observations for the task of WSA using manually-extracted features and analyzing the performance of traditional machine learning algorithms for word sense alignment as a classification problem. Despite the increasing popularity of deep learning methods which provide state-of-the-art results in various NLP fields, we believe that evaluating the performance of feature-engineered approaches is an initial and essential step to reflect the difficulties of the task and also, the expectations from the future approaches.

Regarding the WSA task, the comparison of the definitions of the lexical entries is the most obvious and effective method for establishing similarity between senses in two dictionaries and is the primary method that humans would use. As such, it makes sense to focus our efforts on developing an artificial intelligence approach for the task of estimating the similarities of definitions, which is a kind of semantic textual similarity as explored in tasks at SemEval (Agirre et al., 2016b). Moreover, we move beyond similarity to also predict the taxonomic type of the relationship between senses. Our methodology in MWSA relies on the two sub-tasks of semantic similarity detection and semantic similarity induction. To this end, three main approaches are explored. Firstly use simple text features to provide a baseline for the task. Secondly, a few classification models are created for the alignment task by focusing on the binary, SKOS and SKOS+none classification tasks. We also used data augmentation techniques based on the properties of the semantic relations to create more data instances and also, applied a representation learning technique using RBM. While the first classification objective only predicts whether two sense definitions can be aligned or not, the two other tasks can also induce the semantic relation between sense definitions. And finally, deep learning methods including BERT are introduced and the experiment results are compared.

As the datasets are publicly available, we carried out a shared task on the task of monolingual word sense alignment across dictionaries as part of the GLOBALEX 2020 – Linked Lexicography workshop at the 12th Language Resources and Evaluation Conference (LREC 2020) which took place on Tuesday, May 12 2020 in Marseille (France). The results of this shared task are also presented and compared with the contributions of the thesis.

Based on the results of various experiments, the following findings can be enumerated:

(A) The MWSA can be effectively addressed using methods based on classification and fine-tuning of embeddings and also, by incorporating knowledge graphs like ConceptNet (Speer et al., 2016). Although all the proposed systems outperform the baseline, the performance of the proposed techniques does not essentially follow the same pattern across datasets and languages.

(B) The alignment of sense definitions with a narrower, broader or related is more challenging than exact and none. This is mainly due to the imbalance of data and also, the difficulty of the task, even for human annotators, to determine the type of semantic relation.

(C) Among the explored methods, those based on fine-tuning contextual embeddings like BERT perform more efficiently. However, this is yet to be fully analyzed given that the experiments presented in the thesis that rely on deep learning only address a few languages in the benchmark. In some other cases, the classification-based techniques perform better.

(D) In how senses are defined, structural differences and dissimilarities in the content are the major factors that determine the level of difficulty of an alignment case. This can, for instance, be due to the presence of a hierarchical sense organization or additional excessive descriptions in a definition, such as synonyms and near-synonyms along with cross-references.

It is worth noting that initially, we also focused on alignment approaches based on graph analysis. To this end, we proposed the weighted bipartite b-matching algorithm in Chapter 4 to find the optimal combination of alignments in such a way that the most similar items are linked according to a few conditions. Given that this contribution was made before the creation of the benchmark, the performance of this technique is reported using another dataset limited to senses from English WordNet and Wiktionary.

### 7.1.3 Alignment of dictionaries at the entry level — Chapter 4

In addition to the alignment of dictionaries at the sense level, this thesis also partially focuses on the alignment of dictionaries at the entry-level. In this task, given a set of bilingual dictionaries, the objective is to generate new translation pairs by inducing among the existing ones without using any external resources that provide a direct translation between the source and the target dictionaries. Additionally, the alignment task should be carried out based on the senses of each entry (see Figure 4.6 for an example of 'spring' (noun)). This task is addressed mainly using graph-based methods that leverage the structural information to generate new translation pairs. Two techniques are proposed based on paths and cycles between source and target words in two dictionaries and using other dictionaries as pivots.

Our experiments were carried out in the context of the TIAD workshops. The results indicate that structural information extracted from bilingual dictionaries are useful to employ graph-based methods for the translation inference task. To this end, cycle-based and path-based approaches are suggested. The basic idea of these methods relies on the transitivity relation that leads an entry in a language to another entry in another language, as in 'book'-'*livre*' (French), '*livre*'-'*Buch*' (German), then 'book'-'*Buch*'. Although these techniques are effective, they are limited to the translation pairs provided in the dictionaries and cannot deal with missing translations, unconnected entries and highly polysemous words. This should be addressed as future work by incorporating external resources.

### 7.1.4 Naisc — Section 6.3

From a practical point of view, the contributions of this thesis have been integrated into an existing framework and tool for creating mappings between two dictionaries, called Naisc[2]. Naisc's architecture is intended as an experimental framework into which many components can be integrated. Although many of the techniques included in Naisc can also be used to create multilingual linking between dictionaries and also linking between other kinds of datasets, our focus within this thesis is on only the monolingual word sense alignment task.

## 7.2 REVISITING THE RESEARCH QUESTIONS

Revisiting the research questions that were introduced in Section 1.3 of Chapter 1, we can provide the following responses according to the contributions of the thesis:

**RQ1.** Do graph-based methods improve the performance of lexicographical alignment systems?

**RQ1.1** Given the structural properties of lexicographical data, it is an effective way to model lexical and semantic data as graphs with nodes as lexicalized items, e.g. words or senses, and edges as translations or the type of semantic relation. Graph-based methods can also be ideal to set constraints on the alignment tasks as in bijective or taxonomic alignment proposed in Section 4.2.

**RQ1.2** Both path-based and cycle-based techniques in graph analysis can be effective to generate translation pairs across bilingual dictionaries. According to

2 https://github.com/insight-centre/naisc

the experiments, the path-based approach performs better than the cycle-based one thanks to the wider range of nodes that the first methods visit to calculate a more reliable confidence score for translation pairs. However, the algorithmic efficiency of these techniques hinders their employment in large datasets as they require traversing graphs. Therefore, more optimizations are required.

**RQ2.** How can the output of a monolingual word sense alignment system be evaluated in a multilingual context?
As described in Chapter 5, a benchmark is created that contains a set of 17 datasets of expert-made resources in 15 languages. Although individual datasets in the benchmark provide monolingual data, the benchmark can be used for many languages, making it a valuable resource that fills the gap regarding lesser-resourced languages, like Russian and Serbian, which are included in the benchmark.

**RQ2.1** In the context of historical and modern dictionaries, we study the case of Danish and Portuguese and show that beyond differences due to orthographic reforms, semantic change and semantic shift are the most important issues when aligning lexicographical data. Different practices in lexicography may also be a reason that hinders the alignment task.

**RQ2.2** These datasets focus on the alignment task and the annotation of sense definitions based on various semantic relations, namely exact, narrower, broader, related and none. Given that the annotations are carried out by lexicographers and language experts, the benchmark incorporates human knowledge and expertise to evaluate future efforts in this field.

**RQ3.** What features in sense definitions can be used to create techniques for word sense alignment?
A few techniques are proposed in Chapter 6 for automatic alignment of monolingual word senses and definitions. One of the proposed alignment systems uses manually-crafted features such as length ratios, semantic similarities based on word embeddings and knowledge graphs. These features are effective to capture the semantic similarity and relation of a definition pair.

**RQ3.1** Among the proposed methods of the shared task (see Section 6.7), those based on fine-tuned contextual embeddings such as BERT perform better (see Table 6.10). In addition, incorporating external resources such as knowledge graphs are essential to determine semantic relations between senses. In addition to textual features that rely on the text of the definitions, structural features can be employed for the alignment task to increase the coverage of the predictions and set constraints.

**RQ3.2** In addition to the differences in the structure of senses in two dictionaries, the content of sense definitions affects the alignment task due to additional information such as cross-references or citations and merging various senses together, as in synonyms and near-synonyms. We analyze the impact of definition lengths in Section 6.6.3 and observe that truncating sense definitions improves the estimation of similarity between a definitions pair and therefore, can provide better alignments.

## 7.3 LIMITATIONS AND FUTURE DIRECTIONS

Despite the various methods developed within this Ph.D. project and described in the thesis along with the open-source lexicographical tools, particularly Naisc, that are developed with the ELEXIS project, there are a few limitations that would be compelling to be addressed in the future. Some of the main limitations of the current work are presented as follows:

(A) The alignment task defined in this thesis is limited to the following constraints:

- (a) When aligning at the sense level, only entries with the same grammatical categories are aligned. Therefore, sense definitions are not aligned per se regardless of the lemma that they belong to.
- (b) Due to the complexity in extracting senses, we did not take lexical combinatorics, such as multiword expressions, compound forms and collocations, into account in the alignment process.
- (c) The alignment task is defined at the monolingual level, i.e. monolingual dictionaries are aligned, and not at the multilingual level, i.e. no cross-lingual alignment is made.
- (d) Although sense definitions are the most important components of the micro-structure of an entry in a dictionary, they are not the only information provided. Therefore, other information such as usage examples, etymologies and cross-references could be included in the alignment task.
- (e) The alignment task is limited to the micro-structure of dictionaries and not words and their senses in context as in word sense disambiguation. The relation of the latter task and WSA could be explored to achieve synergies, given that word sense disambiguation has been more widely studied according to the literature.

(B) From an experimental point of view, one major limitation of the classification-based approach (see Section 6.6.2) is the usage of crafted features. We believe that as a future work further techniques can be used, particularly thanks to the current advances in word representations and neural networks. On the other hand, predicting certain semantic relations, such as related, narrower and broader, are deemed more challenging than the other relations, namely exact and none. One way to address this is to merge taxonomic information as in knowledge graphs like Wikidata with lexicographical resources like Wiktionary.

(C) As we initially stated, in the alignment of dictionaries where there is a hierarchy of senses, also known as vertical polysemy (Koskela, 2011), all senses are brought together at the same level, meaning that senses with sub-senses and super-senses are all considered as independent senses. This is a simplistic but pragmatic solution without further analyzing the task of determining how different levels of senses should be merged in a significant manner to capture the sense meaning completely. In order to understand the impact of this "leveling" technique, we look at the experiment results of Danish, as it is the only language in the MWSA benchmark whose sense information was implemented in a hierarchy. Having said that, the sense hierarchy structure can explicitly provide information about the semantic relationship between senses and therefore should preferably be considered in later experiments with the data.

Regarding future research directions, various fields can be explored to address the limitations of the current work. A few of them are proposed as follows:

(A) Despite the efforts to use various techniques for the alignment task, ranging from graph-based methods to those based on fine-tuning contextual embedding, there are various other methods that are yet to be explored. Among these, knowledge graphs as resources and techniques based on artificial intelligence are eminently promising to be incorporated in the alignment task.

(B) As thoroughly studied in this thesis, creation and maintenance of lexical semantic resources is a task that requires much time, cost and energy. Except for a small proportion of languages in the world, many languages are not privileged enough to financially or intellectually invest in the creation of such resources. Therefore, considering other languages other than richly-resourced European ones would be a natural step ahead to increase the usage and fair processing of world languages, particularly in the context of translation inference. Furthermore, this can pave the way for less Anglocentric language technology and bring other languages into the scene.

(C) Adopting more semantic-aware formats should not only be of importance for computational linguists but also lexicographers and those working in digital humanities. The current structure of Wiktionary, for instance, relies extensively on content and less on the structure and structural format such as Ontolex-Lemon (McCrae et al., 2017b). As a result, processing such a valuable open-source resource is a cumbersome task, except in a few efforts such as Dbnary (Sérasset, 2015). To promote interoperability, it is essential to organize lexical and semantic data in a systematic way so that retrieving and processing them will be facilitated.

(D) Converting printed historical dictionaries into structured electronic forms is an expensive and burdensome task. As future work, it is suggested to explore unsupervised methods to detect sense and definition boundaries in dictionaries such as ODS (see Section 6.6.3). Moreover, further methods can be created to automatically detect the type of the semantic relationship that may exist between two senses, also of non-identical lemmas, and also study to which degree manual markups of the meta-information in the ODS improve the method.

(E) Further linguistically-motivated approaches can be taken to create more robust alignment techniques; for instance, the syntactic analysis of a definition can be beneficial to detect the *genus* and *differentia* in a definition, therefore, facilitating the alignment task in some cases.

(F) Among the most promising future techniques that can be used for the alignment task are zero-shot classification (Ye and Guo, 2017), fine-tuning large language models and more advanced graph-based methods, such as modified adsorption analysis (Talukdar and Crammer, 2009) and graph neural networks (Wu et al., 2020) for translation inference. Given that current contextual embeddings have been shown to be able to distinguish word senses to some extent (see Section 2.10), it may be interesting to use the embeddings of the lemma along with the definitions as a piece of extra information for the alignment task.

(G) The MWSA task can be considered as a paraphrasis generation task (Yang et al., 2019) as well and can be seen from the perspective of natural text generation (Koncel-Kedziorski et al., 2019) too.

(H) Practices in lexicography should be more compatible with developments in NLP and language technology, particularly when it comes to publishing data using semantic web technologies and linguistic linked open data.